\documentclass[10pt,twocolumn,letterpaper]{article}

\usepackage{cvpr}
\usepackage{times}
\usepackage{epsfig}
\usepackage{graphicx}
\usepackage{amsmath}
\usepackage{amssymb}
\usepackage{microtype} % legal way to save space

\usepackage{tabularx}
\usepackage{booktabs}
\usepackage{rotating}
\usepackage{authblk}
\usepackage[toc,page]{appendix}

\newcommand{\myparagraph}[1]{\vspace{0.1em}\noindent\textbf{#1}}

\newcommand{\myurl}{\url{http://pose.mpi-inf.mpg.de}}

\usepackage[pagebackref=true,breaklinks=true,letterpaper=true,colorlinks,bookmarks=false]{hyperref}

\newcommand{\singb}{\textit{SP}}

\newcommand{\multb}{\textit{MP}}

\newcommand{\multbu}{\textit{MP UB}}

\newcommand{\detroi}{\textit{det ROI}}
\newcommand{\gtroi}{\textit{GT ROI}}

\newcommand{\cygtroi}{\textit{Chen\&Yuille~SP GT ROI}}

\newcommand{\rcnn}{\textit{AFR-CNN}}
\newcommand{\dense}{\textit{Dense-CNN}}
\newcommand{\deepcuts}{\textit{DeepCuts}}
\newcommand{\deepcut}{\textit{DeepCut}}

\cvprfinalcopy % *** Uncomment this line for the final submission

\def\cvprPaperID{538} % *** Enter the CVPR Paper ID here

\ifcvprfinal\fi
\begin{document}

%%%%%%%%% TITLE
\title{DeepCut: Joint Subset Partition and Labeling for Multi Person Pose Estimation}

%% \author{First Author\\
%% Institution1\\
%% Institution1 address\\
%% {\tt\small firstauthor@i1.org}
% For a paper whose authors are all at the same institution,
% omit the following lines up until the closing ``}''.
% Additional authors and addresses can be added with ``\and'',
% just like the second author.
% To save space, use either the email address or home page, not both
%% \and
%% Second Author\\
%% Institution2\\
%% First line of institution2 address\\
%% {\tt\small secondauthor@i2.org}
%% }

\author[1]{Leonid Pishchulin}
\author[1]{Eldar Insafutdinov}
\author[1]{Siyu Tang}
\author[1]{Bjoern Andres}
\author[1,3]{\\Mykhaylo Andriluka}
\author[2]{Peter Gehler}
\author[1]{Bernt Schiele}

\affil[1]{Max Planck Institute for Informatics, Germany}
\affil[2]{Max Planck Institute for Intelligent Systems, Germany}
\affil[3]{Stanford University, USA}

\maketitle
%\thispagestyle{empty}

%%%%%%%%% ABSTRACT
\begin{abstract}
  This paper considers the task of articulated human pose estimation
  of multiple people in real world images. We propose an approach that
  jointly solves the tasks of detection and pose estimation: it infers
  the number of persons in a scene, identifies occluded body parts,
  and disambiguates body parts between people in close proximity of
  each other. This joint formulation is in contrast to previous
  strategies, that address the problem by first detecting people and
  subsequently estimating their body pose. We propose a partitioning
  and labeling formulation of a set of body-part hypotheses generated
  with CNN-based part detectors. Our formulation, an instance of an
  integer linear program, implicitly performs non-maximum suppression
  on the set of part candidates and groups them to form configurations
  of body parts respecting geometric and appearance constraints.
  Experiments on four different datasets demonstrate state-of-the-art
  results for both single person and multi person pose
  estimation\footnote{Models and code available at \myurl}.
  %the single person only ``LSP'' benchmark, and
  %on scenes with multiple people from ``MPII Human Pose''.
\end{abstract}

%%%%%%%%% BODY TEXT
\section{Introduction}
\begin{figure}
  \centering
  \begin{tabular}{c c c }
%%   \includegraphics[width=0.31\linewidth]{figures/teaser/imgidx_0156_init_graph.pdf}&
%%   \includegraphics[width=0.31\linewidth]{figures/teaser/imgidx_0156_graph.pdf}&
%%   \includegraphics[width=0.31\linewidth]{figures/teaser/imgidx_0156_sticks.pdf}
%%   \\
  \includegraphics[width=0.31\linewidth]{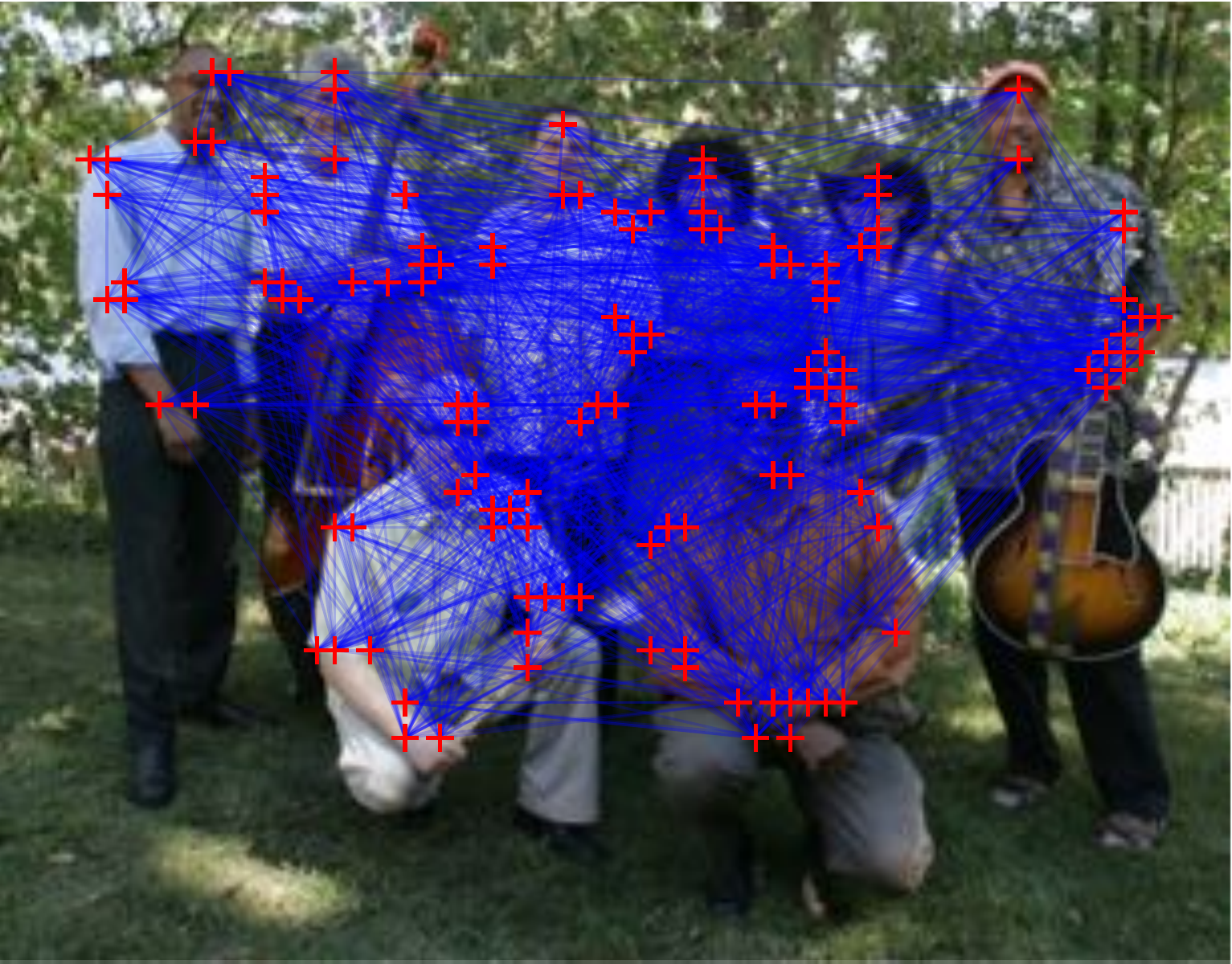}&
  \includegraphics[width=0.31\linewidth]{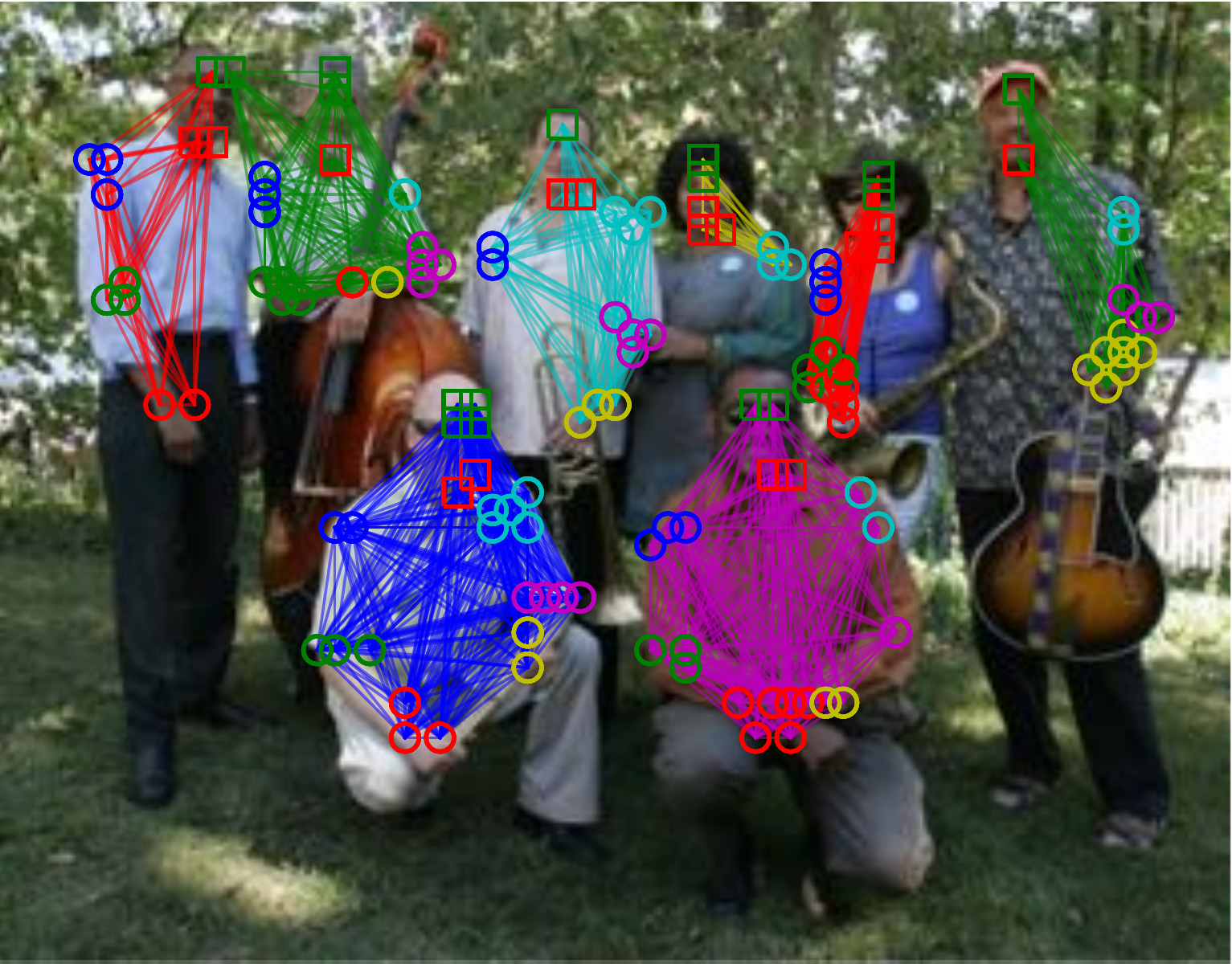}&
  \includegraphics[width=0.31\linewidth]{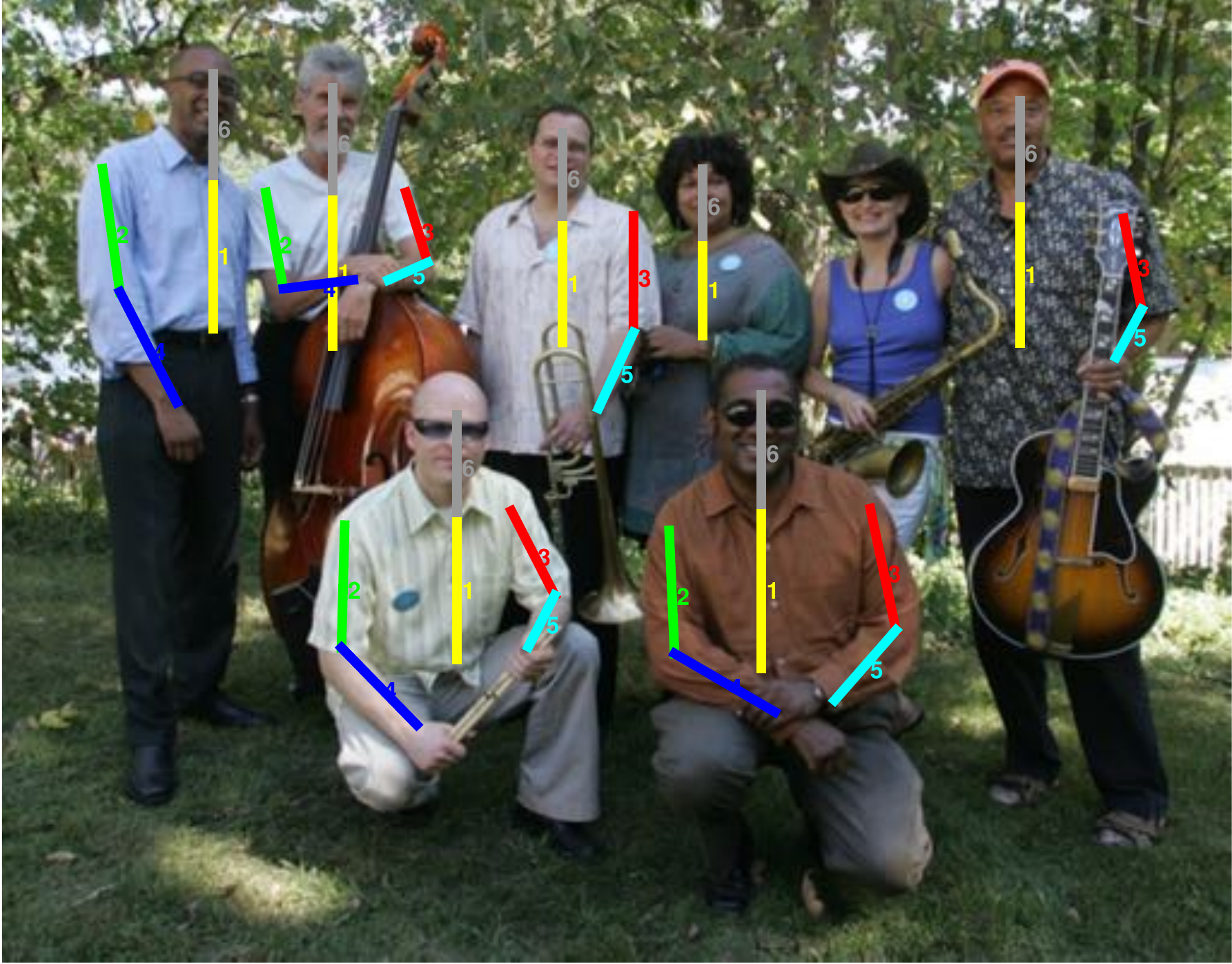}
  \\
  \includegraphics[width=0.31\linewidth]{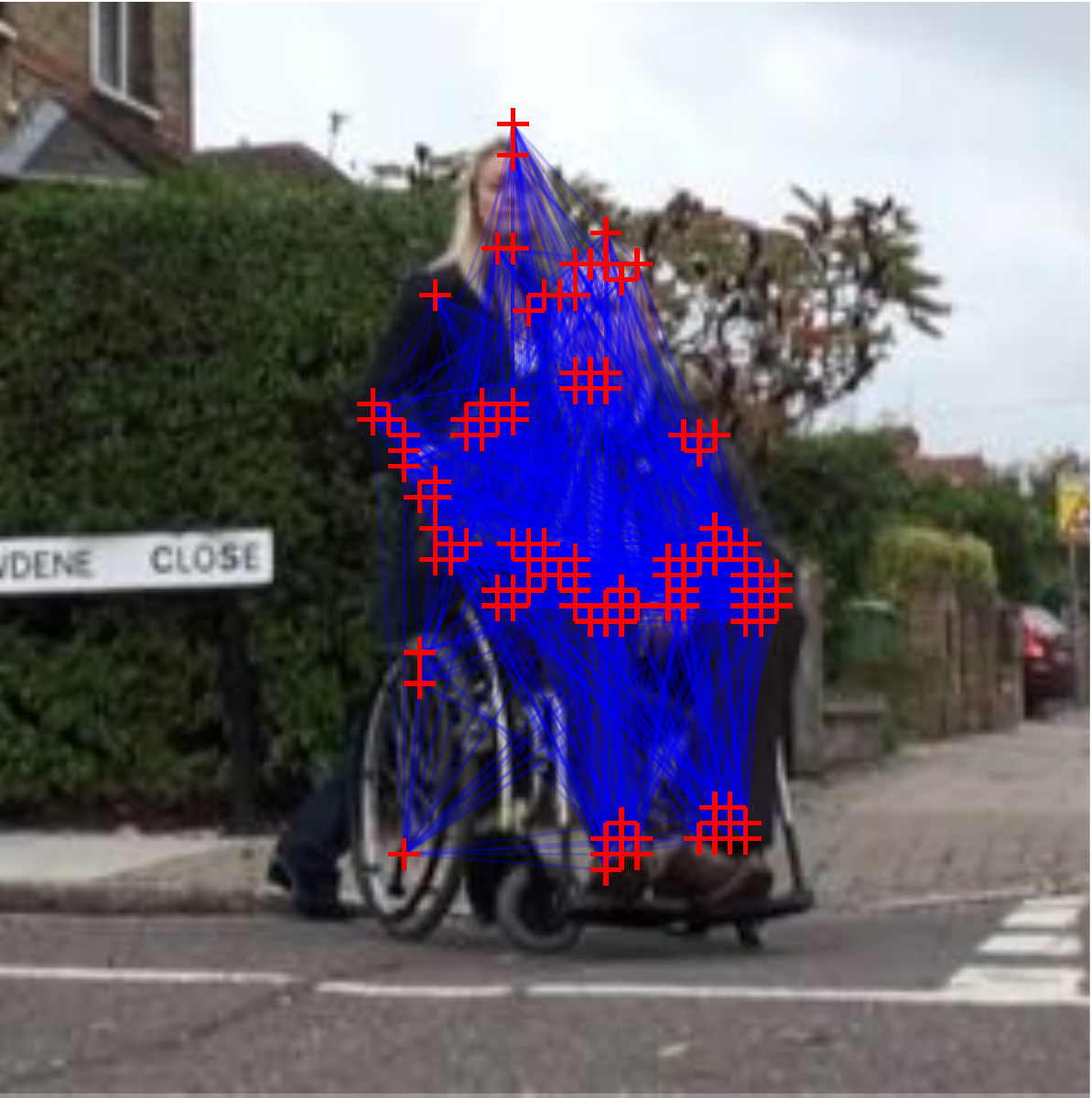}&
  \includegraphics[width=0.31\linewidth]{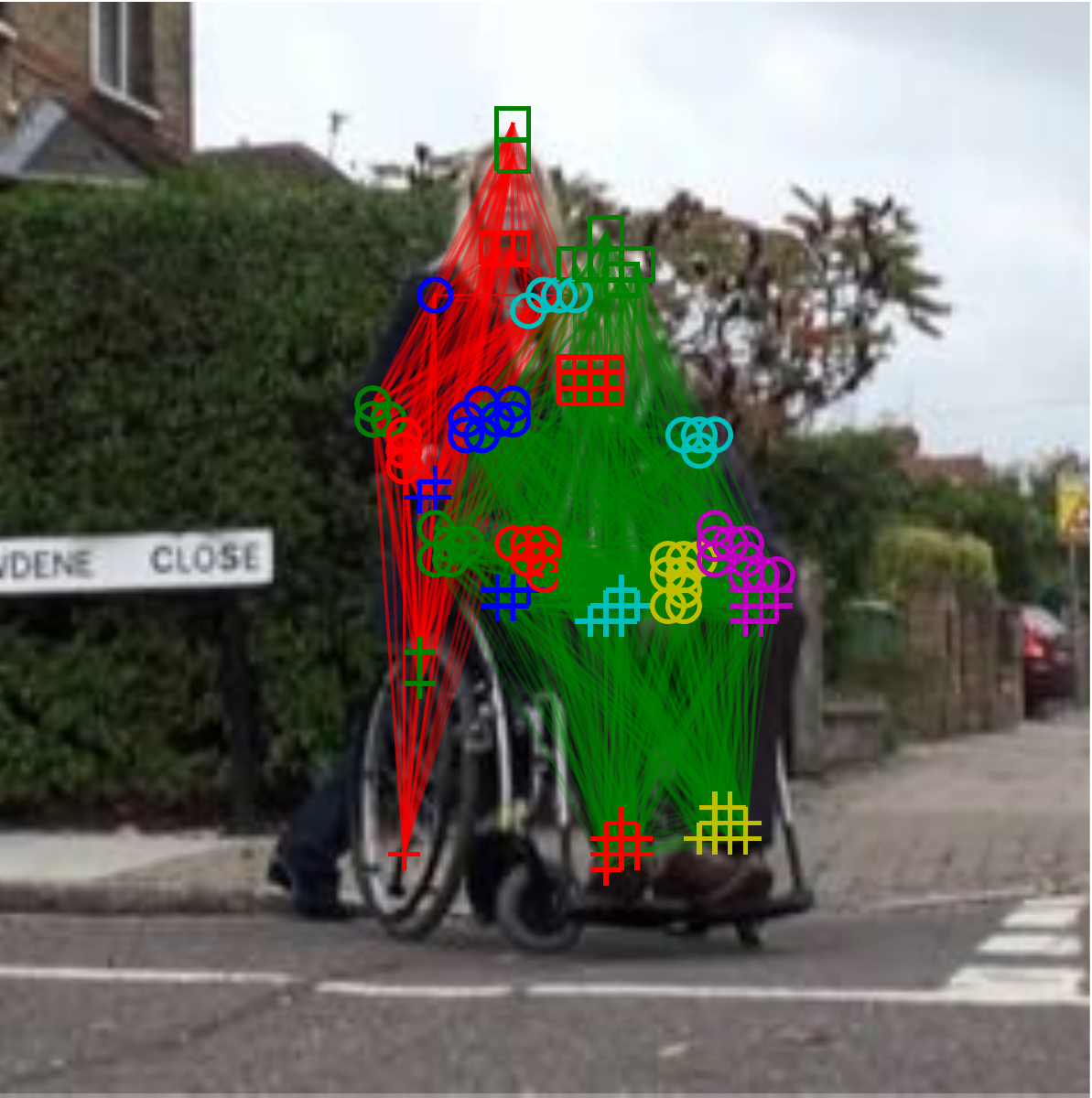}&
  \includegraphics[width=0.31\linewidth]{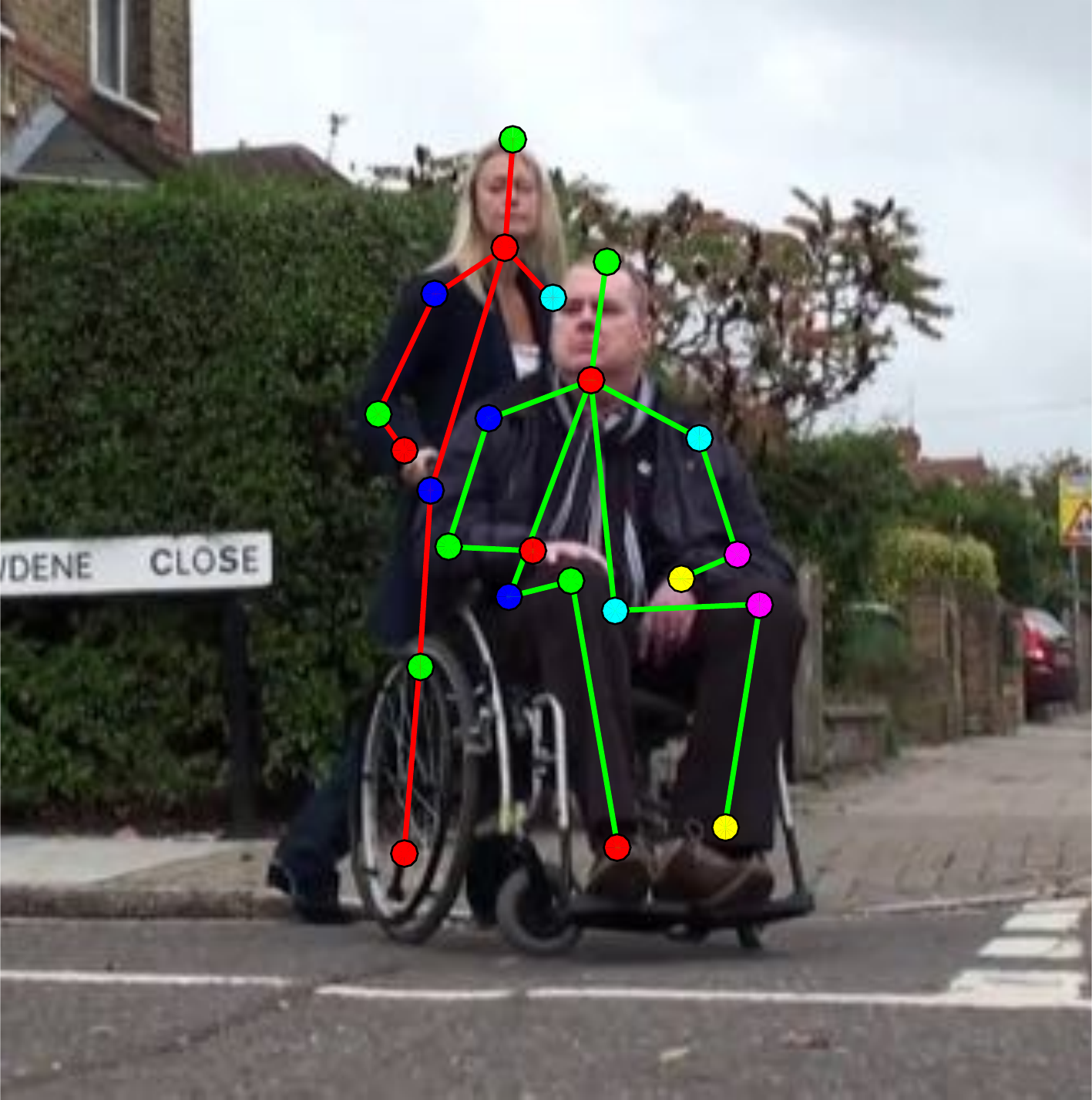} 
  \\
  \includegraphics[width=0.31\linewidth]{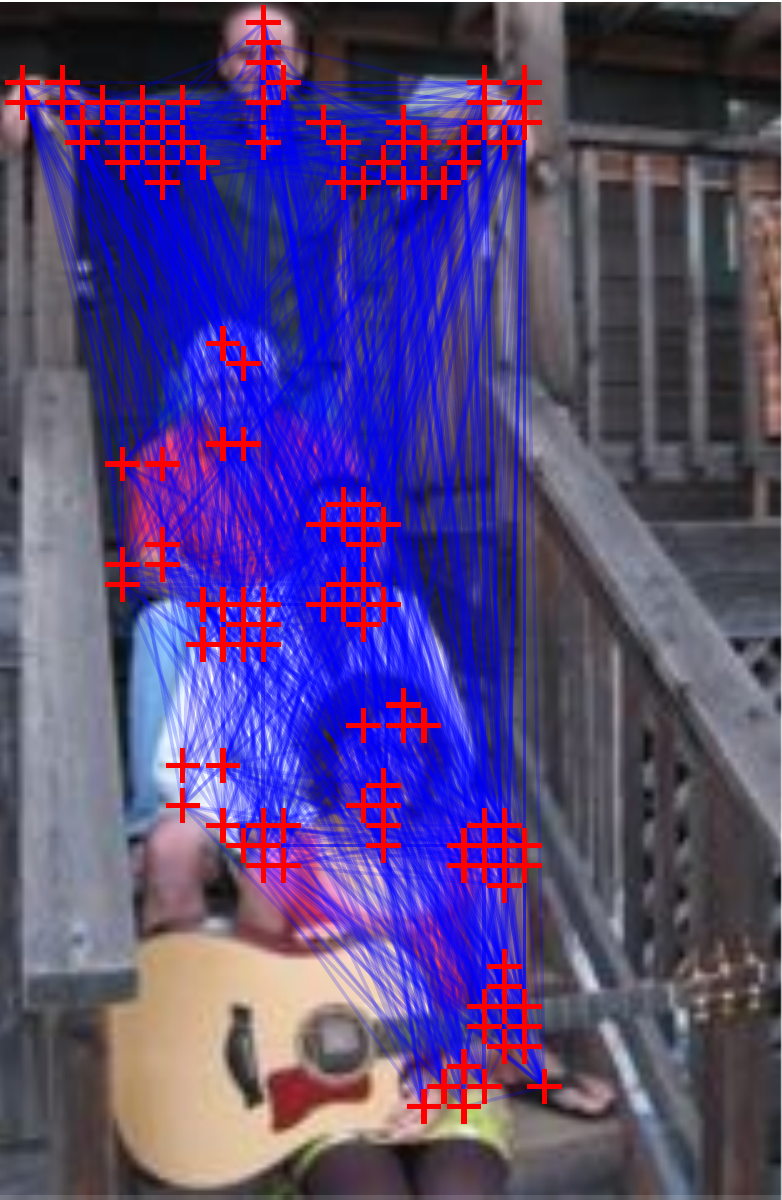}&
  \includegraphics[width=0.31\linewidth]{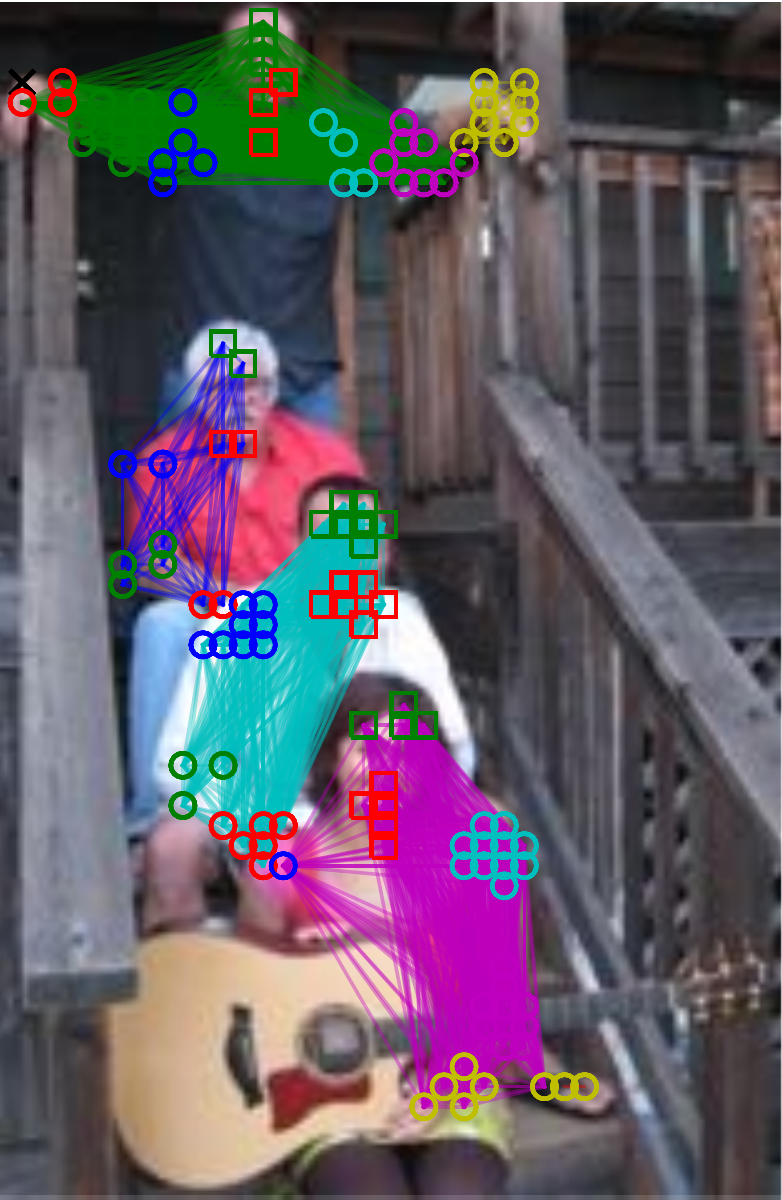}&
  \includegraphics[width=0.31\linewidth]{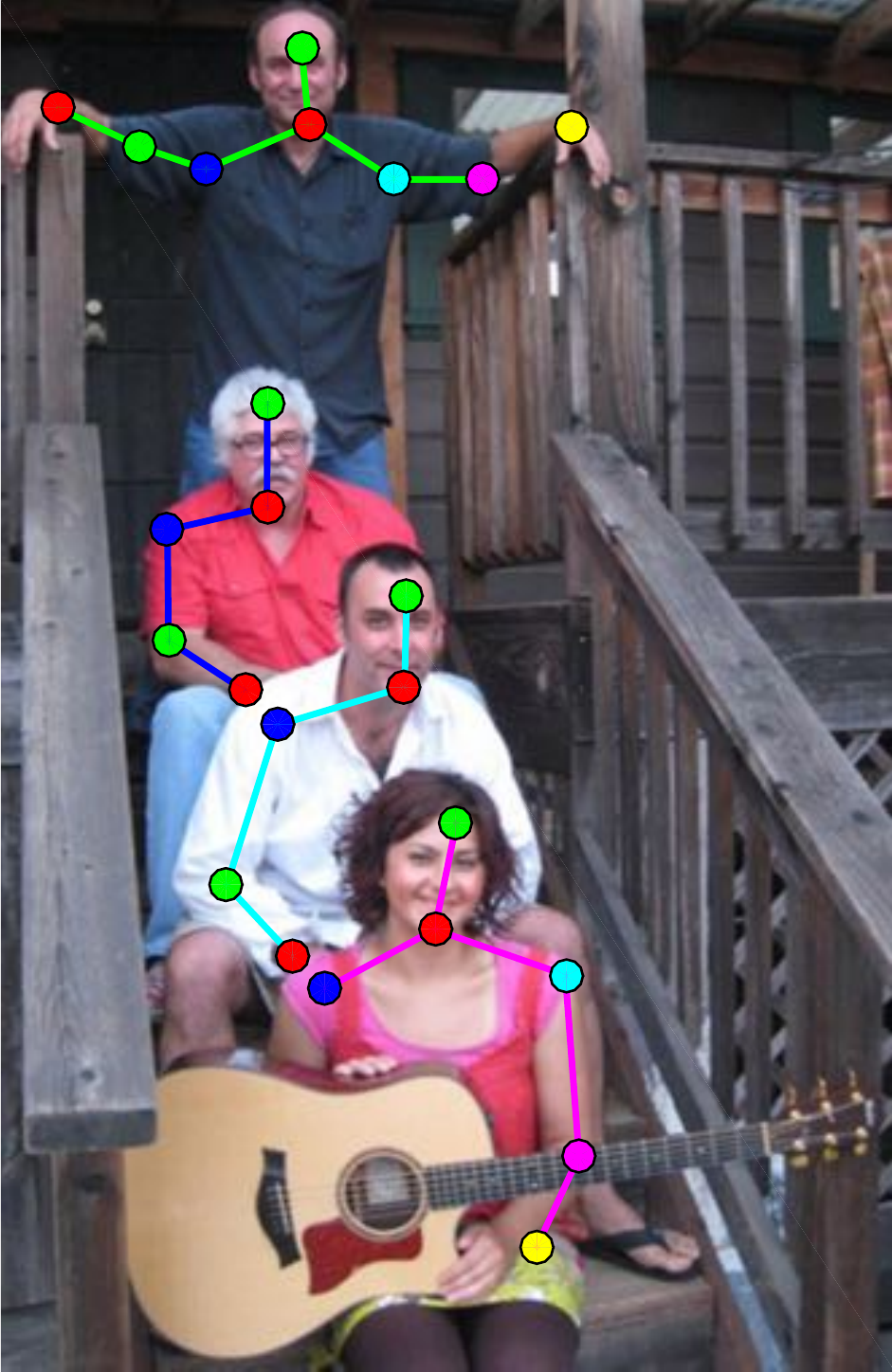}
  \\
%%   \includegraphics[width=0.31\linewidth]{figures/teaser/imgidx_0114_init_graph.pdf}&
%%   \includegraphics[width=0.31\linewidth]{figures/teaser/imgidx_0114_graph.pdf}&
%%   \includegraphics[width=0.31\linewidth]{figures/teaser/imgidx_0114_sticks.pdf}
%%   \\

%% %%   \includegraphics[width=0.31\linewidth]{figures/teaser/imgidx_0146_init_graph.pdf}&
%% %%   \includegraphics[width=0.31\linewidth]{figures/teaser/imgidx_0146_graph.pdf}&
%% %%   \includegraphics[width=0.31\linewidth]{figures/teaser/imgidx_0146_sticks.pdf}
%% %%   \\
%%   \includegraphics[width=0.31\linewidth]{figures/teaser/imgidx_0161_init_graph.pdf}&
%%   \includegraphics[width=0.31\linewidth]{figures/teaser/imgidx_0161_graph.pdf}&
%%   \includegraphics[width=0.31\linewidth]{figures/teaser/imgidx_0161_sticks.pdf}
%%   \\
%%   \includegraphics[width=0.31\linewidth]{figures/teaser/imgidx_0169_init_graph.pdf}&
%%   \includegraphics[width=0.31\linewidth]{figures/teaser/imgidx_0169_graph.pdf}&
%%   \includegraphics[width=0.31\linewidth]{figures/teaser/imgidx_0169_sticks.pdf}
%%   \\
  (a) & (b) & (c) \\   
  \end{tabular} 

  \caption{Method overview: (a) initial detections (= part
  candidates) and pairwise terms (graph) 
  between all detections that (b) are jointly clustered belonging to
  one person (one colored subgraph = one person) and each part is labeled
  corresponding to its part class (different colors and symbols
  correspond to different body parts); (c) shows the predicted pose
  sticks.}
 \vspace{-1.0em}
  \label{fig:overview}
\end{figure}

%\input{figure_teaser}
%In this paper we focus
%on the task of detecting and estimating poses of multiple people in
%uncontrolled real-world images.
Human body pose estimation methods have become
increasingly reliable.
% in localizing body joints.
Powerful body part
detectors \cite{Tompson:2015:EOL} in combination with
tree-structured body models \cite{tompson14nips,chen14nips} show
impressive results on diverse datasets
\cite{johnson11cvpr,andriluka14cvpr,sapp13cvpr}.
% \pgsuggestinstead{.
  These benchmarks promote pose estimation of
  single pre-localized persons but exclude
  scenes with multiple people.
%}{where people are pre-localized and it
%is required to estimate the pose of one person only. However these
%benchmarks specifically exclude scenes with multiple people that
%might overlap each other (\eg.~ as in Fig.~\ref{fig:teaserfig})}.
%\pgsuggest{
%
  This problem definition has been a driver for % the development of
  progress, but also falls short on representing a
  realistic sample of real-world images. Many photographs contain multiple
  people of interest (see Fig~\ref{fig:overview})
  and it is unclear whether single pose
  approaches generalize directly. We argue that
  the multi person case deserves more attention since it is an important real-world task.
  % that is of interest
  %for users and different applications.
  % shows samples of the MPII dataset~\cite{andriluka14cvpr}.

\tabcolsep 1.5pt

  Key challenges inherent to multi person pose estimation are the partial visibility
  of some people,
  %either due to truncation
  %or partial occlusion, and further,
  significant overlap of bounding box regions of people,
%
%In such conditions the number of
%body parts that can be predicted based on the image varies from person
%to person.
%\pgcomment{I do not understand that last sentence}
  and the a-priori unknown number of people in an image.
  The problem thus is to infer the number of persons, assign part detections
  to person instances while respecting geometric and
  appearance constraints.
%}{Since the number
%  of people in the image is not known in advance, this creates a
%  challenging problem where part detections need to be uniquely
%  assigned to unknown number of person instances while respecting
% geometric and appearance constraints.}
  Most strategies use a two-stage inference process
  \cite{pishchulin12cvpr,Gkioxari:2014:UKP,Sun:2011:APM} to first
  detect and then independently estimate poses. This is
  unsuited for cases when people are in close proximity since they
  permit simultaneous assignment of the same body-part candidates to
  multiple people hypotheses.
  %Therefore, in this paper, we approach
  %the pose estimation problem by jointly estimating the number of
  %people in the image, infer their spatial arrangement and reason
  %about body part visibility.

  As a principled solution for multi person pose estimation a model is
  proposed that jointly estimates poses of all people present in an
  image by minimizing a joint objective.  The formulation is based on
  partitioning and labeling an initial pool of body part candidates
  into subsets that correspond to sets of mutually consistent
  body-part candidates and abide to mutual consistency and exclusion
  constraints. The proposed method has a number of appealing
  properties. (1) The formulation is able to deal with an unknown
  number of people, and also infers this number by linking part
  hypotheses. (2) The formulation allows to either deactivate or merge
  part hypotheses in the initial set of part candidates hence
  effectively performing non-maximum suppression (NMS). In contrast to
  NMS performed on individual part candidates, the model incorporates
  evidence from all other parts making the process more reliable.  (3)
  The problem is cast in the form of an Integer Linear Program
  (ILP). Although the problem is NP-hard, the ILP formulation
  facilitates the computation of bounds and feasible solutions with a
  certified optimality gap.

%% Although this problem is NP-hard, the bound
%%   obtained from its dual allows to trade off run-time with a certified optimality gap.
%   { \textcolor{red}{(3) Our
%   approach is an integer linear program, the use
%   of robust optimization techniques allows for guarantees
%   of the optimality of the result with a relative optimality gap below 1\%.}}
% \pgcomment{This (3) is a weak statement. You are
%     basically suggesting that usually there is no sound mathematical
%     foundation.}

% We rely on recent advances in object detection and use CNN-based detector of \cite{} to generate
% initial set of body-part candidates.

% - infer number of people as number of clusters
% - infer visibility / turn off candidates inconsistent with the rest
% - can merge multiple candidates for the same part (NMS), NMS is done while taking info from other
% parts into account (in contrast to standard NMS that would only consider detection score)
% - based on sound mathematical framework, optimization is well understood, can be globally optimized
% (approximate solutions such as Kernighan-Lin available and can be used to speed up inference further \cite{}), other methods that propose global objective require greedy/iterative
% optimization /cite{Ladicky}

% The formulation includes a mechanism of either merging together similar candidates or ignoring
% candidates that are inconsistent with the final solution, thus implicitly performing a non-maximum
% suppression on body-part hypothesis set.

  This paper makes the following contributions. The main
  contribution is the derivation of a joint detection and pose estimation
  formulation cast as an integer linear program.
  Further, two CNN variants are proposed to generate
  representative sets of body part candidates. These, combined with the
  model, obtain state-of-the-art results for both single-person and
  multi-person pose estimation on different datasets.

\myparagraph{Related work.} 
Most work on pose estimation targets the 
single person case. Methods progressed from
simple part detectors and elaborate body models
\cite{Ren:2005:RHB,Ramanan:2006:LPI,Jiang:2008:GPE} to tree-structured
pictorial structures (PS) models with strong
part detectors
\cite{pishchulin13iccv,yang12pami,chen14nips,sapp13cvpr}. 
Impressive results are obtained predicting locations
of parts with convolutional neural networks (CNN)~\cite{Toshev:2014:DHP,Tompson:2015:EOL}. 
%For example
%\cite{Tompson:2015:EOL} proposes a model that does not rely on
%explicit body modeling, and instead encodes appearance of part
%configurations via a convolutional multi-scale image representation.
While body models are not a necessary component for effective
part localization, constraints among parts allow to assemble
independent detections into body configurations as 
demonstrated in~\cite{chen14nips} by combining CNN-based body part
detectors with a body model \cite{yang12pami}.
% while primarily
%focusing on the single-person case.

A popular approach to multi-person pose estimation is to detect people first and then estimate body
pose independently 
\cite{Sun:2011:APM,pishchulin12cvpr,yang12pami,Gkioxari:2014:UKP}. \cite{yang12pami} proposes a
flexible mixture-of-parts model for detection and pose estimation. 
\cite{yang12pami} obtains multiple pose hypotheses 
%by computing locations of body limbs 
corresponding to
different root part positions %(\eg , torso) 
and then performing non-maximum suppression.
\cite{Gkioxari:2014:UKP} detects people using a flexible configuration of
poselets and  
%. Given a person detection 
the body pose is predicted as a weighted average 
%over predictions
%of each activated 
of activated poselets. \cite{pishchulin12cvpr} detects people and then predicts poses of each
person using a PS model. \cite{Belagiannis:2014:3DP} estimates poses of multiple
people in 3D by constructing a shared space of 3D body part hypotheses, but uses 2D person detections
to establish the number of people in the scene. 
%While aiming to predict poses of multiple people these
These approaches are limited to cases with people sufficiently far from each other that do not have
overlapping body parts.

Our work is closely related to \cite{eichner10eccv,Ladicky:2013:HPE}
who also propose a joint objective to estimate poses of multiple
people. \cite{eichner10eccv} proposes a multi-person PS model that
explicitly models depth ordering and person-person occlusions. Our
formulation is not limited by a number of occlusion states among
people. \cite{Ladicky:2013:HPE} proposes a joint model for pose
estimation and body segmentation coupling pose estimates of
individuals by image segmentation.
\cite{eichner10eccv,Ladicky:2013:HPE} uses a person detector
to generate initial hypotheses for the joint
model. \cite{Ladicky:2013:HPE} resorts to a greedy approach of adding
one person hypothesis at a time until the joint objective can be
reduced, whereas our formulation can be solved with a certified optimality gap. In
addition \cite{Ladicky:2013:HPE} relies on expensive labeling of body
part segmentation, which the proposed approach does not require.

Similarly to \cite{Chen:2015:POC} we aim to distinguish between
visible and occluded body parts. \cite{Chen:2015:POC} primarily
focuse on the single-person case and handles multi-person scenes
akin to \cite{yang12pami}. 
% by performing part-based non-maximum
%suppression on the set of pose estimates. 
We consider the
 more difficult problem of full-body pose estimation,
whereas \cite{eichner10eccv,Chen:2015:POC} focus on upper-body poses
and consider a simplified case of people seen from the front.

Our work is related to early work on pose estimation that also relies
on integer linear programming to assemble candidate body part
hypotheses into valid configurations \cite{Jiang:2008:GPE}. Their
single person method employs a tree graph augmented with weaker
non-tree repulsive edges and expects the same number of parts. In
contrast, our novel formulation relies on fully connected model
%is proposed that extends
%beyond \cite{Jiang:2008:GPE} 
to deal with unknown number of people per image and body parts per
person.
%%  and multiple people, deal with a
%% variable number of parts and multiple candidate hypotheses per body
%% part.

The Minimum Cost Multicut Problem
\cite{chopra-1993,deza-1997},
known in machine learning as correlation clustering
\cite{bansal-2004},
has been used in computer vision for image segmentation 
\cite{alush-2012,andres-2011,kim-2014,yarkony-2012}
but has not been used before in the context of pose estimation.
It is known to be NP-hard
\cite{demaine-2006}.

%\cite{Gkioxari:2014:UKP} use agglomerative clustering to generate a set of final person detections
%from an initial candidate set. %This is similar to our approach (TODO).

% \begin{itemize}
% \item approaches that consider one person only and don't model multiple people jointly, assume that location/scale known 

% \item recent Yuille CVPR'15 - they do occlusion only, we focus on inferring poses of multiple people
%   - our model handles occlusions as well.

% \item We are the Family - Eichner ECCV, we deal with more complex poses, build on a stronger detector

% \item Leonid's CVPR'12, we jointly reason about multiple people, they first detect and then estimate
%   people poses.

% \item joint detection and pose estimation, Ramanan, Savarese, Gkioxari

% \item Ladicky CVPR'13 - similar idea, but based on segmentation, greedy selection of number of candidate person hypothesis. require segmentation labels at training time !!!

% \item Belagianis CVPR'13, multiple people in 3D, but greedy two step process (detect first)
% \end{itemize}

% Relation to agglomerative clustering used in Gkioxari et al.

\section{Problem Formulation}
\label{section:problem}
In this section,
the problem of estimating articulated poses of an unknown number of people in an image
is cast as an optimization problem.
The goal of this formulation is to state three problems jointly:
1.~The selection of a subset of body parts from a set $D$ of \emph{body part candidates},
estimated from an image as described in Section~\ref{section:unary}
and depicted as nodes of a graph in Fig.~\ref{fig:overview}(a).
2.~The \emph{labeling} of each selected body part with one of $C$ \emph{body part classes},
e.g., ``arm'', ``leg'', ``torso'',
as depicted in Fig.~\ref{fig:overview}(c).
3.~The \emph{partitioning} of body parts that belong to the same person,
as depicted in Fig.~\ref{fig:overview}(b).
% pg: make sure the references (a),(b),(c) are correct

\subsection{Feasible Solutions}
\label{section:feasible-solutions}
%
% In order to encode solutions to these three problems,
% we consider as feasible solutions of the optimization problem
% triples $(x,y,z)$ of 01-labelings
% $x: D \times C \to \{0,1\}$,
% $y: \tbinom{D}{2} \to \{0,1\}$ and
% $z: \tbinom{D}{2} \times C^2 \to \{0,1\}$.
We encode labelings of the three problems jointly through
triples $(x,y,z)$ of binary random variables with domains
$x\in\{0,1\}^{D\times C}, y\in\{0,1\}^{\tbinom{D}{2}}$ and $z\in\{0,1\}^{\tbinom{D}{2} \times C^2}$.
Here, $x_{dc} = 1$ indicates that body part candidate $d$ is of class $c$,
$y_{dd'} = 1$ indicates that the body part candidates $d$ and $d'$ belong to the same person,
% and $z_{dd'cc'}$ is constrained such that $z_{dd'cc'} = x_{dc} x_{d'c'} y_{dd'}$.
and $z_{dd'cc'}$ are auxiliary variables to relate $x$ and $y$ through $z_{dd'cc'} = x_{dc} x_{d'c'} y_{dd'}$.
Thus, $z_{dd'cc'} = 1$ indicates that
body part candidate $d$ is of class $c$ ($x_{dc}=1$),
body part candidate $d'$ is of class $c'$ ($x_{d'c'}=1$),
and body part candidates $d$ and $d'$ belong to the same person ($y_{dd'}=1$).

In order to constrain the 01-labelings $(x,y,z)$ to well-defined articulated poses of one or more people,
we impose the linear inequalities
\eqref{eq:constraint-map}--\eqref{eq:constraint-cycle}
stated below.
%
% pg: Maybe the next section can be ordered more nicely?
%
Here, the inequalities
\eqref{eq:constraint-map}
guarantee that every body part is labeled with at most one body part class.
(If it is labeled with no body part class, it is suppressed).
The inequalities
\eqref{eq:constraint-suppression-1}
guarantee that distinct body parts $d$ and $d'$ belong to the same person only if neither $d$ nor $d'$ is suppressed.
The inequalities
\eqref{eq:constraint-cycle}
guarantee, for any three pairwise distinct body parts, $d$, $d'$ and $d''$, if $d$ and $d'$ are the same person (as indicated by $y_{dd'} = 1$) and $d'$ and $d''$ are the same person (as indicated by $y_{d'd''} = 1$), then also $d$ and $d''$ are the same person ($y_{dd''} = 1$),
that is, transitivity,
cf.~\cite{chopra-1993}.
Finally, the inequalities
\eqref{eq:constraint-aux-1}
guarantee, for any $dd' \in \tbinom{D}{2}$ and any $cc' \in C^2$ that $z_{dd'cc'} = x_{dc} x_{d'c'} y_{dd'}$.
These constraints allow us to write an objective function as a linear form in $z$ that would otherwise be written as a cubic form in $x$ and $y$.
We denote by $X_{DC}$ the set of all $(x,y,z)$ that satisfy all inequalities,
i.e., the set of feasible solutions.
\begin{align}
\forall d \in D \forall cc' \in \tbinom{C}{2}:
    & \quad x_{dc} + x_{dc'} \leq 1
    \label{eq:constraint-map}\\
\forall dd' \in \tbinom{D}{2}:
    & \quad y_{dd'} \leq \sum_{c \in C} x_{dc}
    \nonumber\\
    & \quad y_{dd'} \leq \sum_{c \in C} x_{d'c}
    \label{eq:constraint-suppression-1}\\
\forall dd'd'' \in \tbinom{D}{3}:
    & \quad y_{dd'} + y_{d'd''} - 1 \leq y_{dd''}
    \label{eq:constraint-cycle}\\
\forall dd' \in \tbinom{D}{2} \forall cc' \in C^2:
    & \quad x_{dc} + x_{d'c'} + y_{dd'} - 2 \leq z_{dd'cc'}
    \nonumber\\
    & \quad z_{dd'cc'} \leq x_{dc}
    \nonumber\\
    & \quad z_{dd'cc'} \leq x_{d'c'}
    \nonumber\\
    & \quad z_{dd'cc'} \leq y_{dd'}
    \label{eq:constraint-aux-1}
\end{align}

When at most one person is in an image,
we further constrain the feasible solutions to a well-defined pose of a single person.
This is achieved by an additional class of inequalities which guarantee, for any two distinct body parts that are not suppressed, that they must be clustered together:
\begin{align}
\forall dd' \in \tbinom{D}{2} \forall cc' \in C^2:
    \quad x_{dc} + x_{d'c'} - 1 \leq y_{dd'}
    \label{eq:constraint-single-person}
\end{align}

\subsection{Objective Function}
% pg: I'd remove this, just stating what follows.
%
% We define an objective function on the feasible set with respect to the probabilities introduced below.
% The estimation of these probabilities from images is discussed in
% Section~\ref{sec:unary-and-pairwise}.

% p\in (0,1) or [0,1]
For every pair $(d, c) \in D \times C$, we will estimate a probability $p_{dc} \in [0,1]$ of the body part $d$ being of class $c$.
In the context of CRFs, these probabilities are called \emph{part unaries} and we will detail their estimation in Section~\ref{section:unary}.

For every $dd' \in \tbinom{D}{2}$ and every $cc' \in C^2$, we consider a probability
$p_{dd'cc'} \in (0,1)$ of the conditional probability of $d$
and $d'$ belonging to the same person, given that $d$ and $d'$ are
body parts of classes $c$ and $c'$, respectively.
For $c \neq c'$, these probabilities $p_{dd'cc'}$ are the \emph{pairwise terms} in a graphical model of the human body.
In contrast to the classic pictorial structures model, our model allows for a {\em fully connected graph} where each body part is connected to all other parts in the entire set $D$ by a pairwise term.
%
% pg: do not understand this.
For $c = c'$, $p_{dd'cc'}$ is the probability of the part candidates
$d$ and $d'$ representing the same part of the same person. This
facilitates \textit{clustering} of multiple part candidates of the
same part of the same person and a \textit{repulsive} property that
prevents nearby part candidates of the same type to be associated to
different people.
%On the other hand, transitivity property~\eqref{eq:constraint-cycle} facilitates \textit{partitioning} of body parts that belong to the same person.

% Our model also implements the ``repulsive interaction'' between the body parts. 
% R1 "pairwise terms between body parts of people adjacent to each other": the model already includes
% such terms, e.g. the graph in Fig. 2 (d) shows connections between body parts of adjacent
% people. The "repulsive interaction" is implemented via cost defined in Eq.14. Note that for the same
% part class the features used to define this cost are based on image proximity and bounding box
% overlap (see Eq. 12). Assigning two detections d and d' of same class to different people requires
% setting z_dd'cc' to 0 (see definition of z on line 284). This will have high cost for detections in
% close proximity.

The optimization problem that we call the \emph{subset partition and labeling problem} is the ILP that minimizes over the set of feasible solutions $X_{DC}$:
\begin{align}
\min_{(x,y,z) \in X_{DC}} \ \
    & \langle \alpha, x \rangle + \langle \beta, z \rangle,
    \label{eq:ilp}
\end{align}
%
% with $X_{DC}$ the set of feasible solutions defined in
% Section~\ref{section:feasible-solutions}
where we used the short-hand notation
\begin{align}
\alpha_{dc}
    & := \log \frac{1 - p_{dc}}{p_{dc}}
    \label{eq:cost-definition-1} \\
\beta_{dd'cc'}
    & := \log \frac{1 - p_{dd'cc'}}{p_{dd'cc'}}
    \label{eq:cost-definition-2}\\
%\end{align}
%%
%and
%%
%\begin{align}
\langle \alpha, x \rangle
    & := \sum_{d \in D} \sum_{c \in C} \alpha_{dc} \, x_{dc}\\
\langle \beta, z \rangle
    & := \sum_{d d' \in \tbinom{D}{2}} \sum_{c,c' \in C} \beta_{dd'cc'} \, z_{dd'cc'}
\enspace .
\label{eq:beta-z}
\end{align}

The objective \eqref{eq:ilp}--\eqref{eq:beta-z} is the MAP estimate of
a probability measure of joint detections $x$ and clusterings $y,z$ of
body parts, where prior probabilities $p_{dc}$ and $p_{dd'cc'}$ are
estimated \emph{independently} from data, and the likelihood is a
positive constant if $(x,y,z)$ satisfies
\eqref{eq:constraint-map}--\eqref{eq:constraint-aux-1}, and is 0,
otherwise. The exact form \eqref{eq:ilp}--\eqref{eq:beta-z} is
obtained when minimizing the negative logarithm of this probability measure.

\subsection{Optimization}
In order to obtain feasible solutions of the ILP
\eqref{eq:ilp}
with guaranteed bounds,
we separate the inequalities \eqref{eq:constraint-map}--\eqref{eq:constraint-single-person}
in the branch-and-cut loop of the state-of-the-art ILP solver Gurobi.
More precisely, we solve a sequence of relaxations of the problem
\eqref{eq:ilp},
starting with the (trivial) unconstrained problem.
Each problem is solved using the cuts proposed by Gurobi.
Once an integer feasible solution is found,
we identify violated inequalities
\eqref{eq:constraint-map}--\eqref{eq:constraint-single-person},
if any, by breadth-first-search, add these to the constraint pool and re-solve the tightened relaxation.
Once an integer solution satisfying all inequalities is found,
together with a lower bound that certifies an optimality gap below 1\%,
we terminate.

\section{Pairwise Probabilities}
\label{sec:unary-and-pairwise}
Here we describe the estimation of the pairwise terms.
% , while unary probabilities are described in
% Sec.~\ref{sec:unary-prob}.
We define pairwise features $f_{dd'}$ for
the variable $z_{dd'cc'}$ (Sec.~\ref{section:problem}).
% where $dd' \in \tbinom{D}{2}$ and $cc'\in C^2$.
Each part detection $d$ includes the probabilities $f_{p_{dc}}$
(Sec.~\ref{sec:unary-prob}), its location $(x_d,y_d)$, scale $h_d$ and
bounding box $B_d$ coordinates.  Given two detections $d$ and $d'$,
and the corresponding features $( f_{p_{dc}}, x_d, y_d, h_d, B_d)$ and
$(f_{p_{d'c}}, x_{d'}, y_{d'}, h_{d'}, B_{d'})$, we define two sets of
auxiliary variables for $z_{dd'cc'}$, one set for $c = c'$ (same body
part class clustering) and one for $c\neq c'$ (across two body part
classes labeling).  These features capture the proximity, kinematic
relation and appearance similarity between body parts.
% \pgcomment{These are designed as to capture...}

\myparagraph{The same body part class ($c=c'$).}
% \paragraph{$c = c'$} We leverage the spatial relations between the
%detections $d$ and $d'$ for the pairwise term within the same body
%part class.
Two detections denoting the same body part of the same person should
be in close proximity to each other. We introduce the following
auxiliary variables that capture the spatial relations: $\Delta x =
|x_d-x_{d'}|/\bar{h}$, $\Delta y = |y_d-y_{d'}|/\bar{h}$, $\Delta h =
|h_d-h_{d'}|/\bar{h}$, $IOUnion$, $IOMin$, $IOMax$. The latter three
are intersections over union/minimum/maximum of the two detection
boxes, respectively, and $\bar{h} = (h_d + h_{d'})/2$.
%% \begin{alignat}{4}
%% \small
%% \smallskip
%%   \Delta x &= \frac{|x_d-x_{d'}|}{\bar{h}}   &\quad IOUnion &= \frac{|B_d \cap B_{d'}|}{|B_d \cup B_{d'}|}     \nonumber \\
%%   \Delta y &= \frac{|y_d-y_{d'}|}{\bar{h}}  &\quad IOMin &= \frac{|B_d \cap B_{d'}|}{\min(|B_d|, |B_{d'}|)}     \\
%%   \Delta h &= \frac{|h_d-h_{d'}|}{\bar{h}}   &\quad IOMax &= \frac{|B_d \cap B_{d'}|}{\max(|B_d|, |B_{d'}|)} \nonumber
%% \end{alignat}
%% where $\bar{h} = \frac{(h_d + h_{d'})}{2}$, $IOUnion$, $IOMin$ and
%% $IOMax$ are intersections over union/minimum/maximum of the two
%% detection bounding box areas, respectively.

{\it Non-linear Mapping.}
We augment the feature representation by appending quadratic and exponential terms.
The final pairwise feature  $f_{dd'}$ for the variable $z_{dd'cc}$ is
$(\Delta x, \Delta y, \Delta h,IOUnion, IOMin, IOMax,{(\Delta x)}^2, \\ \ldots, {(IOMax)}^2, \exp{(-{\Delta x})}, \ldots, \exp{(-{IOMax})})$.

\myparagraph{Two different body part classes ($c\neq c'$).}
% \paragraph{$c\neq c'$}
We encode the kinematic body constraints into the pairwise feature by
introducing auxiliary variables $ S_{dd'}$ and $R_{dd'}$, where
$S_{dd'}$ and $R_{dd'}$ are the Euclidean distance and the angle
between two detections, respectively. To capture the joint
distribution of $S_{dd'}$ and $R_{dd'}$, instead of using $S_{dd'}$
and $R_{dd'}$ directly, we employ the posterior probability
$p(z_{dd'cc'} = 1|S_{dd'},R_{dd'})$ as pairwise feature for
$z_{dd'cc'}$ to encode the geometric relations between the body part
class $c$ and $c'$.  More specifically, assuming the prior probability
$p(z_{dd'cc'} = 1) = p(z_{dd'cc'} = 0) = 0.5$, the posterior
probability of detection $d$ and $d'$ have the body part label $c$ and
$c'$, namely $z_{dd'cc'} = 1$, is
\vspace{-0.1cm}
\begin{align}
\small
\noindent
&p(z_{dd'cc'} = 1|S_{dd'},R_{dd'}) \nonumber \\ &=
\frac{p(S_{dd'},R_{dd'}|z_{dd'cc'} = 1)}{p(S_{dd'},R_{dd'}|z_{dd'cc'}
  = 1) + p(S_{dd'},R_{dd'}|z_{dd'cc'} = 0)}, \nonumber
\enspace
\end{align}
\vspace{-0.1cm}

\noindent where $p(S_{dd'},R_{dd'}|z_{dd'cc'} = 1)$ is obtained by
conducting a normalized 2D histogram of $S_{dd'}$ and $R_{dd'}$ from
positive training examples, analogous to the negative likelihood
$p(S_{dd'},R_{dd'}|z_{dd'cc'} = 0)$. In Sec.~\ref{sec:multicut:single}
we also experiment with encoding the appearance into the pairwise
feature by concatenating the feature $f_{p_{dc}}$ from $d$ and
$f_{p_{d'c}}$ from $d'$, as $f_{p_{dc}}$ is the output of the CNN-based part
detectors. The final pairwise feature is $(p(z_{dd'cc'} =
1|S_{dd'},R_{dd'}),f_{p_{dc}},f_{p_{d'c}})$.

%Regarding to the unary feature, we implicitly employ the CNN feature
%since we directly use the R-CNN softmax output as the probability
%estimation for $x_{dc}$, which is trained based on the CNN feature.
%We discuss it in the next section.
\subsection{Probability Estimation} % {\todo{should it be here?}}}
\label{sec:probability-estimation}
% the unary cost $\alpha_{dc}$ and the pairwise cost $\beta_{dd'cc'}$
% are defined in terms of the probability ratio.
The coefficients $\alpha$ and $\beta $ of the objective function
(Eq.~\ref{eq:ilp}) are defined by the probability ratio
in the log space (Eq.~\ref{eq:cost-definition-1} and
Eq.~\ref{eq:cost-definition-2}).  Here we describe the estimation of
the corresponding probability density: {\it (1)} For every pair of
detection and part classes, namely for any $(d, c) \in D \times C$, we
estimate a probability $p_{dc} \in (0,1)$ of the detection $d$ being a
body part of class $c$.  {\it (2)} For every combination of two
distinct detections and two body part classes, namely for any $dd' \in
\tbinom{D}{2}$ and any $cc' \in C^2$, we estimate a probability
$p_{dd'cc'} \in (0,1)$ of $d$ and $d'$ belonging to the same person,
meanwhile $d$ and $d'$ are body parts of classes $c$ and $c'$,
respectively.

\myparagraph{Learning.} 
%% The output of the R-CNN softmax layer
%% contains the jointly estimated probability of the detection $d$ being
%% each body part class $c$, which we directly use as the probability
%% estimation for $p_{dc}$.  Learning the parameters of the softmax
%% classifier from the training data is done in the R-CNN framework in
%% the standard way.
%%As for $p_{dd'cc'}$, 
Given the features $f_{dd'}$ and a Gaussian prior $p(\theta_{cc'}) =
\mathcal{N}(0,\sigma^2)$ on the parameters, logistic model is
%logistic model
\begin{align}
\label{eq:z-probability}
p(z_{dd'cc'} = 1 | f_{dd'}, \theta_{cc'}) & = \frac{1}{1 + \exp(-\langle \theta_{cc'}, f_{dd'} \rangle)}.
\end{align}
$(|C|\times (|C| + 1))/2 $ parameters are estimated using ML.
%Maximum Likelihood. % on the training set.

%}{Estimating the maximally probable model parameter $\theta_{cc'}$
%  from the training data is formulated as s logistic regression
%  problem. We perform the logistic regression for learning the model
%  parameter $\theta_{cc'}$ for every pair of body part classes, in
%  total $(|C|\times (|C| + 1))/2 $.}
\myparagraph{Inference}
Given two detections $d$ and $d'$,
the coefficients $\alpha_{dc}$ for $x_{dc}$ and $\alpha_{d'c}$ for $x_{d'c}$ are obtained by Eq.~\ref{eq:cost-definition-1},
the coefficient $\beta_{dd'cc'}$ for $z_{dd'cc'}$ has the form
\begin{align}
\small
  \beta_{dd'cc'} = \log \frac{1 - p_{dd'cc'}}{p_{dd'cc'}} = -\langle f_{dd'},  \theta_{cc'}\rangle.
  \label{cost-probability-feature}
\end{align}
%with pairwise features $f_{dd'}$, and 
%learned model parameters $\theta_{cc'}$ from logistic regression.
Model parameters $\theta_{cc'}$ are learned using logistic regression.
% \subsection{Implementation Details}
% \label{sec:features:details}
% Using all $2000$ detections in our model is prohibitive due to
% exponential complexity of the fully connected part graph. We thus
% select a representative subset of detections

% \begin{itemize}

% \item selecting representative set of unaries
% \item min probability for unaries
% \end{itemize}

\section{Body Part Detectors}
\label{section:unary}
\setlength{\belowdisplayskip}{1pt} \setlength{\belowdisplayshortskip}{1pt}
\setlength{\abovedisplayskip}{1pt} \setlength{\abovedisplayshortskip}{1pt}

%% Here we introduce deep learning-based part detection models. First, we
%% describe our proposal-based model inspired by Fast
%% R-CNN~\cite{girshickICCV15fastrcnn}
%% (\ref{sec:R-CNN-for-pose-estimation}). Then, we introduce our novel
%% model that outputs dense part detection scoremaps at the fine image
%% resolution (\ref{sec:Dense-unary}). Finally, we evaluate both
%% detection models on two prominent single person pose estimation
%% benchmarks and show significant performance improvements over the
%% state of the art.

We first introduce our deep learning-based part detection models and
then evaluate them on two prominent benchmarks thereby significantly
outperforming state of the art.

\subsection{Adapted Fast R-CNN ($\rcnn$)}
\label{sec:R-CNN-for-pose-estimation}
To obtain strong part detectors we adapt Fast
R-CNN~\cite{girshickICCV15fastrcnn}.
%% FR-CNN takes as input an image and set of class-independent region
%% proposals generated by selective search~\cite{Uijlings13}. First, an
%% image is processed with several network layers to produce a
%% convolutional scoremap. Next, for each proposal window a fixed-length
%% feature vector is extracted from the feature map and fed into
%% fully-connected (FC) layers. The outputs are softmax probabilities over
%% all classes and refined bounding boxes. To adapt FR-CNN for part
%% detection we alter it in two ways: {\it 1)} proposal generation and
%% {\it 2)} detection region size. The adapted version is called~$\rcnn$
%% throughout the paper.
FR-CNN takes as input an image and set of class-independent region
proposals~\cite{Uijlings13}
%% generated by selective search~\cite{Uijlings13}
and outputs the softmax probabilities over all classes and refined
bounding boxes. To adapt FR-CNN for part detection we alter it in two
ways: {\it 1)} proposal generation and {\it 2)} detection region
size. The adapted version is called~$\rcnn$ throughout the paper.

\myparagraph{Detection proposals.} Generating object proposals is
essential for FR-CNN, meanwhile detecting body parts is challenging
due to their small size and high intra-class variability.
%caused by articulation, shape, appearance and viewpoint changes.
We use DPM-based part detectors \cite{pishchulin13iccv} for proposal
generation.
%% The rotation space of each body part is discretized into $16$
%% orientation bins, obtained by clustering of absolute rotations of
%% training samples. Training samples are assigned to the corresponding
%% rotation bin and a $16$ component deformable part model (DPM) is
%% trained from image patches extracted around body joints. This results
%% in body part detectors based on rotation-dependent mixtures. Finally,
We collect $K$ top-scoring detections by each part detector in a
common pool of $N$ part-independent proposals and use these proposals as
input to \rcnn. $N$ is $2,000$ in case of single and $20,000$ in case
of multiple people.

\myparagraph{Larger context.} Increasing the size of DPM detections by
upscaling every bounding box by a fixed factor allows to capture more
context around each part. In Sec.~\ref{sec:experiments:unary} we
evaluate the influence of upscaling and show that using larger context
around parts is crucial for best performance.

\myparagraph{Details.} Following standard FR-CNN training procedure
ImageNet models are finetuned on pose estimation task.
%% Using data augmentation and selecting optimal parameters
%% (Sec.~\ref{sec:unary:rcnn}) , such as learning rate, number of SGD
%% iterations, assignment of positive/negative labels, and bounding box
%% regression in FR-CNN significantly improves the
%% performance. Furthermore, using deeper architectures, such as
%% VGG~\cite{Simonyan14c} lead to better results compared to smaller
%% networks like AlexNet~\cite{krizhevsky12nips}. We evaluate different
%% network architectures and several important parameters in
%% Sec.~\ref{sec:experiments:unary} and provide more detailed evaluation
%% in the supplementary material. Finally,
Center of a predicted bounding box is used for body part location
prediction. See Appendix~\ref{seq:supplemental:lsp} for detailed parameter analysis.

\subsection{Dense Architecture (\dense)}
\label{sec:Dense-unary}
Using proposals for body part detection may be sub-optimal.
We thus develop a fully convolutional architecture for computing part
probability scoremaps.
%% One of the major drawbacks of using RCNN is reliance on
%% proposals. On the other hand approach described in
%% \cite{tompson14nips} and \cite{Tompson:2015:EOL} uses fully
%% convolutional CNN for part detectors.

\myparagraph{Stride.} We build on VGG~\cite{Simonyan14c}. Fully
convolutional VGG has stride of 32 px -- too coarse for precise part
localization. We thus use hole algorithm~\cite{chen14semantic} to
reduce the stride to 8 px.

\myparagraph{Scale.} Selecting image scale is crucial. We
%% In RCNN we can
%% vary the size of proposals, whereas in CNN the size of receptive field
%% remains fixed (224 pixels for VGG), so
 found that scaling to a standing height of $340$ px performs
 best: %$400\times400$ 
 VGG receptive field sees entire
 body to disambiguate body parts.

\myparagraph{Loss function.}
%% Part detection can be viewed as a multi-class classification problem, where
%% typically softmax is used with multinomial logistic loss. In such formulation
%% each class corresponds to a part and we additionally have one class for no part
%% (background).
We start with a softmax loss that outputs probabilities for each body
part and background. The downside is inability
%% to model situations where for a particular location more than one part
%% can probabilities greater than 0.5, such as in case of
%% foreshortening.
to assign probabilities above $0.5$ to several close-by body parts. We
thus re-formulate the detection as multi-label classification, where
at each location a separate set of probability distributions is
estimated for each part.
%% \cite{tompson14nips}
%% uses MSE objective function, however our experiments showed
%% significant performance degradation (see supplementary materials).
We use sigmoid activation function on the output neurons and cross
entropy loss. We found this loss to perform better than softmax and
converge much faster compared to MSE~\cite{tompson14nips}. Target
training scoremap for each joint is constructed by assigning a
positive label 1 at each location within $15$ px to the ground truth,
and negative label 0 otherwise.

%% as follows: at each location for each
%% joint a positive label 1 is assigned if the location is within $15$ px
%% to the ground truth, and negative label 0 otherwise. Locations with
%% all 0 are the negatives.

\myparagraph{Location refinement.} In order to improve location
precision we follow \cite{girshickICCV15fastrcnn}: we add a location
refinement FC layer after the FC7 and use the relative offsets
$(\Delta x,\Delta y)$ from a scoremap location to the ground truth as
targets.

%% While scoremaps provide sufficient
%% resolution, location precision can be improved.
%% \cite{Tompson:2015:EOL} train additional net to produce fine
%% scoremaps. We follow an alternative and simpler route
%% \cite{girshickICCV15fastrcnn}: we add a location refinement FC layer
%% after the FC7 and use the relative offsets $(\Delta x,\Delta y)$ from a scoremap
%% location to the ground truth as targets.

\myparagraph{Regression to other parts.} Similar to location
refinement we add an extra term to the objective function where for
each part we regress onto all other part locations. We found this
auxiliary task to improve the performance
(c.f. Sec.~\ref{sec:experiments:unary}).
%% We envision these predictions
%% to improve the spatial model as well and leave this for the future
%% work.

%% While we currently discard
%% these extra predictions, they could be used for constructing more
%% robust pairwise terms in the future work.

\myparagraph{Training.} We follow best practices and use SGD for CNN
training. In each iteration we forward-pass a single image. After FC6 we
select all positive and random negative samples to keep the pos/neg
ratio as 25\%/75\%. We finetune VGG from Imagenet model to pose
estimation task and use training data augmentation. We train for 430k
iterations with the following learning rates (lr): 10k at lr=0.001,
180k at lr=0.002, 120k at lr=0.0002 and 120k at
lr=0.0001. Pre-training at smaller lr prevents the gradients from
diverging.
%% We also randomly scale image within range $[0.85;1.15]$ for data
%% augmentation.

%% In summary, by carefully re-purposing and tuning readily available components we achieve state of the art unary performance without employing highly specialized
%% architectures as in \cite{Tompson:2015:EOL}.

%\renewcommand{\tabcolsep}{0.15cm}
\tabcolsep 1.5pt
\begin{table}[tbp]
 \scriptsize
  \centering
  \begin{tabular}{@{} l c ccc cc cc|c@{}}
    \toprule
    Setting& Head   & Sho  & Elb & Wri & Hip & Knee & Ank & PCK & AUC\\
    \midrule
    %\cmidrule(r){1-1} \cmidrule(lr){2-9} \cmidrule(lr){10-10}
    oracle 2000& 98.8  & 98.8  & 97.4  & 96.4  & 97.4  & 98.3 & 97.7 & 97.8 & 84.0\\

    \midrule
    DPM scale 1& 48.8  & 25.1  & 14.4  & 10.2  & 13.6  & 21.8 & 27.1 & 23.0 & 13.6\\

    \midrule    
    %AlN s1 lr.001(80k) 140k IoU.5 & 82.2  & 67.0  & 49.6  & 45.4  & 53.1  & 52.9 & 48.2 & 56.9 & 35.9 \\
AlexNet scale 1 & 82.2  & 67.0  & 49.6  & 45.4  & 53.1  & 52.9 & 48.2 & 56.9 & 35.9 \\

    AlexNet scale 4 & 85.7  & 74.4  & 61.3  & 53.2  & 64.1  & 63.1 & 53.8 & 65.1 & 39.0 \\

    %\input{experiments/pck-torso-expidx26.tex}
    %\input{experiments/pck-torso-expidx34.tex}
    %\input{experiments/pck-torso-expidx98.tex}
    %AlN s4 lr.004(320k) 1M IoU.4 aug& 88.1  & 79.3  & 68.9  & 62.6  & 73.5  & 69.3 & 64.7 & 72.4 & 44.6 \\
\quad + optimal params & 88.1  & 79.3  & 68.9  & 62.6  & 73.5  & 69.3 & 64.7 & 72.4 & 44.6 \\

    \midrule
    %VGG s4 lr.003(160k) 320k IoU.4 aug& 91.0  & 84.2  & 74.6  & 67.7  & 77.4  & 77.3 & 72.8 & 77.9 & 50.0 \\
VGG scale 4 optimal params & 91.0  & 84.2  & 74.6  & 67.7  & 77.4  & 77.3 & 72.8 & 77.9 & 50.0 \\

    %\input{experiments/pck-torso-expidx179.tex}
    %\quad + finetune LSP lr0.0005 (10k) 40k & \textbf{95.4}  & \textbf{86.5}  & \textbf{77.8}  & \textbf{74.0}  & \textbf{84.5}  & \textbf{78.8} & \textbf{82.6} & \textbf{82.8} &\textbf{57.0}\\
\quad + finetune LSP & \textbf{95.4}  & \textbf{86.5}  & \textbf{77.8}  & \textbf{74.0}  & \textbf{84.5}  & \textbf{78.8} & \textbf{82.6} & \textbf{82.8} &\textbf{57.0}\\

    \bottomrule \end{tabular} 
\vspace{0.3em} 
    \caption[]{Unary only performance (PCK) of $\rcnn$ on the LSP
    (Person-Centric) dataset. $\rcnn$ is finetuned from ImageNet to
    MPII (lines 3-6), and then finetuned to LSP (line 7).}
\vspace{-1.0em} 
\label{tab:unary:rcnn}
\end{table}

\subsection{Evaluation of Part Detectors}
\label{sec:experiments:unary}
%% We now evaluate the proposed part detection models on the task of
%% single person pose estimation thereby significantly outperforming the
%% state of the art.

\myparagraph{Datasets.} We train and evaluate on three public
benchmarks: ``Leeds Sports Poses'' (LSP)~\cite{johnson10bmvc}
(person-centric (PC)),
%including $1000$ training and $1000$
%testing images of people doing sports; 
``LSP Extended'' (LSPET)
~\cite{johnson11cvpr}\footnote{To reduce labeling noise we re-annotated original
  high-resolution images and make the data available at \url{http://datasets.d2.mpi-inf.mpg.de/hr-lspet/hr-lspet.zip}}, and  %consisting of $10000$ training images;
%performing ``acrobatics'' and ``parkour''
``MPII Human Pose'' (``Single Person'')~\cite{andriluka14cvpr}.
%% consisting of $19185$ training and $7247$ testing people in every
%% day activities.
%% As in latter case each image may contain several individuals, rough
%% location and scale information is available at test time.
The MPII training set ($19185$ people) is used as default. In some
cases LSP training \textit{and} LSPET are added to MPII (marked as MPII+LSPET in the
experiments).
%% downloaded the original
%% high-resolution images using the provided Flickr links and

\myparagraph{Evaluation measures.} We use the standard 
``PCK''
%% ``Percentage of
%% Correct Keypoints (PCK)'' 
metric~\cite{sapp13cvpr,Toshev:2014:DHP,tompson14nips} and
%% : the body joint is considered to be correctly localized
%% if it's predicted location falls within a threshold distance
%% w.r.t. the ground truth location.
%% On the LSP dataset, we follow
%% \cite{sapp13cvpr,Toshev:2014:DHP,tompson14nips} and use the threshold
%% distance as $0.2$ of the torso diameter computed between the left
%% shoulder and right hip.
evaluation scripts available on the web page of~\cite{andriluka14cvpr}.
%% and thus are directly comparable to other
%% methods.
%On the MPII dataset we use PCK$_h$ measure defined by the
%evaluation protocol~\cite{andriluka14cvpr}.
%% On the MPII dataset, we use
%% $0.5$ of the head size as a threshold, as defined by the evaluation
%% protocol~\cite{andriluka14cvpr} (PCKh measure).
In addition, %to PCK at fixed threshold, 
we report ``Area under
Curve'' (AUC) computed for the entire range of PCK thresholds.

\myparagraph{$\rcnn$.} Evaluation of~$\rcnn$ on LSP is shown in
Tab.~\ref{tab:unary:rcnn}. Oracle selecting per part the closest
%% to the ground
%% truth out of 
from $2,000$ proposals achieves $97.8$\% PCK, as proposals cover majority
of the ground truth locations. Choosing a single proposal per part
using DPM score achieves $23.0$\% PCK -- not surprising given the
difficulty of the body part detection problem. Re-scoring the
proposals using $\rcnn$ with AlexNet~\cite{krizhevsky12nips}
dramatically improves the performance to $56.9$\% PCK, as CNN learns
richer image representations.
%This is
%achieved using basis scale $1$ ($\approx$ head size) of proposals.
%% and training with initial
%% learning rate (lr) of $0.001$ for $80$k iterations, after which lr was
%% reduced by $0.1$, for a total of $140$k iterations. In addition, we
%% use bounding box regression and default IoU threshold of $0.5$ for
%% pos/neg label assignment~\cite{girshickICCV15fastrcnn}.
Extending the regions by $4$x (1x $\approx$ head size) achieves 65.1\%
PCK, as it incorporates more context including the information about
symmetric parts and allows to implicitly encode higher-order part
relations. % into the part
%%detector.
%% No improvements observed for larger scales.
%% Augmenting the training set by horizontally flipping and jittering
%% training samples, increasing the number of positives by reducing
%% positive IoU threshold to $0.4$.  increasing the lr, lr reduction step
%% and total number of iterations altogether improves the performance to
%% $72.4$\% PCK. We refer to the supplementary material for detailed
%% analysis.
Using data augmentation and slightly tuning training parameters
improves the performance to $72.4$\% PCK. We refer to the
Appendix~\ref{seq:supplemental:lsp} for detailed analysis.
%% All results above are achieved by finetuning the AlexNet
%% architecture from the ImageNet model to the MPII training
%% set. Further finetuning the MPII-finetuned model to the LSP
%% training set increases the performance to $77.9$\% PCK. This is due
%% to the fact that image statistics of more diverse MPII dataset
%% differ from the LSP biased towards a few sport activities. Using
%% the deeper VGG architecture improves over more shallow AlexNet
%% achieving remarkable 82.8\% PCK. Strong increase in AUC (57.0
%% vs. 50\%) characterizes the improvement for smaller PCK evaluation
%% thresholds. Switching off bounding box regression results into
%% performance drop (81.3\% PCK, 53.2\% AUC) thus showing the
%% importance of the bounding box regression for better part
%% localization. Overall $\rcnn$ obtains very good results on LSP
%% dataset by far outperforming the state of the art
%% (c.f. Tab.~\ref{tab:multicut:lsp}, rows $7-9$). Unary-only
%% evaluation on MPII Single Person shows competitive performance
%% (Tab.~\ref{tab:multicut:mpii}, row $1$).
Deeper VGG architecture improves over smaller AlexNet reaching
$77.9$\% PCK. All results so far are achieved by finetuning the
ImageNet models on MPII. Further finetuning to LSP leads to remarkable
82.8\% PCK: CNN learns LSP-specific image representations.
%% image statistics of more diverse MPII dataset differ from
%% LSP biased towards a few sport activities. 
Strong increase in AUC (57.0 vs. 50\%) is due to improvements for
smaller PCK thresholds. Using no bounding box regression leads to
performance drop (81.3\% PCK, 53.2\% AUC): location refinement is
crucial for better localization. Overall $\rcnn$ obtains very good
results on LSP by far outperforming the state of the art
(c.f. Tab.~\ref{tab:multicut:lsp}, rows $7-9$). Evaluation on MPII
shows competitive performance (Tab.~\ref{tab:multicut:mpii}, row $1$).

\tabcolsep 1.5pt
\begin{table}[tbp]
 \scriptsize
  \centering
  \begin{tabular}{@{} l c ccc ccc c|c@{}}
    \toprule
    Setting& Head   & Sho  & Elb & Wri & Hip & Knee & Ank & PCK & AUC\\
    \midrule
    %Fast R-CNN MPII & 91.0  & 84.3  & 74.7  & 67.7  & 77.1  & 77.3 & 72.7 & 77.8 & 49.8 \\
    %\quad+ finetune LSP & 95.4  & 86.5  & 77.8  & 74.0  & 84.5  & 82.6 & 78.8 & 82.8 & 56.7\\
    %\midrule
    MPII softmax & 91.5  & 85.3  & 78.0  & 72.4  & 81.7  & 80.7 & 75.7 & 80.8 & 51.9 \\

    %Dense MPII sigm reg &91.1 & 86.1 & 79.6 & 74.8 & 84.7 & 81.8 & 79.0 & 82.4 & 57.1 \\
    \quad+ LSPET & 94.6  & 86.8  & 79.9  & 75.4  & 83.5  & 82.8 & 77.9 & 83.0 & 54.7 \\

    \quad\quad+ sigmoid & 93.5  & 87.2  & 81.0  & 77.0  & 85.5  & 83.3 & 79.3 & 83.8 & 55.6 \\

    \quad\quad\quad\quad + location refinement& 95.0  & 88.4  & 81.5  & 76.4  & 88.0  & 83.3 & 80.8 & 84.8 & 61.5\\

    \quad\quad\quad\quad\quad + auxiliary task & 95.1  & 89.6  & 82.8  & 78.9  & 89.0  & 85.9 & 81.2 & 86.1 & 61.6 \\

    %\midrule 
    \quad\quad\quad\quad\quad\quad + finetune LSP & \textbf{97.2}  & \textbf{90.8}  & \textbf{83.0}  & \textbf{79.3}  & \textbf{90.6}  & \textbf{85.6} & \textbf{83.1} & \textbf{87.1} & \textbf{63.6} \\

    \bottomrule \end{tabular} 
\vspace{0.3em} 
    \caption[]{Unary only performance (PCK) of $\dense$ VGG on LSP
    (PC) dataset. $\dense$ is finetuned from ImageNet to MPII (line
    1), to MPII+LSPET (lines 2-5), and finally to LSP (line 6).}

\vspace{-1.0em} \label{tab:unary:dense}
\end{table}

\myparagraph{$\dense$.} The results are in Tab.~\ref{tab:unary:dense}.
Training with VGG on MPII with softmax loss achieves $80.8$\% PCK
thereby outperforming $\rcnn$ (c.f. Tab.~\ref{tab:unary:rcnn}, row
6). This shows the advantages of fully convolutional training and
evaluation.
%% of all, we obtain state of the art performance on this
%% dataset by using the simplest version of fully convolutional network
%% trained just on MPII dataset.
Expectedly, training on larger MPII+LSPET dataset improves the results
($83.0$ vs. $80.8$\% PCK). Using cross-entropy loss with sigmoid
activations improves the results to $83.8$\% PCK, as it better models
the appearance of close-by parts. Location refinement improves
localization accuracy ($84.8$\% PCK), which becomes more clear when
analyzing AUC ($61.5$ vs. $55.6$\%). Interestingly, regressing to other
parts further improves PCK to $86.1$\% showing a value of training
with the auxiliary task. Finally, finetuning to LSP achieves the best
result of $87.1$\% PCK, which is significantly higher than the best
published results (c.f. Tab.~\ref{tab:multicut:lsp}, rows
$7-9$). Unary-only evaluation on MPII reveals slightly higher AUC
results compared to the state of the art
(Tab.~\ref{tab:multicut:mpii}, row $3-4$).

\subsection{Using Detections in DeepCut Models}
\label{sec:unary-prob}
The SPLP problem is NP-hard, to solve instances of it efficiently we
select a subset of representative detections from the entire set
produced by a model.
%% In particular, we first perform
%% non-maximum suppression and then for each detection $d$, we take the
%% maximum of all the $p_{dc}$ as its confidence. Detection subset then
%% consists of the top $n$ detections with the highest confidence.
In our experiments we use $|D|= 100$ as default detection set size.
% $|D| = 125$ and $|D|= 150$.
%% In addition, the $x_{dc}$ variables are constrained to $0$ during the
%% optimization, if $p_{dc} \leq 0.2$.
In case of the $\rcnn$ we directly use the softmax output as unary
probabilities: $f_{p_{dc}} = (p_{d1},\ldots,p_{dc})$, where $p_{dc}$
is the probability of the detection $d$ being the part class $c$. For
$\dense$ detection model we use the sigmoid detection unary scores.

\section{DeepCut Results}
\label{sec:experiments:multicut}
The aim of this paper is to tackle the multi person case. To that end,
we evaluate the proposed $\deepcut$ models on four diverse
benchmarks. We confirm that both single person ($\singb$) and
multi person ($\multb$) variants (Sec.~\ref{section:problem}) are
effective on standard $\singb$ pose estimation
datasets~\cite{johnson10bmvc,andriluka14cvpr}.
%% described in
%%Sec.~\ref{section:unary}.
%Both types of models
%perform equally well, slightly improving over unaries only and
%significantly outperforming state of the art.
Then, we demonstrate superior performance of $\deepcut~\multb$ on the
multi person pose estimation task.

\begin{figure}
  \centering
  \begin{tabular}{c c}  
  \includegraphics[width=0.48\linewidth]{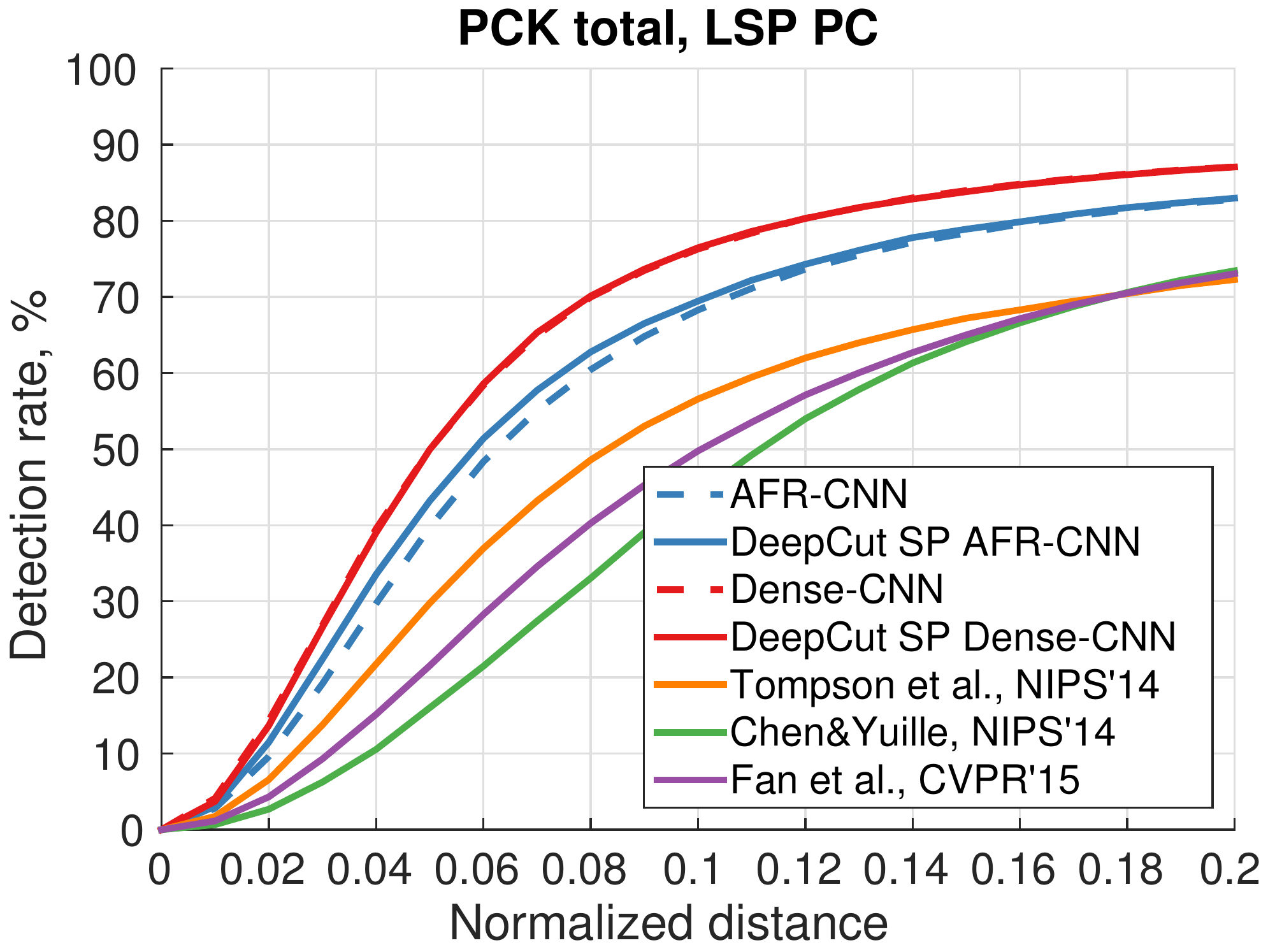}&
  \includegraphics[width=0.48\linewidth]{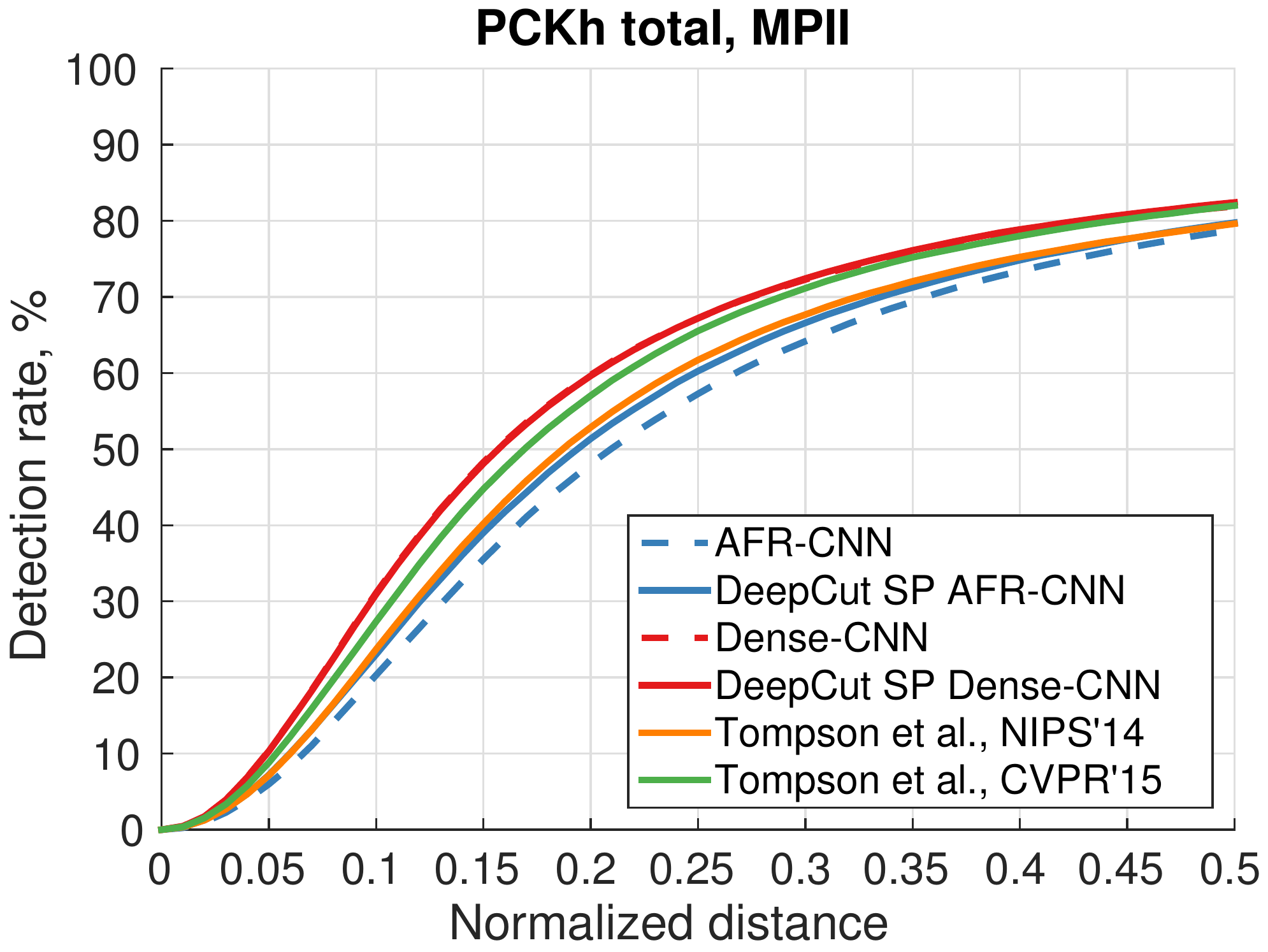}\\
  (a) LSP (PC)& (b) MPII Single Person \\
  \end{tabular}
  \vspace{-0.1em}
%  \caption{Pose estimation performance on (a) LSP (PC) dataset (PCK) and (b) MPII Single Person dataset (PCKh).} 
  \caption{Pose estimation results over all PCK thresholds.} 
    \vspace{-1.2em}
  \label{fig:pck-curves}
\end{figure}

\subsection{Single Person Pose Estimation}
\label{sec:multicut:single}
We now evaluate single person (\textit{SP}) and more general
multi person (\textit{MP}) $\deepcut$ models on LSP and MPII $\singb$
benchmarks described in Sec.~\ref{section:unary}. Since this
evaluation setting implicitly relies on the knowledge that all parts
are present in the image we always output the full number of parts.
% BERNT: I would not include the following sentence as such
% if you want to totally clear we should rephrase this as to make this
% more compelling
%To that end we use
%hypothesis with the highest unary score when the body part predicted
%as occluded.
%\renewcommand{\tabcolsep}{0.15cm}
\tabcolsep 1.5pt
\begin{table}[tbp]
 \scriptsize
  \centering
  \begin{tabular}{@{} l c ccc ccc c|c@{}}
    \toprule
    Setting& Head   & Sho  & Elb & Wri & Hip & Knee & Ank & PCK & AUC\\       
    \midrule
    $\rcnn$ (unary) & 95.4  & 86.5  & 77.8  & 74.0  & 84.5  & 82.6 & 78.8 & 82.8 & 57.0\\

    \quad+ $\deepcut$ \singb& 95.4  & 86.7  & 78.3  & 74.0  & 84.3  & 82.9 & 79.2& 83.0 & 58.4 \\

    \quad\quad+ appearance pairwise & 95.4  & 87.2  & 78.6  & 73.7  & 84.7  & 82.8 & 78.8 & 83.0 & 58.5 \\

    \quad+ $\deepcut$ \multb & 95.2  & 86.7  & 78.2  & 73.5  & 84.6  & 82.8 & 79.0 & 82.9 & 58.0\\

    \midrule
    $\dense$ (unary)                   & \textbf{97.2}  & 90.8  & 83.0  & \textbf{79.3}  & 90.6  & 85.6 & \textbf{83.1} & \textbf{87.1} & \textbf{63.6}\\
    %\quad + $\deepcuts~\singb$ & 97.0  & \textbf{91.1}  & \textbf{83.9}  & 77.8  & 90.7  & \textbf{86.7} & 81.0 & 86.9 & 63.4\\
    \quad + $\deepcut~\singb$ & 97.0  & 91.0  & \textbf{83.8}  & 78.1  & 91.0  & 86.7 & 82.0 & \textbf{87.1} & 63.5\\
    \quad + $\deepcut~\multb$& 96.2  & \textbf{91.2}  & 83.3  & 77.6  & \textbf{91.3}  & \textbf{87.0} & 80.4 & 86.7 & 62.6 \\

    \midrule
    %Tompson et al.~\cite{tompson14nips} (unary)  & 88.3 & 77.3 & 66.1 & 62.7 & 66.6 & 68.3 & 65.0 & 70.6 \\
    Tompson et al.~\cite{tompson14nips}& 90.6  & 79.2  & 67.9  & 63.4  & 69.5  & 71.0 & 64.2 & 72.3 & 47.3\\
    Chen\&Yuille~\cite{chen14nips}& 91.8  & 78.2  & 71.8  & 65.5  & 73.3  & 70.2 & 63.4& 73.4 & 40.1\\
    Fan et al.~\cite{fan15cvpr}$^*$
    & 92.4 & 75.2& 65.3& 64.0& 75.7& 68.3& 70.4& 73.0 & 43.2 \\
    \bottomrule
  \end{tabular}
  
  $^*$ re-evaluated using the standard protocol, for details see project page of \cite{fan15cvpr}
  \caption[]{Pose estimation results (PCK) on LSP (PC) dataset.}
    \vspace{-1.5em}
  \label{tab:multicut:lsp}
\end{table}

\myparagraph{Results on LSP.} We report per-part PCK results
(Tab.~\ref{tab:multicut:lsp}) and results for a variable distance
threshold (Fig.~\ref{fig:pck-curves} (a)). $\deepcut~\singb~\rcnn$
model using $100$ detections improves over unary only ($83.0$
vs. $82.8$\%~PCK, $58.4$ vs. $57$\% AUC), as pairwise connections
filter out some of the high-scoring detections on the background. The
improvement is clear in Fig.~\ref{fig:pck-curves}~(a) for smaller
thresholds. Using part appearance scores in addition to geometrical
features in $c\neq c'$ pairwise terms only slightly improves AUC, as
the appearance of neighboring parts is mostly captured by a relatively
large region centered at each part. 
%% As geometrical only pairwise lead
%% to faster convergence, we use them in the rest of the
%% experiments. 
The performance of $\deepcut~\multb~\rcnn$ matches the
$\singb$ and improves over $\rcnn$ alone: $\deepcut~\multb$ correctly
handles the $\singb$ case. Performance of $\deepcut~\singb~\dense$ is
almost identical to unary only, unlike the results for
$\rcnn$. $\dense$ performance is noticeably higher compared to
$\rcnn$, and ``easy'' cases that could have been corrected by a
spatial model are resolved by stronger part detectors alone.

\myparagraph{Comparison to the state of the art (LSP).}
Tab.~\ref{tab:multicut:lsp} compares results of $\deepcut$ models to
other deep learning methods specifically designed for single person
pose estimation. All $\deepcuts$ significantly outperform the state of
the art, with $\deepcut~\singb~\dense$ model improving by $13.7$\% PCK
over the best known result~\cite{chen14nips}. The improvement is even
more dramatic for lower thresholds (Fig.~\ref{fig:pck-curves}~(a)):
for PCK @ $0.1$ the best model improves by $19.9$\% over Tompson et
al.~\cite{tompson14nips}, by $26.7$\% over Fan et al.~\cite{fan15cvpr},
and by $32.4$\%~PCK over Chen\&Yuille~\cite{chen14nips}. The latter is
interesting, as~\cite{chen14nips} use a stronger spatial model that
predicts the pairwise conditioned on the CNN features, whereas
$\deepcuts$ use geometric-only pairwise connectivity. Including body
part orientation information into $\deepcuts$ should further improve
the results.

\myparagraph{Results on MPII Single Person.} Results are shown in
Tab.~\ref{tab:multicut:mpii} and Fig.~\ref{fig:pck-curves}
(b). $\deepcut~\singb~\rcnn$ noticeably improves over $\rcnn$ alone
($79.8$ vs. $78.8$\%~PCK, $51.1$ vs. $49.0$\% AUC). The improvement is
stronger for smaller thresholds (c.f. Fig.~\ref{fig:pck-curves}), as
spatial model improves part localization. $\dense$ alone trained on
MPII outperforms $\rcnn$ ($81.6$ vs. $78.8$\% PCK), which shows the
advantages of dense training and evaluation. As expected, $\dense$
performs slightly better when trained on the larger
MPII+LSPET. Finally, $\deepcut~\dense~\singb$ is slightly better than
$\dense$ alone leading to the best result on MPII dataset ($82.4$\%
PCK).

\myparagraph{Comparison to the state of the art (MPII).} We compare
the performance of $\deepcut$ models to the best deep learning
approaches from the
literature~\cite{tompson14nips,Tompson:2015:EOL}\footnote{\cite{tompson14nips}
  was re-trained and evaluated on MPII dataset by the
  authors.}.
%% \cite{tompson14nips} is a fully-convolutional deep architecture
%% relying on multi-resolution filter banks that have various receptive
%% field sizes. In addition, this model jointly trains part detectors and
%% spatial model in the same deep architecture. This model was kindly
%% re-trained by the authors on the MPII dataset for fair
%% comparison. \cite{Tompson:2015:EOL} is an extension of
%% \cite{tompson14nips} and it specifically trains a separate deep
%% full-convolutional model for location refinement. The results are
%% shown in Tab.~\ref{tab:multicut:mpii}.
$\deepcut~\singb~\dense$ outperforms
both~\cite{tompson14nips,Tompson:2015:EOL} ($82.4$ vs $79.6$ and $82.0$\%
PCK, respectively). Similar to them $\deepcuts$ rely on dense training
and evaluation of part detectors, but unlike them use single size
receptive field and do not include multi-resolution context
information. Also, appearance and spatial components of $\deepcuts$
are trained piece-wise, unlike~\cite{tompson14nips}. We observe that
performance differences are higher for smaller thresholds
(c.f. Fig.~\ref{fig:pck-curves}~(b)). This is remarkable, as a much
simpler strategy for location refinement is used compared
to~\cite{Tompson:2015:EOL}. Using multi-resolution filters and joint
training should improve the performance.
%% We believe that training a separate deep
%% convolutional model for location refinement should further improve
%% results.

%\renewcommand{\tabcolsep}{0.15cm}
\tabcolsep 1.5pt
\begin{table}[tbp]
 \scriptsize
  \centering
  \begin{tabular}{@{} l c ccc ccc c|c@{}}
    \toprule
    Setting& Head   & Sho  & Elb & Wri & Hip & Knee & Ank & PCK$_h$ & AUC\\
    \midrule
    %Fast R-CNN MPII& 91.5  & 89.8  & 80.6  & 74.2  & 76.7  & 69.4 & 63.0 & 81.6 & 78.7 & 48.7 \\
%Fast R-CNN MPII& 91.5  & 89.8  & 80.6  & 74.2  & 76.7  & 69.4 & 63.0 & 78.7 & 48.7 \\
$\rcnn$ (unary) & 91.5  & 89.7  & 80.5  & 74.4  & 76.9  & 69.6 & 63.1 & 78.8 & 49.0\\

    %Fast R-CNN MPII~\singb& 92.3  & 90.8  & 81.6  & 74.9  & 79.0  & 70.2 & 63.0 & 79.7 & 50.8 \\
\quad+ $\deepcut$ \singb& 92.3  & 90.6  & 81.7  & 74.9  & 79.2  & 70.4 & 63.0 & 79.8 &51.1\\

    \midrule
    $\dense$ (unary)              & 93.5  & 88.6  & 82.2  & 77.1  & 81.7  & 74.4 & \textbf{68.9} & 81.6 & 56.0\\
    \quad+LSPET        & 94.0  & 89.4  & 82.3  & 77.5  & 82.0          & 74.4          & 68.7          & 81.9          & \textbf{56.5} \\
    \quad\quad+$\deepcut$ \singb& 94.1  & 90.2  & 83.4  & 77.3  & \textbf{82.6} & \textbf{75.7} & 68.6          & \textbf{82.4} & \textbf{56.5} \\
    \midrule
    Tompson et al.~\cite{tompson14nips}  &95.8 & 90.3 & 80.5 & 74.3 & 77.6 & 69.7 & 62.8 & 79.6 & 51.8\\
    Tompson et al.~\cite{Tompson:2015:EOL}  &\textbf{96.1} & \textbf{91.9} & \textbf{83.9} &\textbf{77.8} & 80.9 & 72.3 & 64.8 & 82.0 & 54.9\\
    \bottomrule
  \end{tabular}
  \vspace{0.1em}     
  \caption[]{Pose estimation results (PCK$_h$) on MPII Single Person.}
    \vspace{-1.5em}
  \label{tab:multicut:mpii}
\end{table}

\subsection{Multi Person Pose Estimation}
We now evaluate $\deepcut~\multb$ models on the challenging task of
$\multb$ pose estimation with an unknown number of people per image
and visible body parts per person.
%% In this
%% evaluation we use the variants $\mult$ and $\multu$ of our model since
%% they demonstrated performance comparable to more costly variants in
%% previous experiments.  \todo{I don't think "mlt" = MP60 and "mltu" =
%%   MP60,UB are mentioned before?}

\myparagraph{Datasets.} For evaluation we use two public $\multb$
benchmarks: ``We Are Family'' (WAF)~\cite{eichner10eccv} with $350$
training and $175$ testing group shots of people;
%% staying close to each other and often occluding each other, the
%% images contain six people on average;
``MPII Human Pose'' (``Multi-Person'') ~\cite{andriluka14cvpr}
consisting of $3844$ training
%% images with $9687$ people
and $1758$ testing groups
%%  with $4484$ individuals involved into hundreds of every day
%% activities.
of multiple interacting individuals in highly articulated poses with
variable number of parts. On MPII, we use a subset of $288$ testing
images for evaluation.
%% The number, location and scale of
%% individuals is unknown.
%% On average there are three people per image.
%% Sample images are shown in Fig.~\ref{fig:teaserfig}.
We first pre-finetune both $\rcnn$ and $\dense$ from ImageNet to MPII
and MPII+LSPET, respectively, and further finetune each model to WAF
and MPII Multi-Person. For WAF, we re-train the spatial model on WAF
training set.
%% : it leads to better results, as poses in
%% WAF dataset are more constrained compared to MPII.

\myparagraph{WAF evaluation measure.} Approaches are evaluated using
the official toolkit~\cite{eichner10eccv}, thus results are directly
comparable to prior work. The toolkit implements occlusion-aware
``Percentage of Correct Parts ($m$PCP)'' metric. In addition, we
report ``Accuracy of Occlusion Prediction
(AOP)''~\cite{Chen:2015:POC}.

\myparagraph{MPII Multi-Person evaluation measure.} PCK metric is
suitable for $\singb$ pose estimation with known number of parts and
does not penalize for false positives that are not a part of the
ground truth.
% due to occlusion or truncation.
Thus, for $\multb$ pose estimation we use ``Mean Average Precision
(mAP)'' measure, similar to \cite{Sun:2011:APM,yang12pami}. In
contrast to~\cite{Sun:2011:APM,yang12pami} evaluating the detection of
\emph{any} part instance in the image disrespecting inconsistent pose
predictions, we evaluate consistent part configurations. First,
multiple body pose predictions are generated and then assigned to the
ground truth (GT) based on the highest
PCK$_h$~\cite{andriluka14cvpr}. Only single pose can be assigned to
GT. Unassigned predictions are counted as false positives. Finally, AP
for each body part is computed and mAP is reported.

\myparagraph{Baselines.} To assess the performance of $\rcnn$ and
$\dense$ we follow a traditional route from the literature based on
two stage approach: first a set of regions of interest (\emph{ROI}) is
generated and then the $\singb$ pose estimation is performed in the
\emph{ROIs}. This corresponds to unary only performance. \emph{ROI}
are either based on a ground truth ($\gtroi$) or on the people
detector output ($\detroi$).

\tabcolsep 1.5pt
\begin{table}[tbp]
 \scriptsize
  \centering
  \begin{tabular}{@{} l c cc cc |c@{}}
    \toprule
    Setting& Head   & U Arms  & L Arms & Torso & $m$PCP  & AOP \\
    \midrule
    \rcnn~\detroi & 69.8 & 46.0 & 36.7 & 83.7 & 53.1 & 73.9 \\

    \deepcut~\multb~\rcnn & 99.0 & 79.5 & 74.3 & 87.1 & 82.2 & 85.6 \\

    \midrule
    %\input{experiments_dense/pcp-expidx675.tex}
    %\input{experiments_dense/pcp-expidx685.tex}
    %Dense \multb~\etad{11/2} \\
    \dense~\detroi & 76.0 & 46.0 & 40.2 & 83.7 & 55.3 & 73.8\\

    %                               99.3 & 81.6 & 81.4 & 79.2 & 79.7 & 87.1 & 84.7 & 86.5
%Dense FaceDet \textit{MP125} & 96.7 & 81.8 & 81.6 & 78.9 & 79.7 & 83.5 & 83.7 & 85.9 \\
%Dense FaceDet \textit{MP125} & 96.7 & 81.7 & 79.3 & 83.5 & 83.7 & 85.9 \\
\deepcut~\multb~\dense & \textbf{99.3} & \textbf{81.5} & \textbf{79.5} & 87.1 & \textbf{84.7} & \textbf{86.5}\\

    %Dense \multbc~+ torso \etad{11/2} \\
    \midrule
    Ghiasi et. al.~\cite{ghiasi14cvpr}  & -      & -      & -      & -         & 63.6  &  74.0 \\
    Eichner\&Ferrari~\cite{eichner10eccv}   & 97.6   & 68.2  &   48.1     &86.1    &69.4   & 80.0 \\
    Chen\&Yuille~\cite{Chen:2015:POC}  & 98.5 &   77.2 &  71.3 &   \textbf{88.5} &   80.7 &   84.9 \\
    \bottomrule
  \end{tabular}
    \vspace{0.1em}
  \caption[]{Pose estimation results ($m$PCP) on WAF dataset.}
    \vspace{-1.0em}
  \label{tab:multicut:waf}
\end{table}

\myparagraph{Results on WAF.} Results are shown in
Tab.~\ref{tab:multicut:waf}. $\detroi$ is obtained by extending
provided upper body detection boxes.
%% by $2h$ towards the bottom of the
%% image, and then by $0.1\times h$ in all directions, where $h$ is the
%% height of the original bounding box.
$\rcnn~\detroi$ achieves $57.6$\% $m$PCP and $73.9$\%
AOP. $\deepcut~\multb~\rcnn$ significantly improves over
\rcnn~\detroi~achieving $82.2$\% $m$PCP. This improvement is stronger
compared to LSP and MPII due to several reasons. First, $m$PCP
requires consistent prediction of body sticks as opposite to body
joints, and including spatial model enforces consistency. Second,
$m$PCP metric is occlusion-aware. $\deepcuts$ can deactivate
detections for the occluded parts thus effectively reasoning about
occlusion. This is supported by strong increase in AOP ($85.6$
vs. $73.9$\%). Results by $\deepcut~\multb~\dense$ follow the same
tendency achieving the best performance of $84.7$\% $m$PCP and $86.5$\%
AOP. Both increase in $m$PCP and AOP show the advantages of
$\deepcuts$ over traditional \detroi~approaches.

Tab.~\ref{tab:multicut:waf} shows that 
%comparison to prior
%work. 
$\deepcuts$ outperform all prior methods.
%other
%methods. 
%$\deepcuts~\multb~\dense$ outperforms 
Deep learning method~\cite{Chen:2015:POC} is outperformed both for
$m$PCP (84.7 vs. 80.7\%) and AOP (86.5 vs. 84.9\%) measures. This is
remarkable, as $\deepcuts$ reason about part interactions across
several people, whereas~\cite{Chen:2015:POC} primarily focuses on the
single-person case and handles multi-person scenes akin to
\cite{yang12pami}. In contrast to~\cite{Chen:2015:POC}, $\deepcuts$
are not limited by the number of possible occlusion patterns and cover
person-person occlusions and other types as truncation and occlusion
by objects in one formulation. $\deepcuts$ significantly
outperform~\cite{eichner10eccv} while being more general:
%% who achieve good
%% performance considering much weaker part detectors based on
%% handcrafted features.
unlike~\cite{eichner10eccv} $\deepcuts$ do not require person detector
and not limited by a number of occlusion states among people.

\begin{figure*}
  \centering
  \begin{tabular}{c c c c c c}
  \begin{sideways}\bf \small \quad\quad $\detroi$\end{sideways}&
  \includegraphics[height=0.150\linewidth]{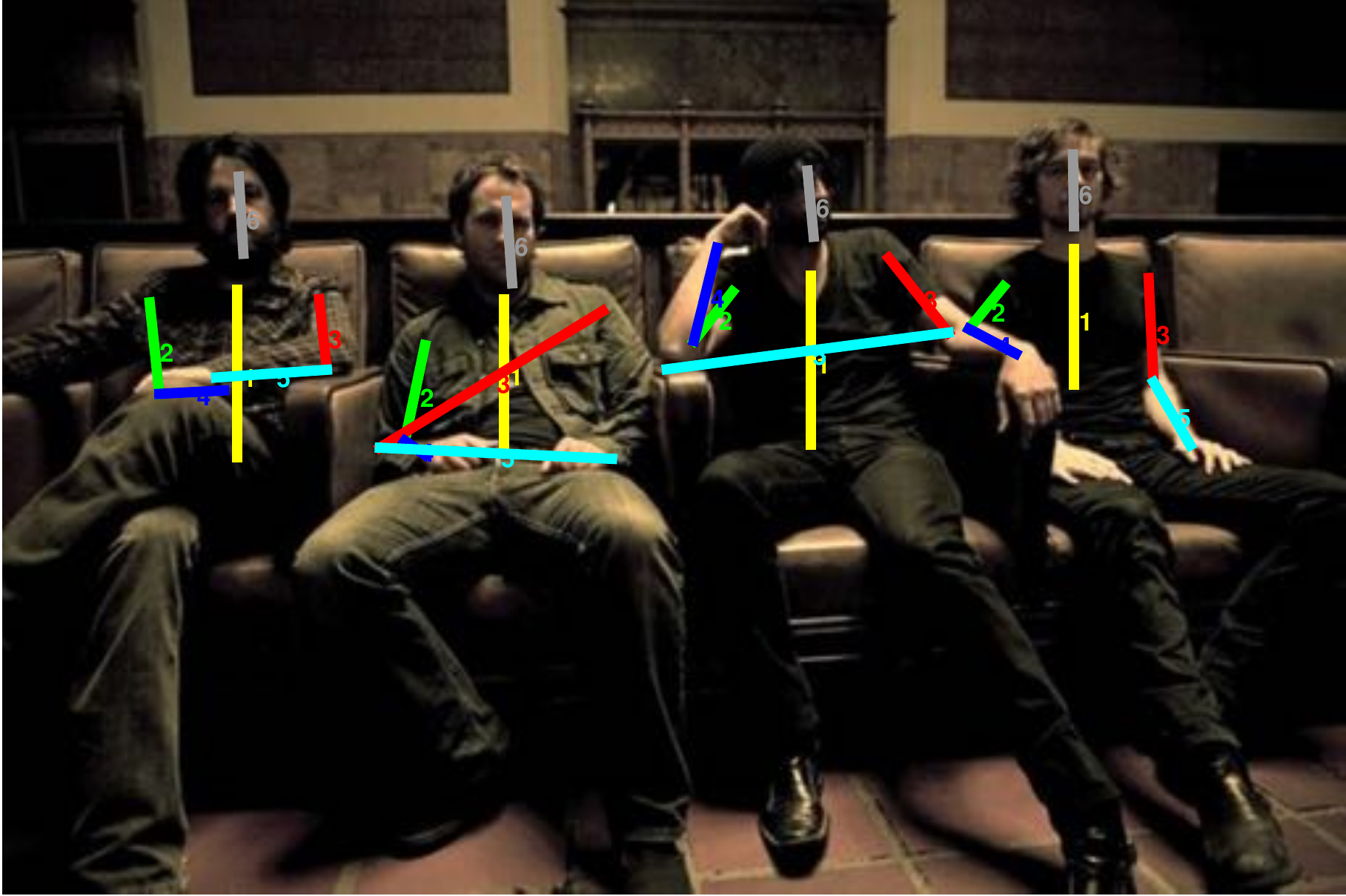}&
  \includegraphics[height=0.150\linewidth]{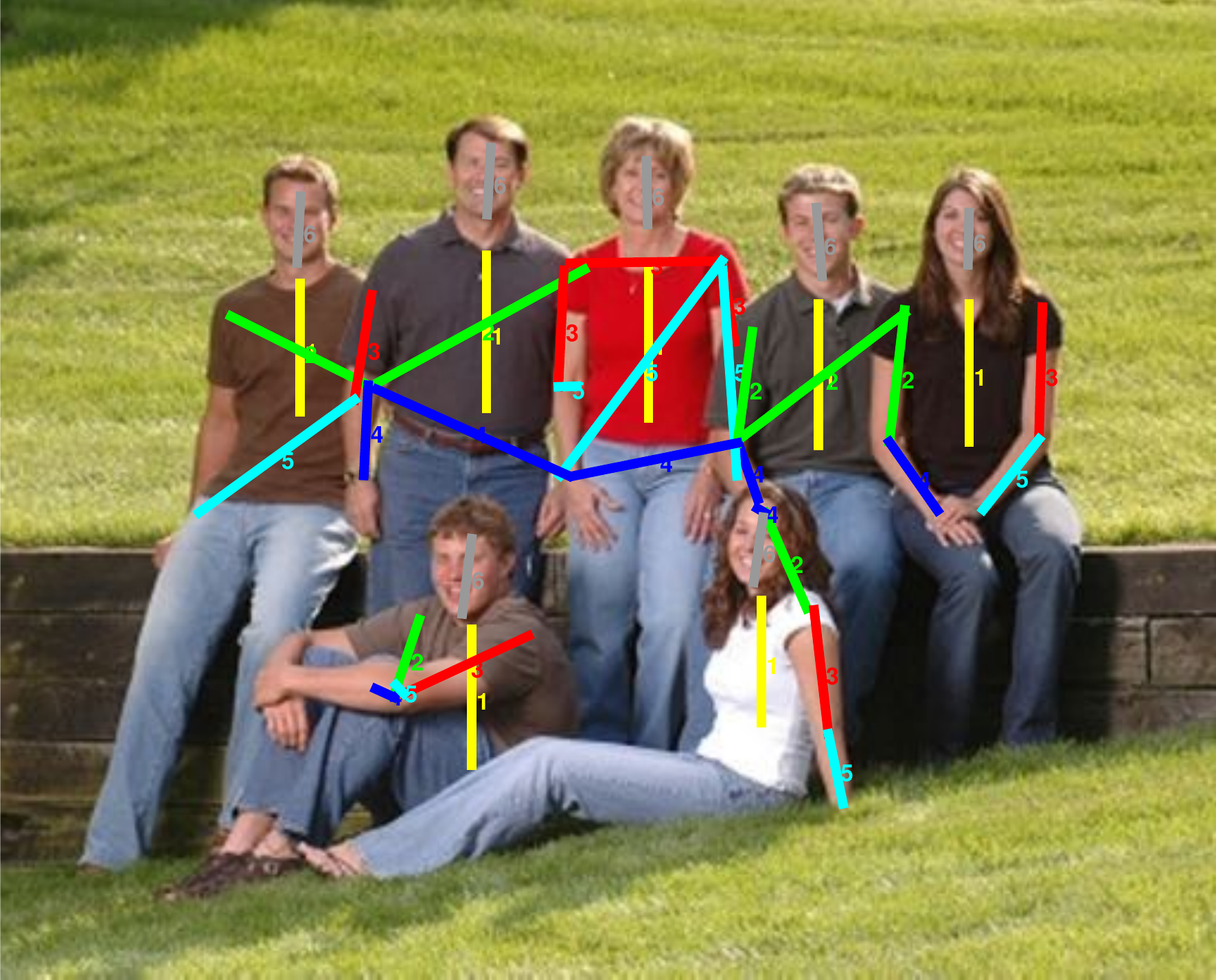}&
  \includegraphics[height=0.150\linewidth]{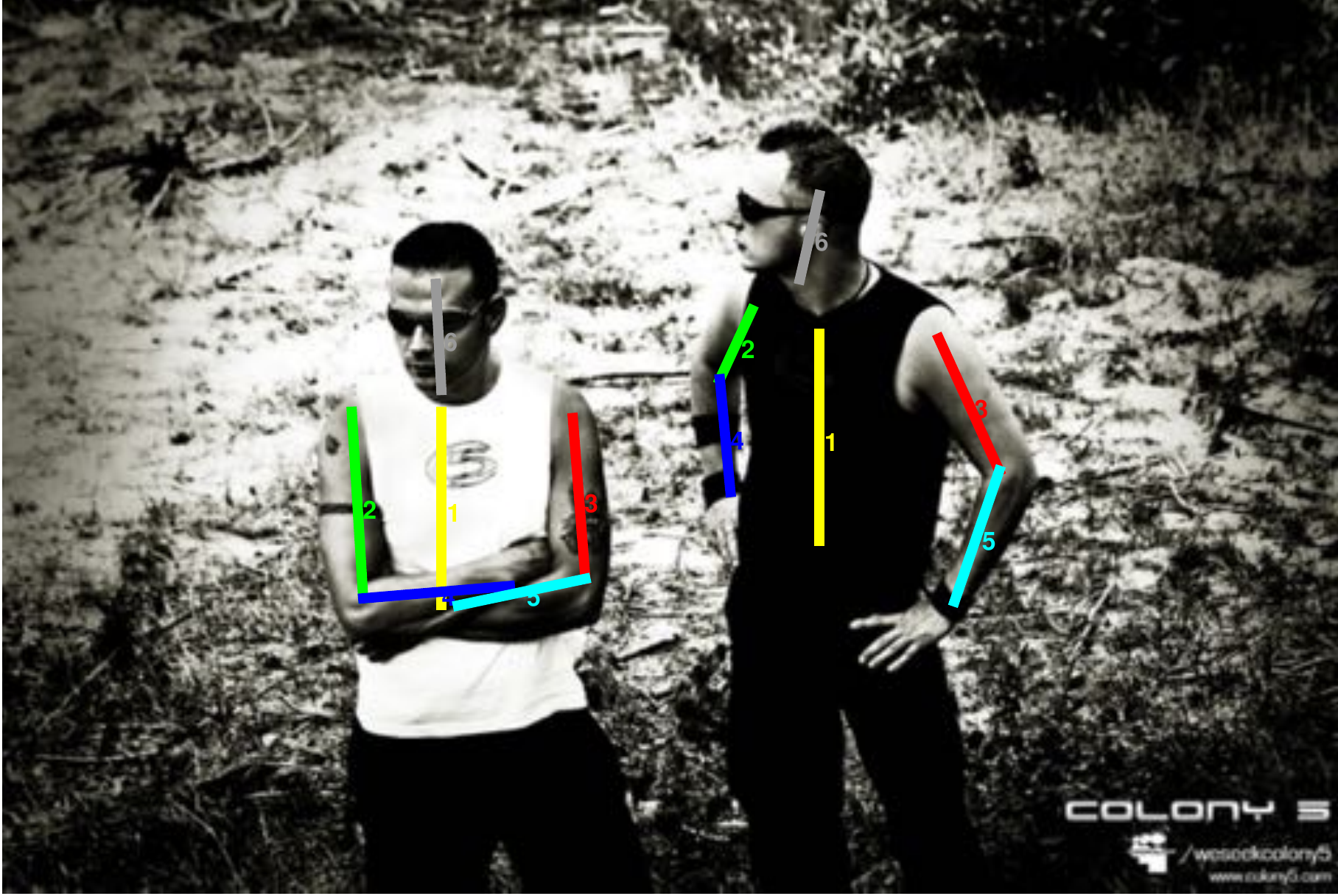}&
  \includegraphics[height=0.150\linewidth]{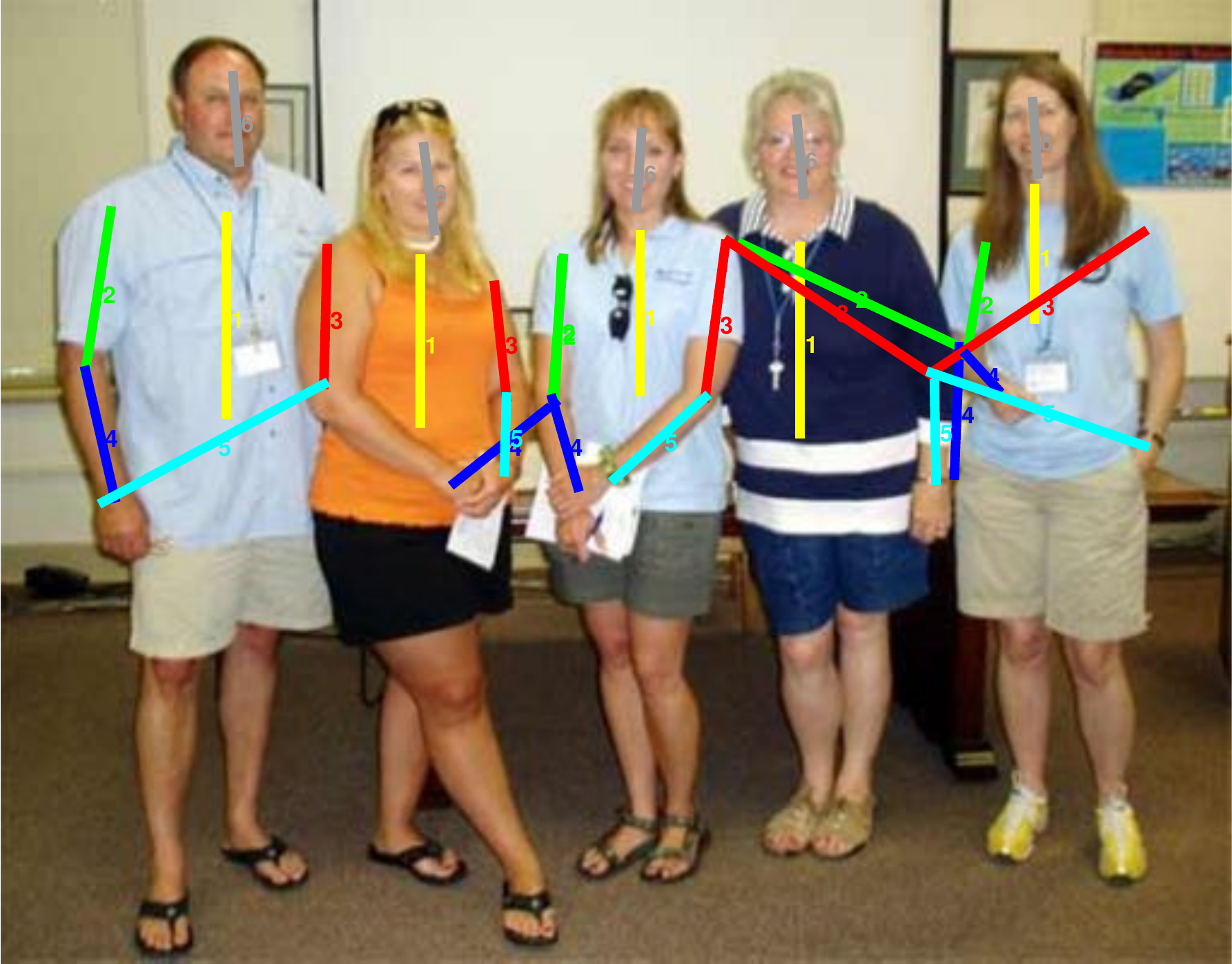}&
  \includegraphics[height=0.150\linewidth]{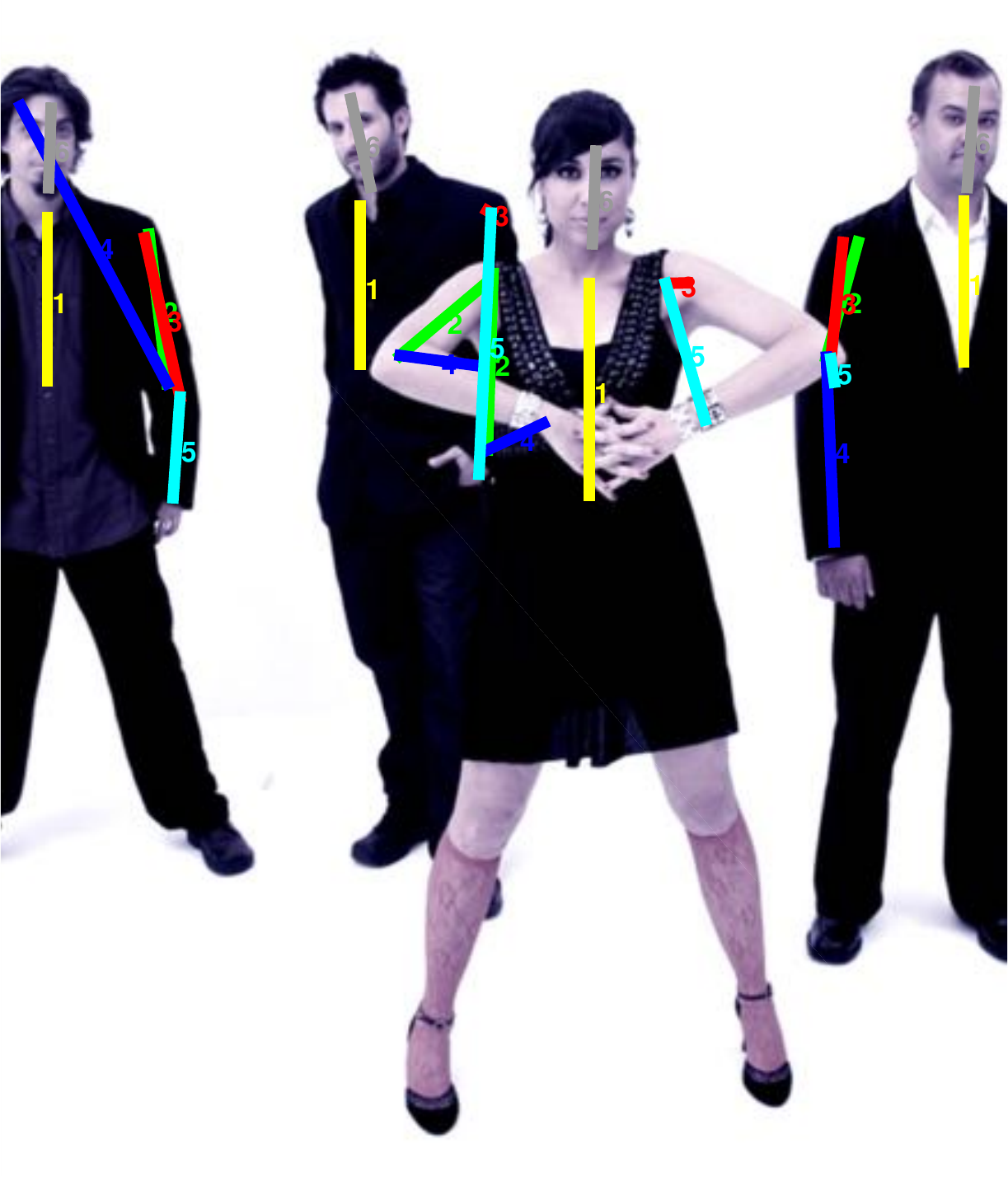}\\
  \begin{sideways}\bf \small\quad $\deepcut~\multb$\end{sideways}&
  \includegraphics[height=0.150\linewidth]{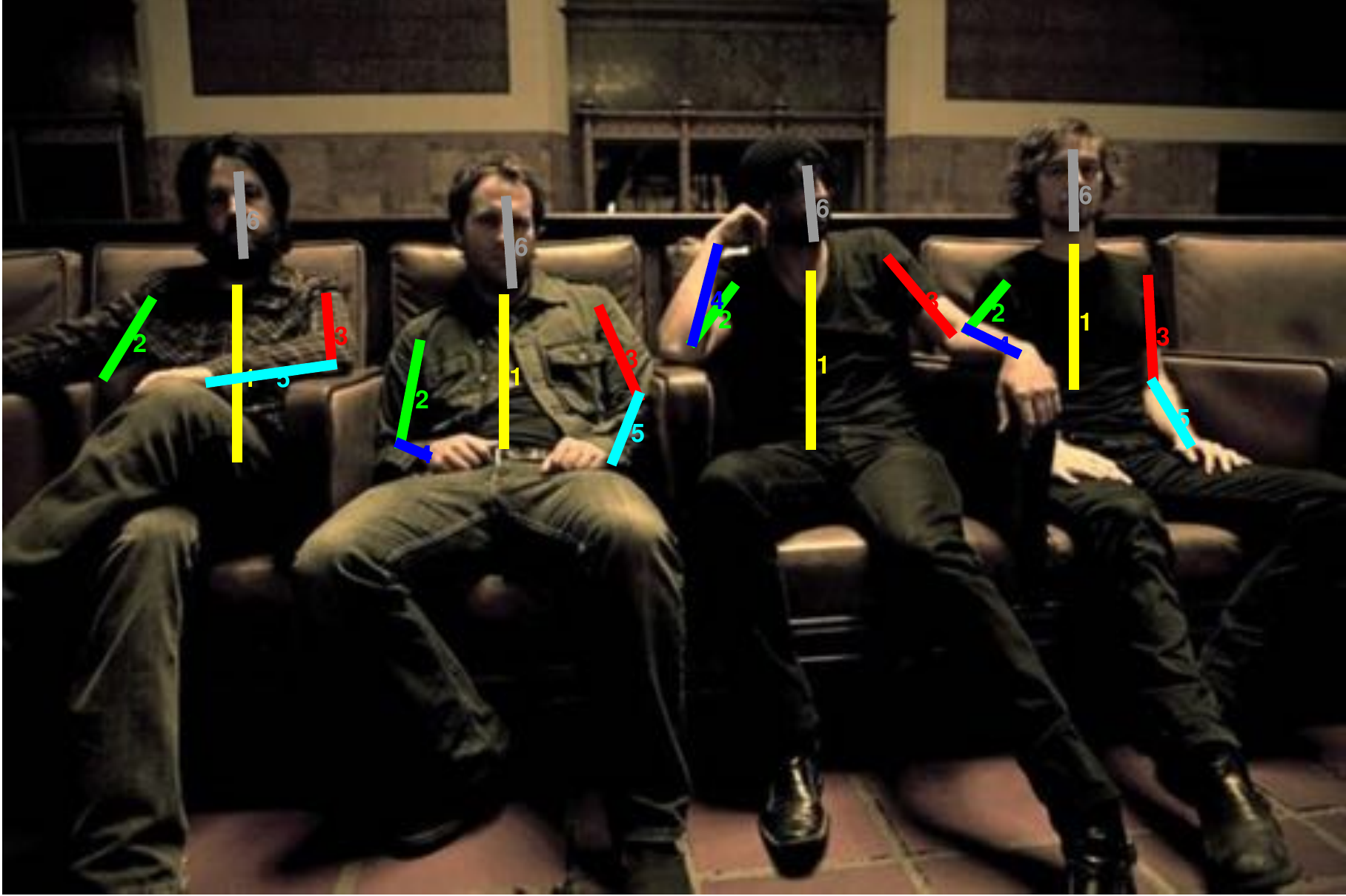}&
  \includegraphics[height=0.150\linewidth]{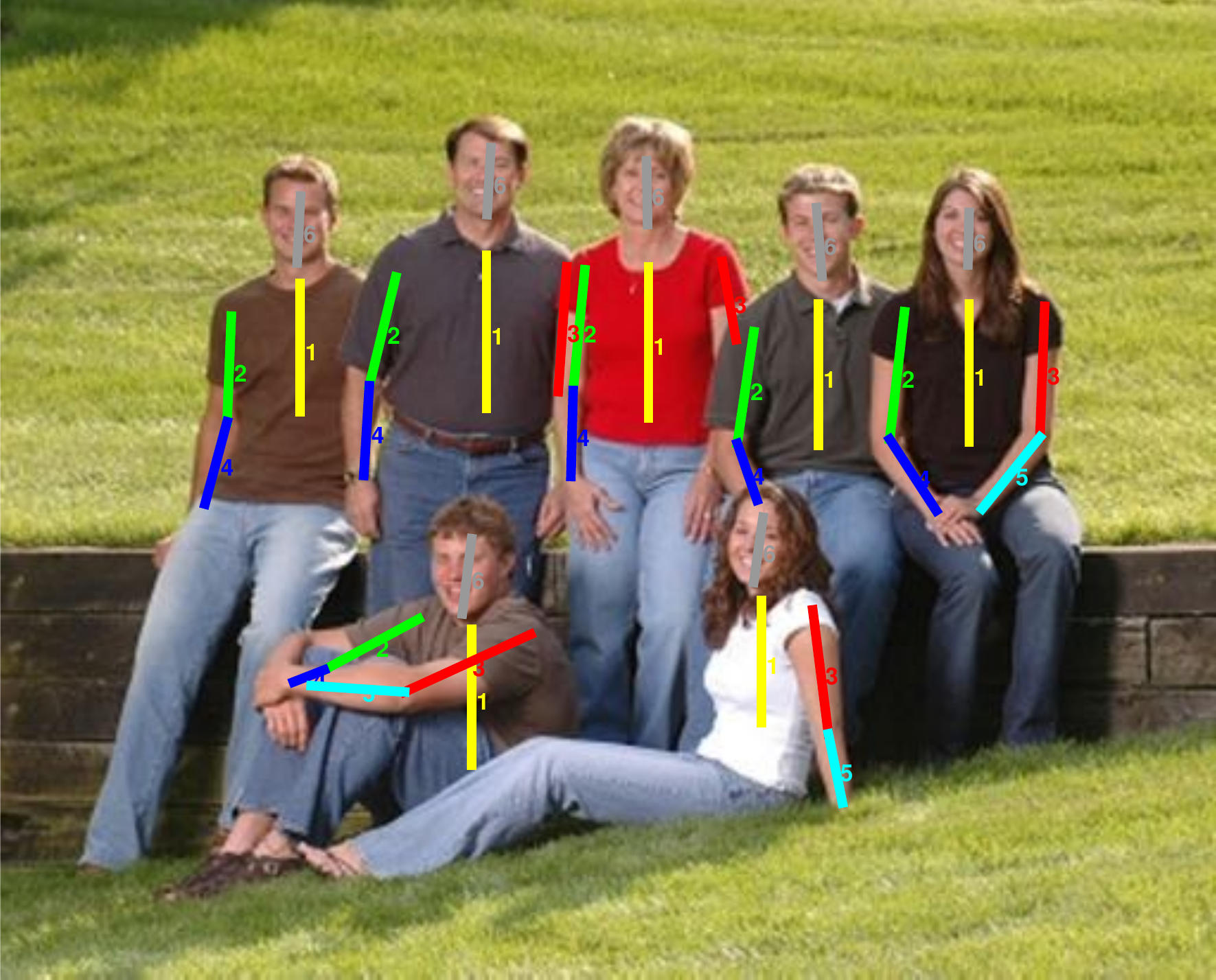}&
  \includegraphics[height=0.150\linewidth]{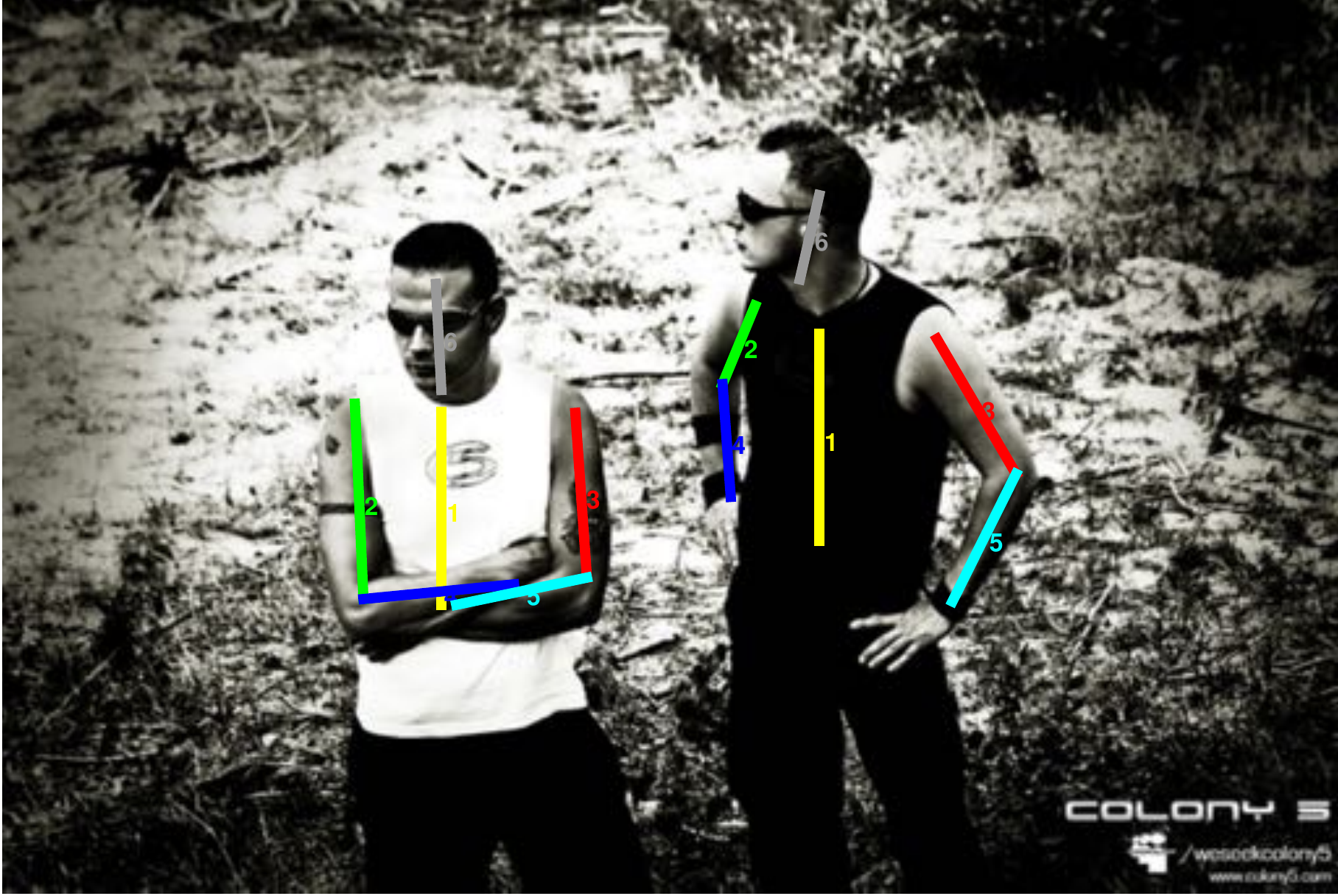}&
  \includegraphics[height=0.150\linewidth]{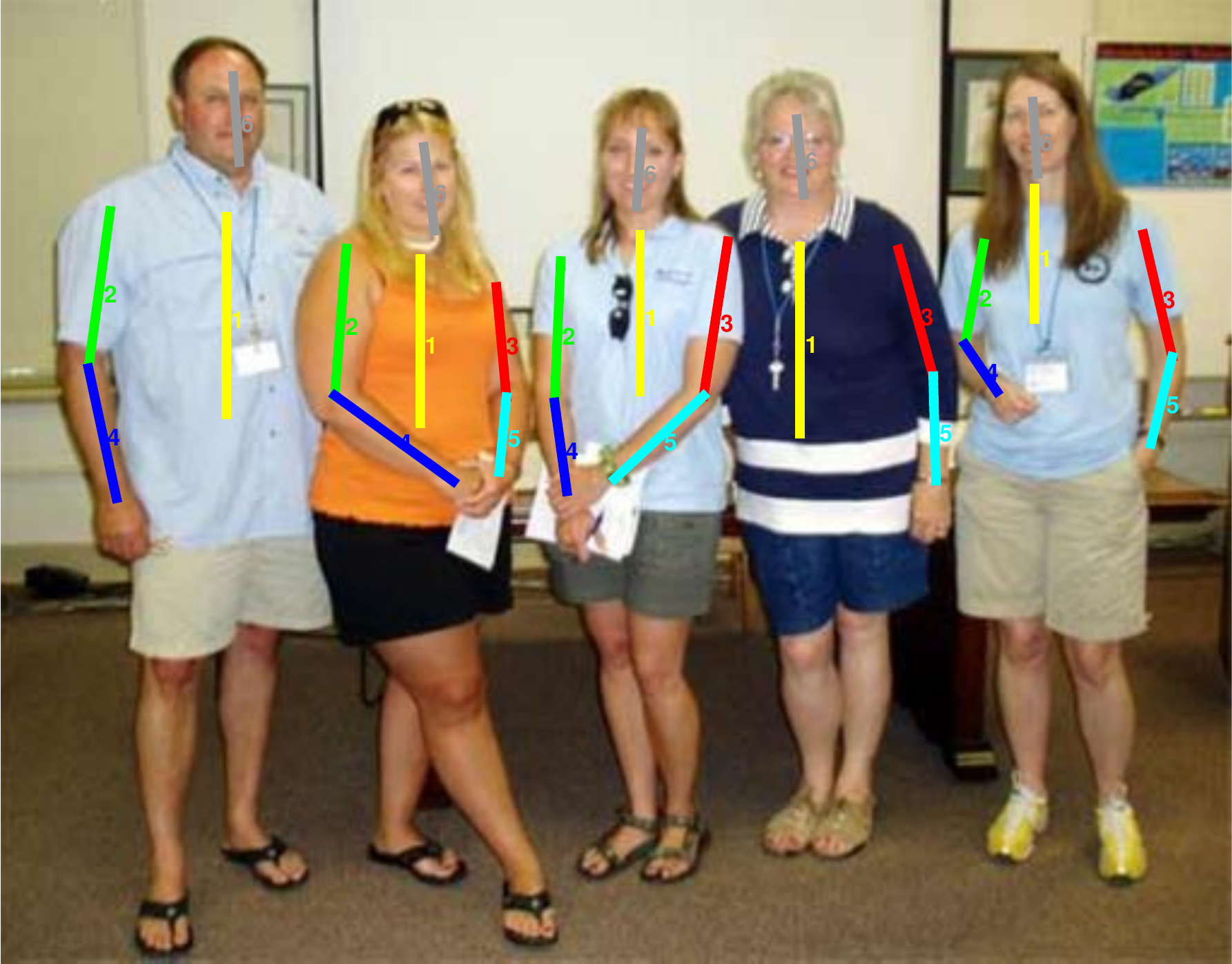}&
  \includegraphics[height=0.150\linewidth]{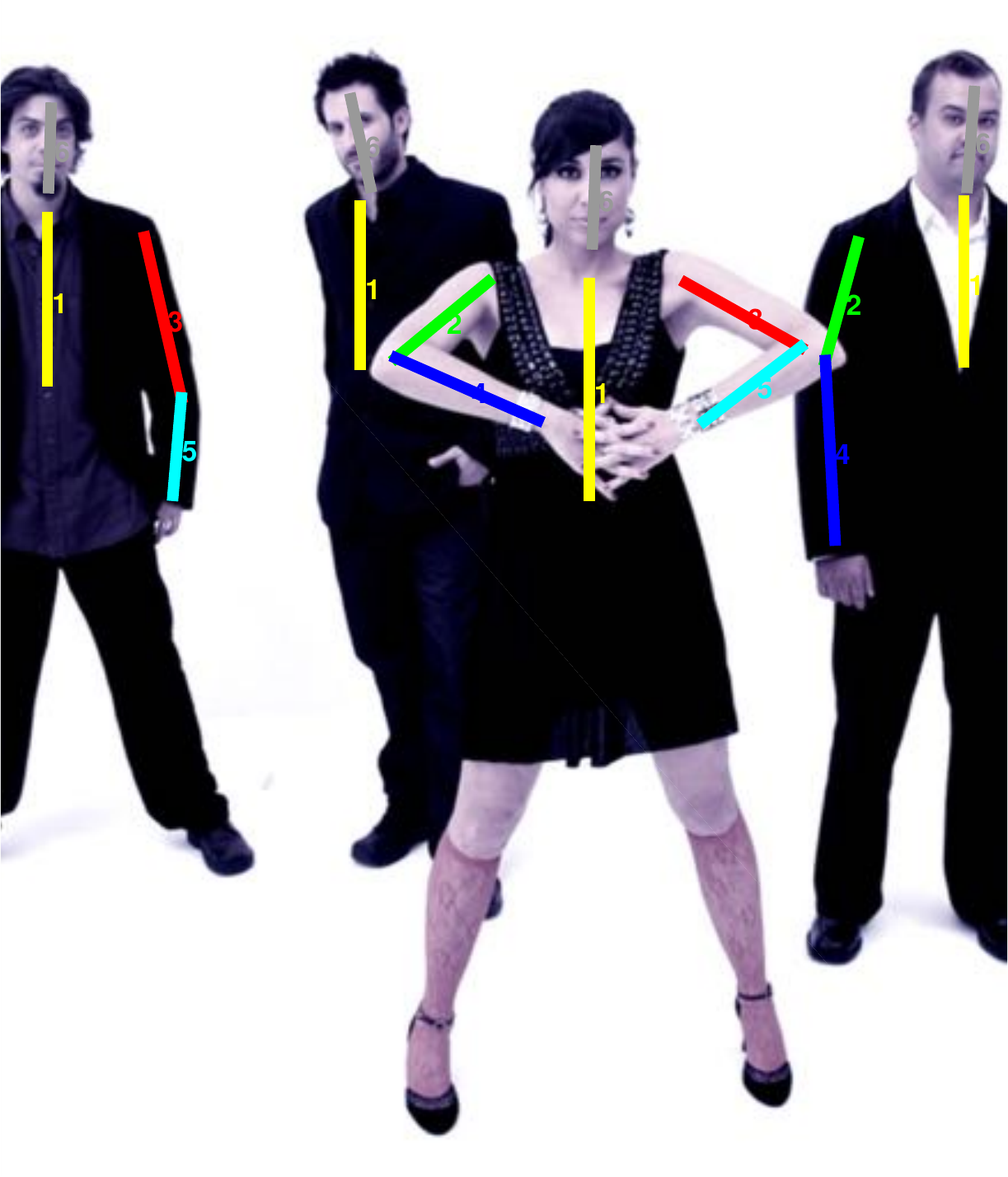}\\
  \begin{sideways}\bf \small Chen\&Yuille~\cite{Chen:2015:POC}\end{sideways}&
  \includegraphics[height=0.150\linewidth]{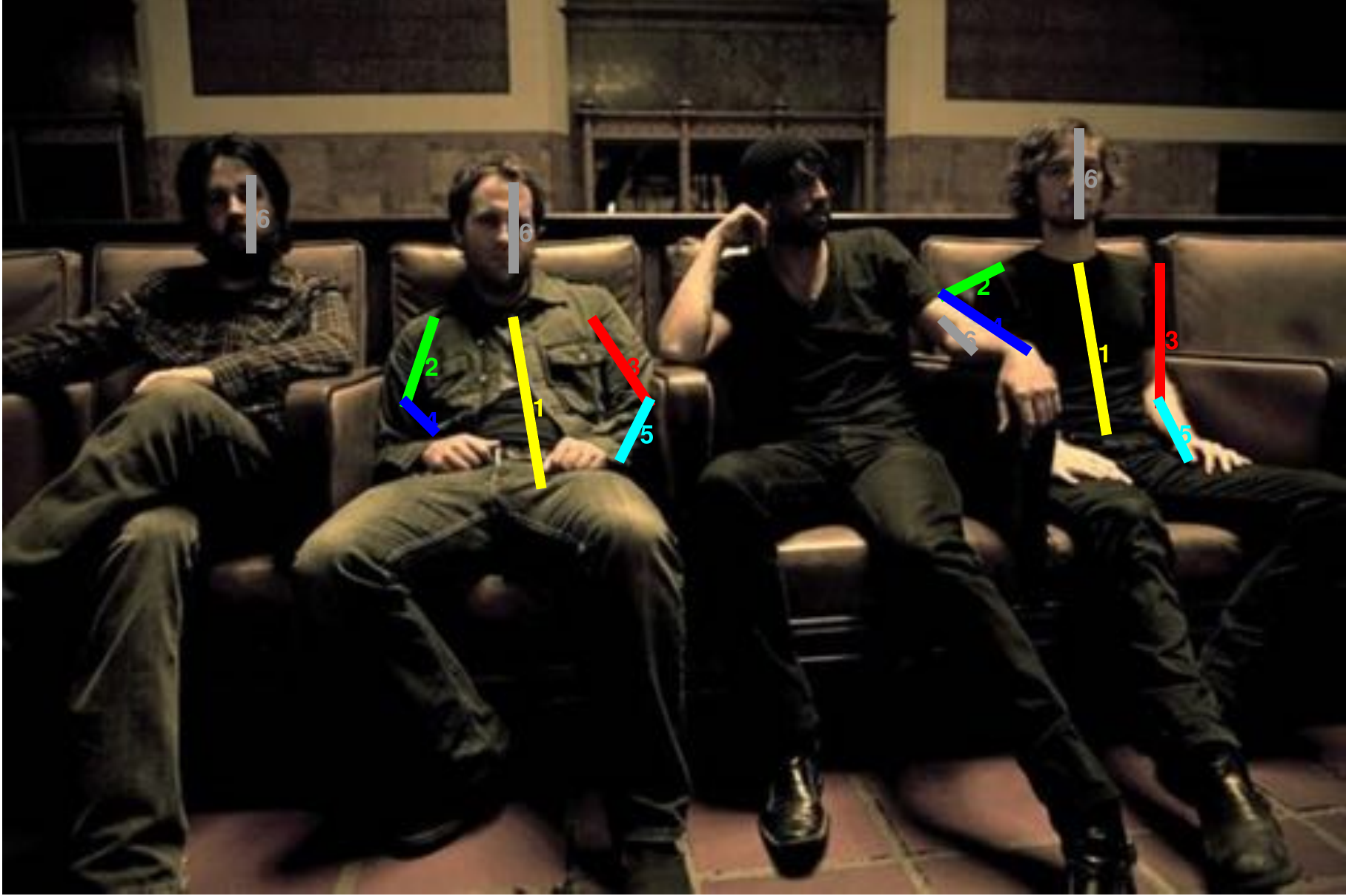}&
  \includegraphics[height=0.150\linewidth]{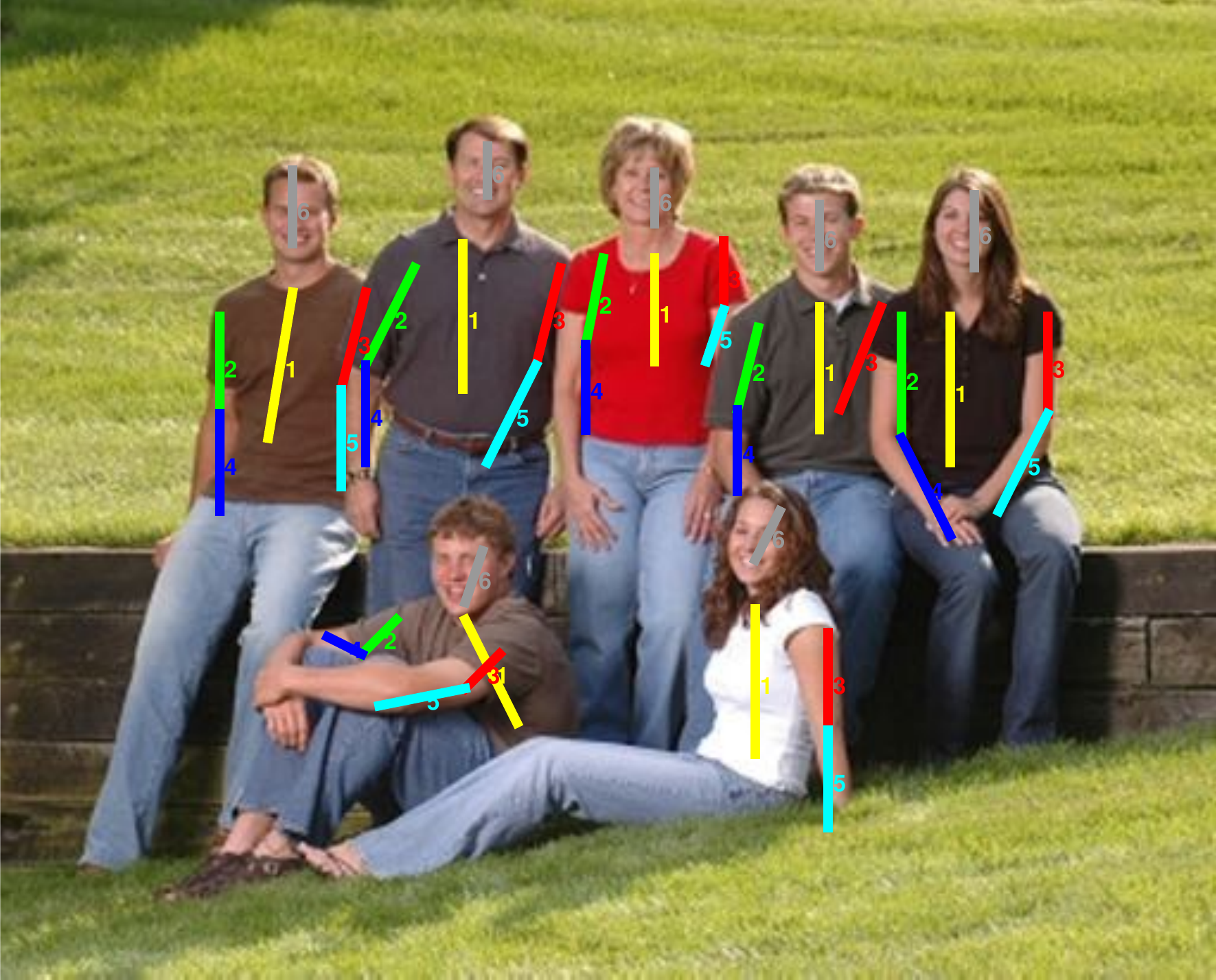}&
  \includegraphics[height=0.150\linewidth]{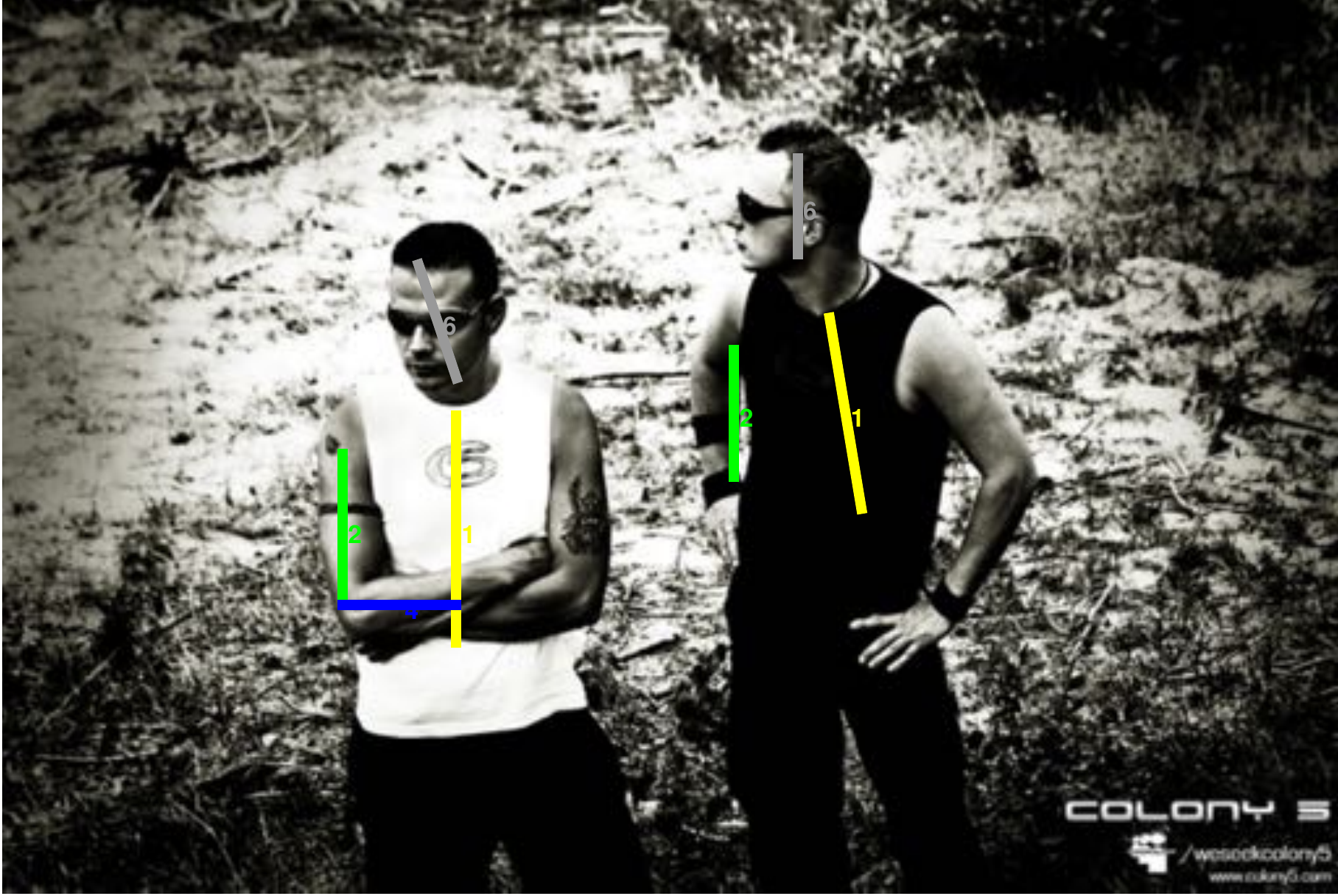}&
  \includegraphics[height=0.150\linewidth]{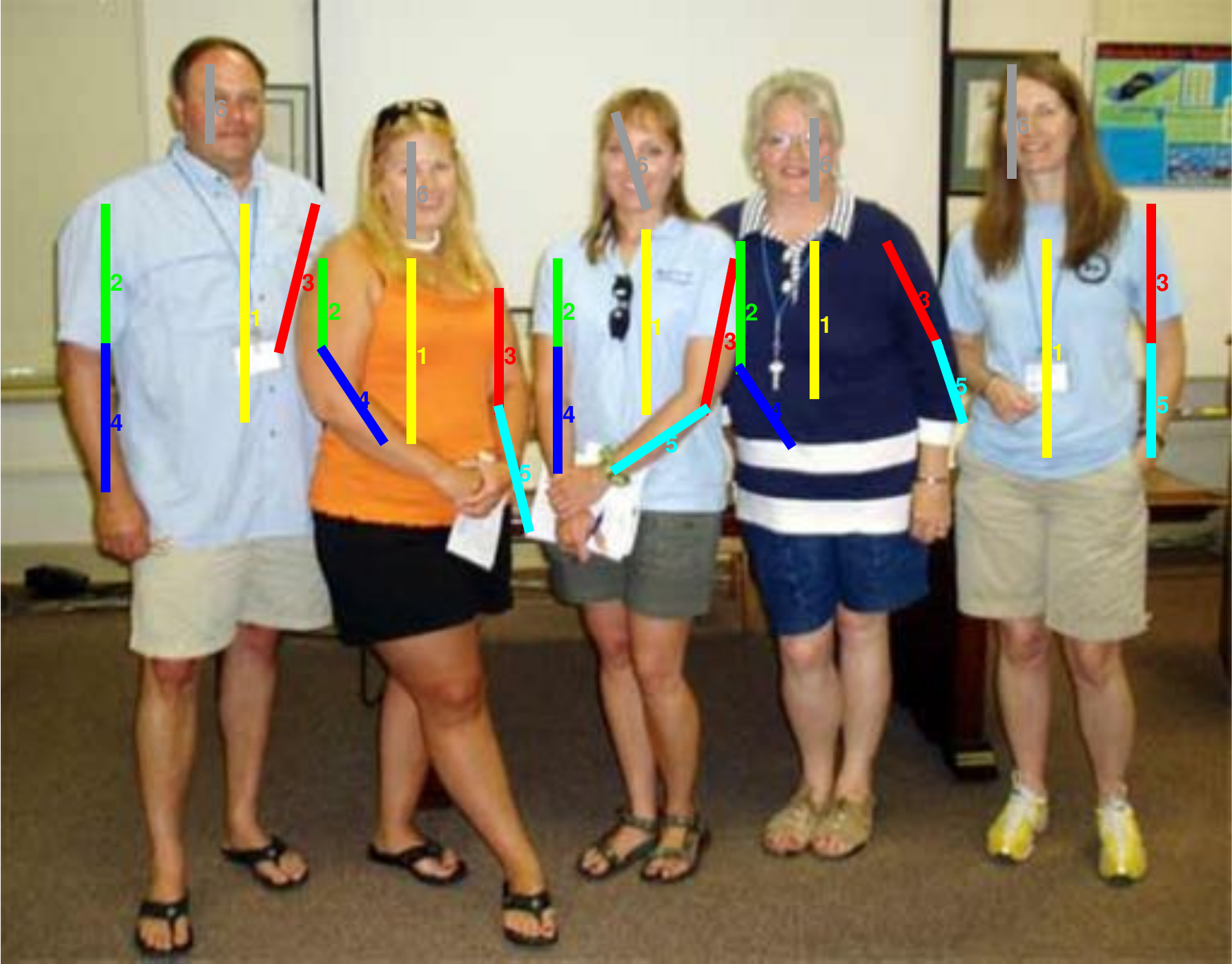}&
  \includegraphics[height=0.150\linewidth]{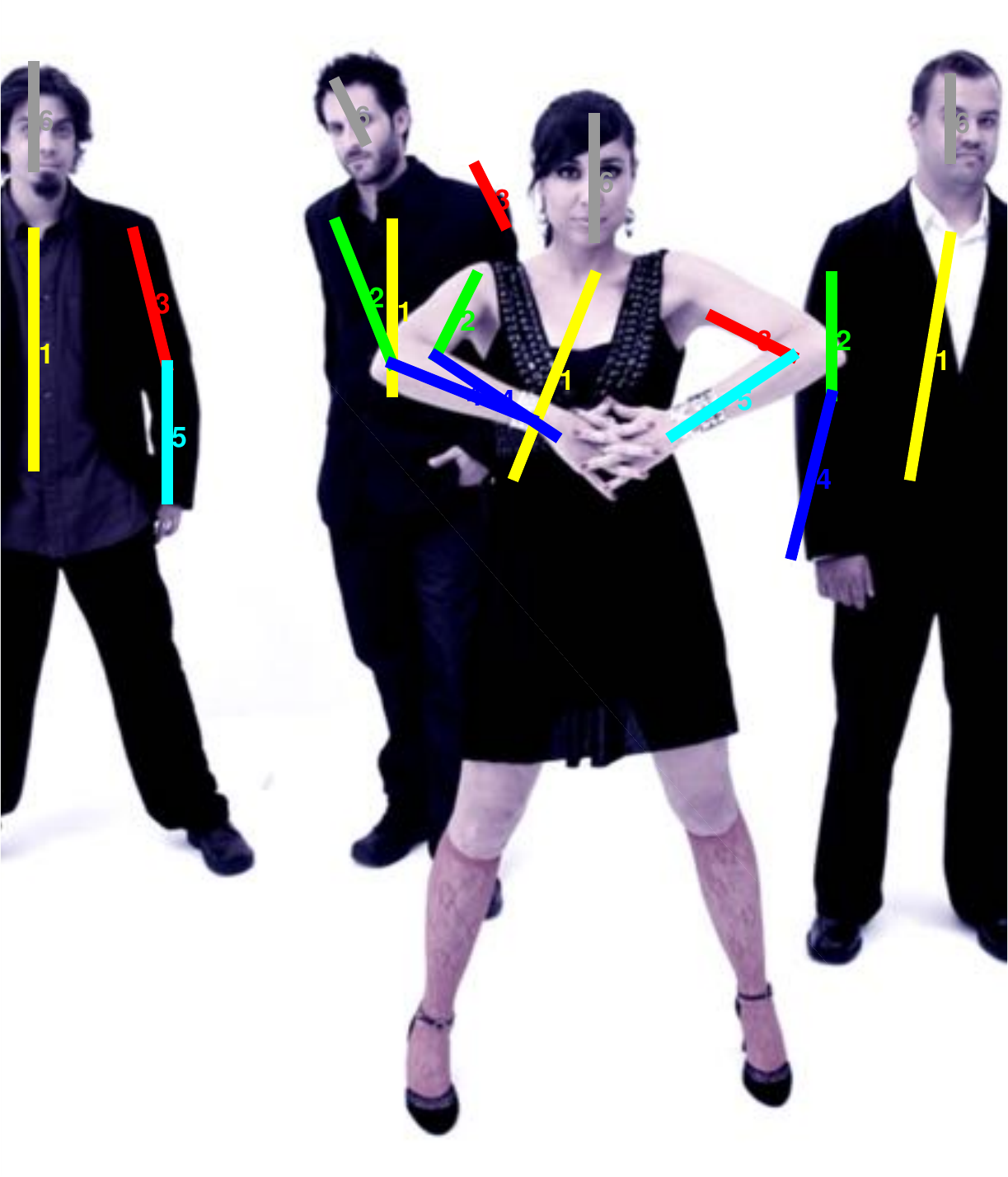}\\
  \end{tabular}
  %\vspace{-0.1em}
\caption{Qualitative comparison of our joint formulation
  $\deepcut~\multb~\dense$ (middle) to the traditional two-stage
  approach $\dense~\detroi$ (top) and the approach of
  Chen\&Yuille~\cite{Chen:2015:POC} (bottom) on WAF dataset. In
  contrast to $\detroi$, $\deepcut~\multb$ is able to disambiguate
  multiple and potentially overlapping persons and correctly assemble
  independent detections into plausible body part configurations. In
  contrast to ~\cite{Chen:2015:POC}, $\deepcut~\multb$ can better
  predict occlusions (image 2 person $1-4$ from the left, top row;
  image 4 person 1, 4; image 5, person 2) and better cope with strong
  articulations and foreshortenings (image 1, person 1, 3; image 2
  person 1 bottom row; image 3, person 1-2). See
  Appendix~\ref{sec:supplemental:waf} for more examples.
  %See supplementary material for more examples.
    %, but may fail                                                                                                                                                                                                                           
    %due to missing recall of part detection set (row 4 person 2, row 5                                                                                                                                                                       
    %person 3, 4).                                                                                                                                                                                                                            
  }
   \vspace{-1.0em}
  \label{fig:qualitative_waf_2}
\end{figure*}

Qualitative comparison to ~\cite{Chen:2015:POC} is provided in
Fig.~\ref{fig:qualitative_waf_2}.

\myparagraph{Results on MPII Multi-Person.}
%% We first describe the way
%% we compute \detroi~and~\gtroi~and then present the results.
%\noindent $\detroi$.
Obtaining a strong detector of highly articulated people having strong
occlusions and truncations is difficult. We employ a neck detector as
a person detector as it turned out to be the most reliable part.
%Optimal IoU-based non-maximum suppression and detection thresholds were
%selected using the training set.
%% , which results
%% in detection performance of 76.7\% AP on the test set, the highest
%% among the part detectors.
Full body bounding box is created around a neck detection and used as~\detroi.
%Then we non-uniformly extend detection bounding box while also taking
%the scale into account
%%  such that it has a size of
%% the upright person. This region is then used
%and use this box as \detroi. 
\gtroi s were provided by the authors~\cite{andriluka14cvpr}.
%% \noindent We now define four baselines based on the \emph{ROI}:
%% \noindent\emph{Unaries from $\detroi$} ($\undetroi$).
%% We extract a single highest scoring location for each body part
%% falling inside the $\detroi$.
%% \noindent\emph{Unaries from $\gtroi$} ($\ungtroi$).
%% Same as above but using the $\gtroi$.
%% \noindent\emph{Chen\&Yuille~\cite{chen14nips} on $\detroi$} ($\cydetroi$).
%% We perform single person pose estimation using the state of the art method of~\cite{chen14nips} applied to $\detroi$
%% crops.
%$\cygtroi$).
As the $\multb$ approach~\cite{Chen:2015:POC} is not public, we
compare to $\singb$ state-of-the-art method~\cite{chen14nips} applied
to $\gtroi$ image crops.

Results are shown in
Tab.~\ref{tab:multicut:mpii-multi}. 
%% \rcnn~$\detroi$ achieves $47.1$\%
%% AP. 
$\deepcut~\multb~\rcnn$ improves over $\rcnn~\detroi$ by 4.3\%
achieving 51.4\% AP. The largest differences are observed for the
ankle, knee, elbow and wrist, as those parts benefit more from the
connections to other parts. 
%We evaluate the performance of the
%$\multbu$ model that includes upper body parts
%only. 
$\deepcut~\multbu~\rcnn$ using upper body parts only slightly improves
over the full body model when compared on common parts (60.5 vs 58.2\%
AP). Similar tendencies are observed for $\dense$s, though improvements
of $\multbu$ over $\multb$ are more significant.
%% with two differences: 1) similar to the
%% other datasets the performance by $\deepcut~\dense$ models is higher
%% compared to $\deepcuts~\rcnn$; and 2) the performance of
%% ($\deepcuts~\multbu~\rcnn$)~is much better compared to $\multb$.%, due
%% %to better wrist estimation. %\todo{why?}

All $\deepcuts$ outperform $\cygtroi$, partially due to stronger part
detectors compared to~\cite{chen14nips}
(c.f. Tab.~\ref{tab:multicut:lsp}). Another reason is that $\cygtroi$
does not model body part occlusion and truncation always predicting
the full set of parts, which is penalized by the AP measure. In
contrast, our formulation allows to deactivate the part hypothesis in
the initial set of part candidates thus effectively performing
non-maximum suppression. In $\deepcuts$ part hypotheses are suppressed
based on the evidence from all other body parts making this process
more reliable.

\tabcolsep 1.5pt
\begin{table}[tbp]
 \scriptsize
  \centering
  \begin{tabular}{@{} l c ccc ccc cc@{}}
    \toprule
    Setting& Head   & Sho  & Elb & Wri & Hip & Knee & Ank & UBody & FBody \\
    \midrule
    %% \input{experiments/ap-det-roi-0.5-0.2-expidx485.tex}
%%     \input{experiments/ap-expidx485.tex}
%%     \input{experiments/ap-det-roi-0.5-0.2-expidx486.tex}
%%     \input{experiments/ap-expidx486.tex}
%%     \midrule
%%     \input{experiments/ap-gt-roi-expidx485.tex}
%%     \input{experiments/ap-gt-roi-expidx486.tex}
%%     \midrule
%%     \input{experiments_dense/ap-expidx695-det.tex}
%%     \input{experiments_dense/ap-expidx695.tex}
%%     \input{experiments_dense/ap-expidx699.tex}
%%     \midrule
%%     \input{experiments_dense/ap-expidx695-GT.tex}

%%     \textcolor{red}{103 images}\\
%%     \input{experiments/ap-det-roi-0.5-0.2-nimg103-expidx485.tex}
%%     \input{experiments/ap-nimg103-expidx485.tex}
%%     \input{experiments/ap-nimg103-expidx486.tex}
%%     \midrule
%%     \input{experiments/ap-det-roi-0.5-0.2-nimg103-expidx498.tex}
%%     \input{experiments/ap-nimg103-expidx498.tex}
%%     \input{experiments/ap-nimg103-expidx500.tex}
%%     \midrule
%%     \input{experiments/ap-gt-roi-nimg103-expidx485.tex}
%%     \input{experiments/ap-gt-roi-nimg103-expidx498.tex} 

%%    \textcolor{red}{162 images}\\
%%     \input{experiments/ap-det-roi-0.5-0.2-nimg162-expidx485.tex}
%%     \input{experiments/ap-nimg162-expidx485.tex}
%%     \input{experiments/ap-nimg162-expidx486.tex}
%%     \midrule
%%     \input{experiments/ap-det-roi-0.5-0.2-nimg162-expidx498.tex}
%%     \input{experiments/ap-nimg162-expidx498.tex}
%%     \input{experiments/ap-nimg162-expidx500.tex}
%%     \midrule
%%     \input{experiments/ap-gt-roi-nimg162-expidx485.tex}
%%     \input{experiments/ap-gt-roi-nimg162-expidx498.tex} 

    \rcnn~\detroi& 71.1  & 65.8  & 49.8  & 34.0  & 47.7  & 36.6 & 20.6 & 55.2 & 47.1 \\

    \rcnn~\multb& 71.8  & 67.8  & 54.9  & 38.1  & 52.0  & 41.2 & 30.4 & 58.2 & 51.4 \\

    \rcnn~\multbu& 75.2  & 71.0  & 56.4  & 39.6  & -  & - & - & 60.5 & - \\

    \midrule
    \dense~\detroi& 77.2  & 71.8  & 55.9  & 42.1  & 53.8  & 39.9 & 27.4 & 61.8 & 53.2 \\

    \dense~\multb& 73.4  & 71.8  & 57.9  & 39.9  & \textbf{56.7}  & \textbf{44.0} & \textbf{32.0} & 60.7 & \textbf{54.1} \\

    \dense~\multbu& \textbf{81.5}  & \textbf{77.3}  & \textbf{65.8}  & \textbf{50.0}  & -  & - & - & \textbf{68.7} & - \\

    \midrule
    \rcnn~\gtroi& 73.2  & 66.5  & 54.6  & 42.3  & 50.1  & 44.3 & 37.8 & 59.1 & 53.1 \\

    \dense~\gtroi& 78.1  & 74.1  & 62.2  & 52.0  & 56.9  & 48.7 & 46.1 & 66.6 & 60.2 \\

    $\cygtroi$& 65.0  & 34.2  & 22.0  & 15.7  & 19.2  & 15.8 & 14.2 & 34.2 & 27.1 \\

    \bottomrule
  \end{tabular}
  \vspace{0.1em}
\caption[]{Pose estimation results (AP) on MPII Multi-Person.}
    \vspace{-1.5em}
  \label{tab:multicut:mpii-multi}
\end{table}

\section{Conclusion}
Articulated pose estimation of multiple people in uncontrolled real
world images is challenging but of real world interest. In this work,
we proposed a new formulation as a joint subset partitioning and
labeling problem (SPLP). Different to previous two-stage strategies
that separate the detection and pose estimation steps, the SPLP model
jointly infers the number of people, their poses, spatial proximity,
and part level occlusions. Empirical results on four diverse and
challenging datasets show significant improvements over all previous
methods not only for the multi person, but also for the single person
pose estimation problem. On multi person WAF dataset we improve by
$30$\% PCP over the traditional two-stage approach. This shows that a
joint formulation is crucial to disambiguate multiple and potentially
overlapping persons. Models and code available at \myurl.

%% \pg{which
%%   is
%% \pg{exclude:
%%   ,which shows the effectiveness of our model.}  \pg{include the
%%   message of the paper, for example: This paper finds that single
%%   person approaches can be used for the multi-person case only to some
%%   extend. A joint formulation of the problem is crucial to disambiguate
%%   multiple and potential overlapping persons.  We have demonstrated
%%   that the SPLP model, by integrating meaningful constraints severely
%%   improves on this important task.}

%% For example on the upper body pose estimation our
%% approach improves by $9.3$ percent points of mAP over the strong
%% CNN-based part detector, which is more than 30\% relative
%% improvement. In the future we plan to improve the performance of our
%% approach by employing more powerful unary and pairwise terms.
%% \pg{That is probably the weakest sentence to end this paper
%%   with. Nobody will ask for future extensions, especially not the most
%%   boring ones.}

\begin{appendices}
\section{Additional Results on LSP dataset}
\label{seq:supplemental:lsp}
We provide additional quantitative results on LSP dataset using
person-centric (PC) and observer-centric (OC) evaluation settings.
\subsection{LSP Person-Centric (PC)}
First, detailed performance analysis is performed when evaluating
various parameters of $\rcnn$ and results are reported using
PCK~\cite{sapp13cvpr} evaluation
measure. Then, performance of the proposed $\rcnn$ and $\dense$ part
detection models is evaluated using strict PCP~\cite{Ferrari:2008:PSS}
measure.

\myparagraph{Detailed $\rcnn$ performance analysis (PCK).} Detailed
parameter analysis of $\rcnn$ is provided in
Tab.~\ref{tab:unary-large:rcnn} and results are reported using PCK
evaluation measure. Respecting parameters for each experiment are
shown in the first column and parameter differences between the
neighboring rows in the table are highlighted in bold. Re-scoring the
$2000$ DPM proposals using $\rcnn$ with
AlexNet~\cite{krizhevsky12nips} leads to $56.9$\% PCK. This is
achieved using basis scale $1$ ($\approx$ head size) of proposals and
training with initial learning rate (lr) of $0.001$ for $80$k
iterations, after which lr is reduced by $0.1$, for a total number of
$140$k SGD iterations. In addition, bounding box regression and
default IoU threshold of $0.5$ for positive/negative label
assignment~\cite{girshickICCV15fastrcnn} have been used. Extending the
regions by $4$x increases the performance to 65.1\% PCK, as it
incorporates more context including the information about symmetric
body parts and allows to implicitly encode higher-order body part
relations into the part detector. No improvements observed for larger
scales. Increasing lr to $0.003$, lr reduction step to $160$k and
training for a larger number of iterations ($240$k) improves the
results to $67.4$, as higher lr allows for for more significant
updates of model parameters when finetuned on the task of human body
part detection. Increasing the number of training examples by reducing
the training IoU threshold to $0.4$ results into slight performance
improvement ($68.8$ vs. $67.4$\% PCK). Further increasing the number
of training samples by horizontally flipping each image and performing
translation and scale jittering of the ground truth training samples
improves the performance to $69.6$\% PCK and $42.3$\% AUC. The
improvement is more pronounced for smaller distance thresholds ($42.3$
vs. $40.9$\% AUC): localization of body parts is improved due to the
increased number of jittered samples that significantly overlap with
the ground truth. Further increasing the lr, lr reduction step and
total number of iterations altogether improves the performance to
$72.4$\% PCK, and very minor improvements are observed when training
longer. All results above are achieved by finetuning the AlexNet
architecture from the ImageNet model on the MPII training set. Further
finetuning the MPII-finetuned model on the LSP training set increases
the performance to $77.9$\% PCK, as the network learns LSP-specific
image representations. Using the deeper VGG~\cite{Simonyan14c}
architecture improves over more shallow AlexNet (77.9 vs. 72.4\% PCK,
50.0 vs. 44.6\% AUC). Funetuning VGG on LSP achieves remarkable 82.8\%
PCK and 57.0\% AUC. Strong increase in AUC (57.0 vs. 50\%)
characterizes the improvement for smaller PCK evaluation
thresholds. Switching off bounding box regression results into
performance drop (81.3\% PCK, 53.2\% AUC) thus showing the importance
of the bounding box regression for better part localization. Overall,
we demonstrate that proper adaptation and tweaking of the
state-of-the-art generic object detector
FR-CNN~\cite{girshickICCV15fastrcnn} leads to a strong body part
detection model that dramatically improves over the vanilla FR-CNN
($82.8$ vs. $56.9$\% PCK, $57.8$ vs. $35.9$\% AUC) and significantly
outperforms the state of the art ($+9.4$\% PCK over the best known PCK
result~\cite{chen14nips} and $+9.7$\% AUC over the best known AUC
result~\cite{tompson14nips}.

\tabcolsep 1.5pt
\begin{table*}[tbp]
 \scriptsize
  \centering
  \begin{tabular}{@{} l c ccc cc cc|c@{}}
    \toprule
    Setting& Head   & Sho  & Elb & Wri & Hip & Knee & Ank & PCK & AUC\\
    \midrule
    %\cmidrule(r){1-1} \cmidrule(lr){2-9} \cmidrule(lr){10-10}
    %\input{experiments/pck-torso-expidx527-2.tex}
    %\midrule
    %\input{experiments/pck-torso-expidx534.tex}
    %\midrule    
    %AlN s1 lr.001(80k) 140k IoU.5 & 82.2  & 67.0  & 49.6  & 45.4  & 53.1  & 52.9 & 48.2 & 56.9 & 35.9 \\
AlexNet scale 1, lr 0.001, lr step 80k, \# iter 140k, IoU pos/neg 0.5 & 82.2  & 67.0  & 49.6  & 45.4  & 53.1  & 52.9 & 48.2 & 56.9 & 35.9 \\

    AlexNet \textbf{scale 4}, lr 0.001, lr step 80k, \# iter 140k, IoU pos/neg 0.5& 85.7  & 74.4  & 61.3  & 53.2  & 64.1  & 63.1 & 53.8 & 65.1 & 39.0 \\

    AlexNet scale 4, \textbf{lr 0.003, lr step 160k, \# iter 240k}, IoU pos/neg 0.5 & 87.0  & 75.1  & 63.0  & 56.3  & 67.0  & 65.7 & 58.0 & 67.4 & 40.8\\

    AlexNet scale 4, lr 0.003, lr step 160k, \# iter 240k, \textbf{IoU pos/neg 0.4} & 87.5  & 76.7  & 64.8  & 56.0  & 68.2  & 68.7 & 59.6 & 68.8 & 40.9\\

    AlexNet scale 4, lr 0.003, lr step 160k, \# iter 240k, IoU pos/neg 0.4, \textbf{data augment} & 87.8  & 77.8  & 66.0  & 58.1  & 70.9  & 66.9 & 59.8 & 69.6 & 42.3\\

    %AlN s4 lr.004(320k) 1M IoU.4 aug& 88.1  & 79.3  & 68.9  & 62.6  & 73.5  & 69.3 & 64.7 & 72.4 & 44.6 \\
AlexNet scale 4, \textbf{lr 0.004, lr step 320k, \# iter 1M}, IoU pos/neg 0.4, data augment & 88.1  & 79.3  & 68.9  & 62.6  & 73.5  & 69.3 & 64.7 & 72.4 & 44.6 \\

    \quad\quad + finetune LSP, lr 0.0005, lr step 10k, \# iter 40k& 92.9  & 81.0  & 72.1  & 66.4  & 80.6  & 77.6 & 75.0 & 77.9 & 51.6\\

    \midrule
    %VGG s4 lr.003(160k) 320k IoU.4 aug& 91.0  & 84.2  & 74.6  & 67.7  & 77.4  & 77.3 & 72.8 & 77.9 & 50.0 \\
VGG scale 4, lr 0.003, lr step 160k, \# iter 320k, IoU pos/neg 0.4, data augment & 91.0  & 84.2  & 74.6  & 67.7  & 77.4  & 77.3 & 72.8 & 77.9 & 50.0 \\

    %\input{experiments/pck-torso-expidx179.tex}
    %\quad + finetune LSP lr0.0005 (10k) 40k & \textbf{95.4}  & \textbf{86.5}  & \textbf{77.8}  & \textbf{74.0}  & \textbf{84.5}  & \textbf{78.8} & \textbf{82.6} & \textbf{82.8} &\textbf{57.0}\\
\quad + finetune LSP lr 0.0005, lr step 10k, \# iter 40k & \textbf{95.4}  & \textbf{86.5}  & \textbf{77.8}  & \textbf{74.0}  & \textbf{84.5}  & \textbf{78.8} & \textbf{82.6} & \textbf{82.8} &\textbf{57.0}\\

    \bottomrule \end{tabular} 
    \vspace{0.3em} 
    \caption[]{PCK performance of $\rcnn$ (unary) on LSP (PC)
    dataset. $\rcnn$ is finetuned from ImageNet on MPII (lines 1-6,
    8), and then finetuned on LSP (lines 7, 9).}
\vspace{-1.5em} 
\label{tab:unary-large:rcnn}
\end{table*}

\myparagraph{Overall performance using PCP evaluation measure.}
Performance when using the strict ``Percentage of Correct Parts
(PCP)''~\cite{Ferrari:2008:PSS} measure is reported in
Tab.~\ref{tab:multicut:lsp-pc-pcp}. In contrast to PCK measure
evaluating the accuracy of predicting body joints, PCP evaluation
metric measures the accuracy of predicting body part sticks. $\rcnn$
achieves $78.3$\% PCP. Similar to PCK results, $\deepcut~\singb~\rcnn$
slightly improves over unary alone, as it enforces more consistent
predictions of body part sticks. Using more general multi-person
$\deepcut~\multb~\rcnn$ model results into similar performance, which
shows the generality of $\deepcut~\multb$
method. $\deepcut~\singb~\dense$ slightly improves over $\dense$ alone
($84.3$ vs. $83.9$\% PCP) achieving the best PCP result on LSP dataset
using PC annotations. This is in contrast to PCK results where
performance differences $\deepcut~\singb~\dense$ vs. $\dense$ alone
are minor.

We now compare the PCP results to the state of the art. The $\deepcut$
models outperform all other methods by a large margin. The best known
PCP result by Chen\&Yuille~\cite{chen14nips} is outperformed by
$10.7$\% PCP. This is interesting, as their deep learning based
method relies on the image conditioned pairwise terms while our
approach uses more simple geometric only connectivity. Interestingly,
$\rcnn$ alone outperforms the approach of Fan et al.~\cite{fan15cvpr}
(78.3 vs. 70.1\% PCP), who build on the previous version of the R-CNN
detector~\cite{girshick2014rcnn}. At the same time, the best
performing dense architecture $\deepcut~\singb~\dense$
outperforms~\cite{fan15cvpr} by $+14.2$\% PCP. Surprisingly,
$\deepcut~\singb~\dense$ dramatically outperforms the method of
Tompson et al.~\cite{tompson14nips} (+17.7\% PCP) that also produces
dense score maps, but additionally includes multi-scale receptive
fields and jointly trains appearance and spatial models in a single
deep learning framework. We envision that both advances can further
improve the performance of $\deepcut$ models. Finally, all proposed
approaches significantly outperform earlier non-deep learning based
methods~\cite{wang13cvpr,pishchulin13iccv} relying on hand-crafted
image features.

\tabcolsep 1.5pt
\begin{table}[tbp]
 \scriptsize
  \centering
  \begin{tabular}{@{} l cc ccc ccc@{}}
    \toprule
    &Torso & Upper & Lower & Upper & Fore- & Head  & PCP \\
    &   & Leg   & Leg& Arm   & arm   &       &       \\
    \midrule
    $\rcnn$ (unary) & 93.2  & 82.7  & 77.7  & 75.5  & 63.5  & 91.2 & 78.3 \\

    \quad+ $\deepcut$ \singb & 93.3  & 83.2  & 77.8  & 76.3  & 63.7  & 91.5 & 78.7 \\

    \quad\quad+ appearance pairwise & 93.4  & 83.5  & 77.8  & 76.6  & 63.8  & 91.8 & 78.9 \\

    \quad+ $\deepcut$ \multb & 93.6  & 83.3  & 77.6  & 76.3  & 63.5  & 91.2 & 78.6 \\

    \midrule
    $\dense$ (unary)          & 96.2  & 87.8  & 81.8  & 81.6  & \textbf{72.3}  & 95.6 & 83.9 \\
    \quad + $\deepcut~\singb$ & \textbf{97.0}  & \textbf{88.8}  & \textbf{82.0}  & \textbf{82.4}  & 71.8  & \textbf{95.8} & \textbf{84.3} \\
    \quad + $\deepcut~\multb$& 96.4  & \textbf{88.8}  & 80.9  & \textbf{82.4}  & 71.3  & 94.9 & 83.8\\
 
    \midrule
    Tompson et al.~\cite{tompson14nips}& 90.3  & 70.4  & 61.1  & 63.0  & 51.2  & 83.7 & 66.6 \\
    Chen\&Yuille~\cite{chen14nips}& 96.0  & 77.2  & 72.2  & 69.7  & 58.1  & 85.6 & 73.6 \\
    Fan et al.~\cite{fan15cvpr}$^*$
    & 95.4  & 77.7  & 69.8  & 62.8  & 49.1  & 86.6 & 70.1 \\
    Pishchulin et al.~\cite{pishchulin13iccv}& 88.7  & 63.6  & 58.4  & 46.0  & 35.2  & 85.1 & 58.0 \\
    Wang\&Li~\cite{wang13cvpr}& 87.5  & 56.0  & 55.8  & 43.1  & 32.1  & 79.1 & 54.1 \\
    \bottomrule
  \end{tabular}
  
  $^*$ re-evaluated using the standard protocol, for details see project page of~\cite{fan15cvpr}
  \caption[]{Pose estimation results (PCP) on LSP (PC) dataset.}
    \vspace{-1.5em}
  \label{tab:multicut:lsp-pc-pcp}
\end{table}

\subsection{LSP Observer-Centric (OC)}

We now evaluate the performance of the proposed part detection models
on LSP dataset using the observer-centric (OC)
annotations~\cite{eichner12accv}. In contrast to the person-centric
(PC) annotations used in all previous experiments, OC annotations do
not penalize for the right/left body part prediction flips and count a
body part to be the right body part, if it is on the right side of the
line connecting pelvis and neck, and a body part to be the left body part
otherwise.

Evaluation is performed using the official OC annotations provided
by~\cite{pishchulin13cvpr,eichner12accv}. Prior to evaluation, we
first finetune the $\rcnn$ and $\dense$ part detection models from
ImageNet on MPII and MPII+LSPET training sets, respectively, (same as
for PC evaluation), and then further finetuned the models on LSP OC
training set.

\begin{figure}
  \centering
  \begin{tabular}{c c}  
  \includegraphics[width=0.6\linewidth]{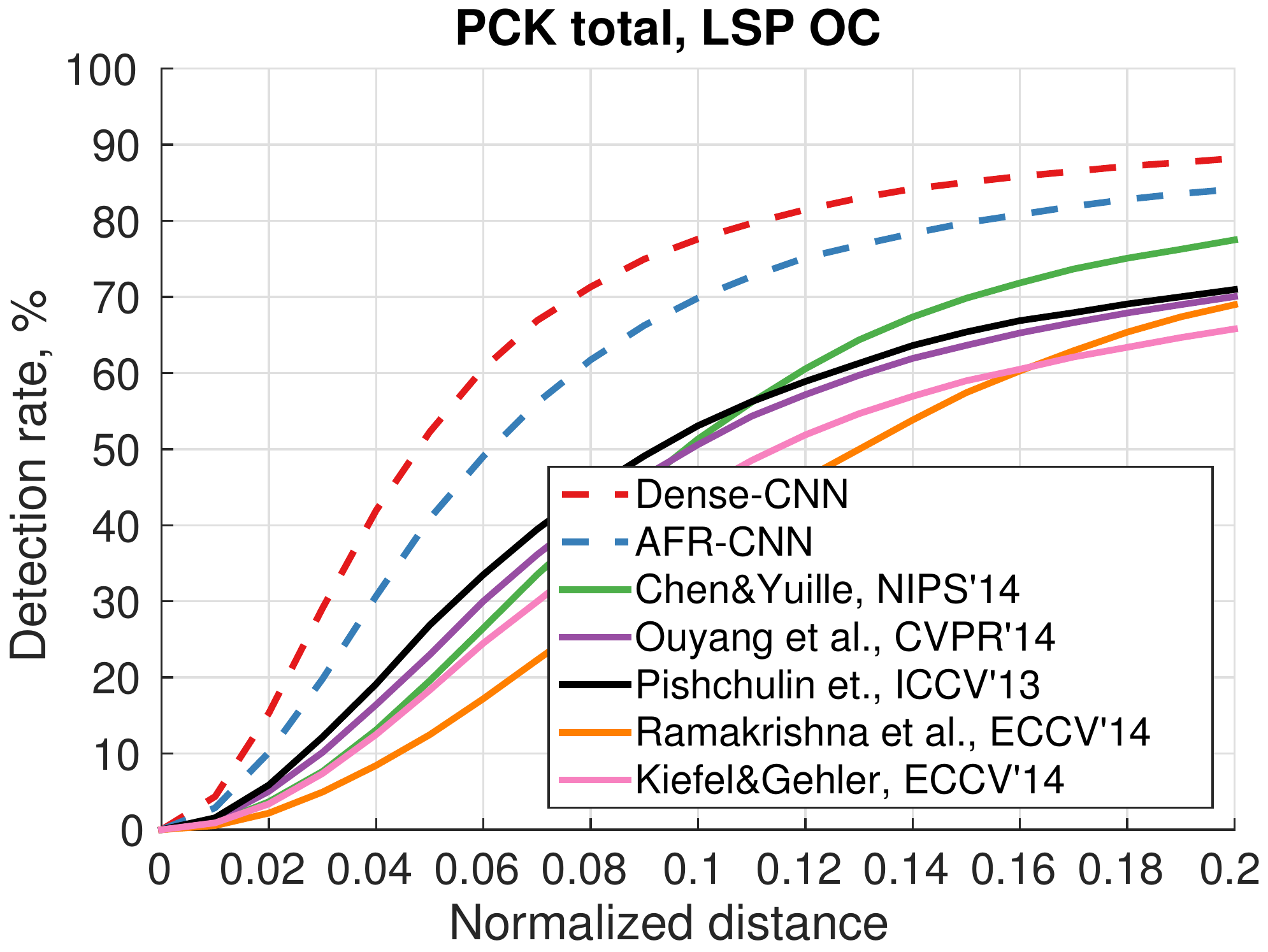}&
  \end{tabular}
  %\vspace{-1.0em}
%  \caption{Pose estimation performance on (a) LSP (PC) dataset (PCK) and (b) MPII Single Person dataset (PCKh).} 
  \caption{Pose estimation results over all PCK thresholds on LSP (OC) dataset.} 
  \vspace{0.2em}
  \label{fig:pck-curves:lsp:oc}
   \vspace{-1.0em}
\end{figure}

\myparagraph{PCK evaluation measure.} Results using OC annotations and
PCK evaluation measure are shown in Tab.~\ref{tab:multicut:lsp:oc} and
in Fig.~\ref{fig:pck-curves:lsp:oc}. $\rcnn$ achieves $84.2$\% PCK and
$58.1$\% AUC. This result is only slightly better compared to $\rcnn$
evaluated using PC annotations (84.2 vs 82.8\% PCK, $58.1$
vs. $57.0$\% AUC). Although PC annotations correspond to a harder
task, only small drop in performance when using PC annotations shows
that the network can learn to accurately predict person's viewpoint and
correctly label left/right limbs in most cases. This is contrast to
earlier approaches based on hand-crafted features whose performance
drops much stronger when evaluated in PC evaluation setting
(e.g.~\cite{pishchulin13iccv} drops from 71.0\% PCK when using OC
annotations to 58.0\% PCK when using PC annotations). Similar to PC
case, $\dense$ detection model outperforms $\rcnn$ (88.2 vs. 84.2\%
PCK and 65.0 vs. 58.1\% AUC). The differences are more pronounced when
examining the entire PCK curve for smaller distance thresholds
(c.f. Fig.~\ref{fig:pck-curves:lsp:oc}).

\tabcolsep 1.5pt
\begin{table}[tbp]
 \scriptsize
  \centering
  \begin{tabular}{@{} l c ccc ccc c|c@{}}
    \toprule
    Setting& Head   & Sho  & Elb & Wri & Hip & Knee & Ank & PCK & AUC\\       
    \midrule
    $\rcnn$ (unary)& 95.3  & 88.3  & 78.5  & 74.2  & 87.3  & 84.2 & 81.2 & 84.2 & 58.1\\
    \midrule
    $\dense$ (unary)  & \textbf{97.4}  & \textbf{92.0}  & \textbf{83.8}  & \textbf{79.0}  & \textbf{93.1}  & \textbf{88.3} & \textbf{83.7} & \textbf{88.2} & \textbf{65.0}\\
    \midrule
    Chen\&Yuille~\cite{chen14nips}& 91.5  & 84.7  & 70.3  & 63.2  & 82.7  & 78.1 & 72.0 & 77.5 & 44.8\\
    Ouyang et al.~\cite{ouyang14cvpr}& 86.5  & 78.2  & 61.7  & 49.3  & 76.9  & 70.0 & 67.6 & 70.0 & 43.1\\
    Pishchulin et.~\cite{pishchulin13iccv}& 87.5  & 77.6  & 61.4  & 47.6  & 79.0  & 75.2 & 68.4 & 71.0& 45.0\\
    Kiefel\&Gehler~\cite{kiefel14eccv} & 83.5  & 73.7  & 55.9  & 36.2  & 73.7  & 70.5 & 66.9 & 65.8 & 38.6\\
    Ramakrishna et al.~\cite{ramakrishna14eccv}& 84.9  & 77.8  & 61.4  & 47.2  & 73.6  & 69.1 & 68.8 & 69.0& 35.2\\
    \bottomrule                   
  \end{tabular}
  \vspace{0.3em} 
  \caption[]{Pose estimation results (PCK) on LSP (OC) dataset.}
  %\vspace{-1.5em} 
  \label{tab:multicut:lsp:oc}
\end{table}

Comparing the performance by $\rcnn$ and $\dense$ to the state of the
art, we observe that both proposed approaches significantly outperform
other methods. Both deep learning based approaches of
Chen\&Yuille~\cite{chen14nips} and Ouyang et al.~\cite{ouyang14cvpr}
are outperformed by $+10.7$ and $+18.2$\% PCK when compared to the
best performing $\dense$. Analysis of PCK curve for the entire range
of PCK distance thresholds reveals even larger performance differences
(c.f. Fig.~\ref{fig:pck-curves:lsp:oc}). The results using OC
annotations confirm our findings from PC evaluation and clearly show
the advantages of the proposed part detection models over the
state-of-the-art deep learning methods~\cite{chen14nips,ouyang14cvpr},
as well as over earlier pose estimation methods based on hand-crafted
image features~\cite{pishchulin13iccv,kiefel14eccv,ramakrishna14eccv}.

\tabcolsep 1.5pt
\begin{table}[tbp]
 \scriptsize
  \centering
  \begin{tabular}{@{} l cc ccc ccc@{}}
    \toprule
    &Torso & Upper & Lower & Upper & Fore- & Head  & PCP \\
    &   & Leg   & Leg& Arm   & arm   &       &       \\
    \midrule
    $\rcnn$ (unary) & 92.9  & 86.3  & 79.8  & 77.0  & 64.2  & 91.8 & 79.9 \\   
    \midrule
    $\dense$ (unary)          & \textbf{96.0}  & \textbf{91.0}  & \textbf{83.5}  & \textbf{82.8}  & \textbf{71.8}  & \textbf{96.2} & \textbf{85.0} \\
    \midrule
    Chen\&Yuille~\cite{chen14nips}& 92.7  & 82.9  & 77.0  & 69.2  & 55.4  & 87.8 & 75.0 \\
    Ouyang et al.~\cite{ouyang14cvpr}& 88.6  & 77.8  & 71.9  & 61.9  & 45.4  & 84.3 & 68.7 \\
    Pishchulin et.~\cite{pishchulin13iccv}& 88.7  & 78.9  & 73.2  & 61.8  & 45.0  & 85.1 & 69.2 \\
    Kiefel\&Gehler~\cite{kiefel14eccv}& 84.3  & 74.5  & 67.6  & 54.1  & 28.3  & 78.3 & 61.2 \\
    Ramakrishna et al.~\cite{ramakrishna14eccv}& 88.1  & 79.0  & 73.6  & 62.8  & 39.5  & 80.4 & 67.8 \\
    \bottomrule
  \end{tabular}
  \caption[]{Pose estimation results (PCP) on LSP (OC) dataset.}
    \vspace{-1.5em}
  \label{tab:multicut:lsp-oc-pcp}
\end{table}

\myparagraph{PCP evaluation measure.} Results using OC annotations and
PCP evaluation measure are shown in
Tab.~\ref{tab:multicut:lsp-oc-pcp}. Overall, the trend is similar to
PC evaluation: both proposed approaches significantly outperform
the state-of-the-art methods with $\dense$ achieving the best result
of 85.0\% PCP thereby improving by $+10$\% PCP over the best
published result~\cite{chen14nips}.

\section{Additional Results on WAF dataset}
\label{sec:supplemental:waf}
\begin{figure*}
  \centering
  \begin{tabular}{c c c c c c c}
    \begin{sideways}\bf \small\quad\quad $\detroi$\end{sideways}&
    \includegraphics[height=0.145\linewidth]{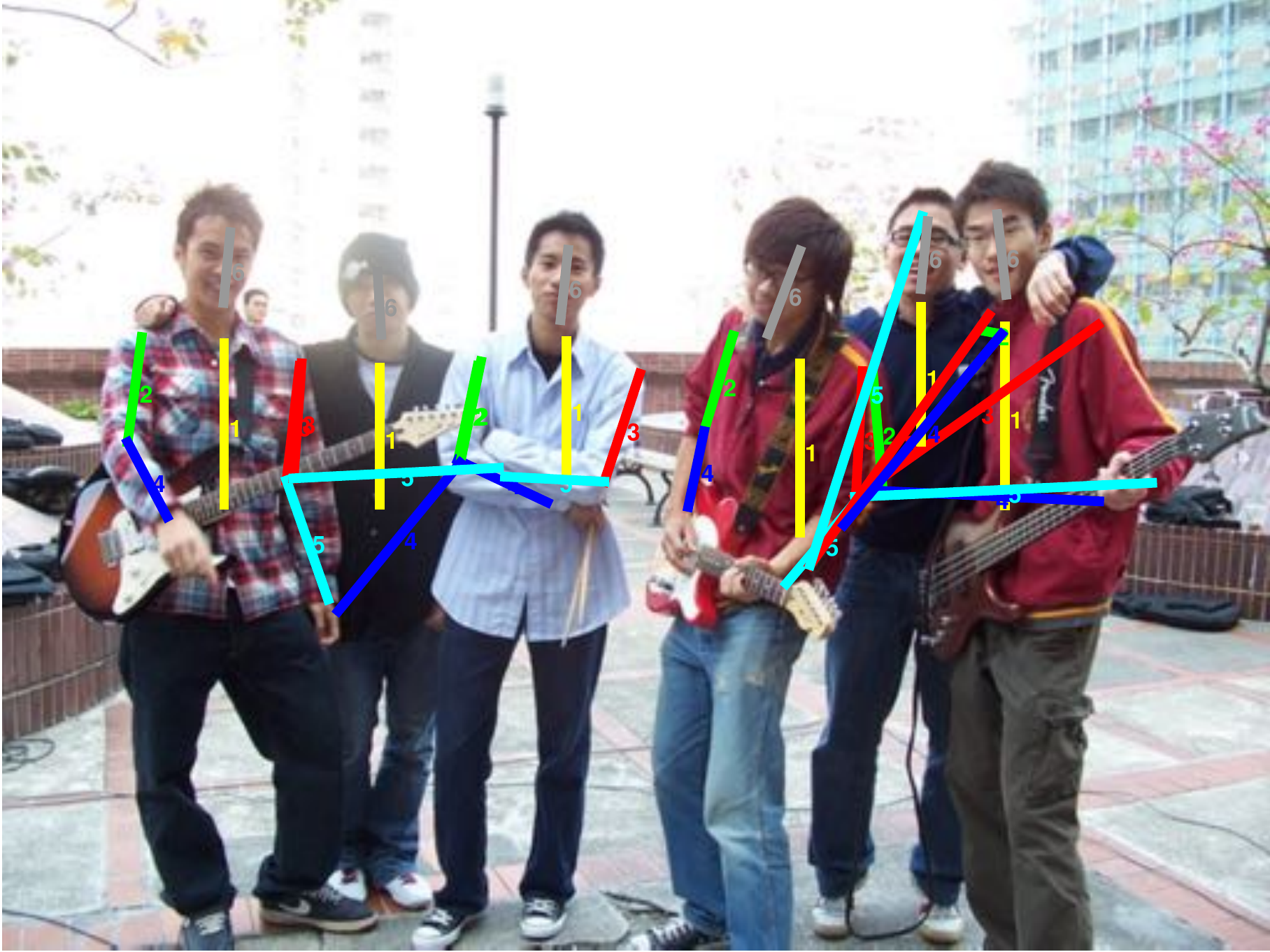}&
    \includegraphics[height=0.145\linewidth]{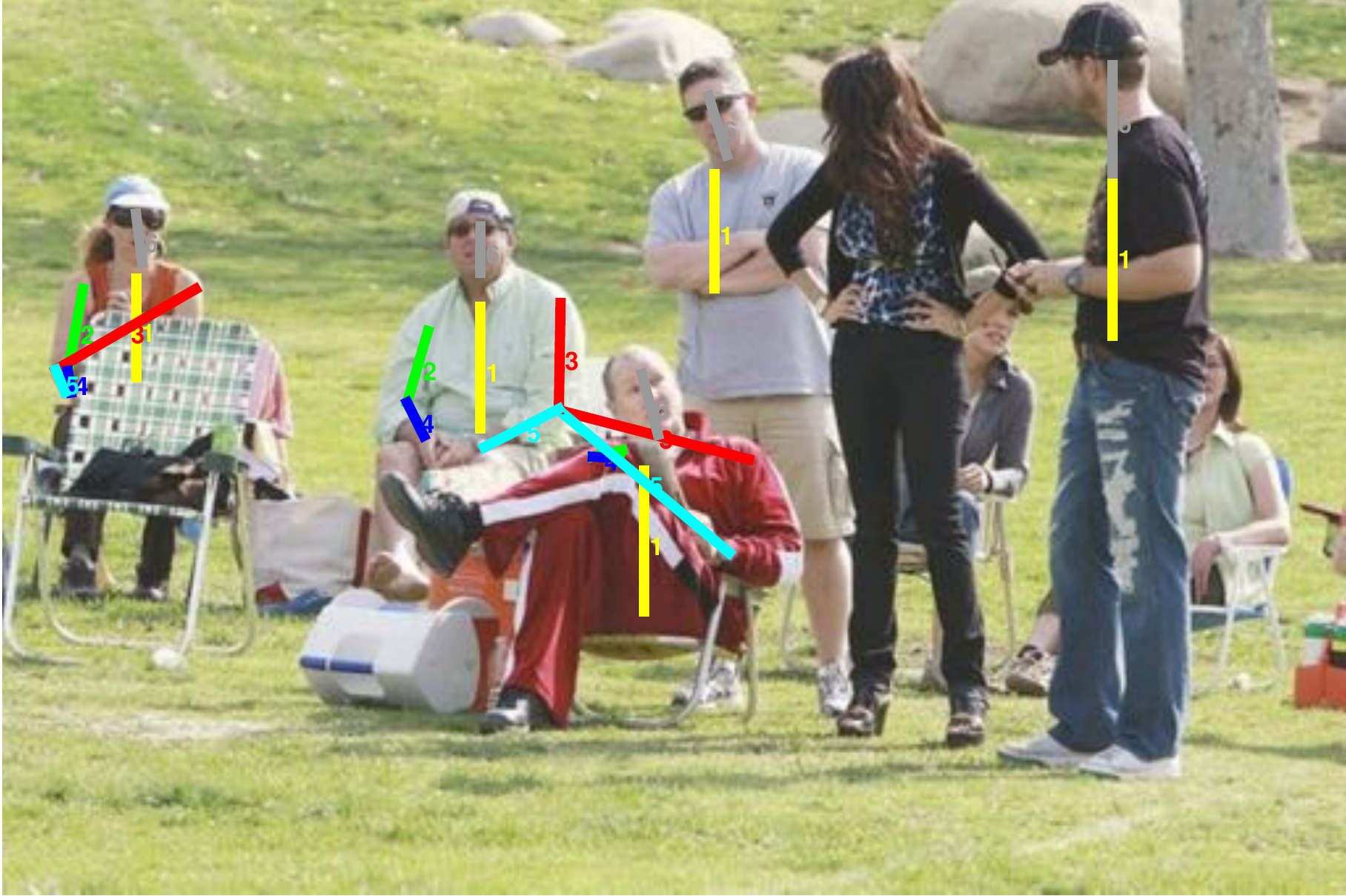}&
    \includegraphics[height=0.145\linewidth]{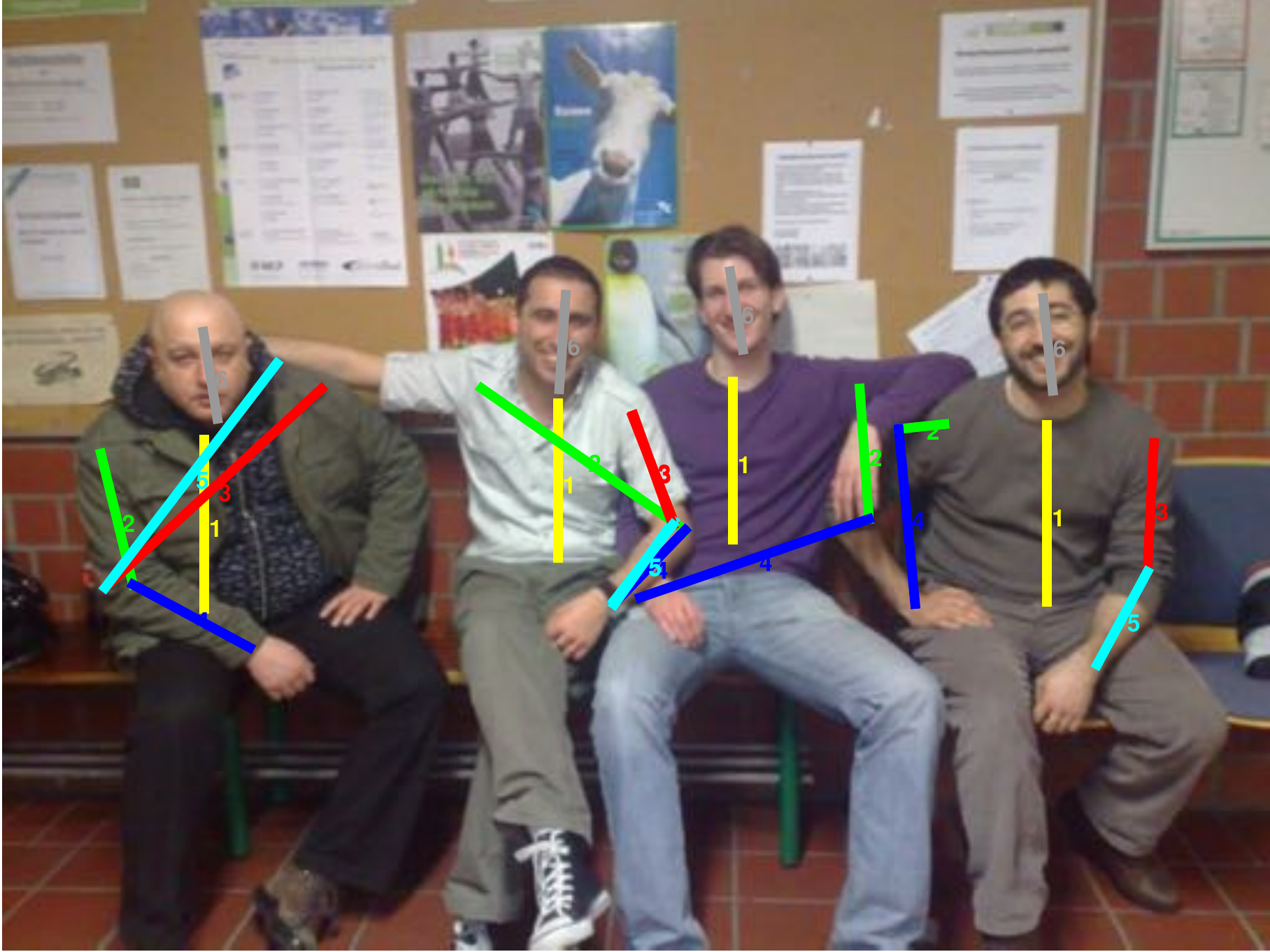}& 
    \includegraphics[height=0.145\linewidth]{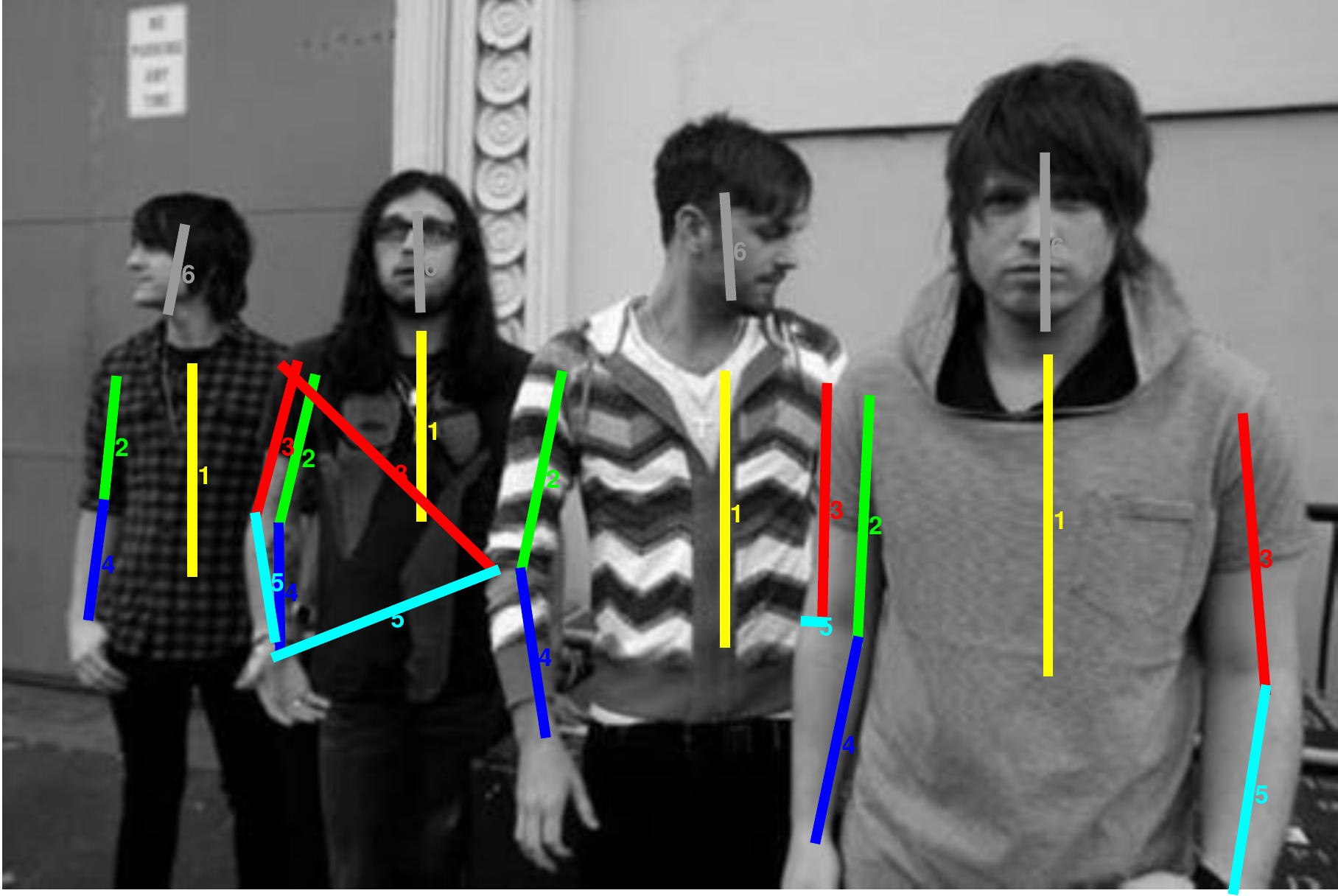}& % 58
    \includegraphics[height=0.145\linewidth]{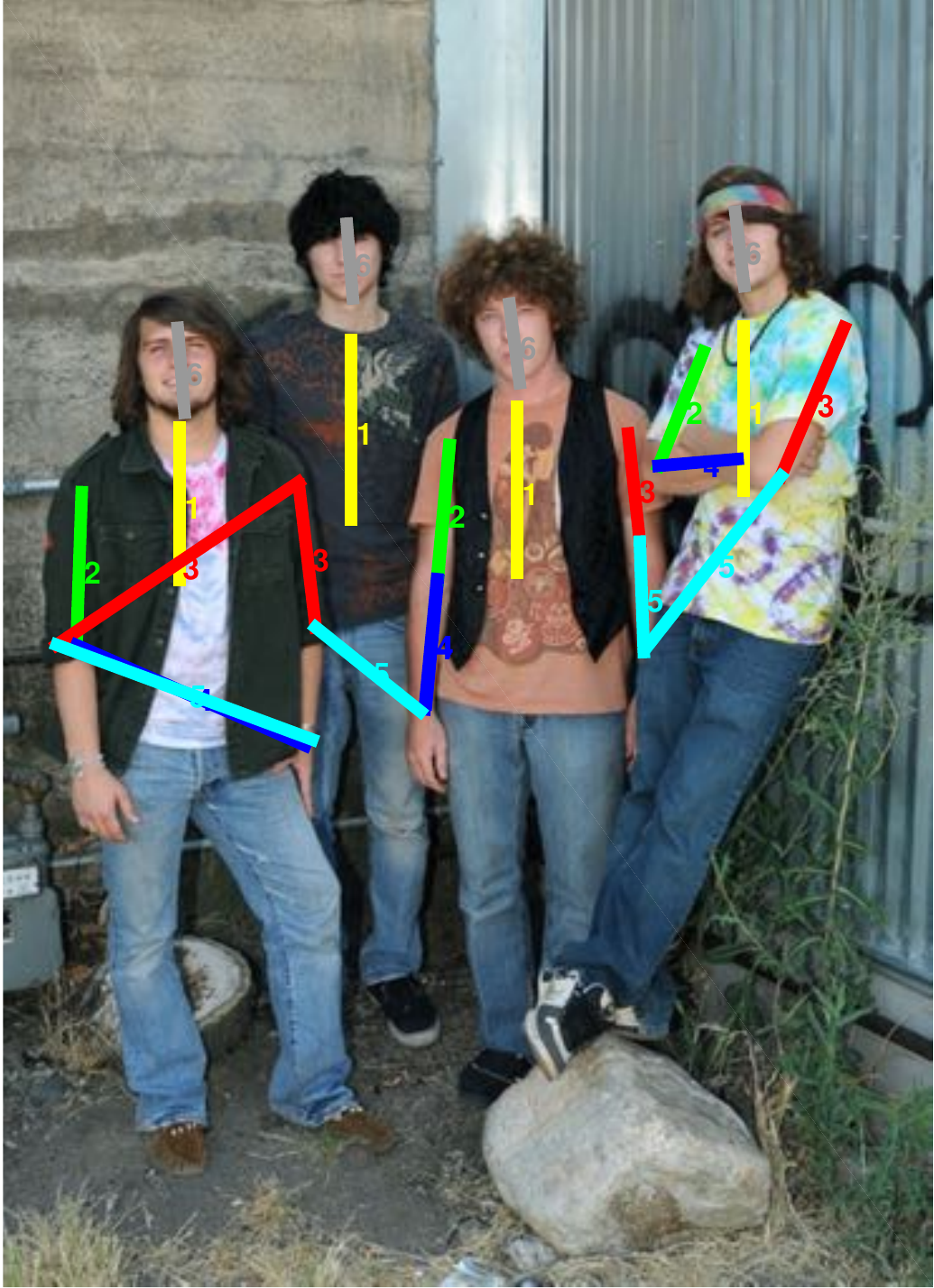}\\ 
    \begin{sideways}\bf \small\quad $\deepcut~\multb$\end{sideways}&
    \includegraphics[height=0.145\linewidth]{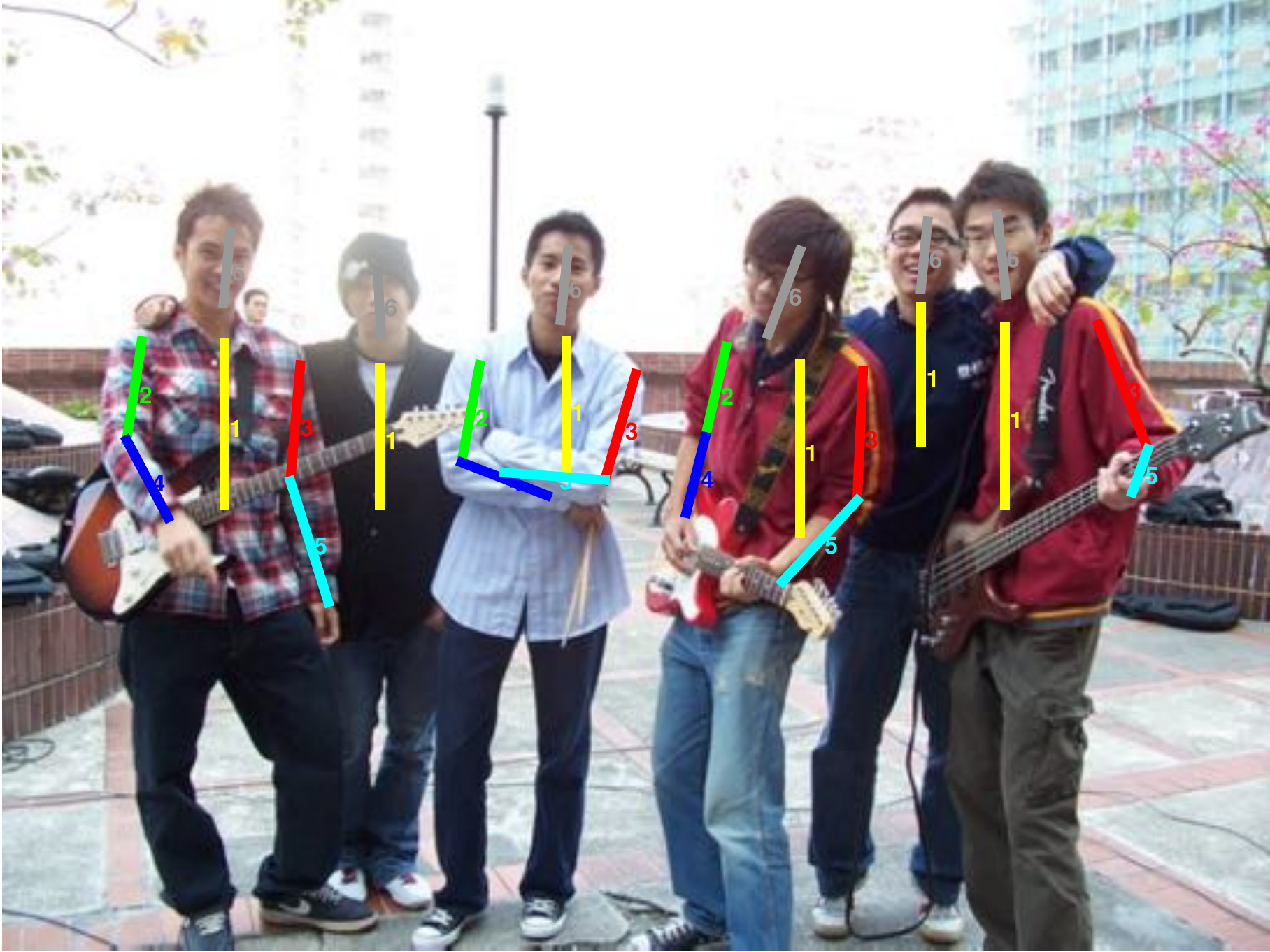}&
    \includegraphics[height=0.145\linewidth]{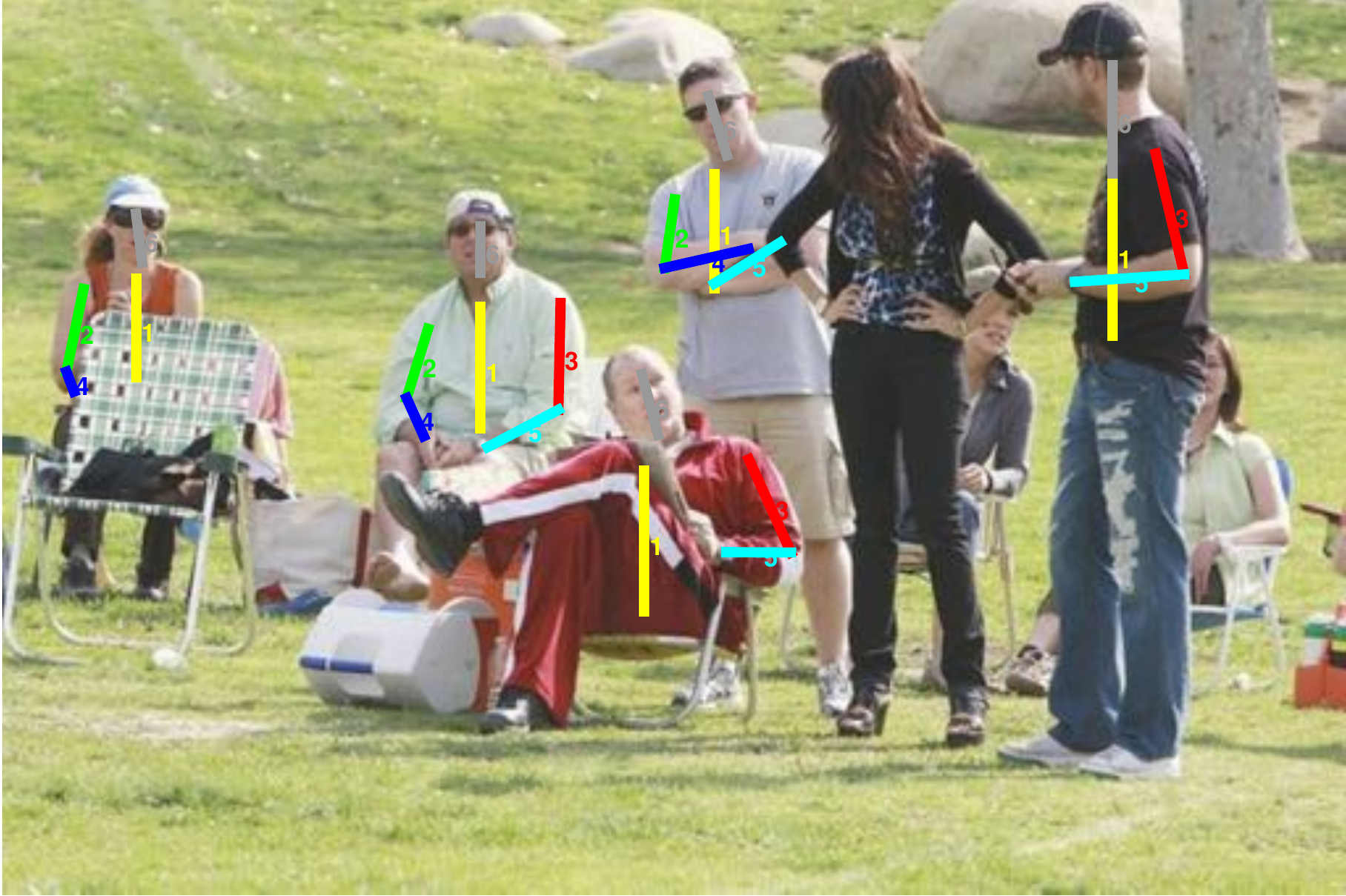}&
    \includegraphics[height=0.145\linewidth]{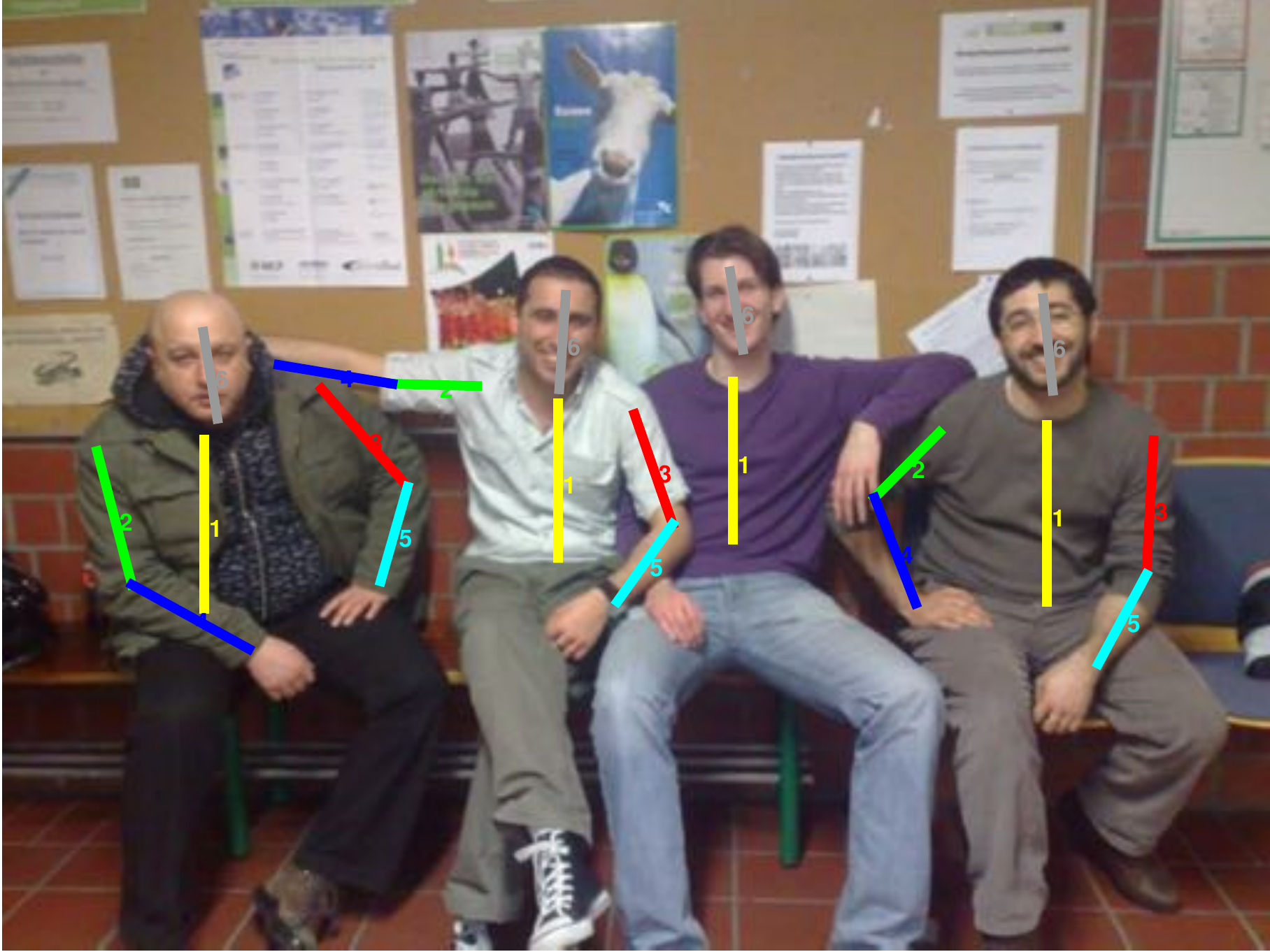}&
    \includegraphics[height=0.145\linewidth]{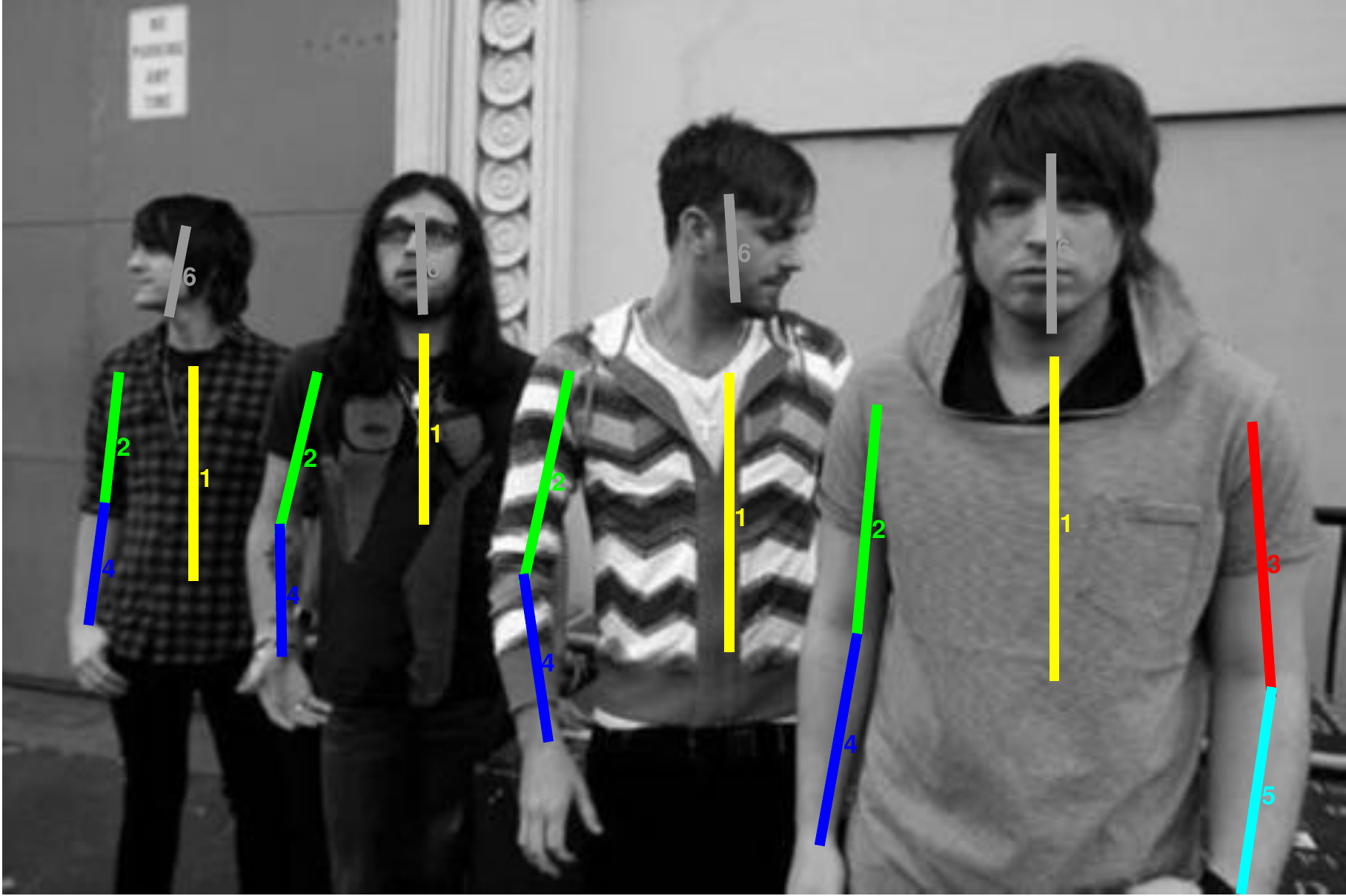}&
    \includegraphics[height=0.145\linewidth]{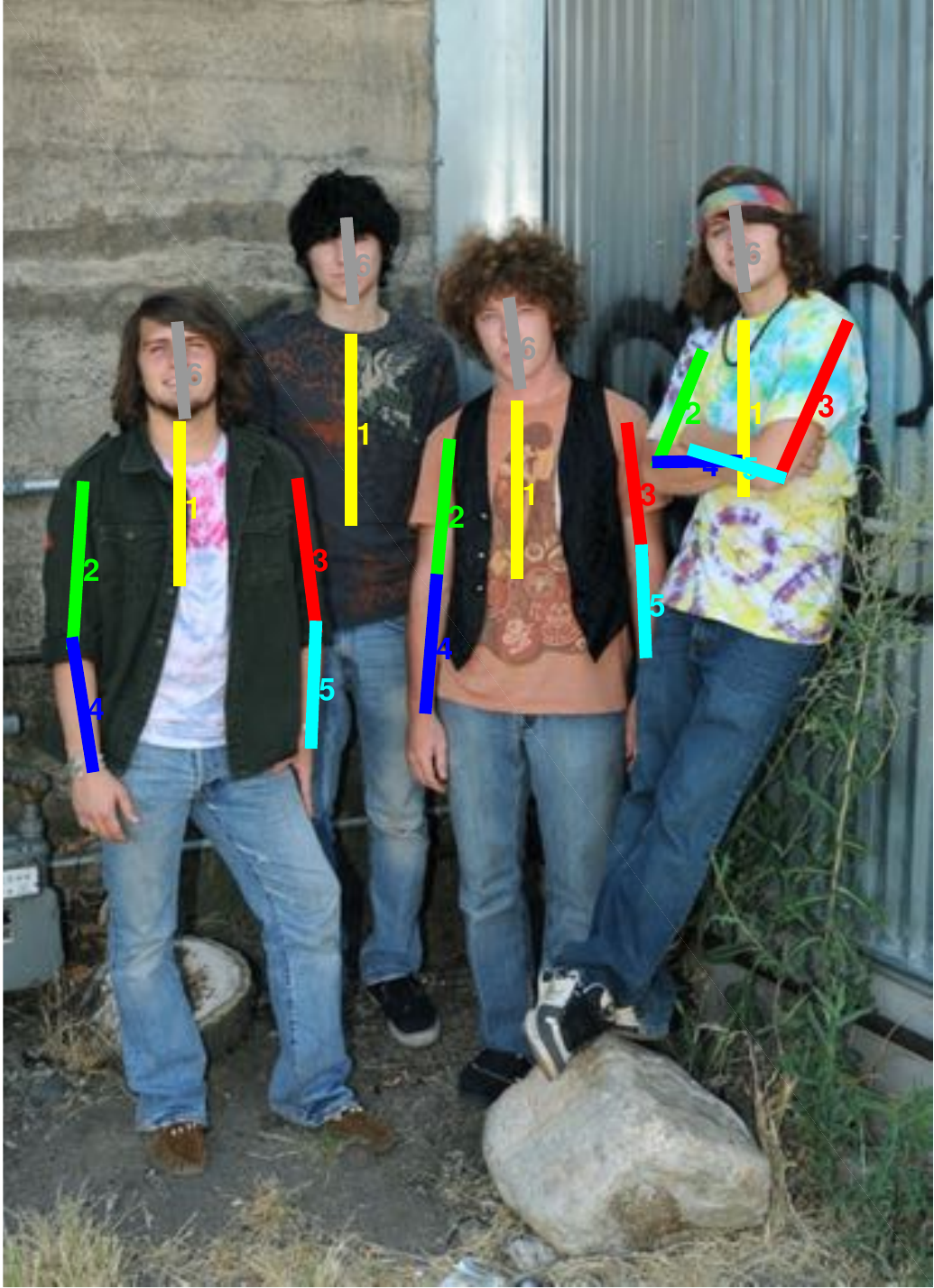}\\
    \begin{sideways}\bf \small Chen\&Yuille~\cite{Chen:2015:POC}\end{sideways}&
    \includegraphics[height=0.145\linewidth]{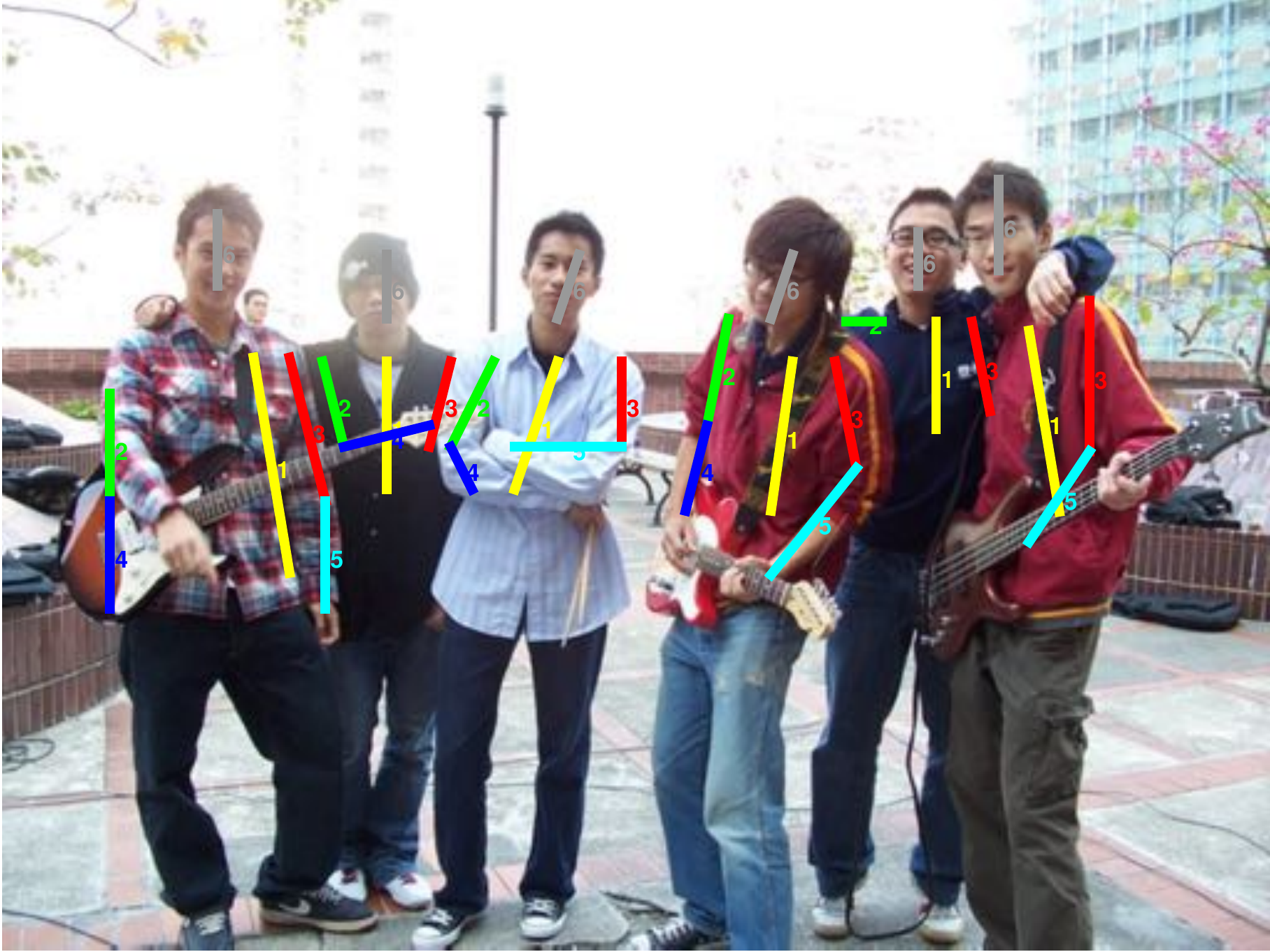}&
    \includegraphics[height=0.145\linewidth]{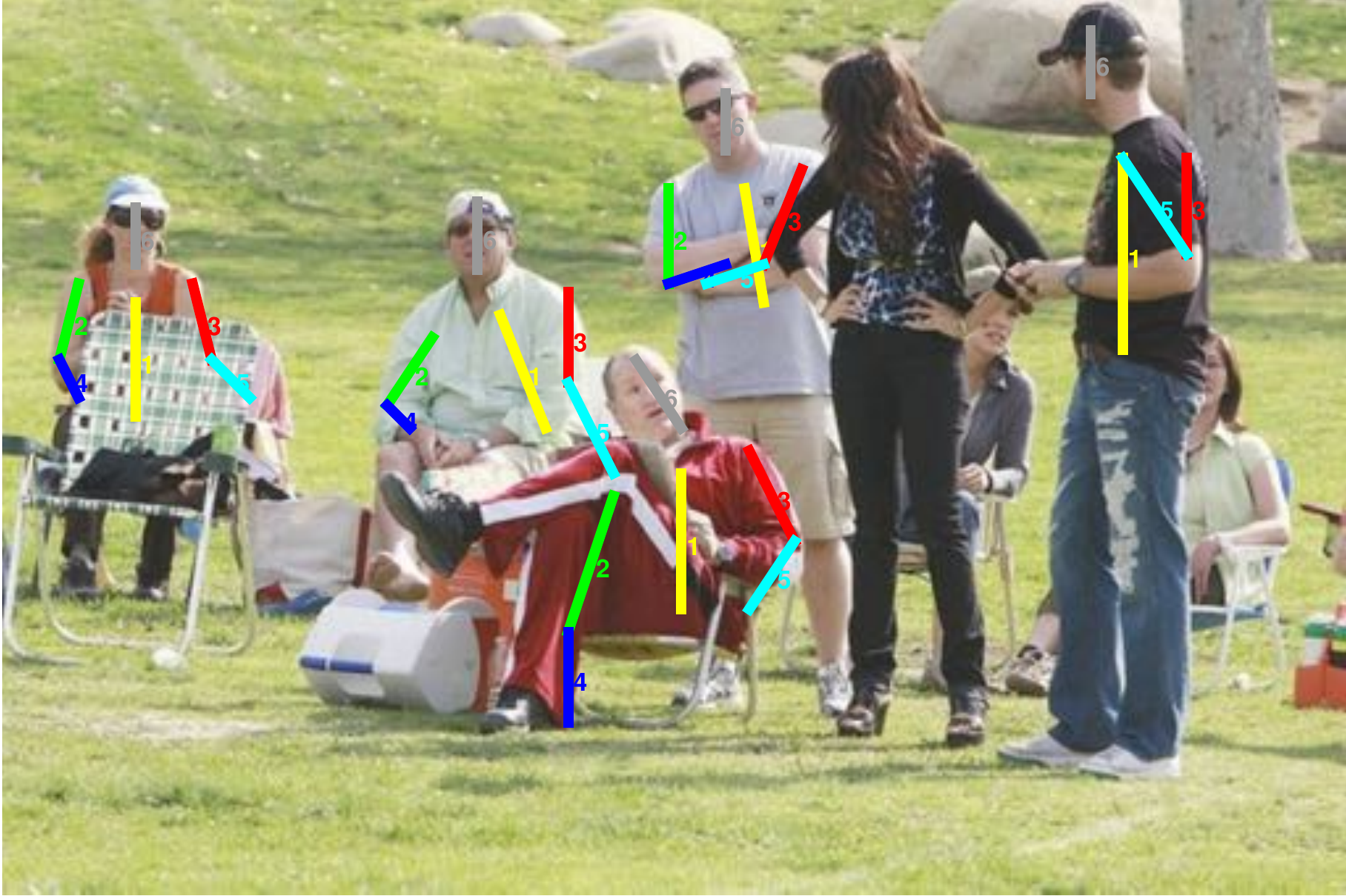}&
    \includegraphics[height=0.145\linewidth]{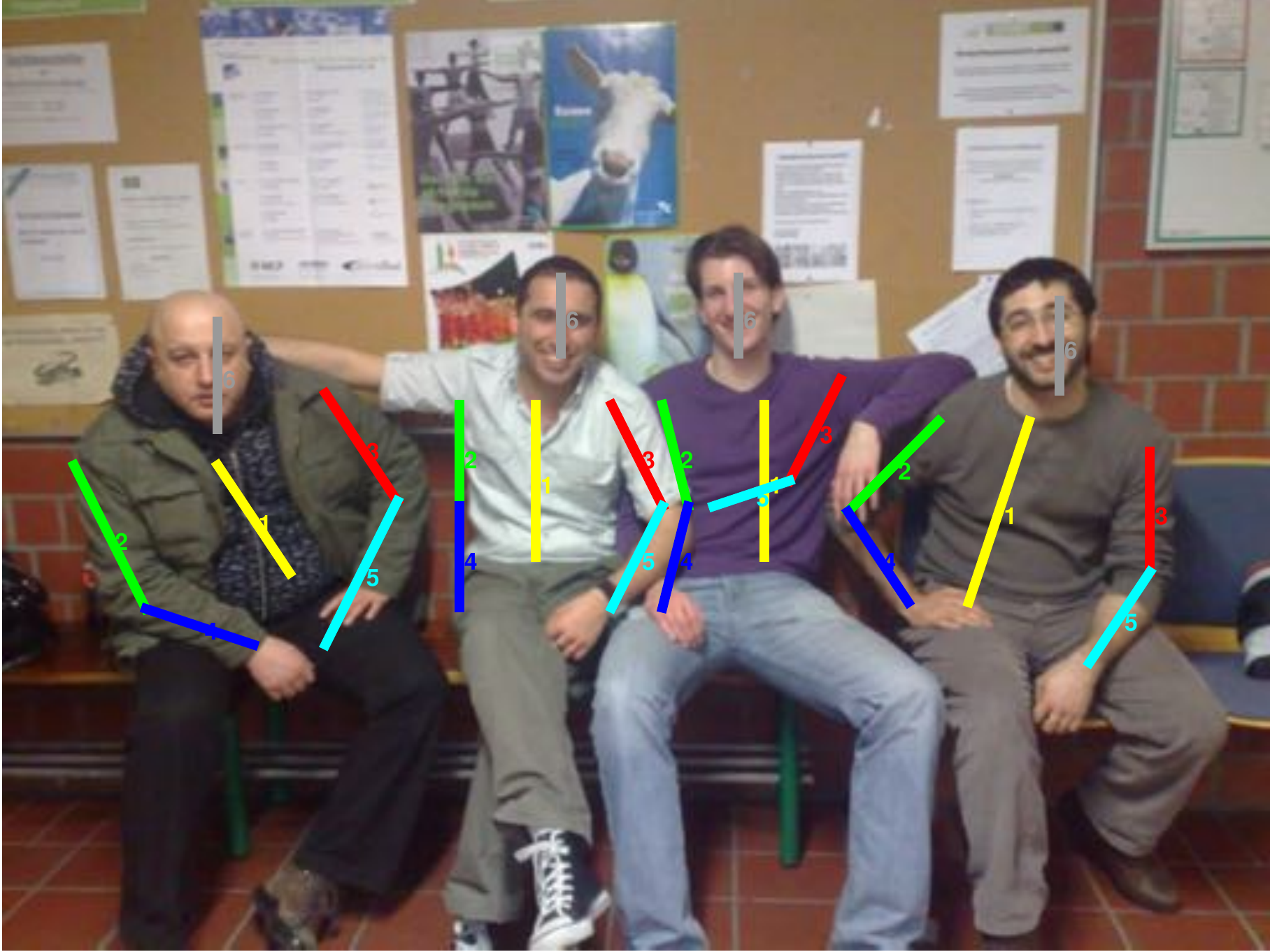}&
    \includegraphics[height=0.145\linewidth]{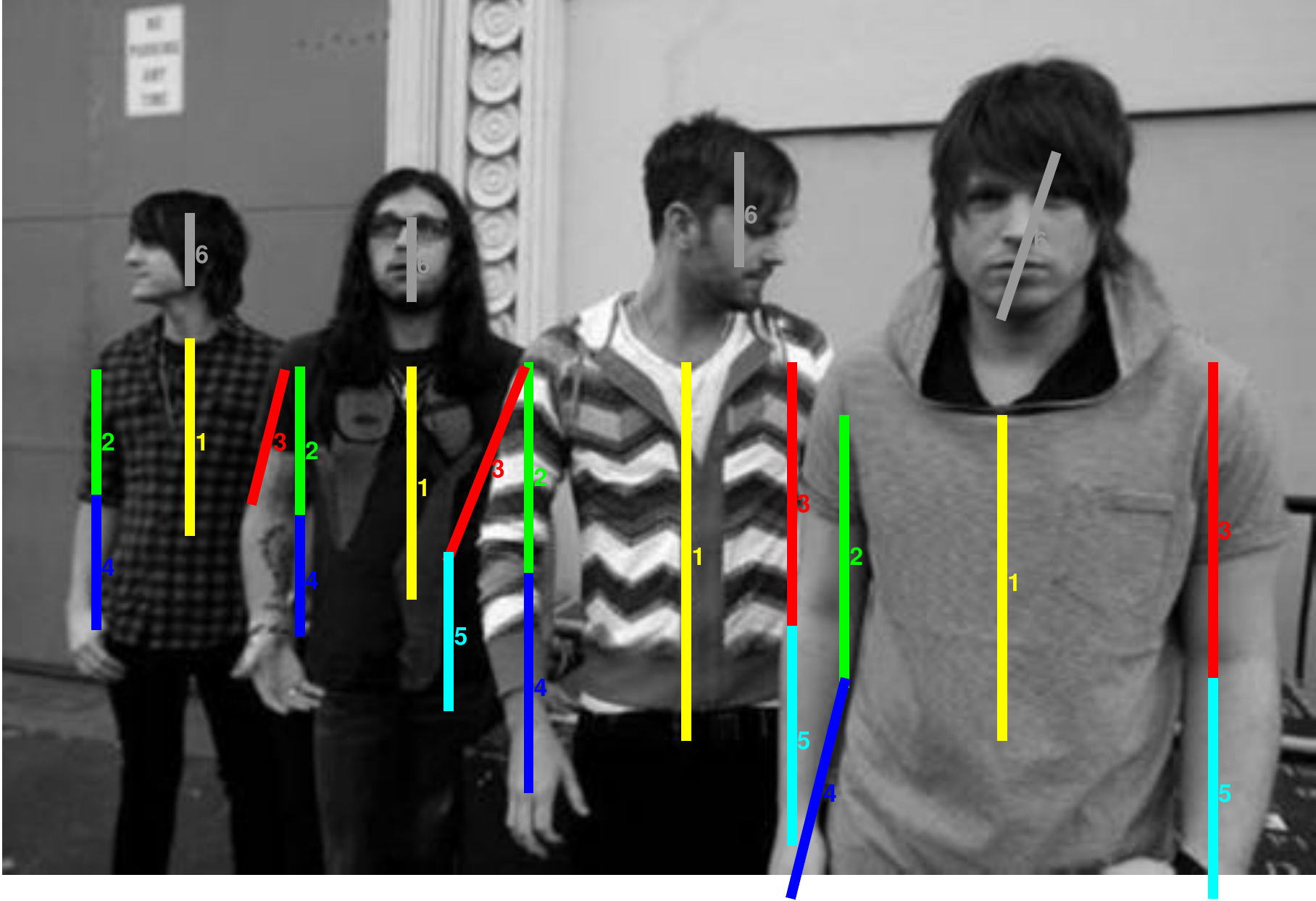}&
    \includegraphics[height=0.145\linewidth]{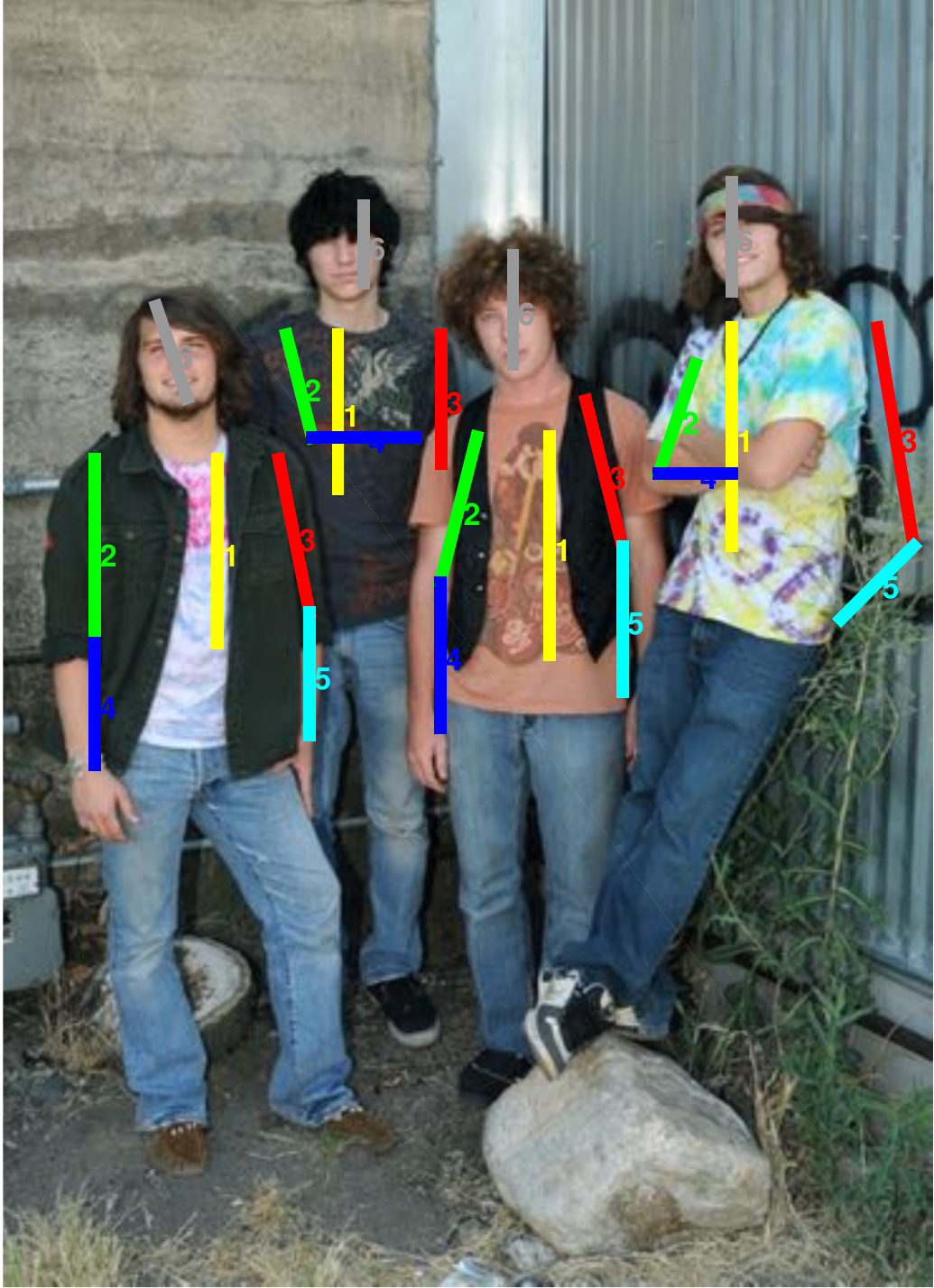}\\
    &1&2&3&4&5\\
 %   &&&&&\\
  \end{tabular}

  \begin{tabular}{c c c c c c c}
    \begin{sideways}\bf \small\quad\quad $\detroi$\end{sideways}&
    \includegraphics[height=0.145\linewidth]{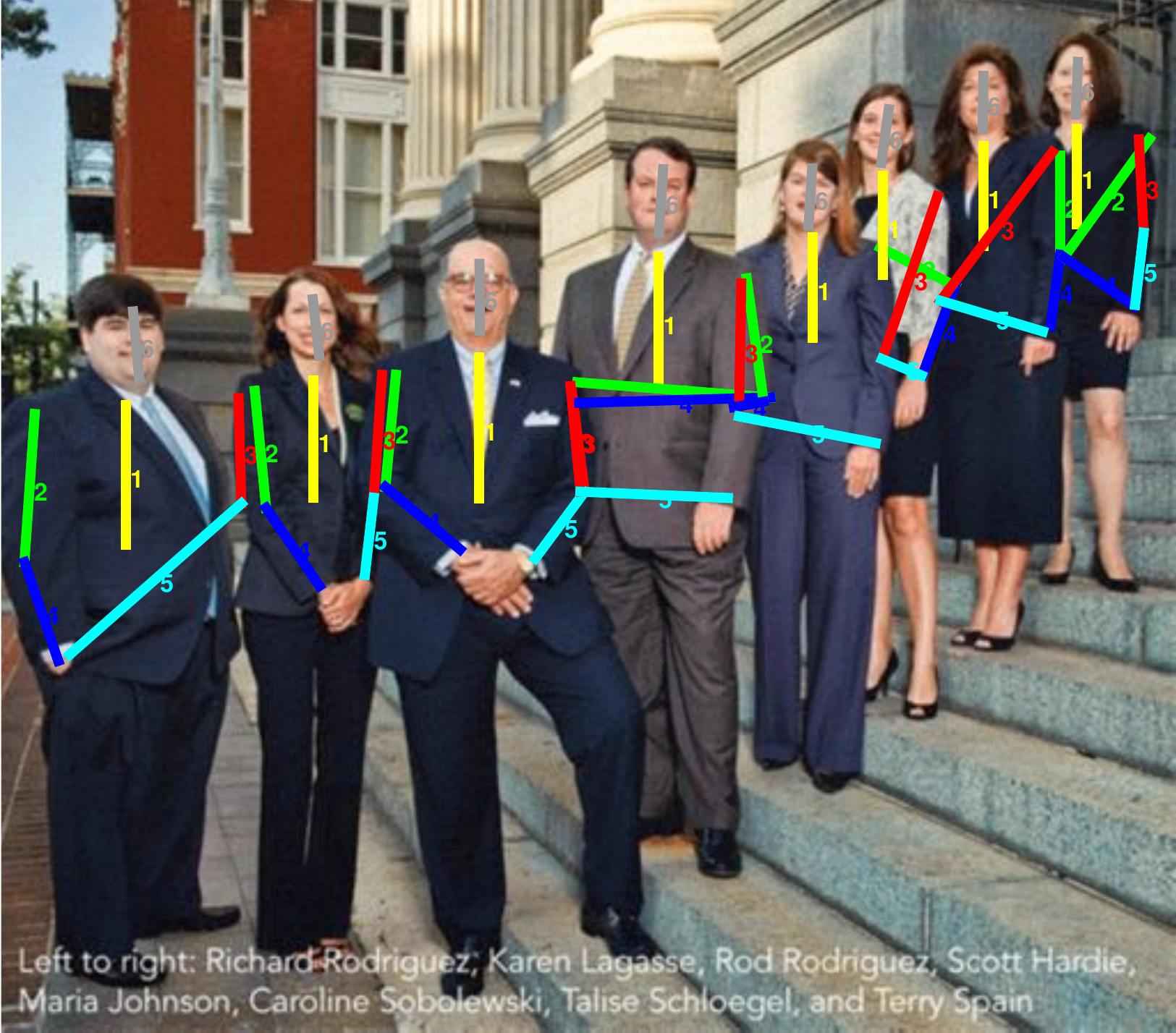}&
    \includegraphics[height=0.145\linewidth]{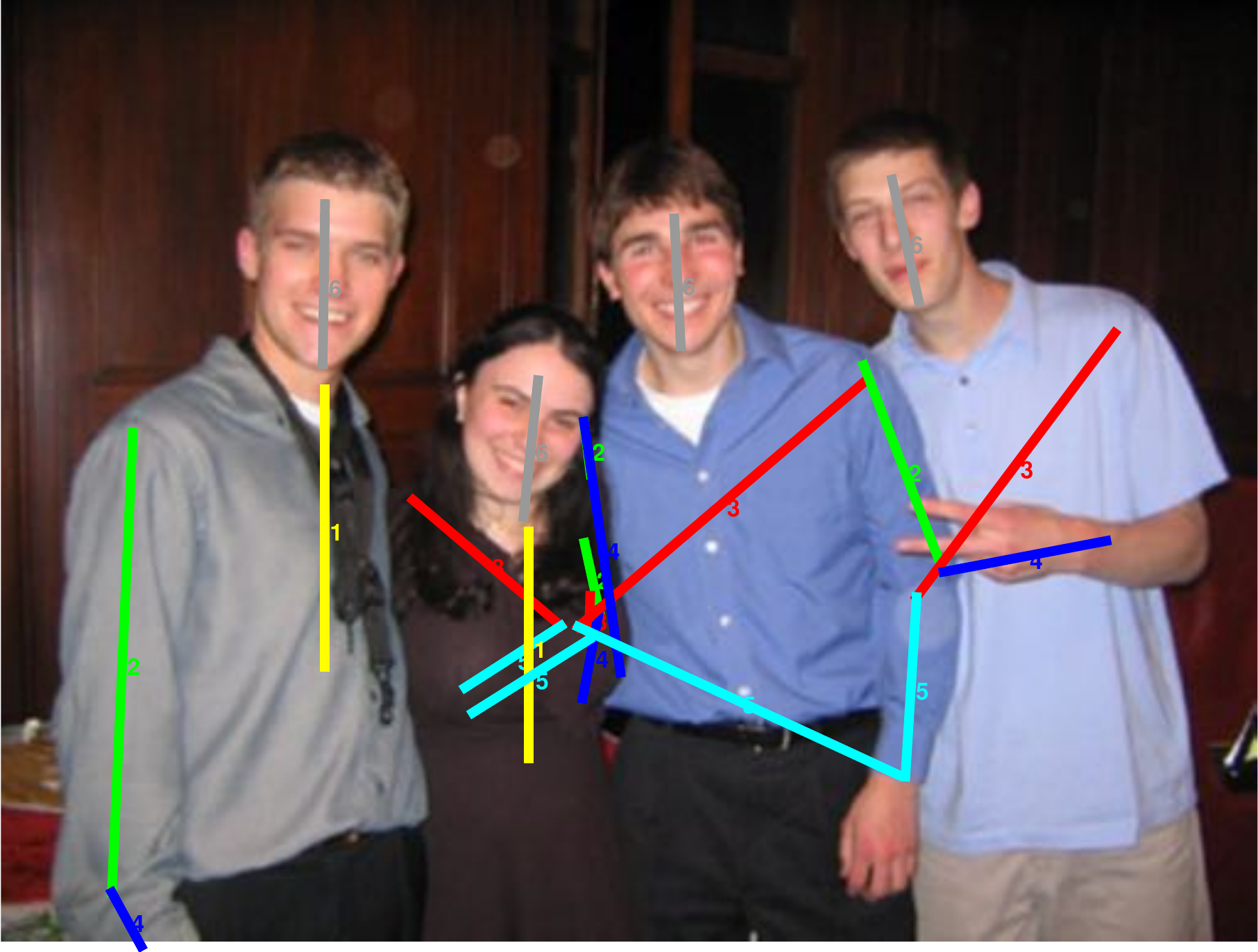}& 
    \includegraphics[height=0.145\linewidth]{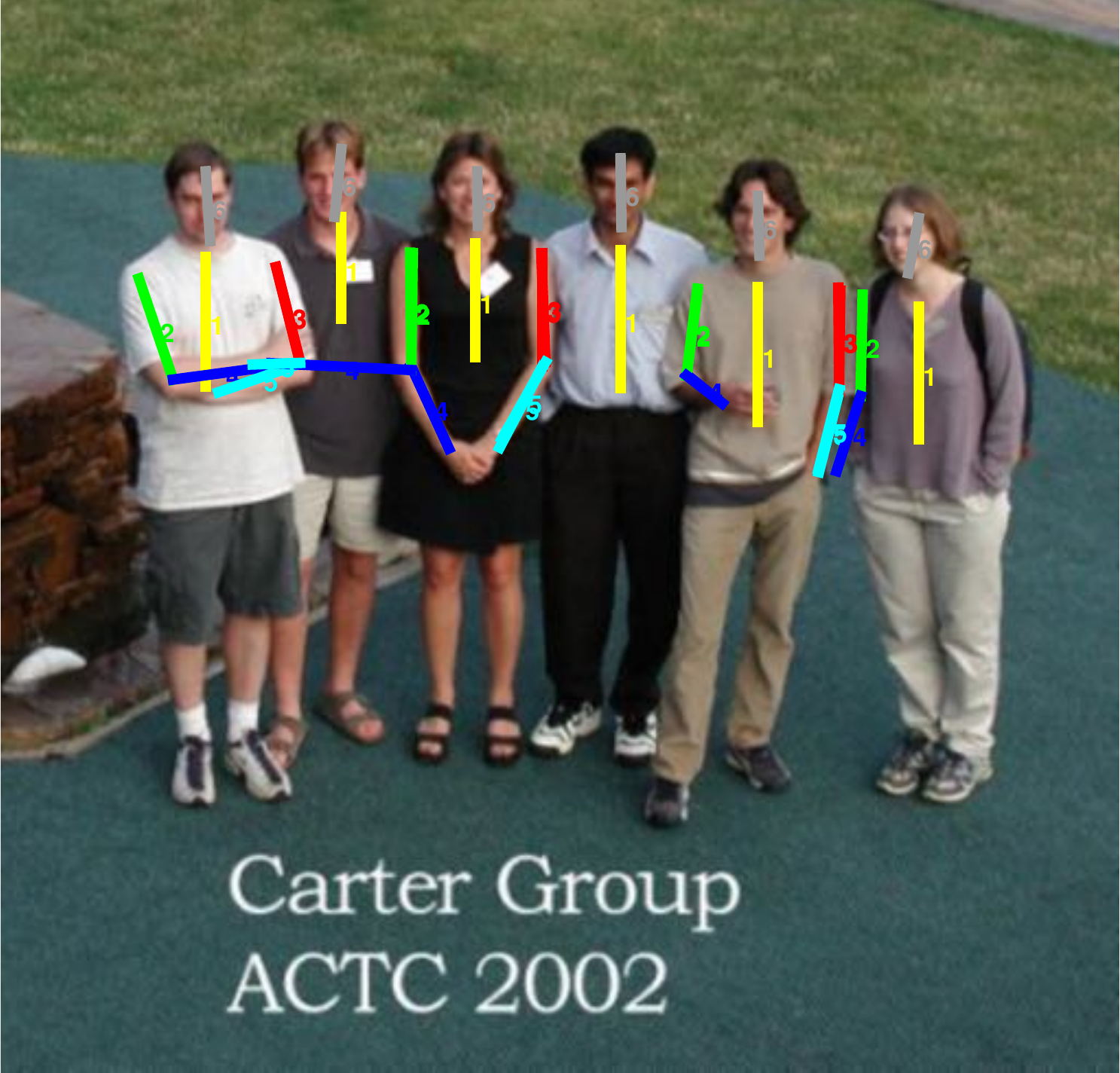}& % 56
    \includegraphics[height=0.145\linewidth]{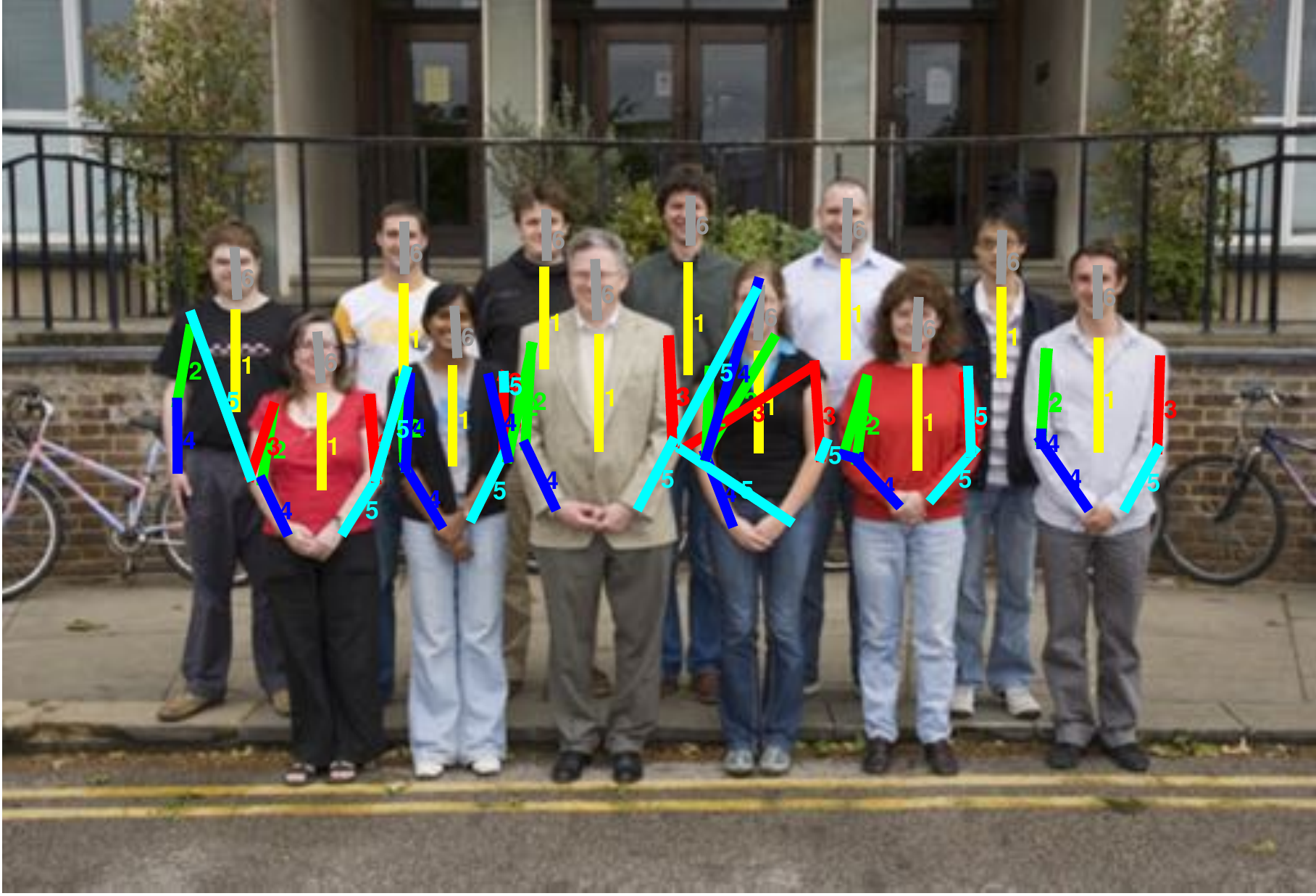}& % 74
    \includegraphics[height=0.145\linewidth]{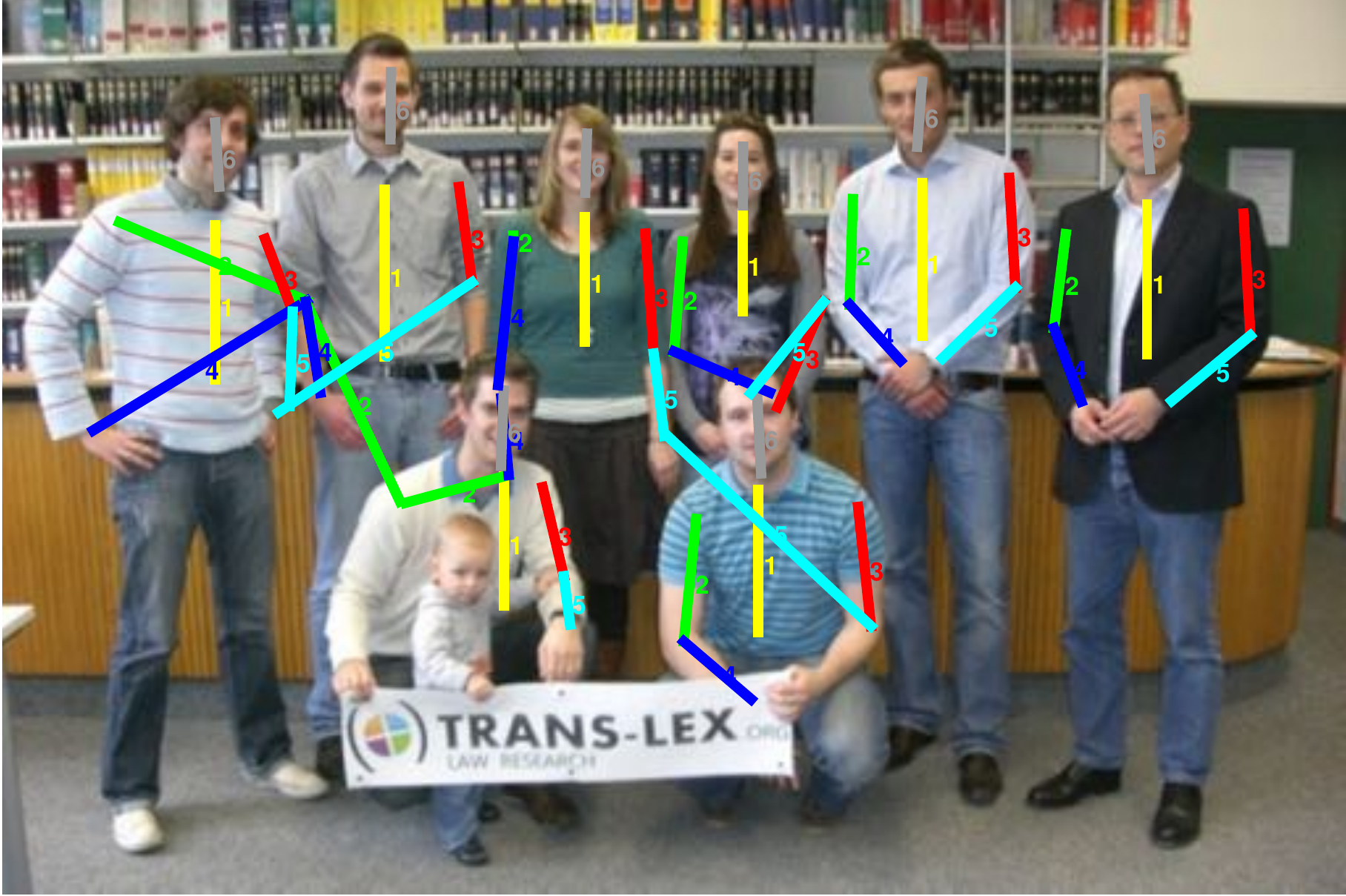}\\
    \begin{sideways}\bf \small\quad $\deepcut~\multb$\end{sideways}&
    \includegraphics[height=0.145\linewidth]{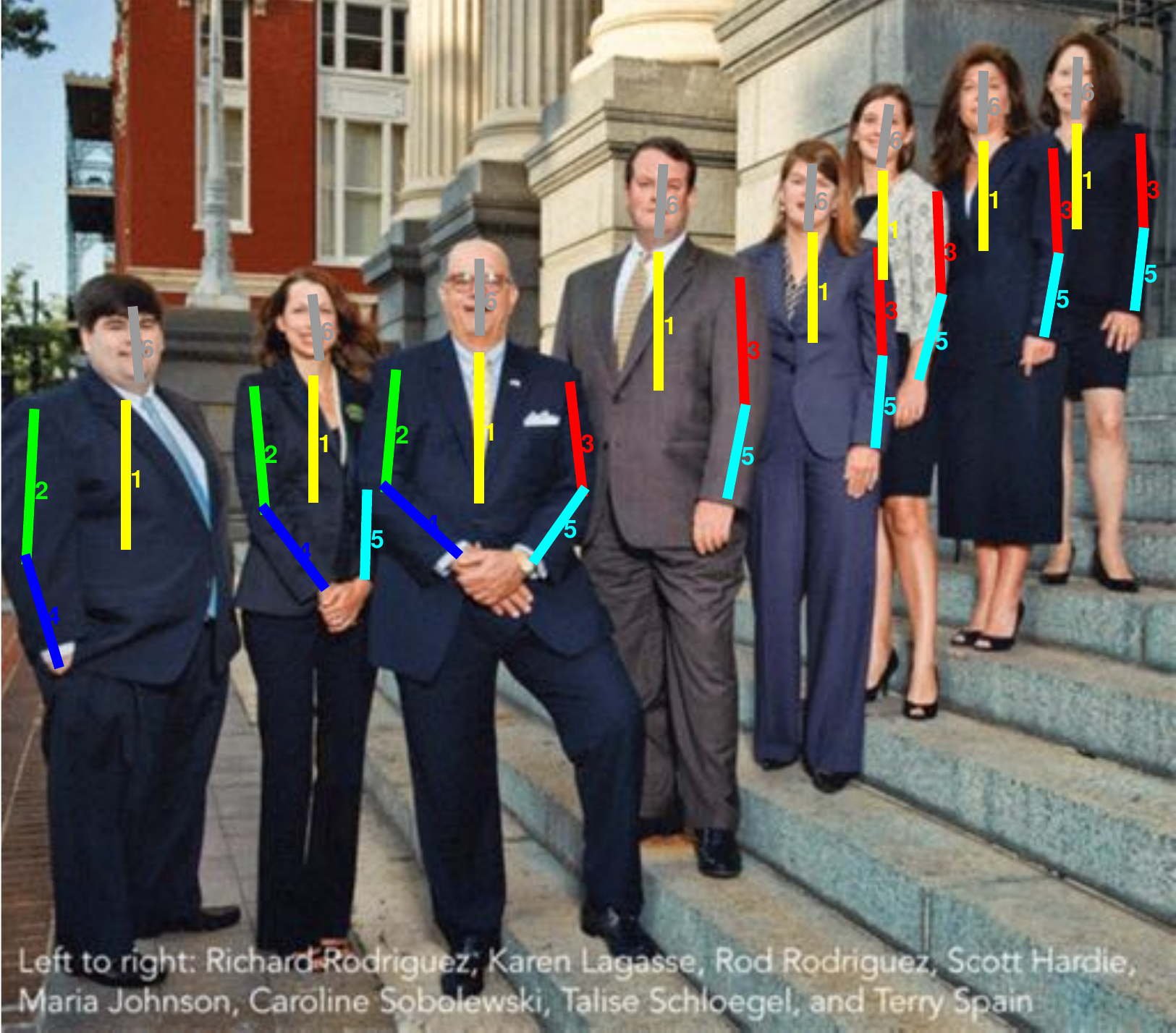}&
    \includegraphics[height=0.145\linewidth]{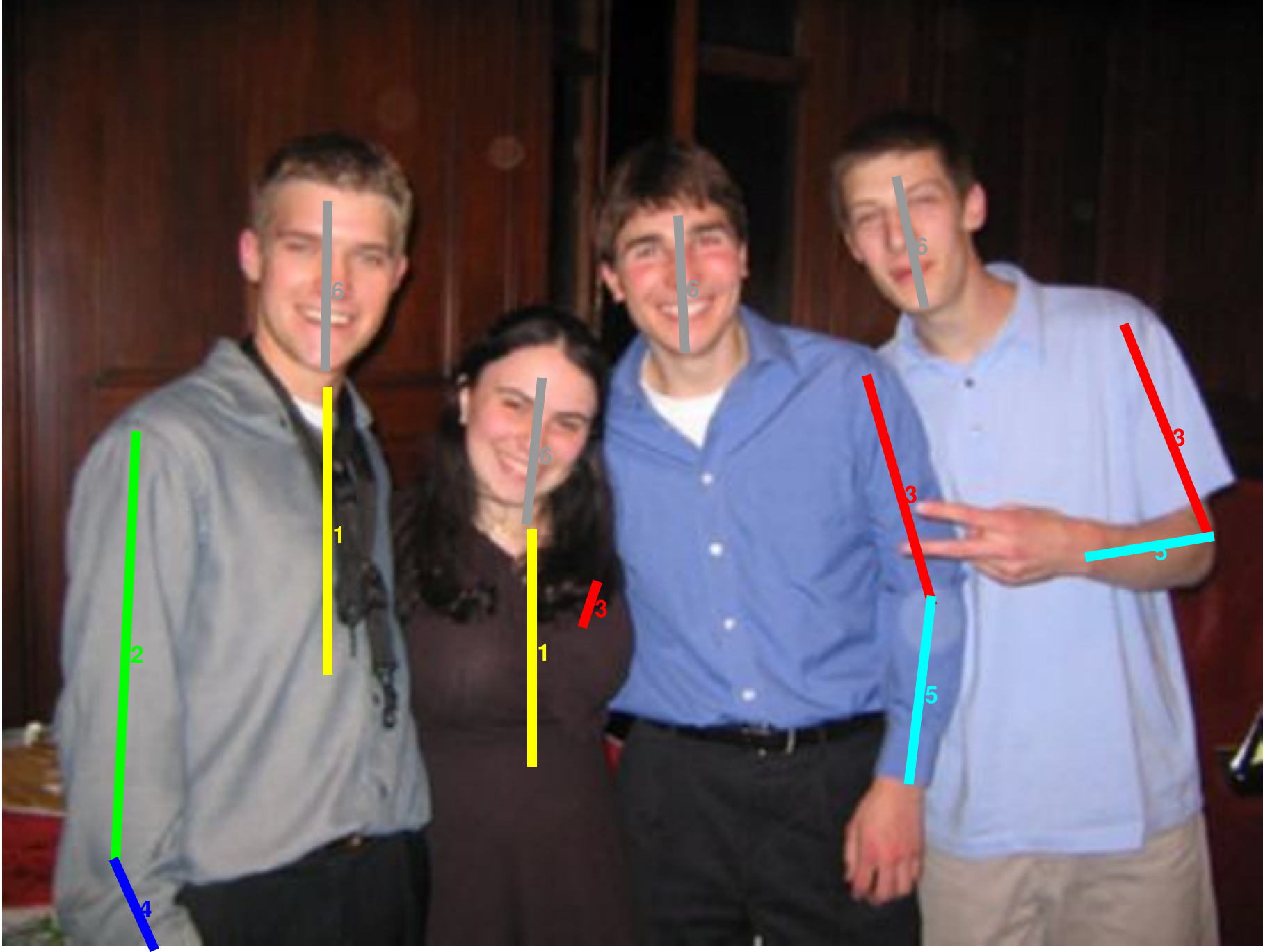}&
    \includegraphics[height=0.145\linewidth]{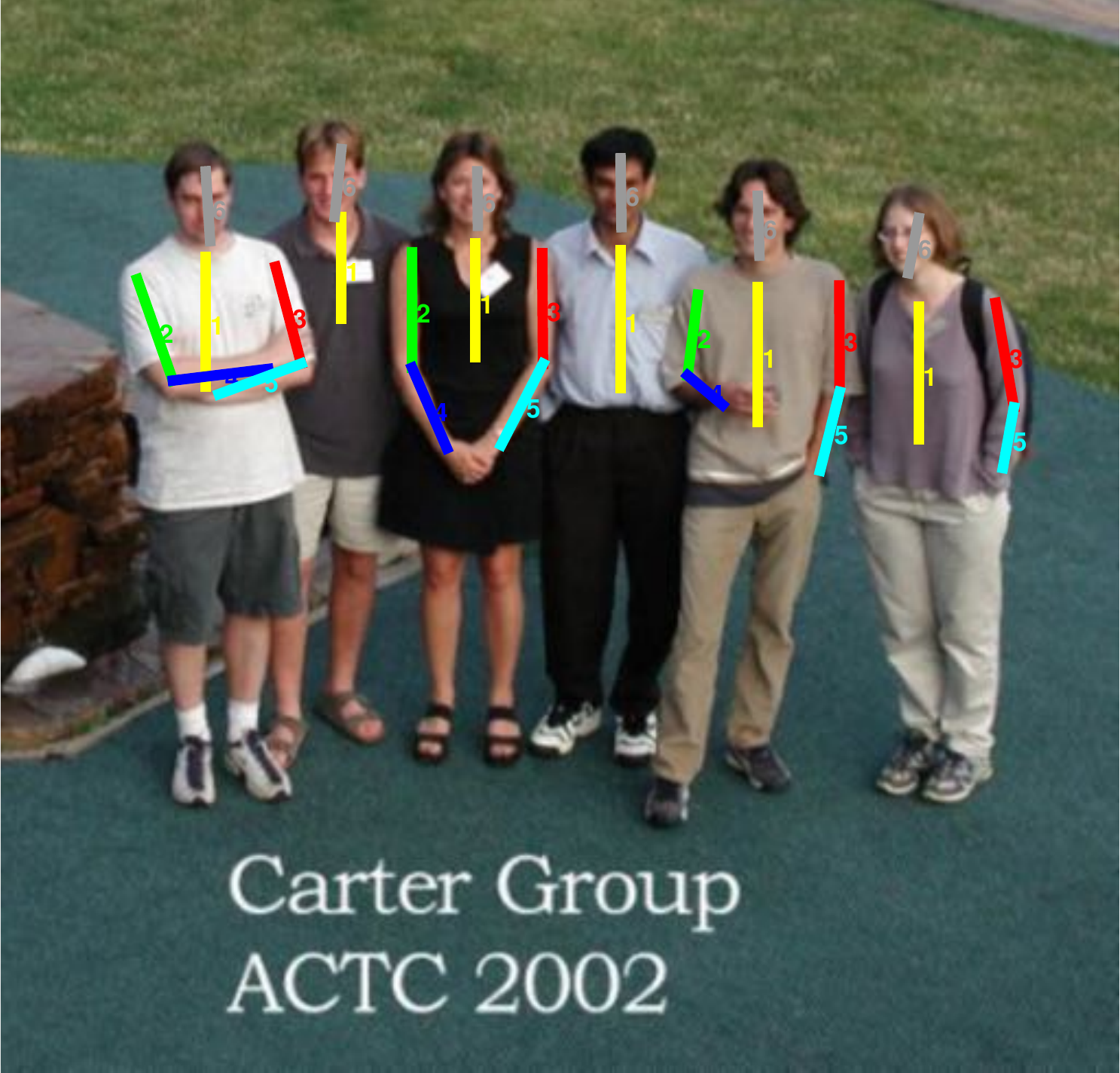}&
    \includegraphics[height=0.145\linewidth]{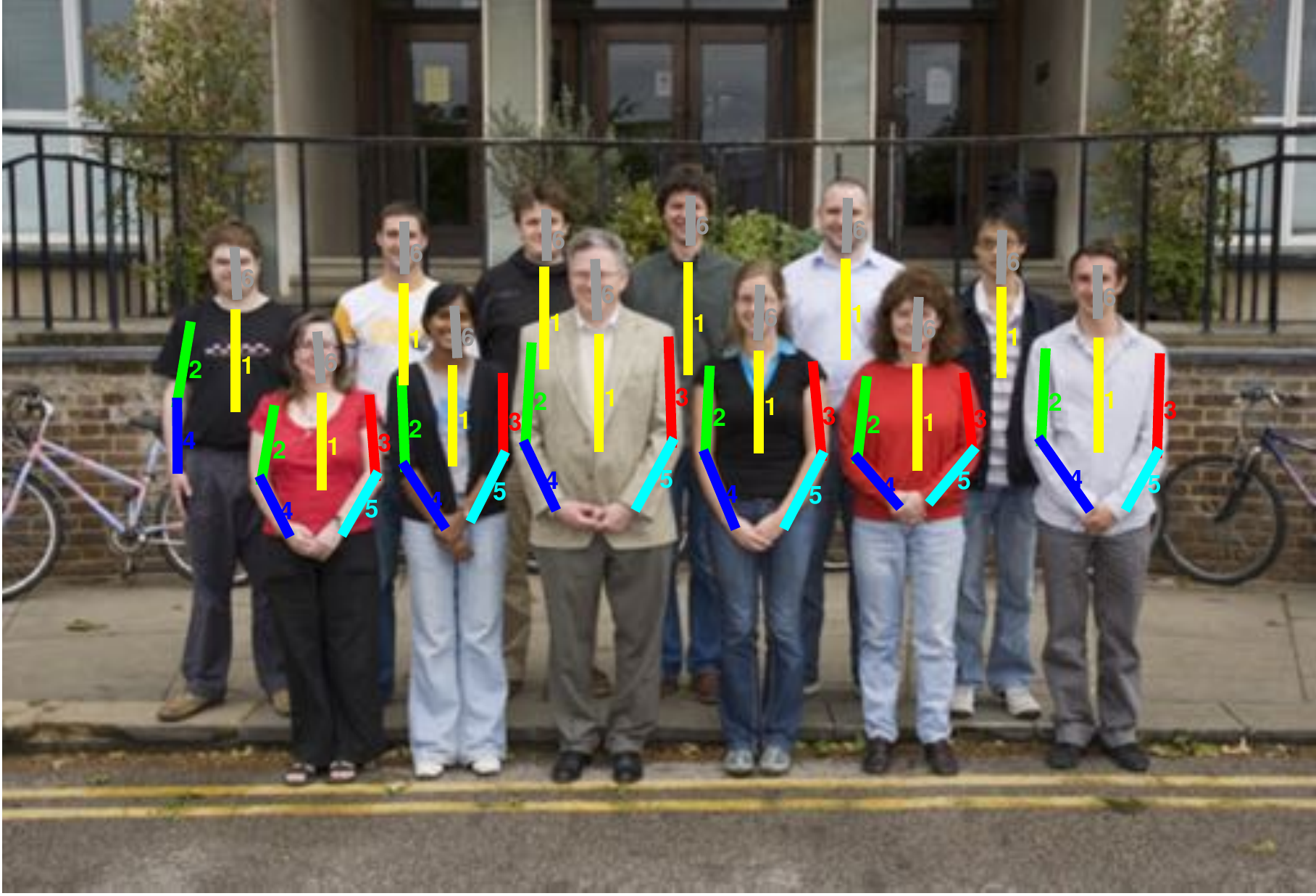}&
    \includegraphics[height=0.145\linewidth]{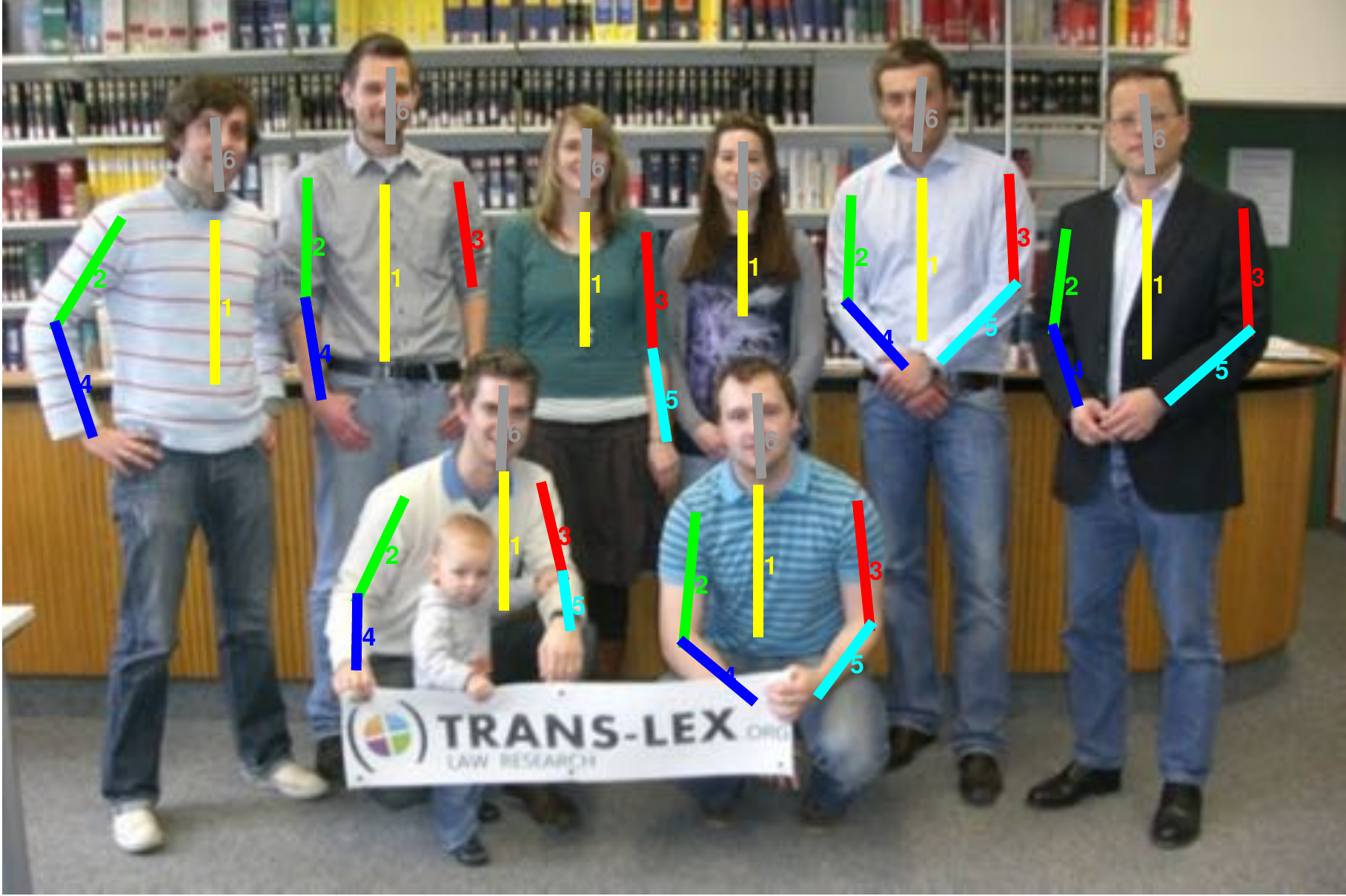}\\
    \begin{sideways}\bf \small Chen\&Yuille~\cite{Chen:2015:POC}\end{sideways}&
    \includegraphics[height=0.145\linewidth]{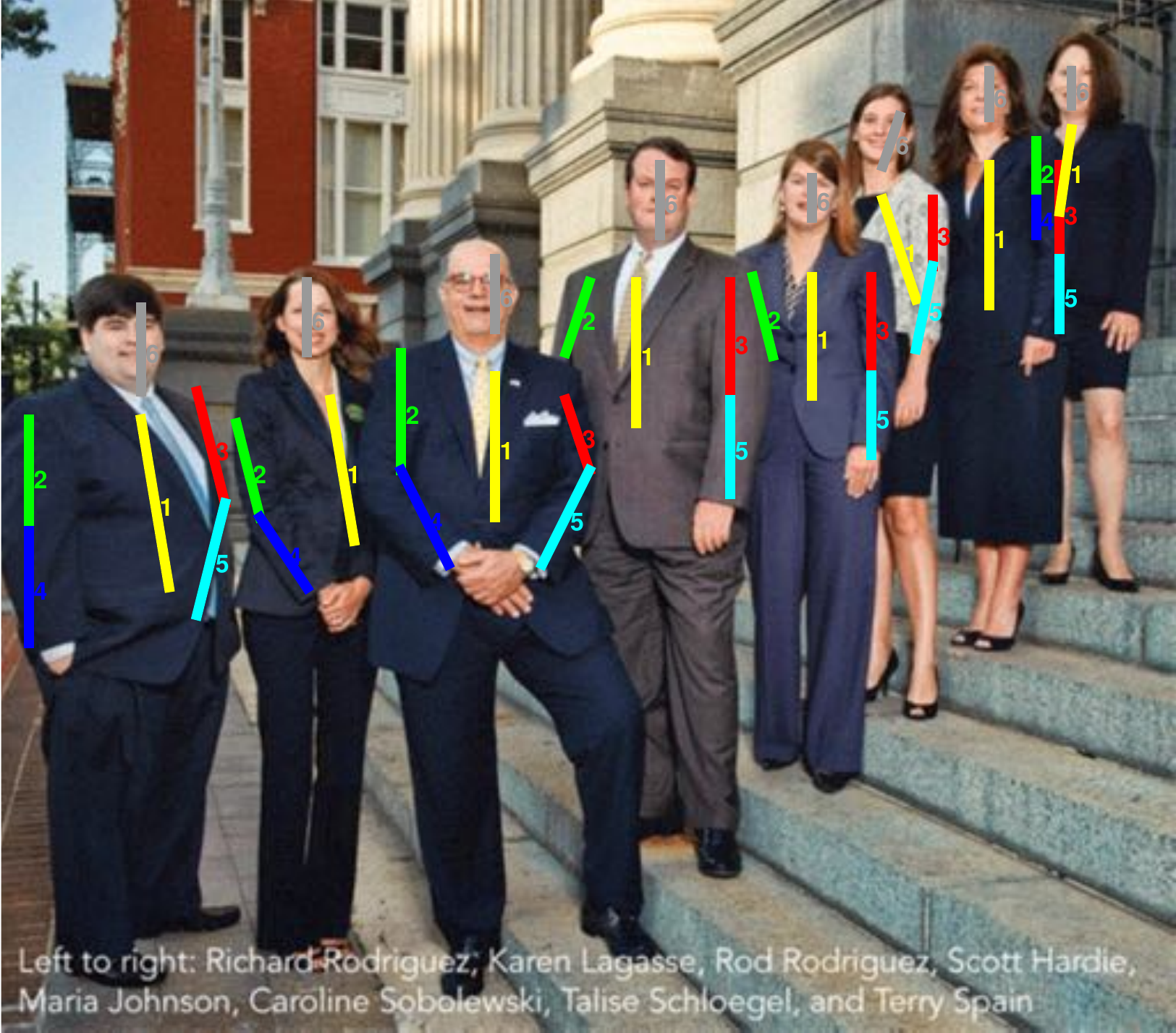}&
    \includegraphics[height=0.145\linewidth]{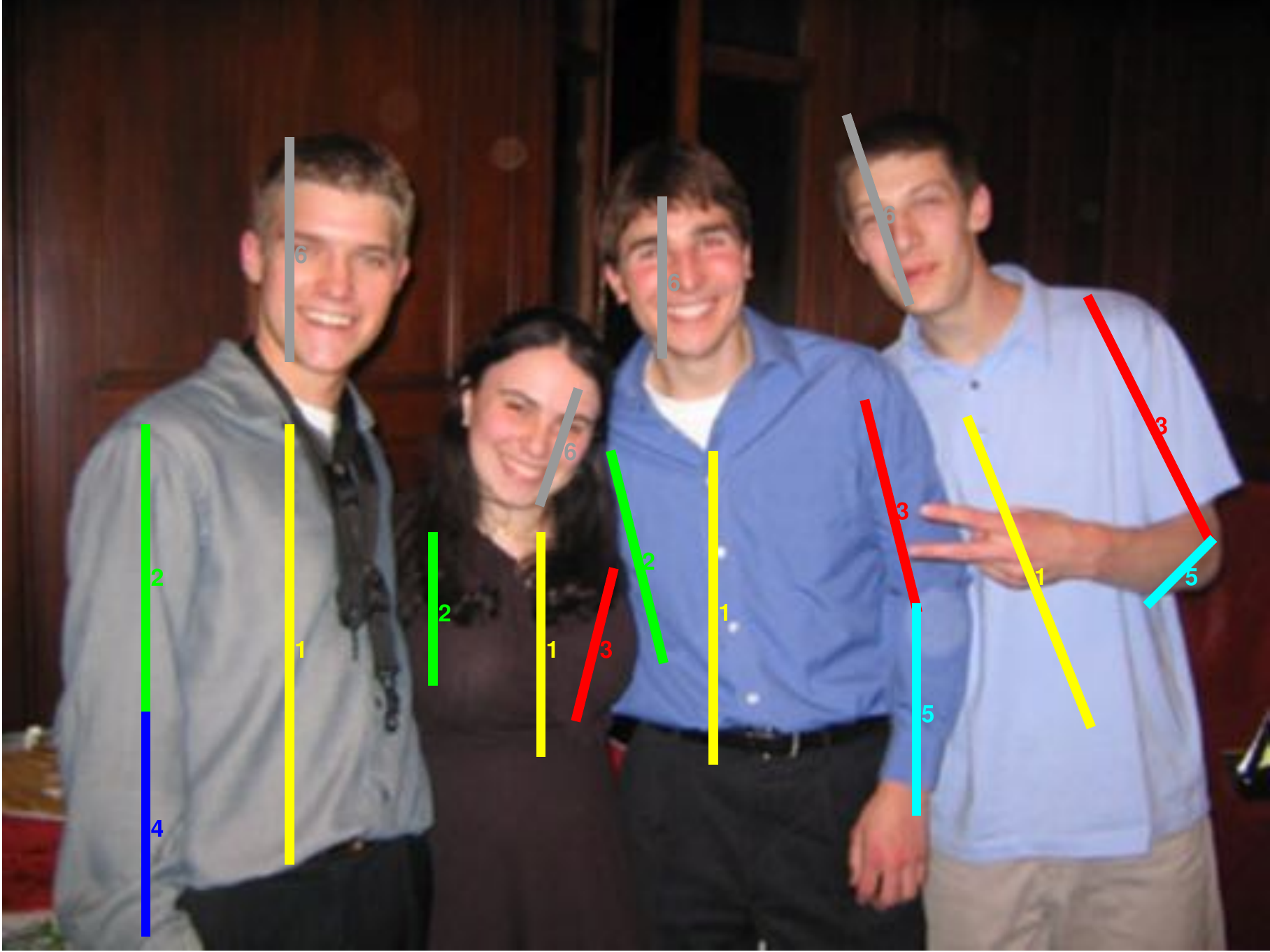}&
    \includegraphics[height=0.145\linewidth]{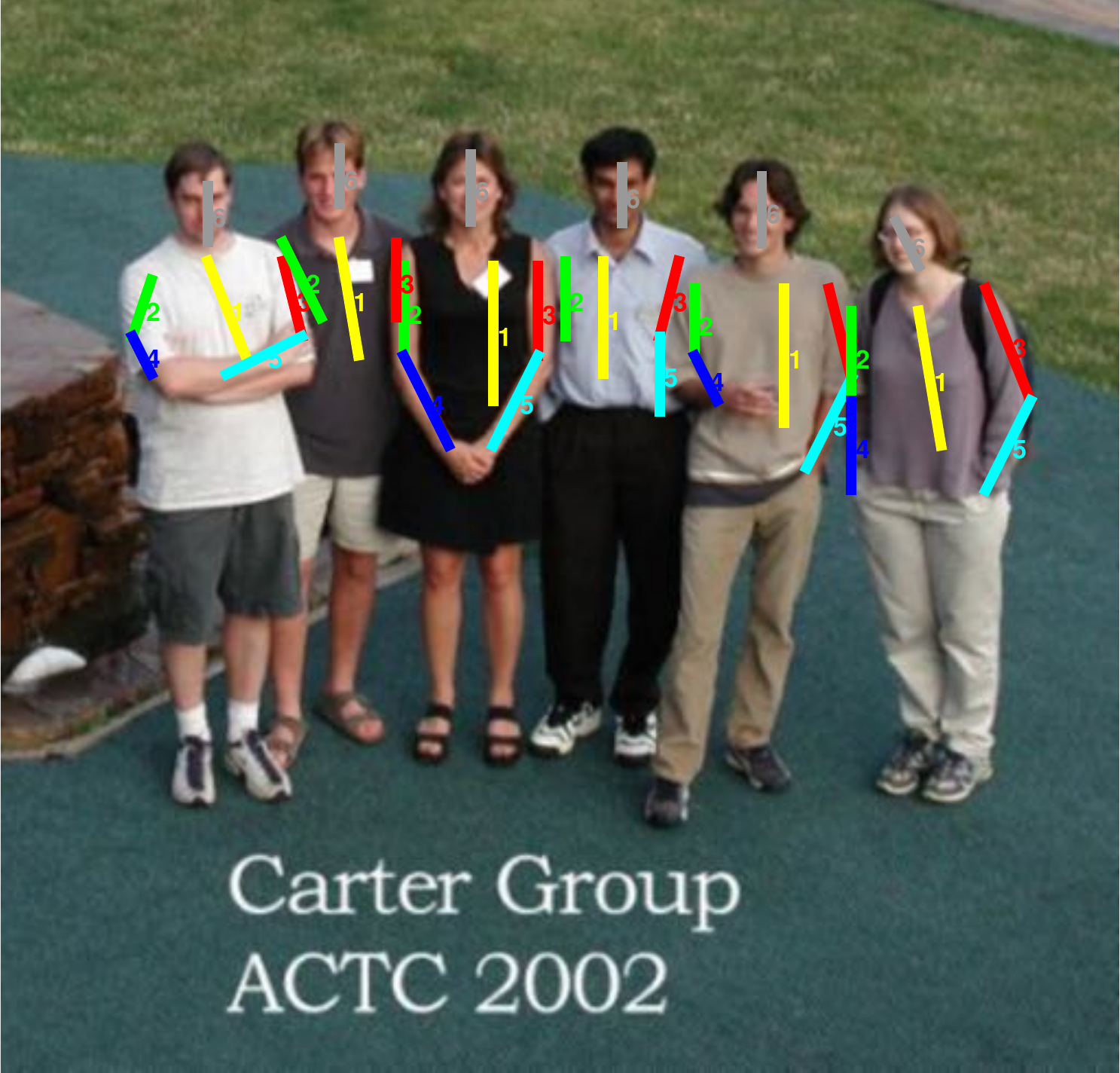}& 
    \includegraphics[height=0.145\linewidth]{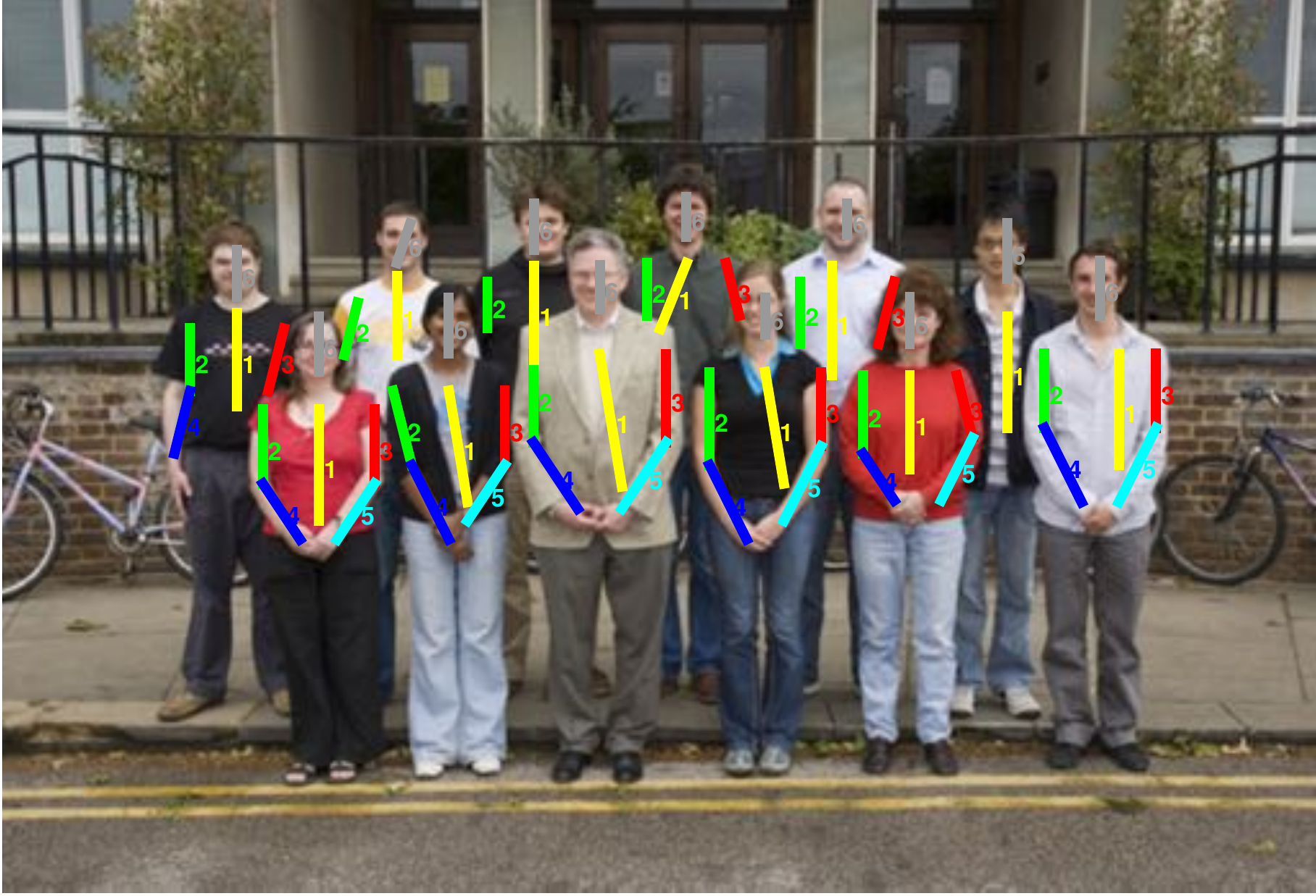}& 
    \includegraphics[height=0.145\linewidth]{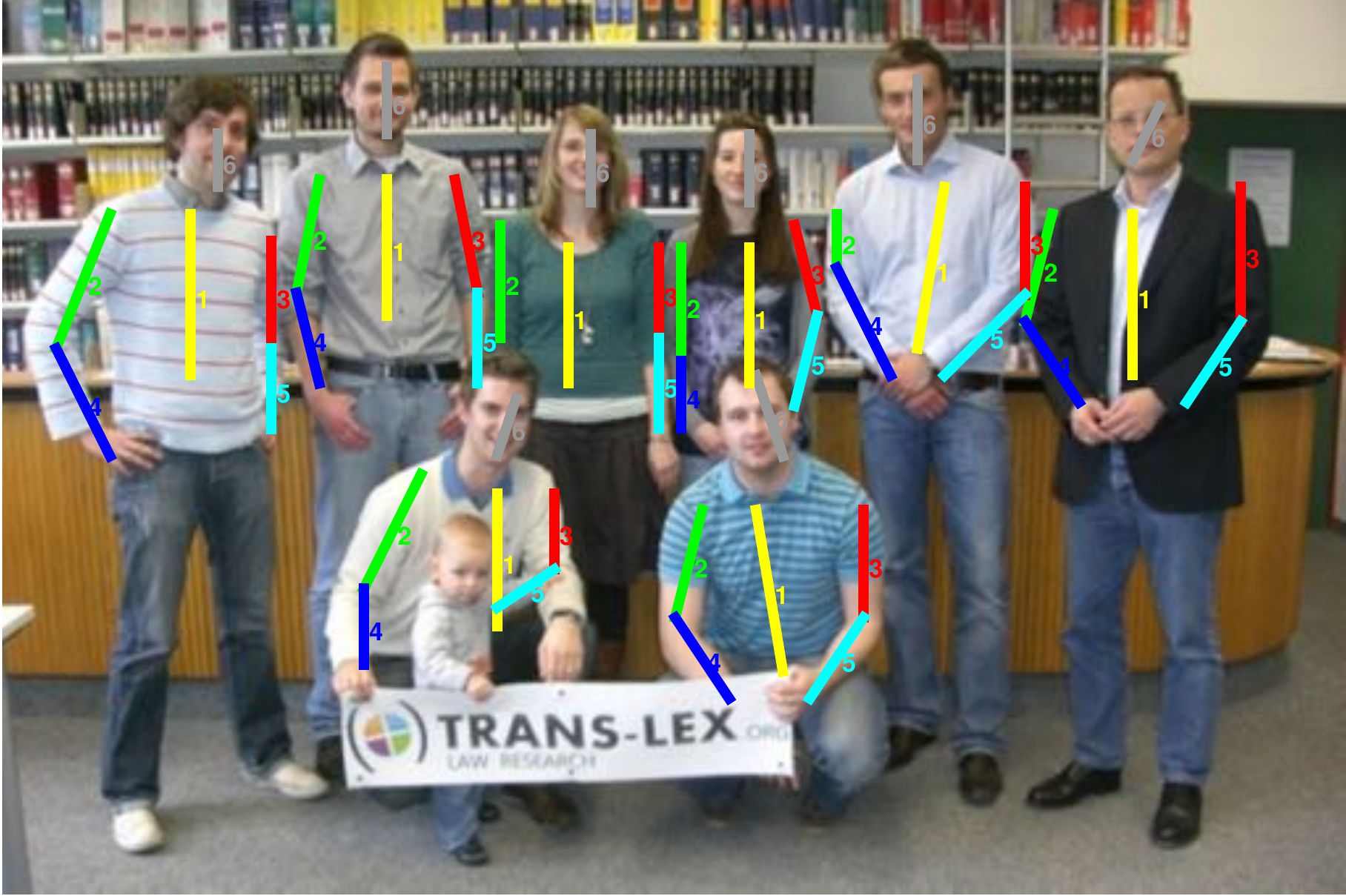}\\
    &6&7&8&9&10\\
  \end{tabular}

%%   \begin{tabular}{c c c c c c c}
%%     \begin{sideways}\bf \small \quad\quad$\detroi$\end{sideways}&
%%     \includegraphics[height=0.145\linewidth]{imgidx_0066_sticks_unary_waf.pdf}& % 32
%%     \includegraphics[height=0.145\linewidth]{imgidx_0078_sticks_unary_waf.pdf}& % 145
%%     \includegraphics[height=0.145\linewidth]{imgidx_0146_sticks_unary_waf.pdf}& 
%%     \includegraphics[height=0.145\linewidth]{imgidx_0124_sticks_unary_waf.pdf}& 
%%     \includegraphics[height=0.145\linewidth]{imgidx_0071_sticks_unary_waf.pdf}\\
%%     \begin{sideways}\bf \small \quad$\deepcut~\multb$\end{sideways}&
%%     \includegraphics[height=0.145\linewidth]{imgidx_0066_sticks_waf.pdf}&
%%     \includegraphics[height=0.145\linewidth]{imgidx_0078_sticks_waf.pdf}&
%%     \includegraphics[height=0.145\linewidth]{imgidx_0146_sticks_waf.pdf}&
%%     \includegraphics[height=0.145\linewidth]{imgidx_0124_sticks_waf.pdf}&
%%     \includegraphics[height=0.145\linewidth]{imgidx_0071_sticks_waf.pdf}\\
%%     \begin{sideways}\bf \small Chen\&Yuille~\cite{Chen:2015:POC}\end{sideways}&
%%     \includegraphics[height=0.145\linewidth]{imgidx_0066_sticks_chen_waf.pdf}&
%%     \includegraphics[height=0.145\linewidth]{imgidx_0078_sticks_chen_waf.pdf}&
%%     \includegraphics[height=0.145\linewidth]{imgidx_0146_sticks_chen_waf.pdf}&
%%     \includegraphics[height=0.145\linewidth]{imgidx_0124_sticks_chen_waf.pdf}& 
%%     \includegraphics[height=0.145\linewidth]{imgidx_0071_sticks_chen_waf.pdf}\\
%%   \end{tabular}

  %\vspace{-1.0em}
  \caption{Qualitative comparison of our joint formulation
    $\deepcut~\multb~\dense$ (rows 2, 5) to the traditional two-stage
    approach $\dense~\detroi$ (rows 1, 4) and to the approach of
    Chen\&Yuille~\cite{Chen:2015:POC} (rows 3, 6) on WAF
    dataset. $\detroi$ does not reason about occlusion and often
    predicts inconsistent body part configurations by linking the
    parts across the nearby staying people (image 4, right shoulder
    and wrist of person 2 are linked to the right elbow of person 3;
    image 5, left elbow of person 4 is linked to the left wrist of
    person 3). In contrast, $\deepcut~\multb$ predicts body part
    occlusions, disambiguates multiple and potentially overlapping
    people and correctly assembles independent detections into
    plausible body part configurations (image 4, left arms of people
    1-3 are correctly predicted to be occluded; image 5, linking of
    body parts across people 3 and 4 is corrected; image 7, occlusion
    of body parts is correctly predicted and visible parts are
    accurately estimated). In contrast to
    Chen\&Yuille~\cite{Chen:2015:POC}, $\deepcut~\multb$ better
    predicts occlusions of person's body parts by the nearby staying
    people (images 1, 3-9), but also by other objects (image 2, left
    arm of person 1 is occluded by the chair). Furthermore,
    $\deepcut~\multb$ is able to better cope with strong articulations
    and foreshortenings (image 1, person 6; image 3, person 2; image
    5, person 4; image 7, person 4; image 8, person 1). Typical
    $\deepcut~\multb$ failure case is shown in image 10: the right
    upper arm of person 3 and both arms of person 4 are not estimated
    due to missing part detection candidates.}
   %\vspace{-1.0em}
  \label{fig:qualitative_waf}
\end{figure*}

Qualitative comparison of our joint formulation
$\deepcut~\multb~\dense$ to the traditional two-stage approach
$\dense~\detroi$ relying on person detector, and to the approach of
Chen\&Yuille~\cite{Chen:2015:POC} on WAF dataset is shown in
Fig.~\ref{fig:qualitative_waf}. See figure caption for visual
performance analysis. 

\section{Additional Results on MPII Multi-Person}
\begin{figure*}
  \centering
  \begin{tabular}{c c c c c c c}
    &
    \includegraphics[height=0.140\linewidth]{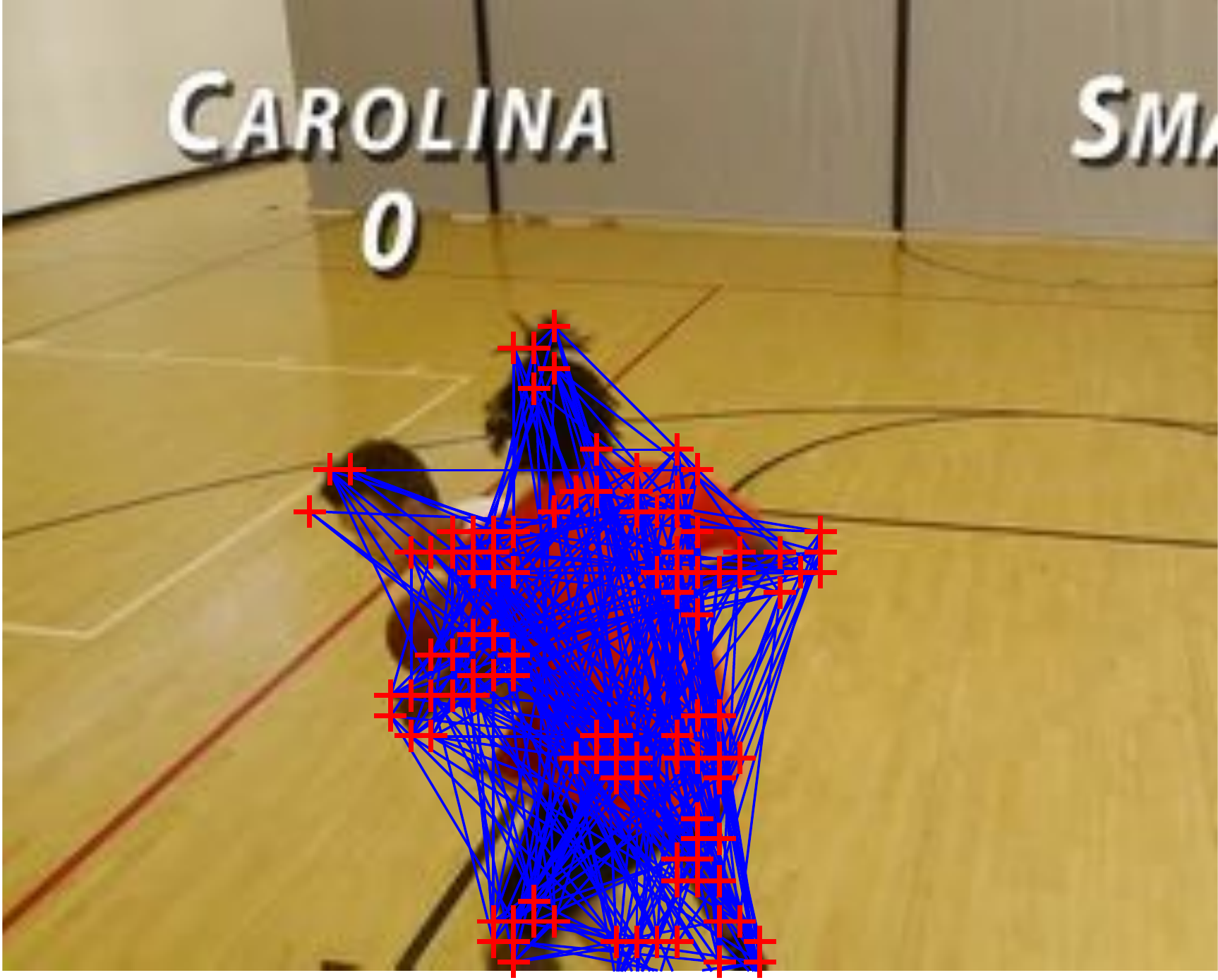}&
    \includegraphics[height=0.140\linewidth]{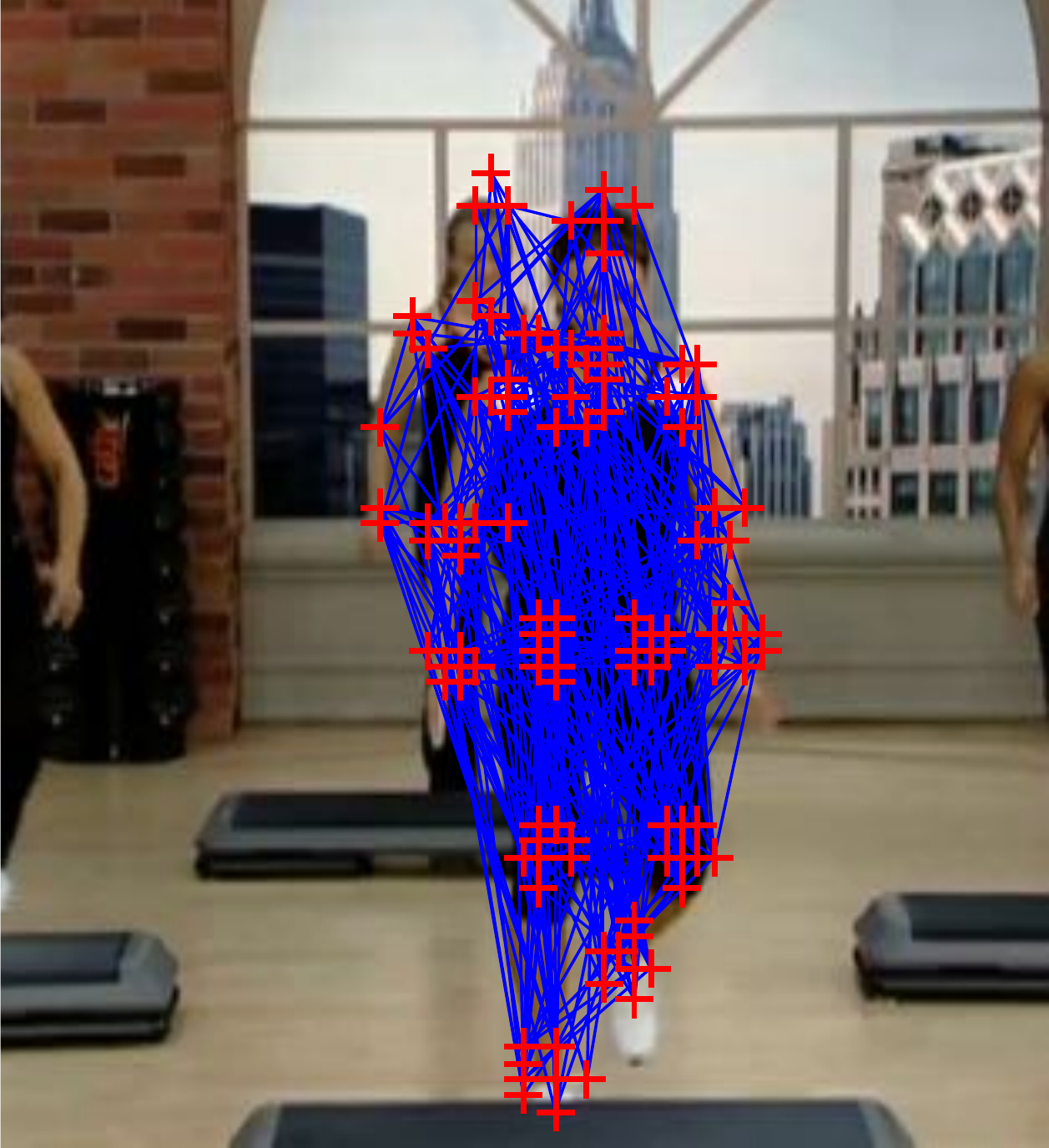}&
    \includegraphics[height=0.140\linewidth]{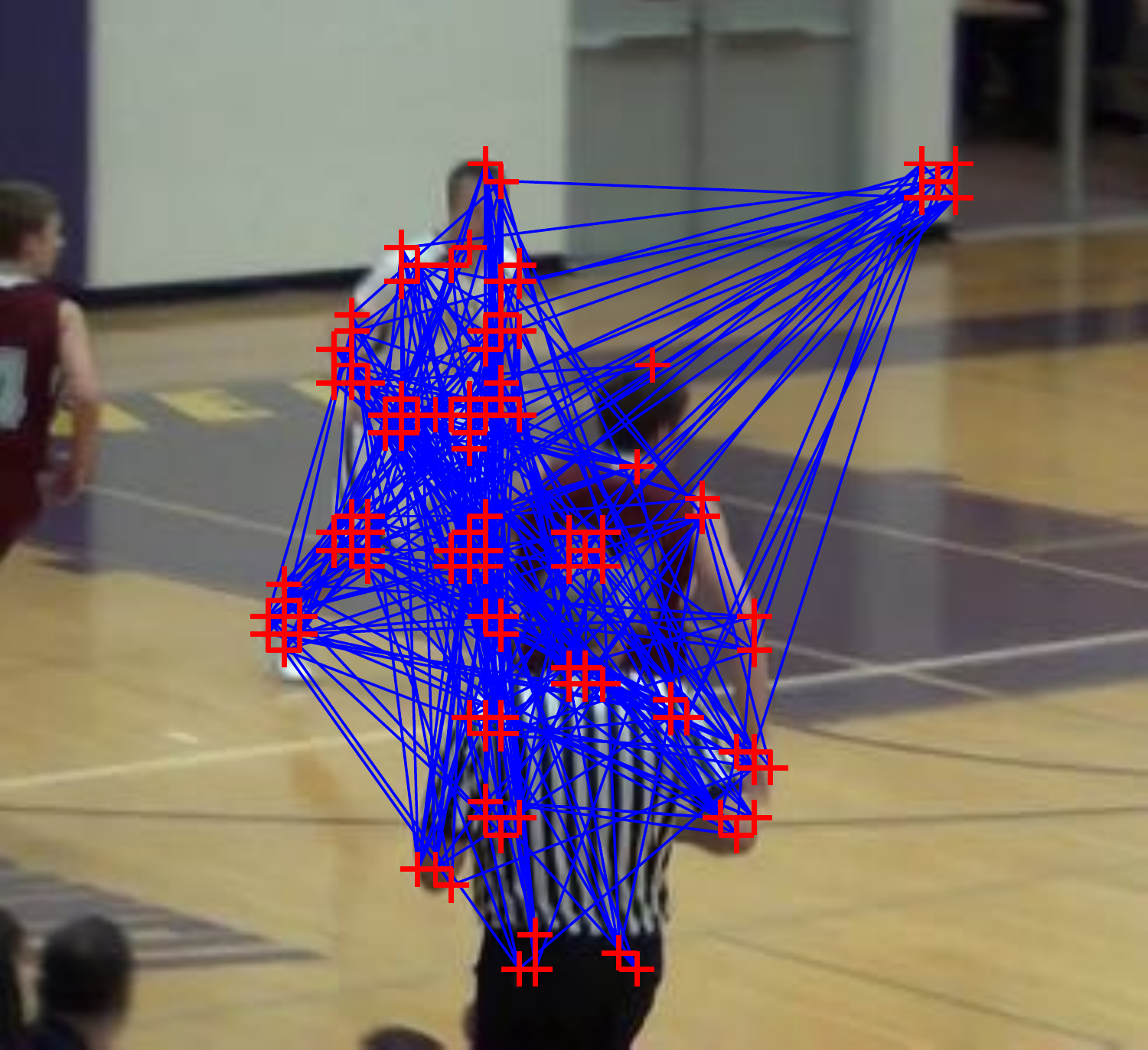}&
    \includegraphics[height=0.140\linewidth]{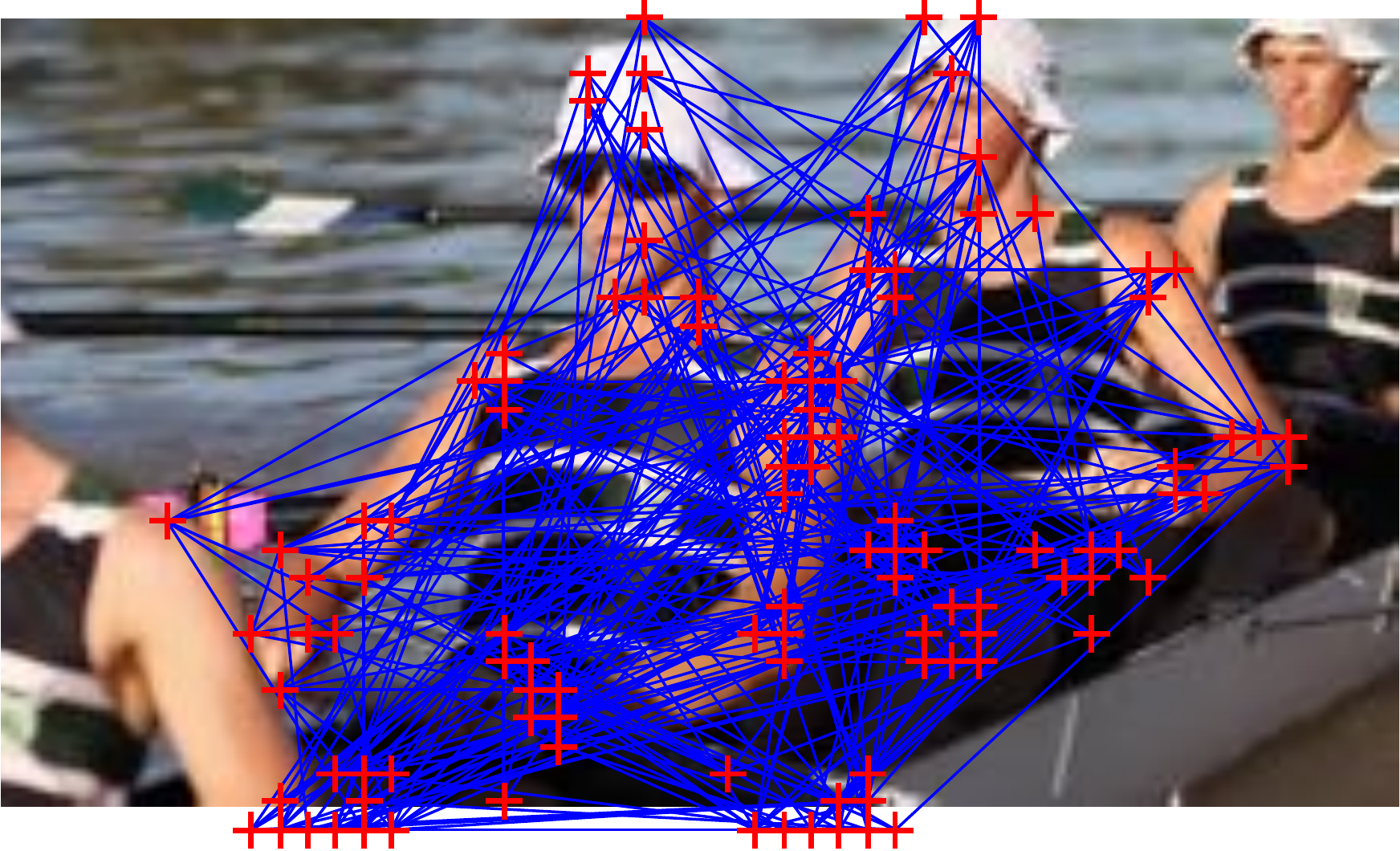}&
    \includegraphics[height=0.140\linewidth]{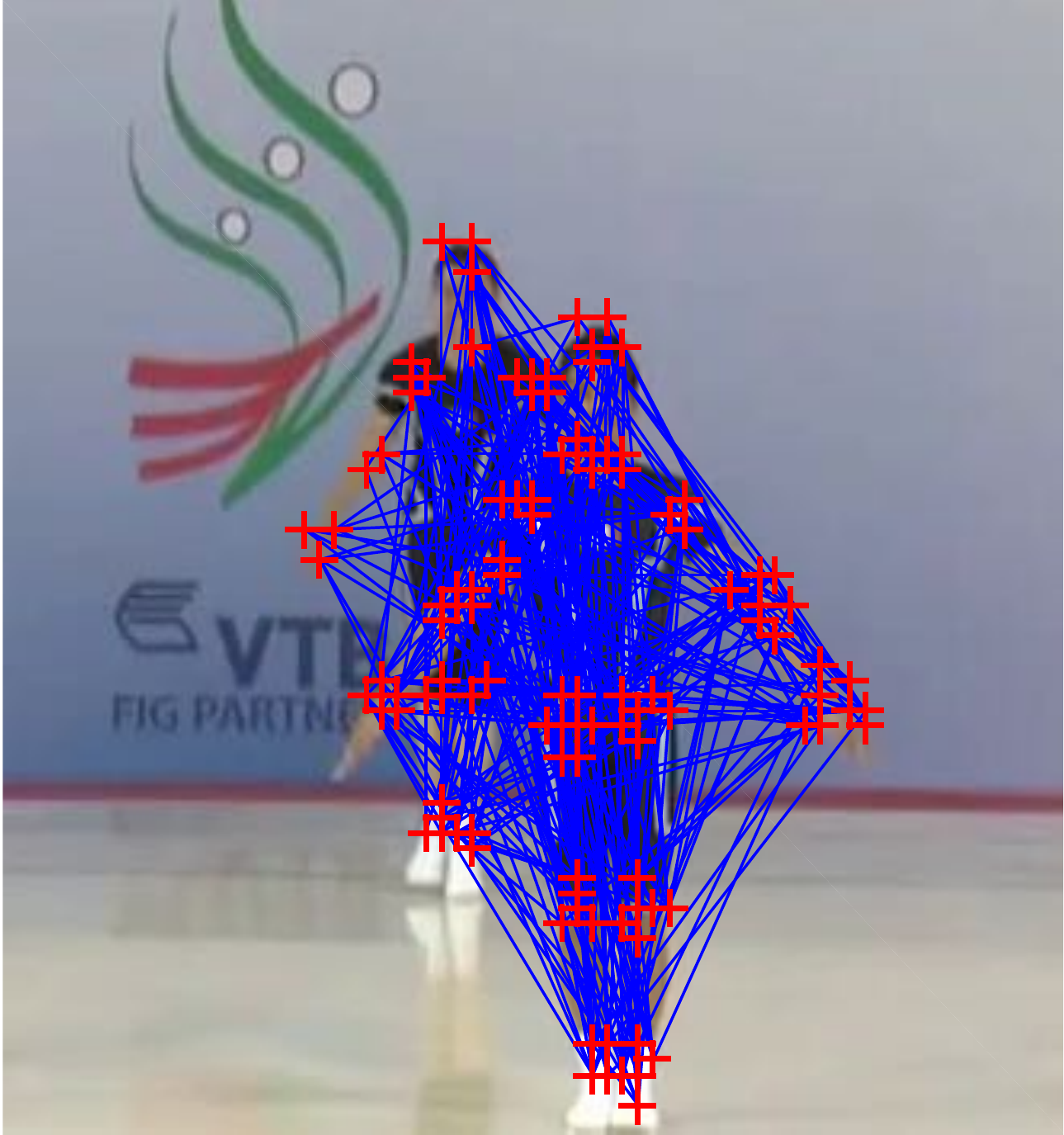}\\
    \begin{sideways}\bf\quad $\deepcut~\multb$\end{sideways}&
    \includegraphics[height=0.140\linewidth]{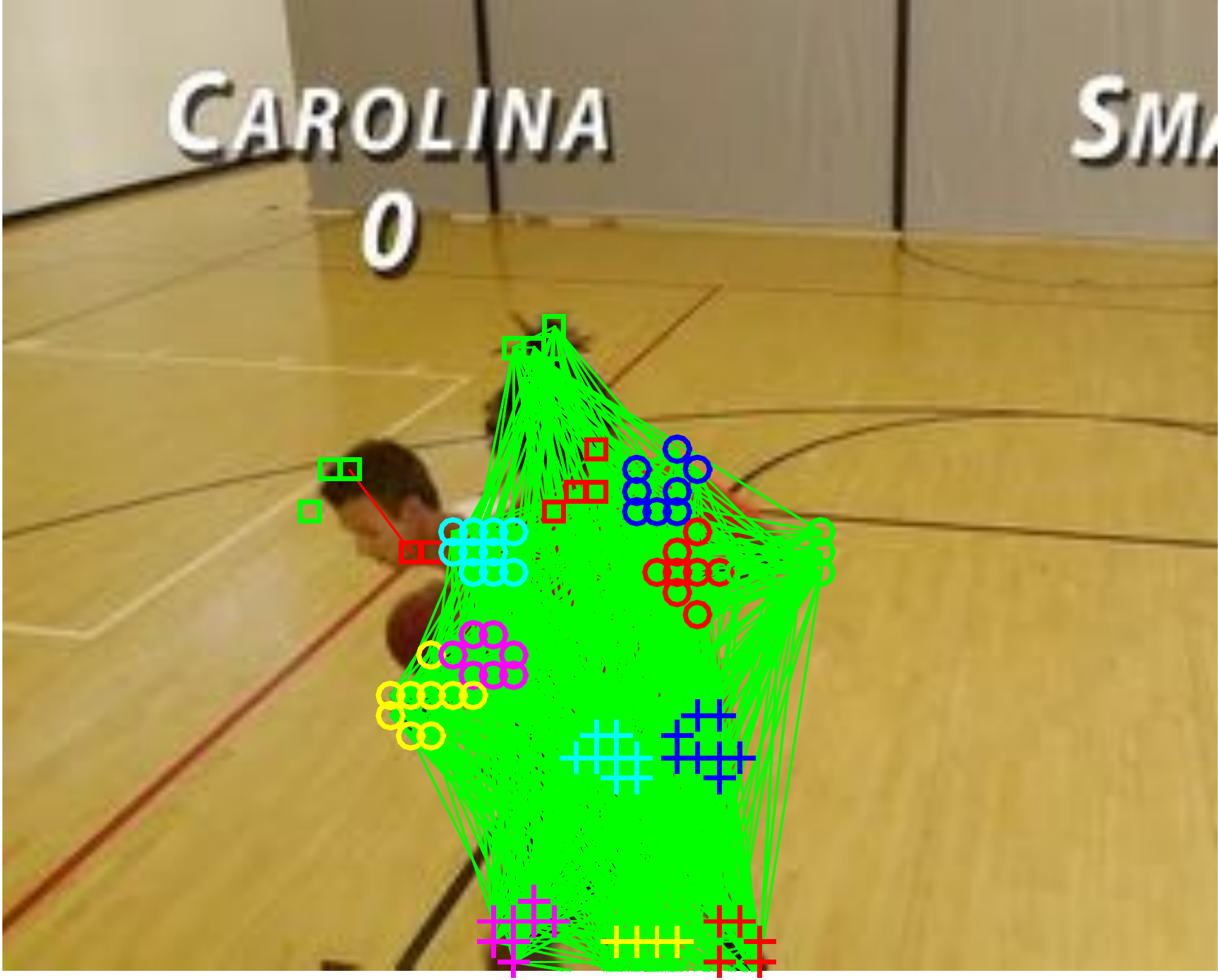}&
    \includegraphics[height=0.140\linewidth]{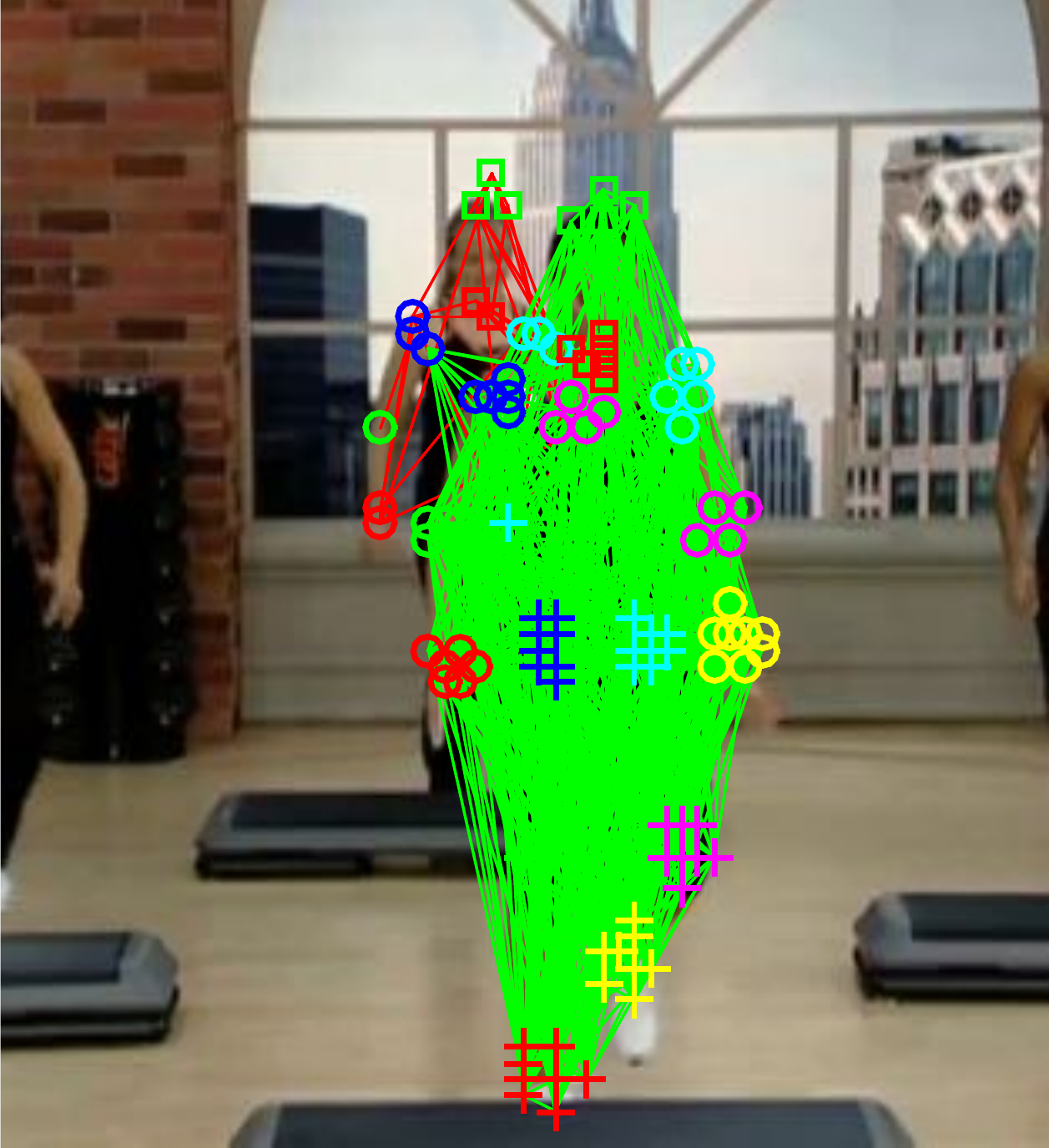}&
    \includegraphics[height=0.140\linewidth]{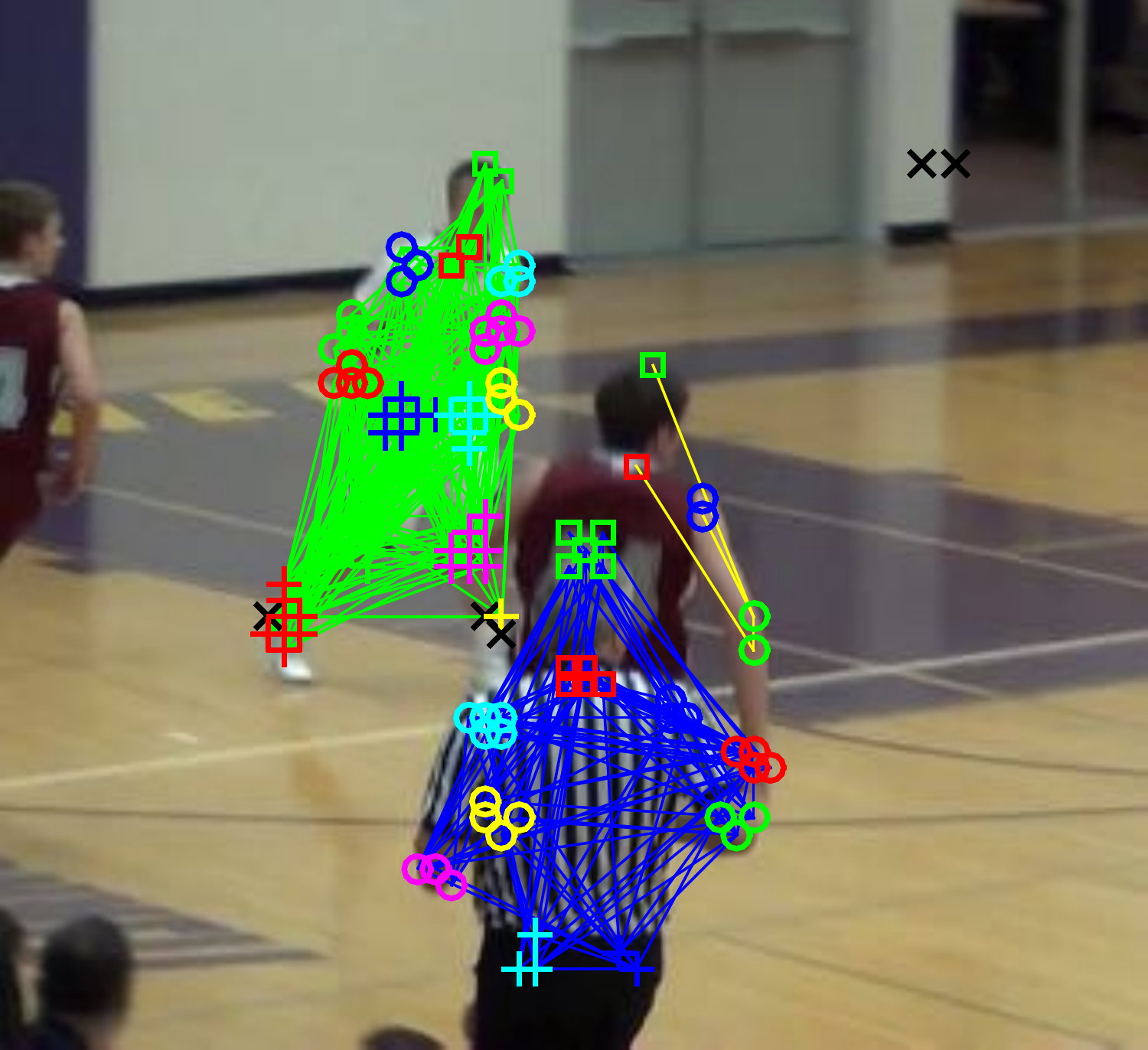}&
    \includegraphics[height=0.140\linewidth]{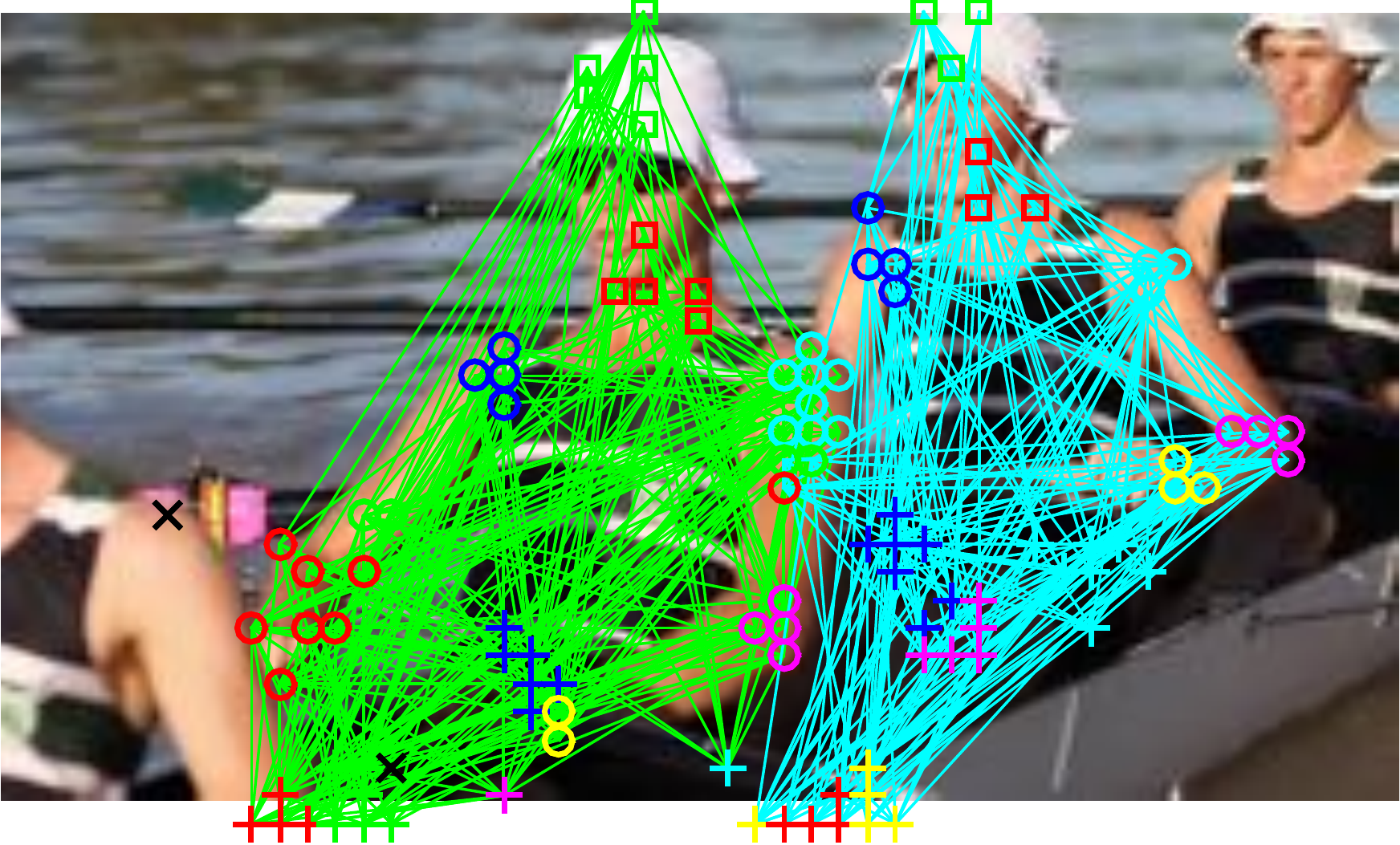}&
    \includegraphics[height=0.140\linewidth]{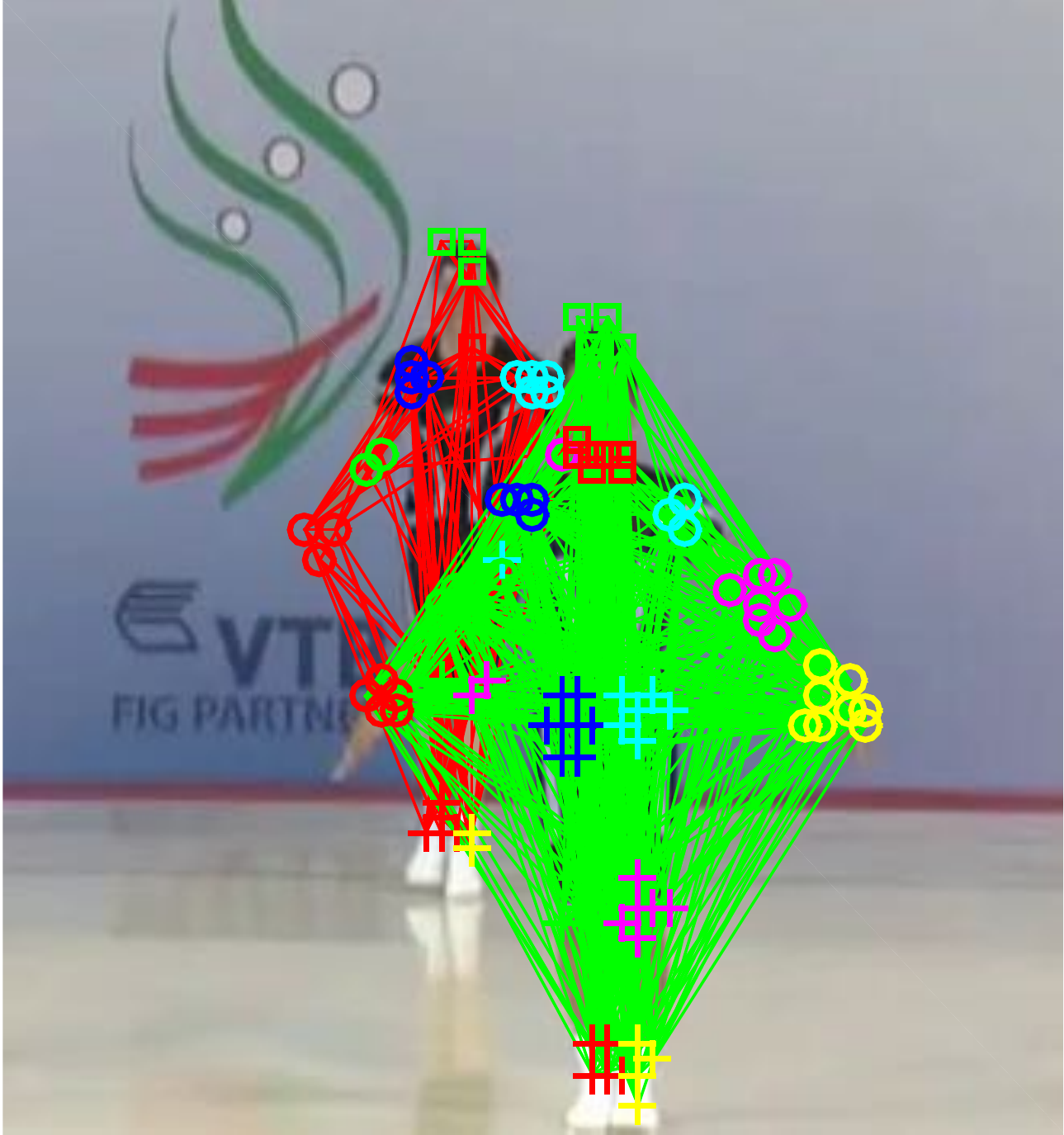}\\
    &
    \includegraphics[height=0.140\linewidth]{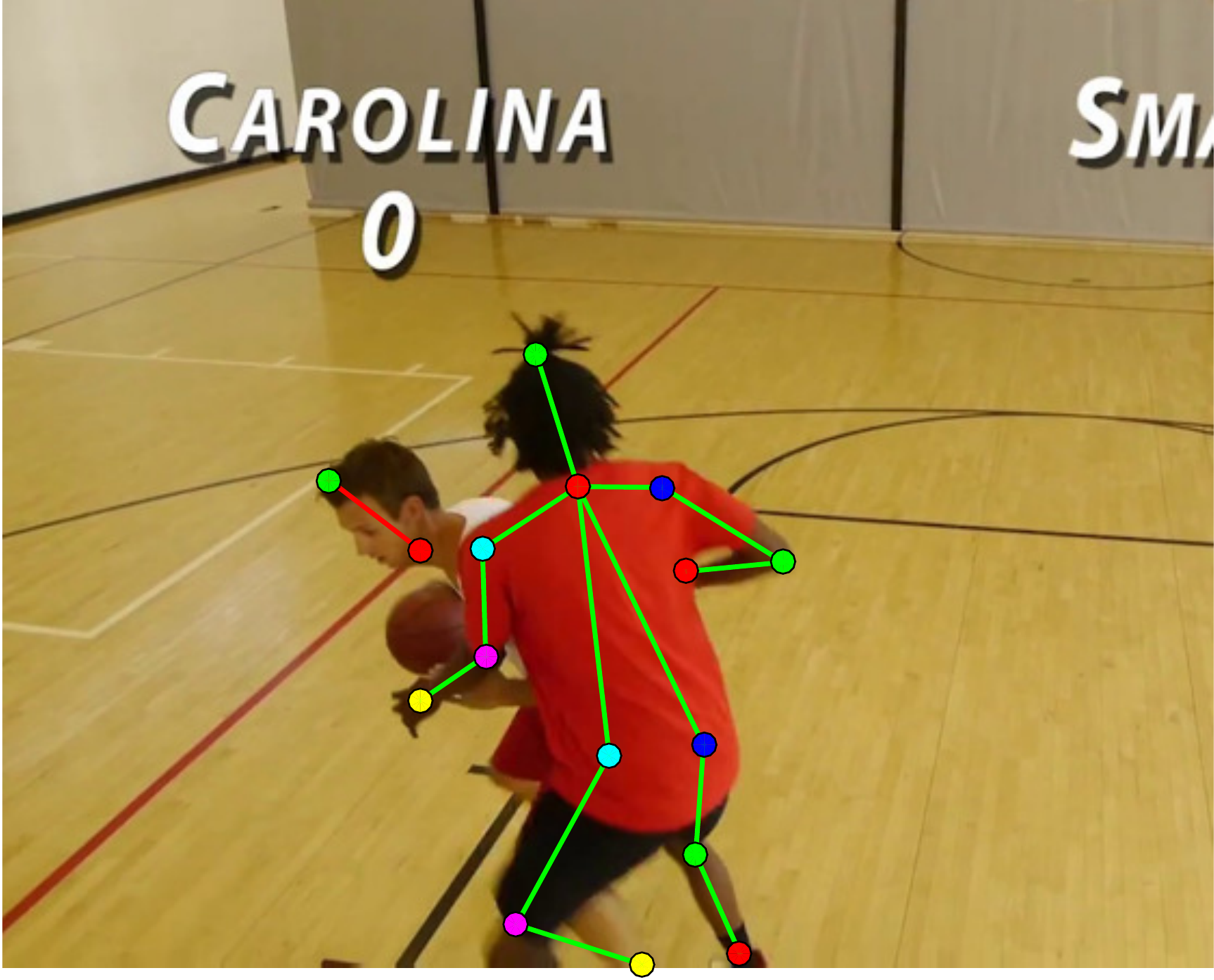}& 
    \includegraphics[height=0.140\linewidth]{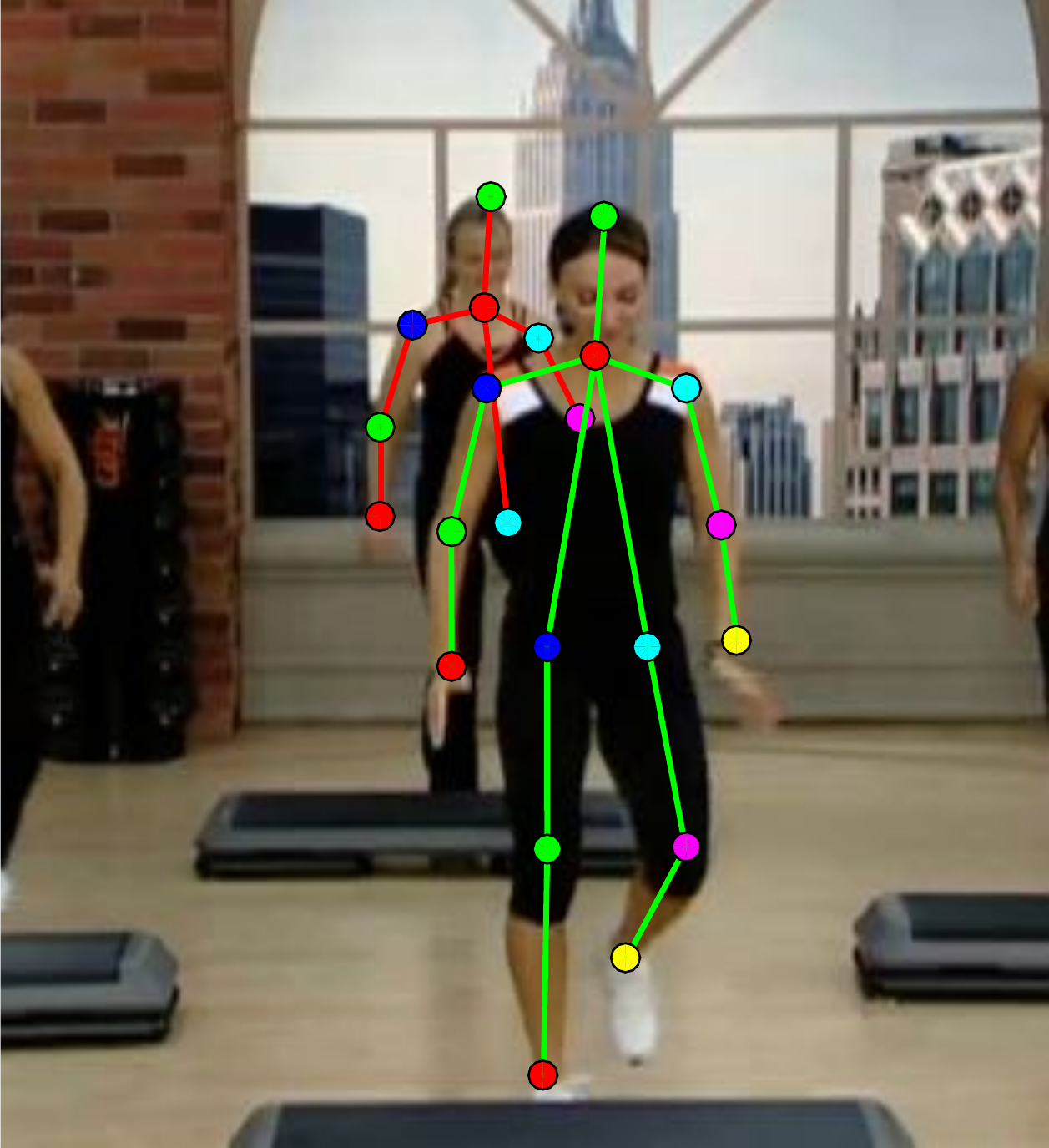}&
    \includegraphics[height=0.140\linewidth]{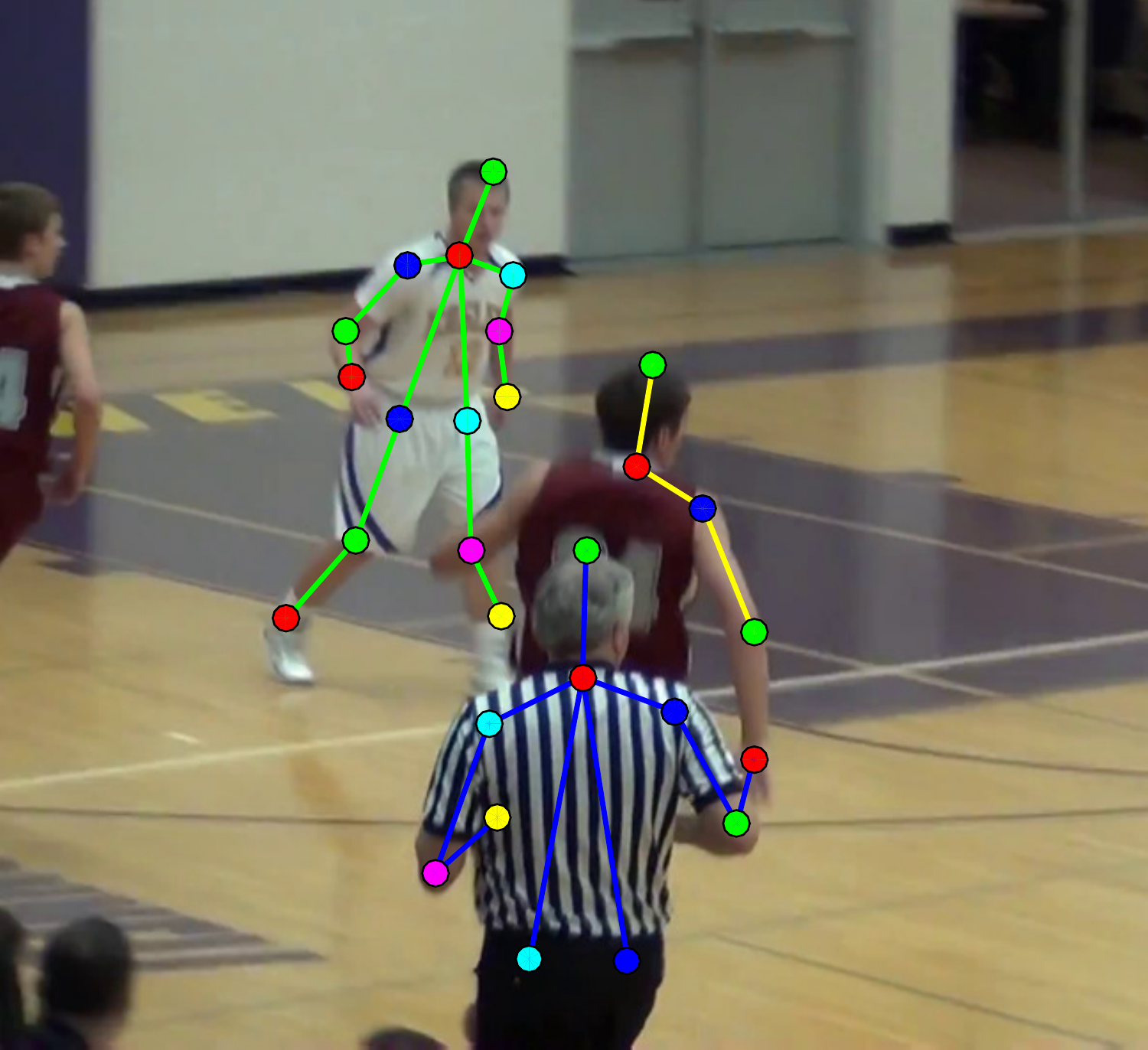}&
    \includegraphics[height=0.140\linewidth]{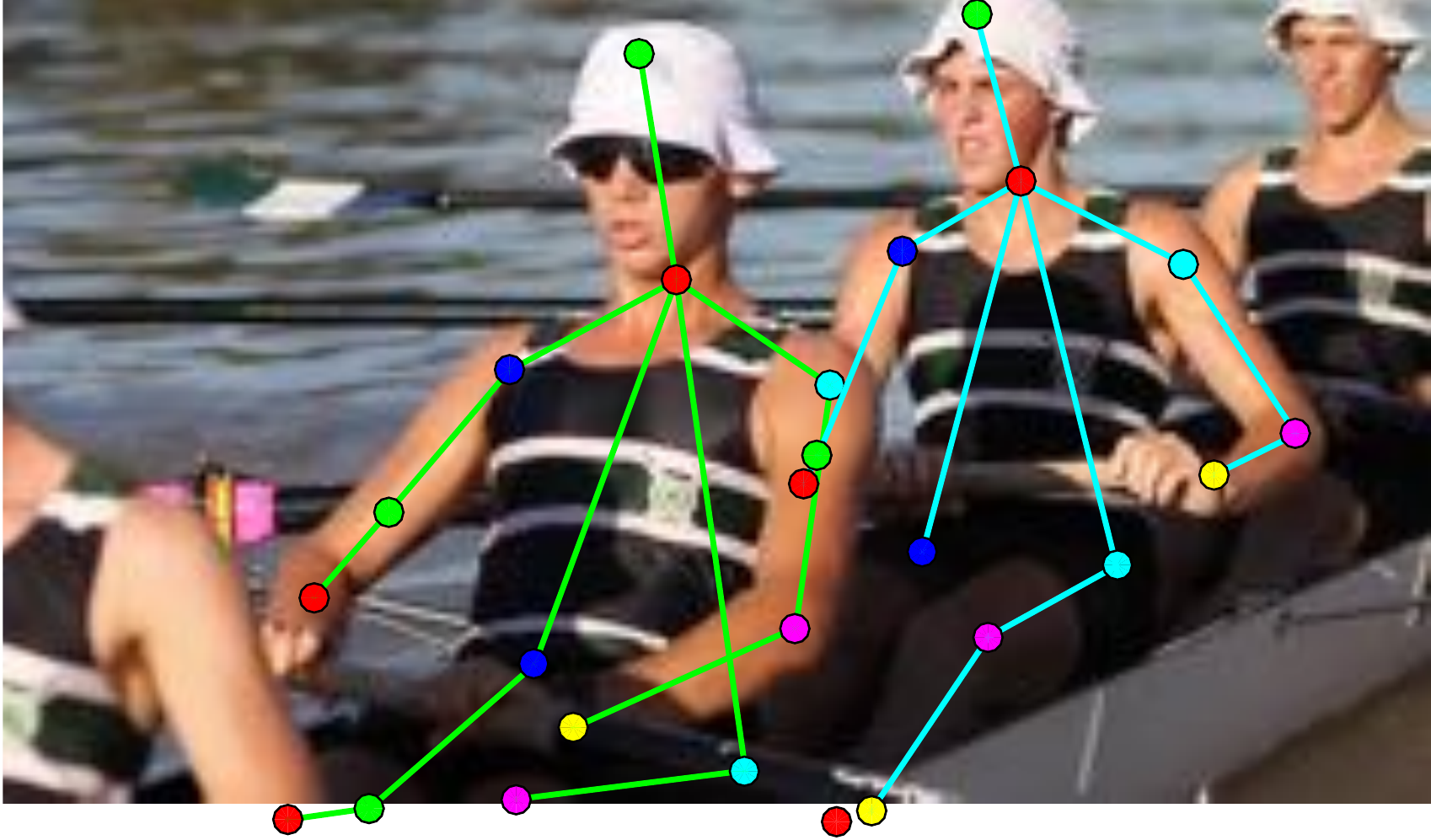}&
    \includegraphics[height=0.140\linewidth]{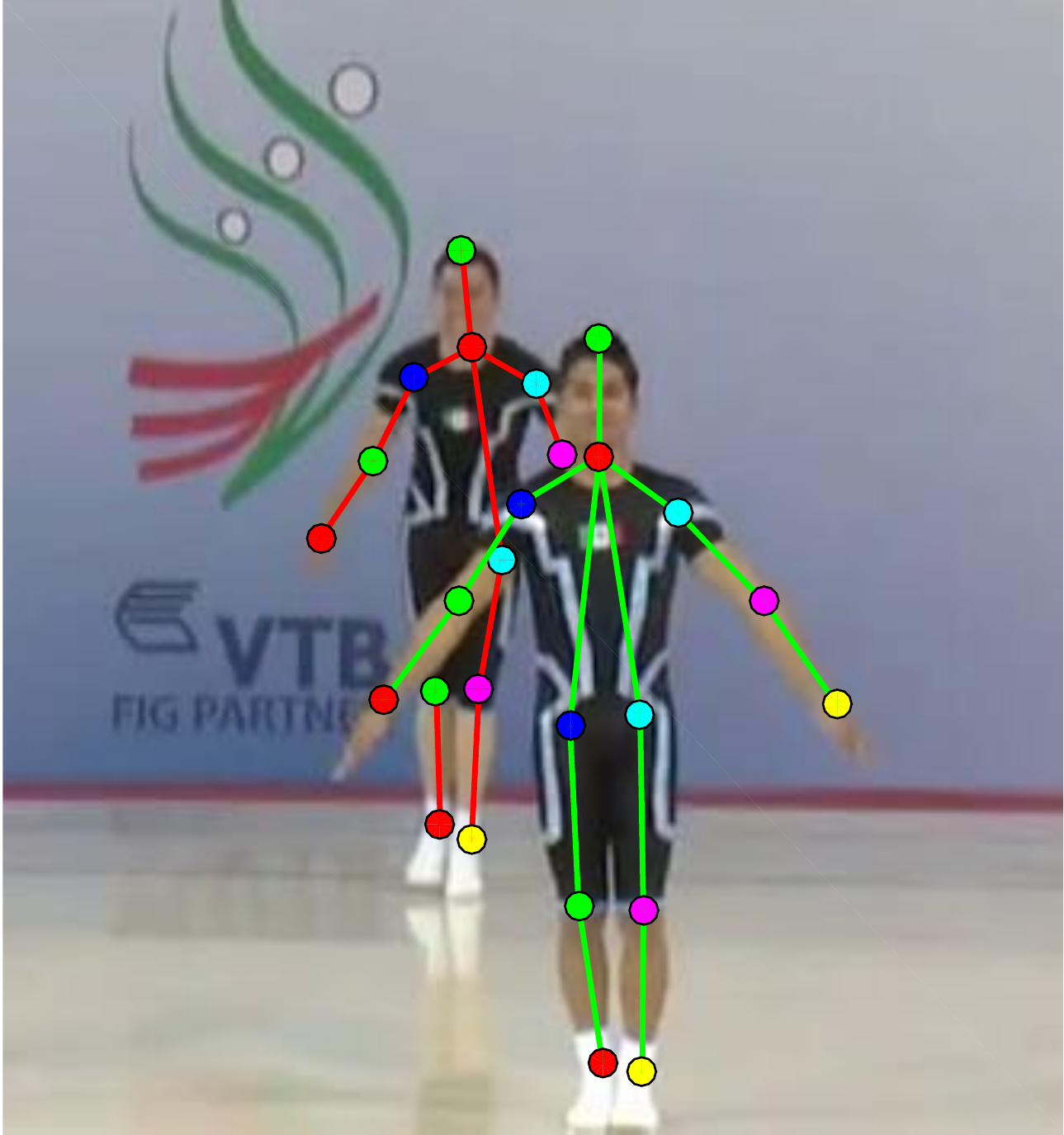}\\
    \midrule\midrule
    \begin{sideways}\bf \quad\quad$\detroi$\end{sideways}&
    \includegraphics[height=0.140\linewidth]{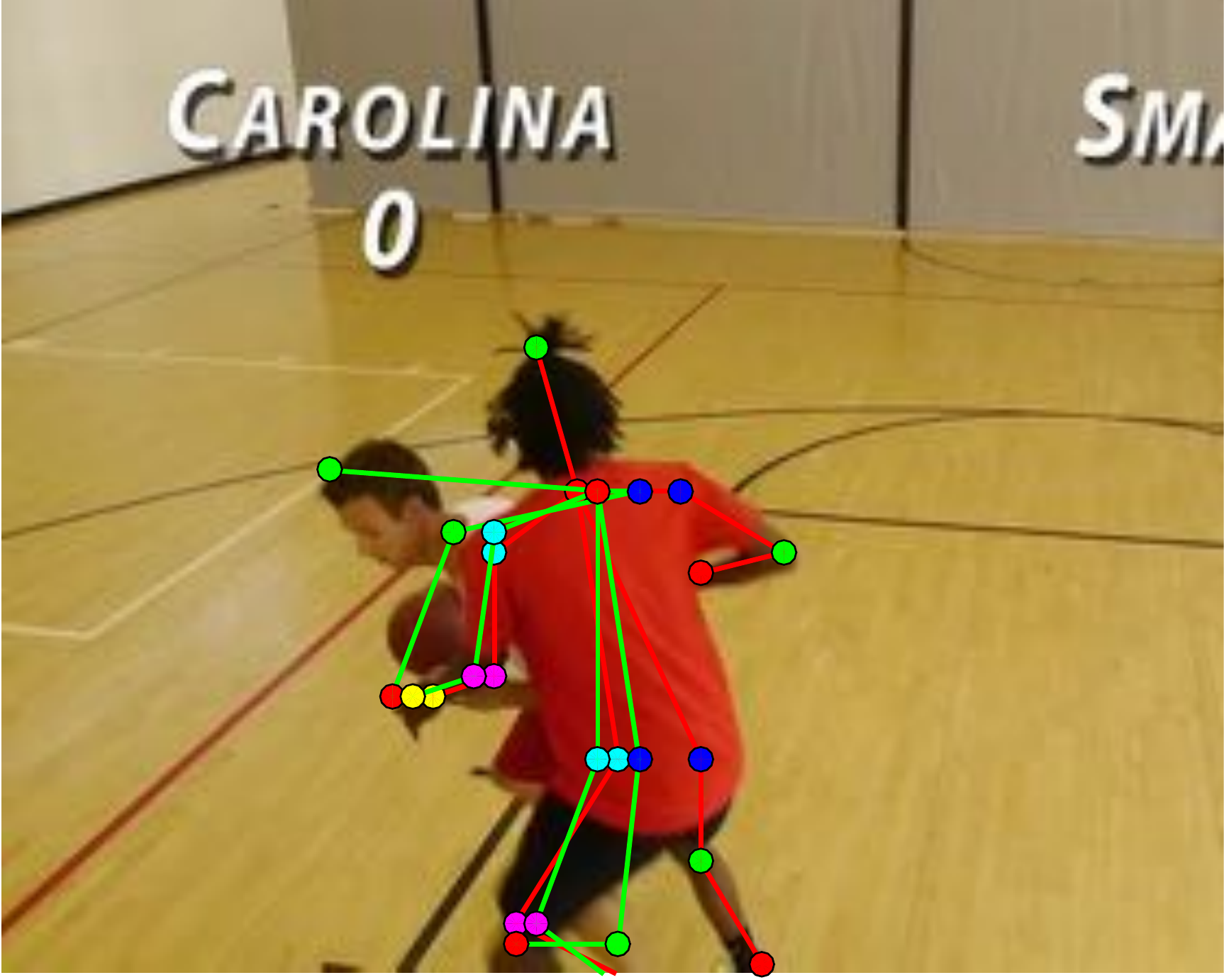}& 
    \includegraphics[height=0.140\linewidth]{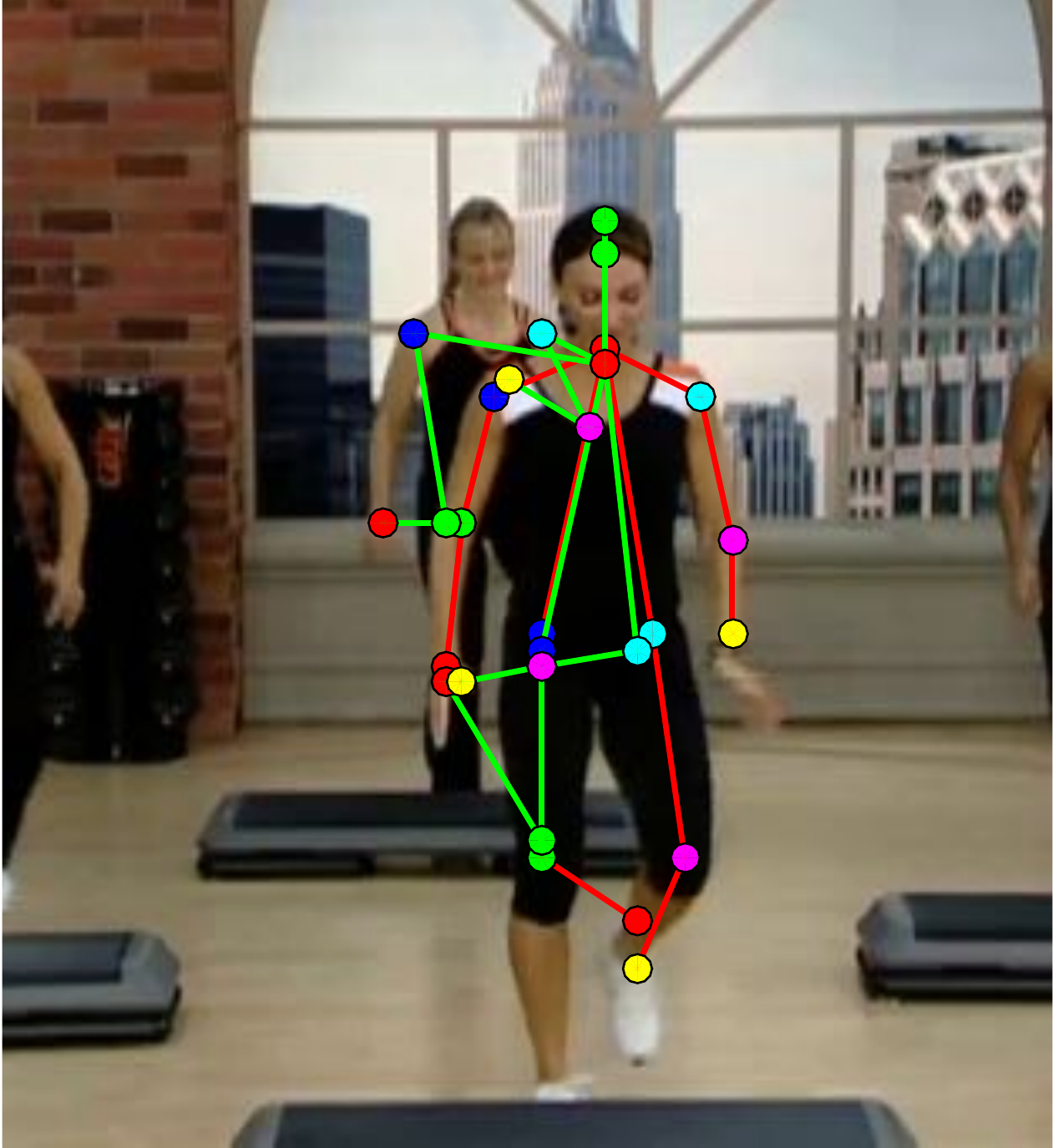}&
    \includegraphics[height=0.140\linewidth]{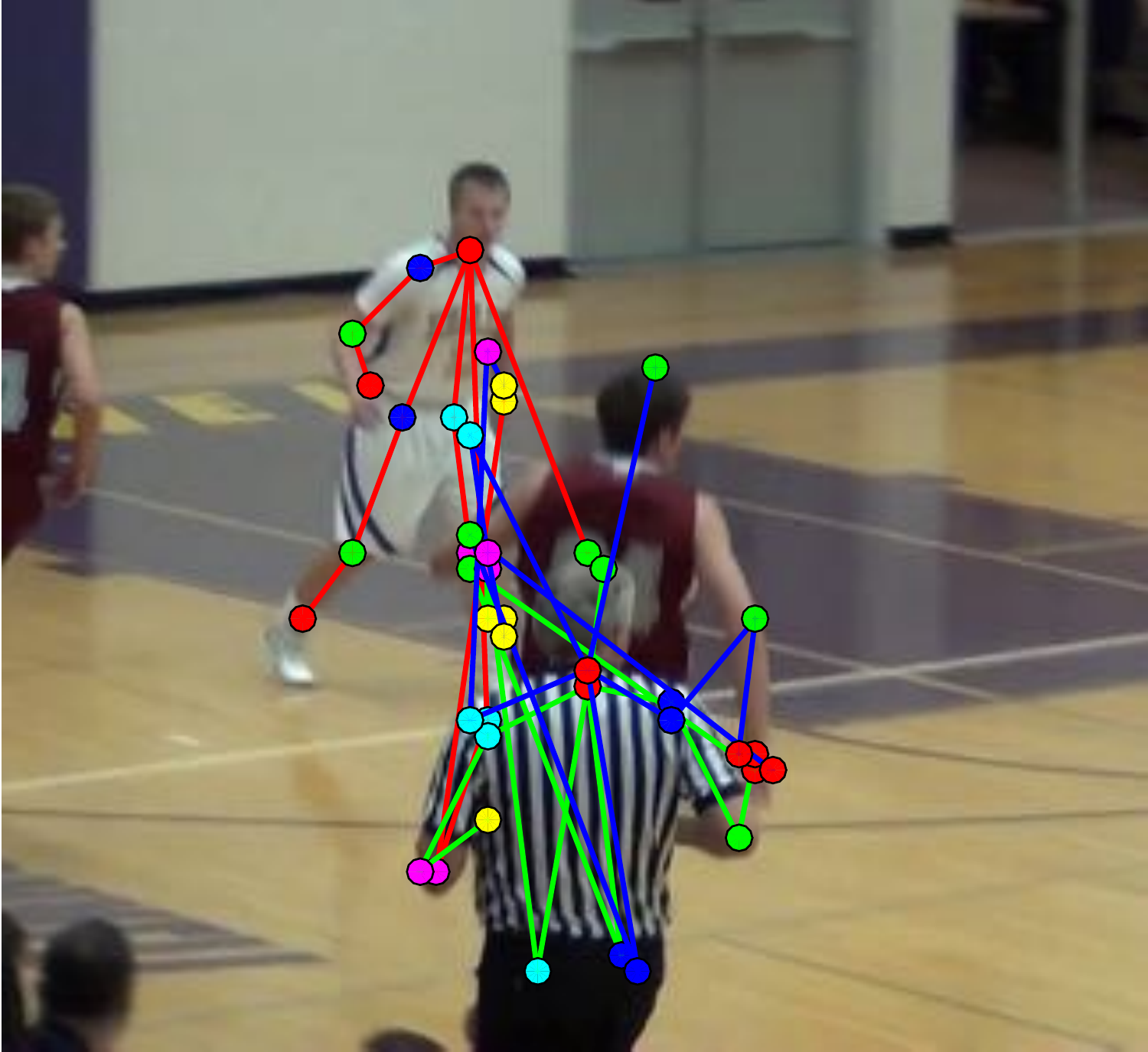}&
    \includegraphics[height=0.140\linewidth]{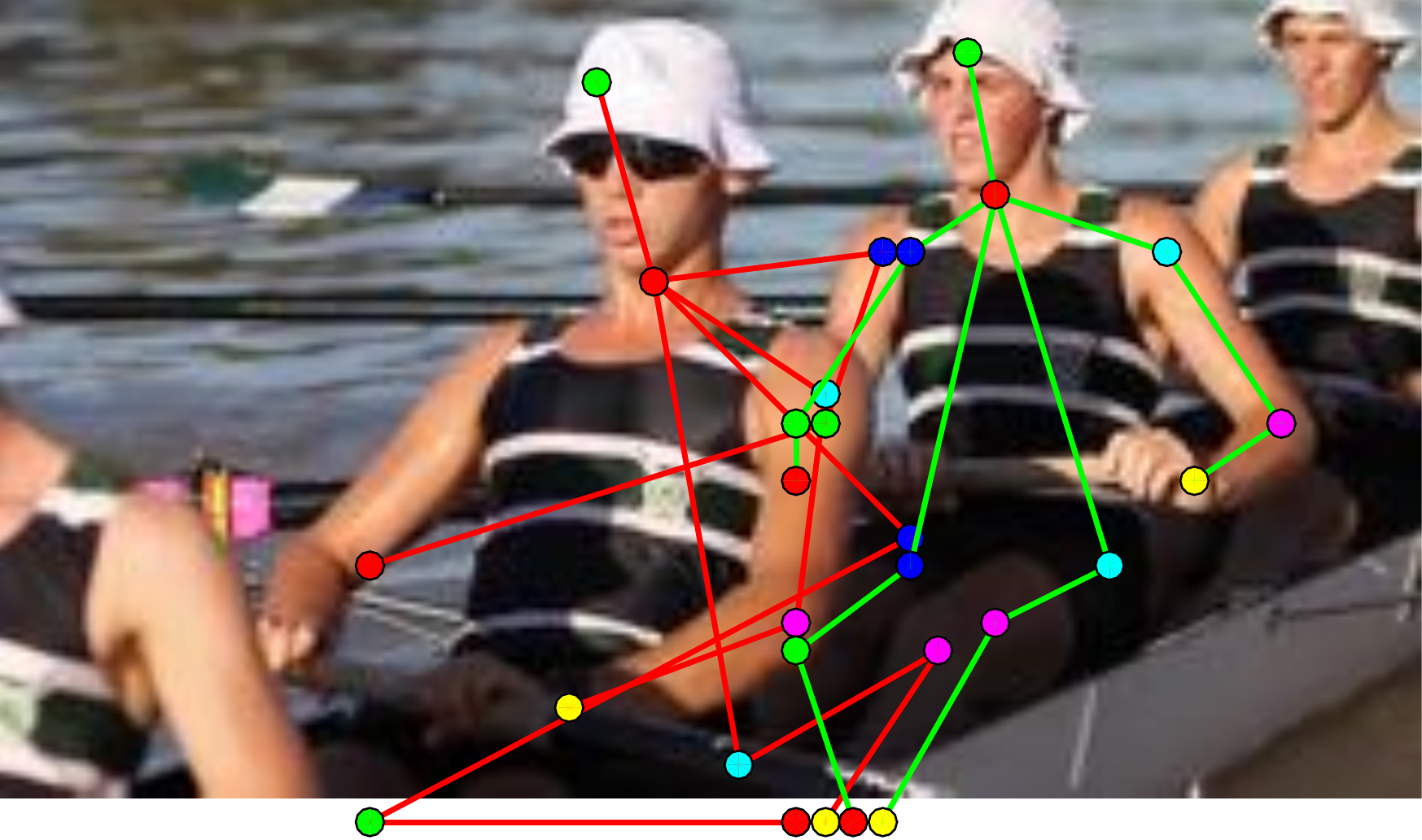}&
    \includegraphics[height=0.140\linewidth]{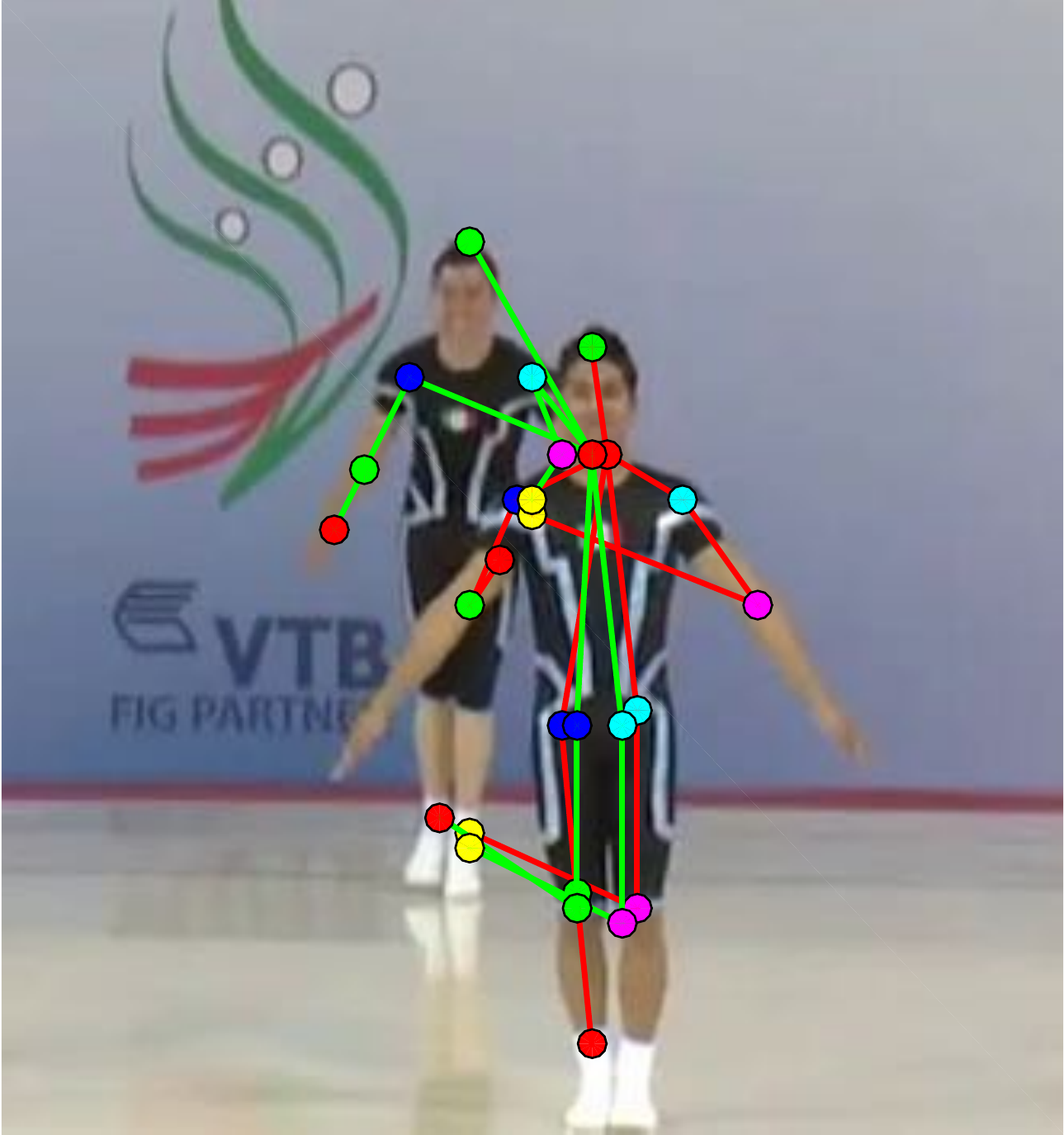}\\
    %&&&&\\
  &1&2&3&4&5\\  
  \end{tabular}

  \begin{tabular}{c c c c c c c}
    \toprule
    &
    \includegraphics[height=0.140\linewidth]{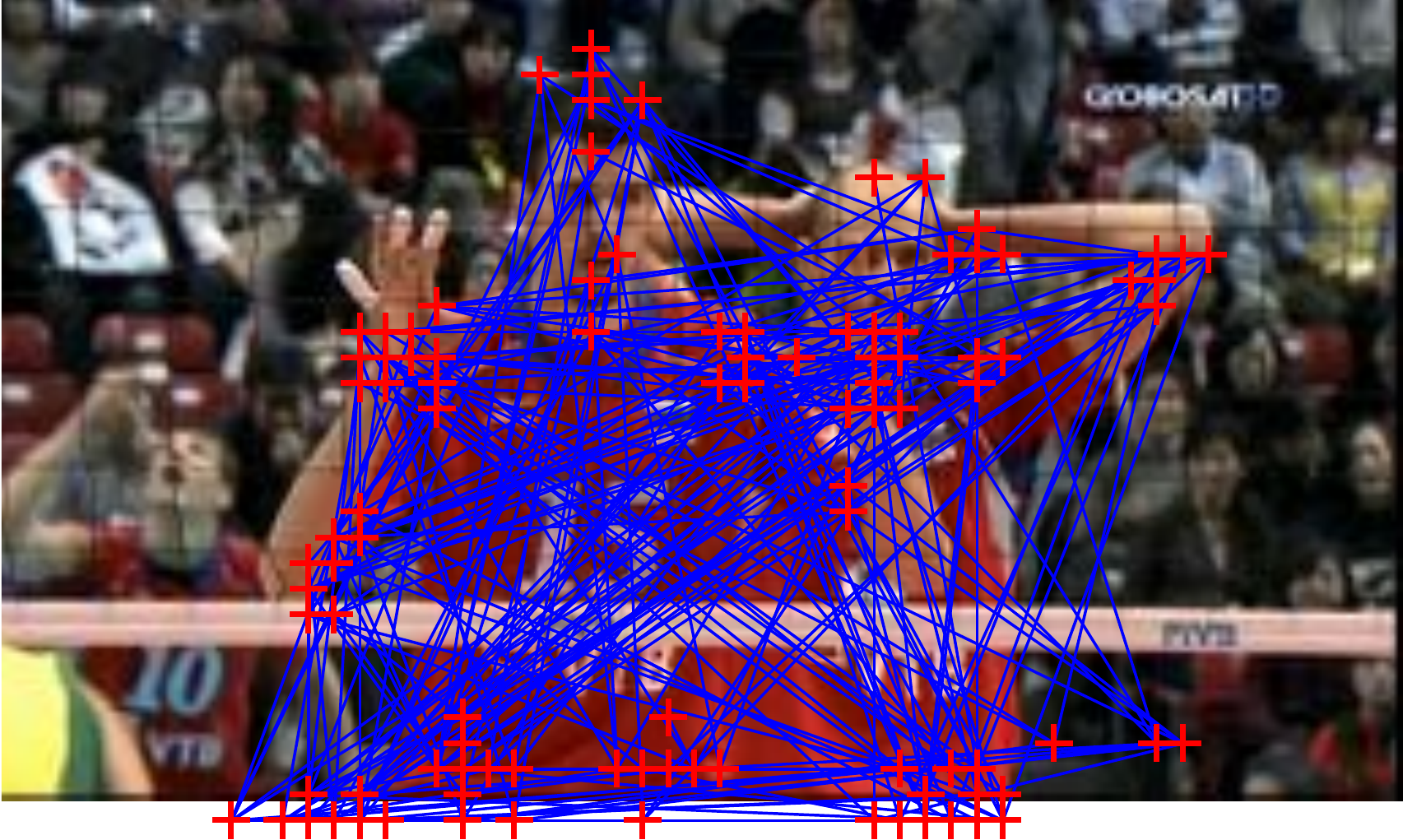}&
    \includegraphics[height=0.140\linewidth]{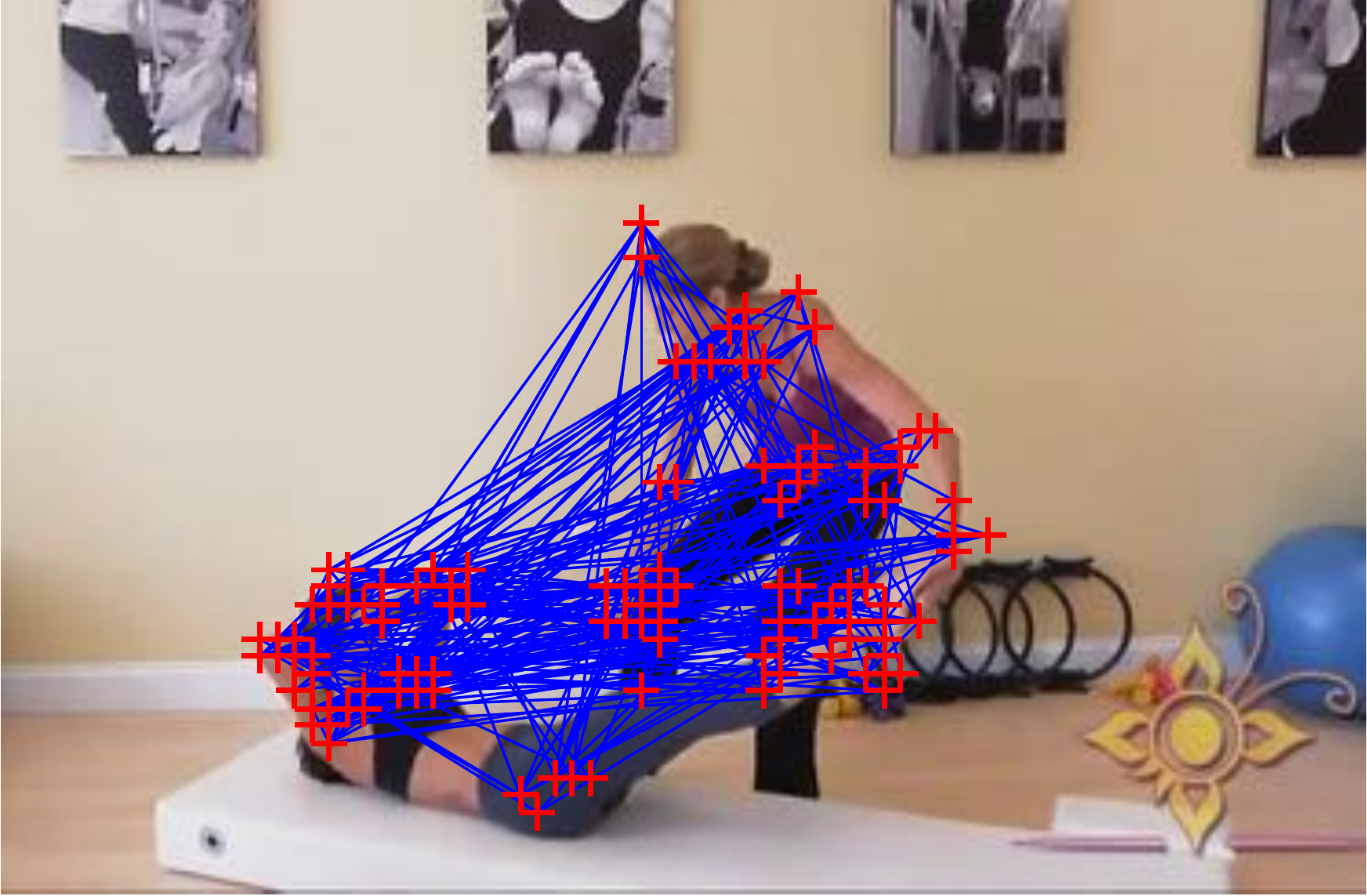}&
    \includegraphics[height=0.140\linewidth]{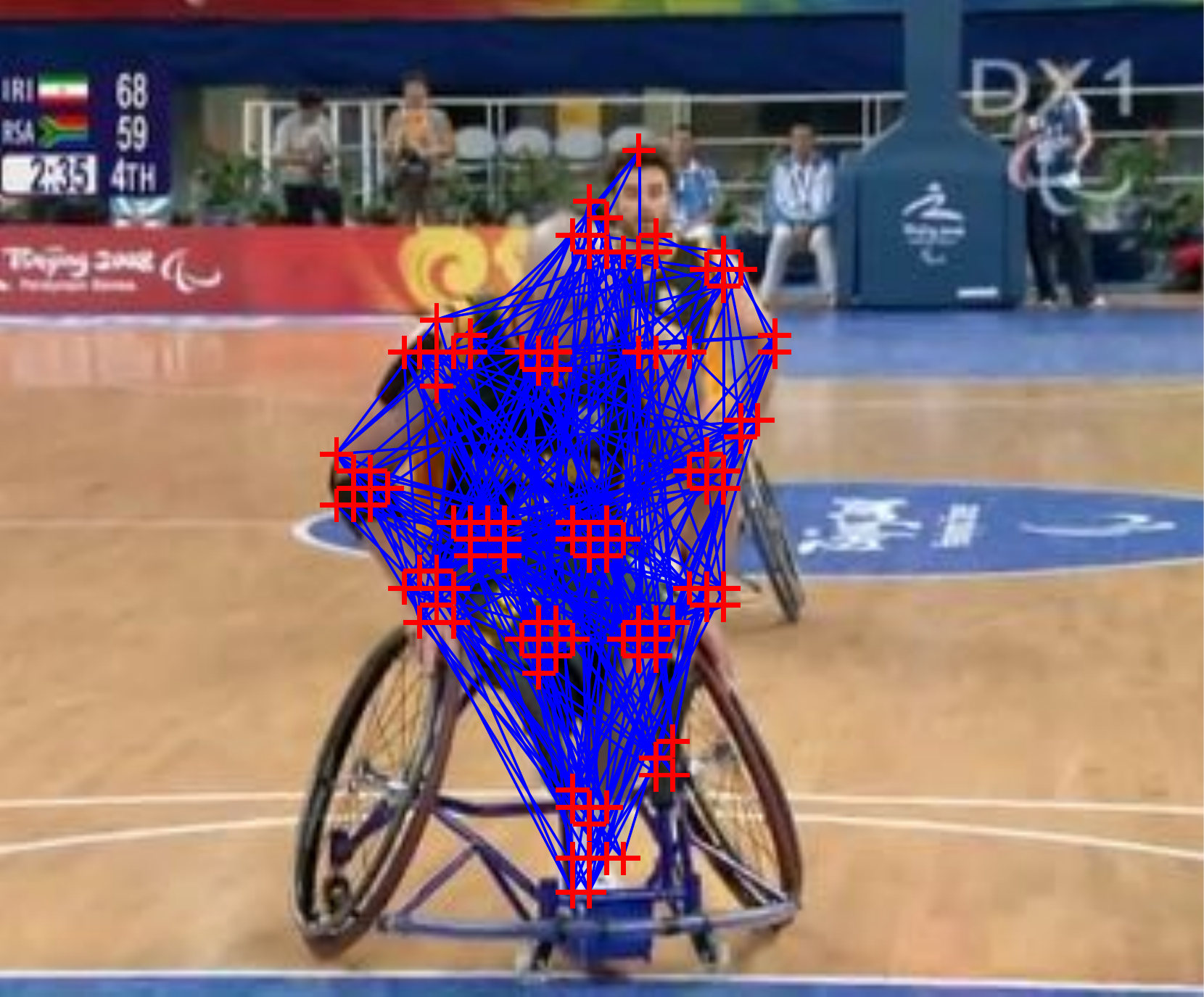}&
    \includegraphics[height=0.140\linewidth]{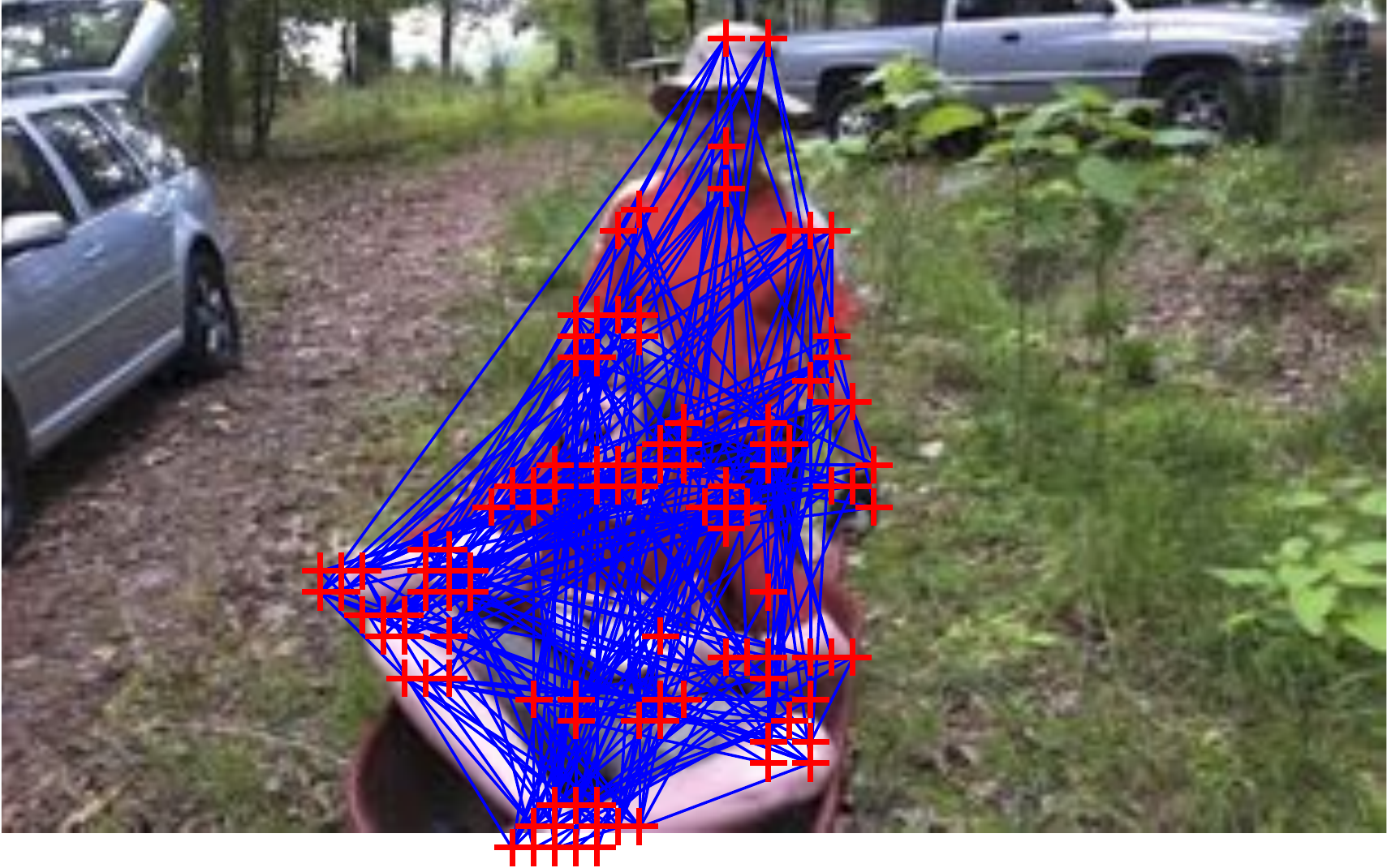}\\
    \begin{sideways}\bf\quad $\deepcut~\multb$\end{sideways}&    
    \includegraphics[height=0.140\linewidth]{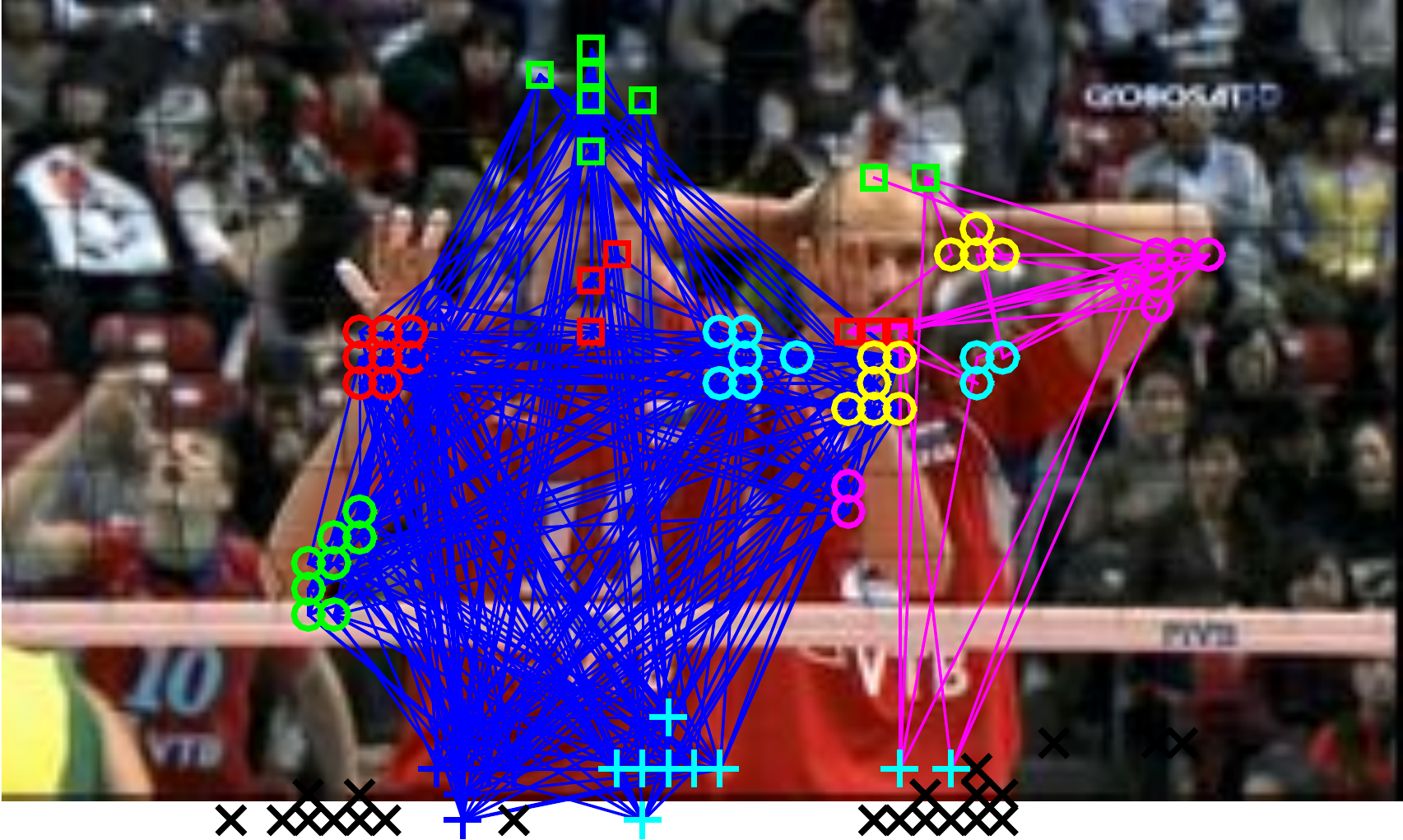}&
    \includegraphics[height=0.140\linewidth]{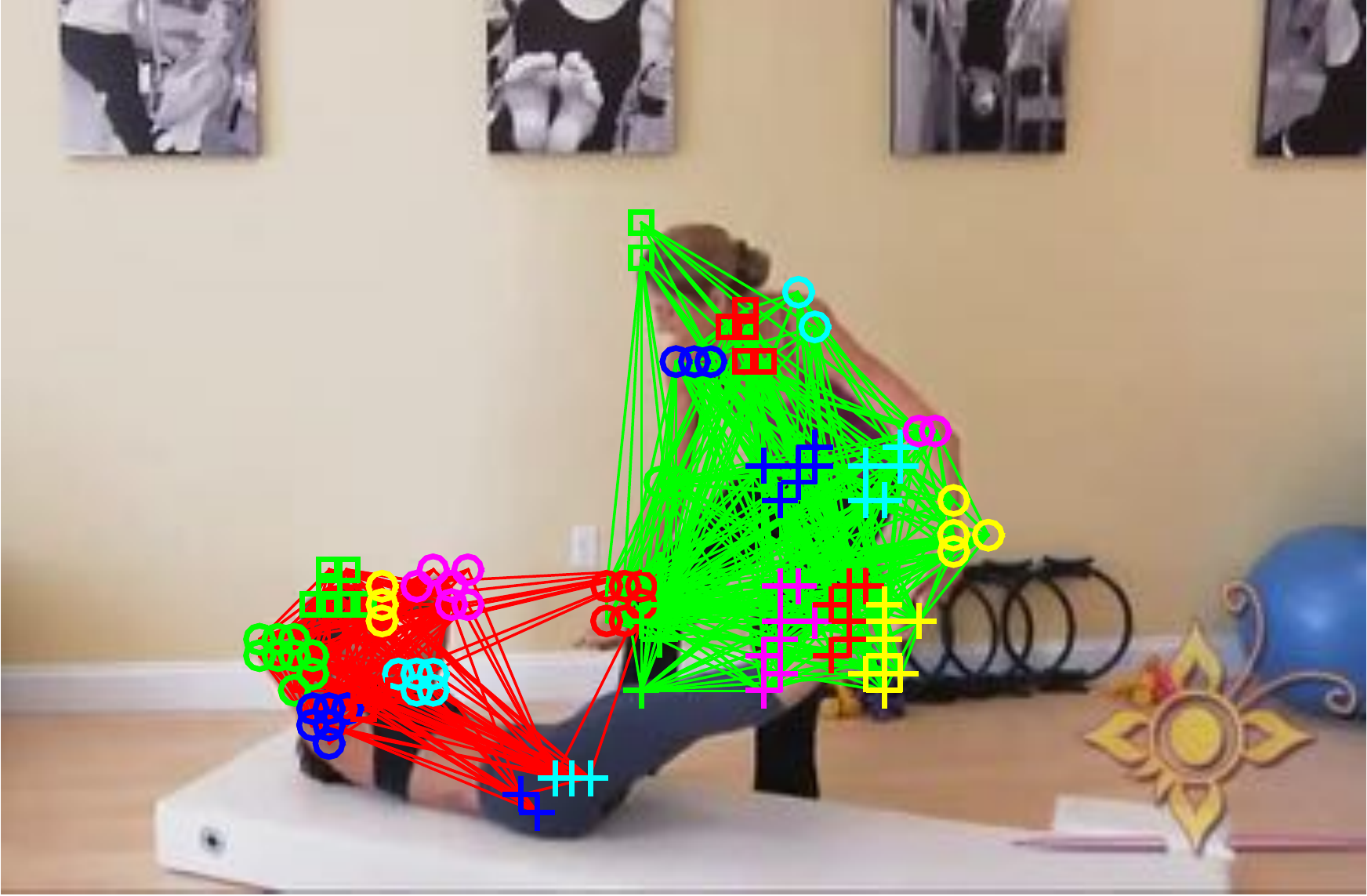}&
    \includegraphics[height=0.140\linewidth]{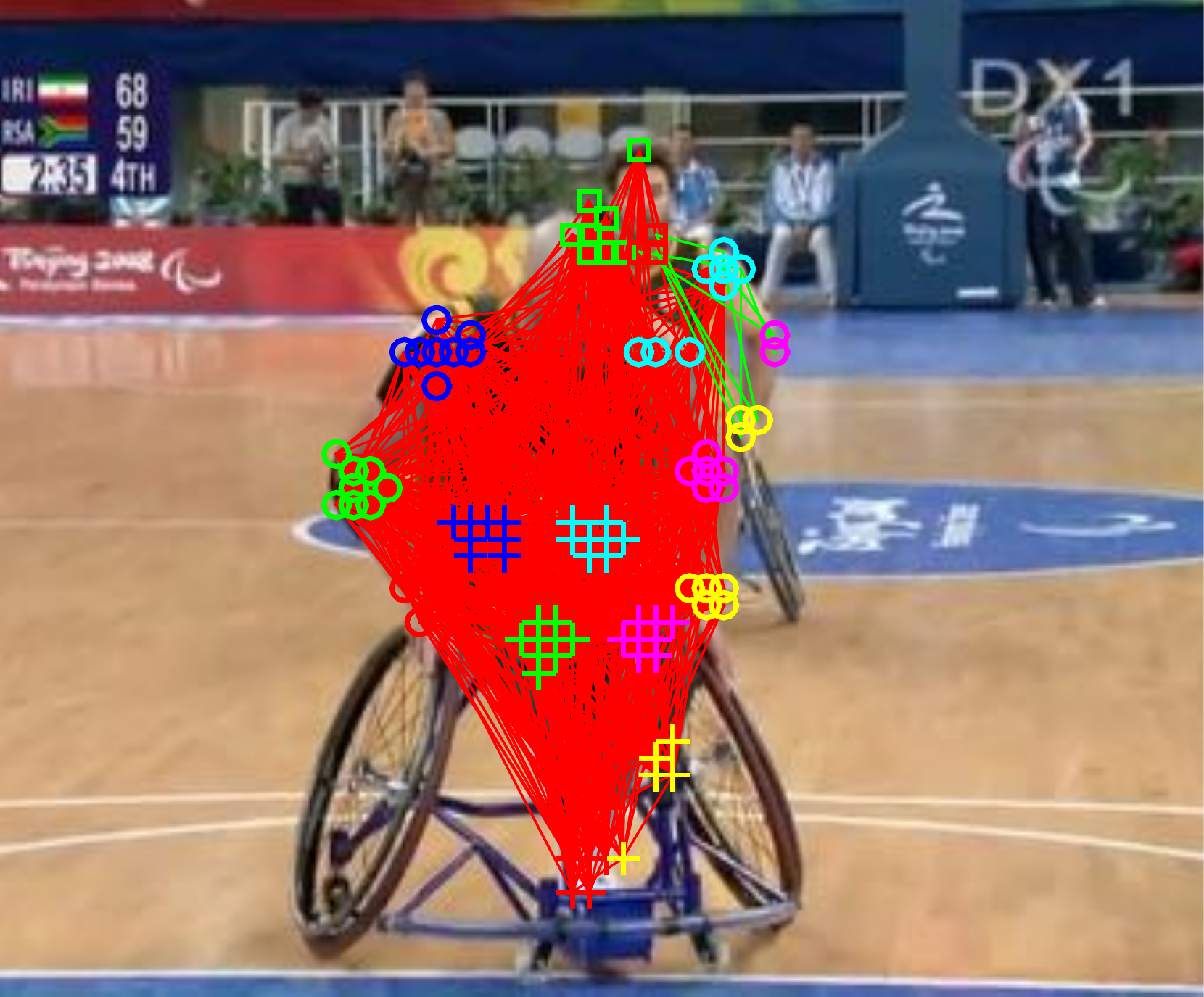}&
    \includegraphics[height=0.140\linewidth]{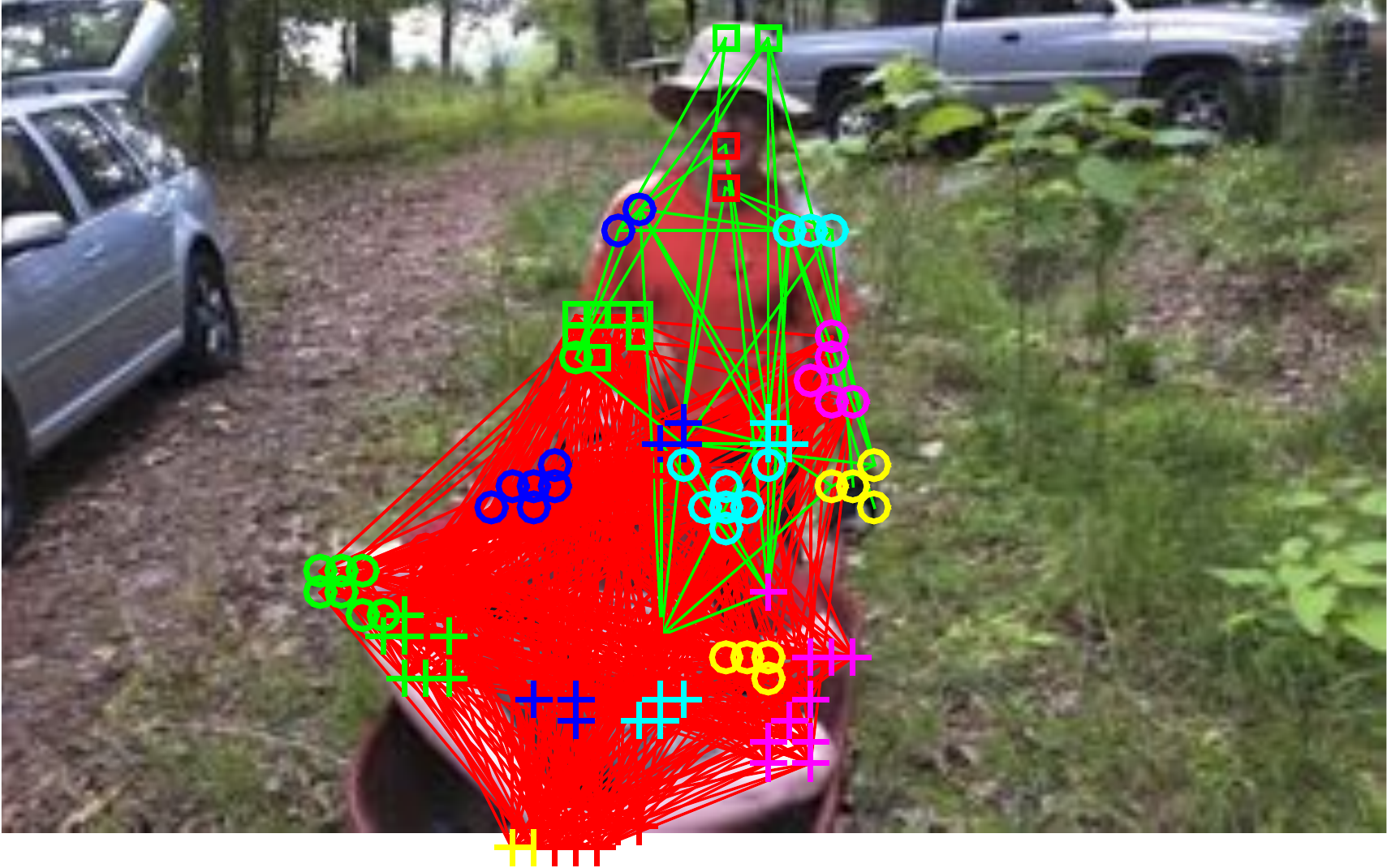}\\
    &
    \includegraphics[height=0.140\linewidth]{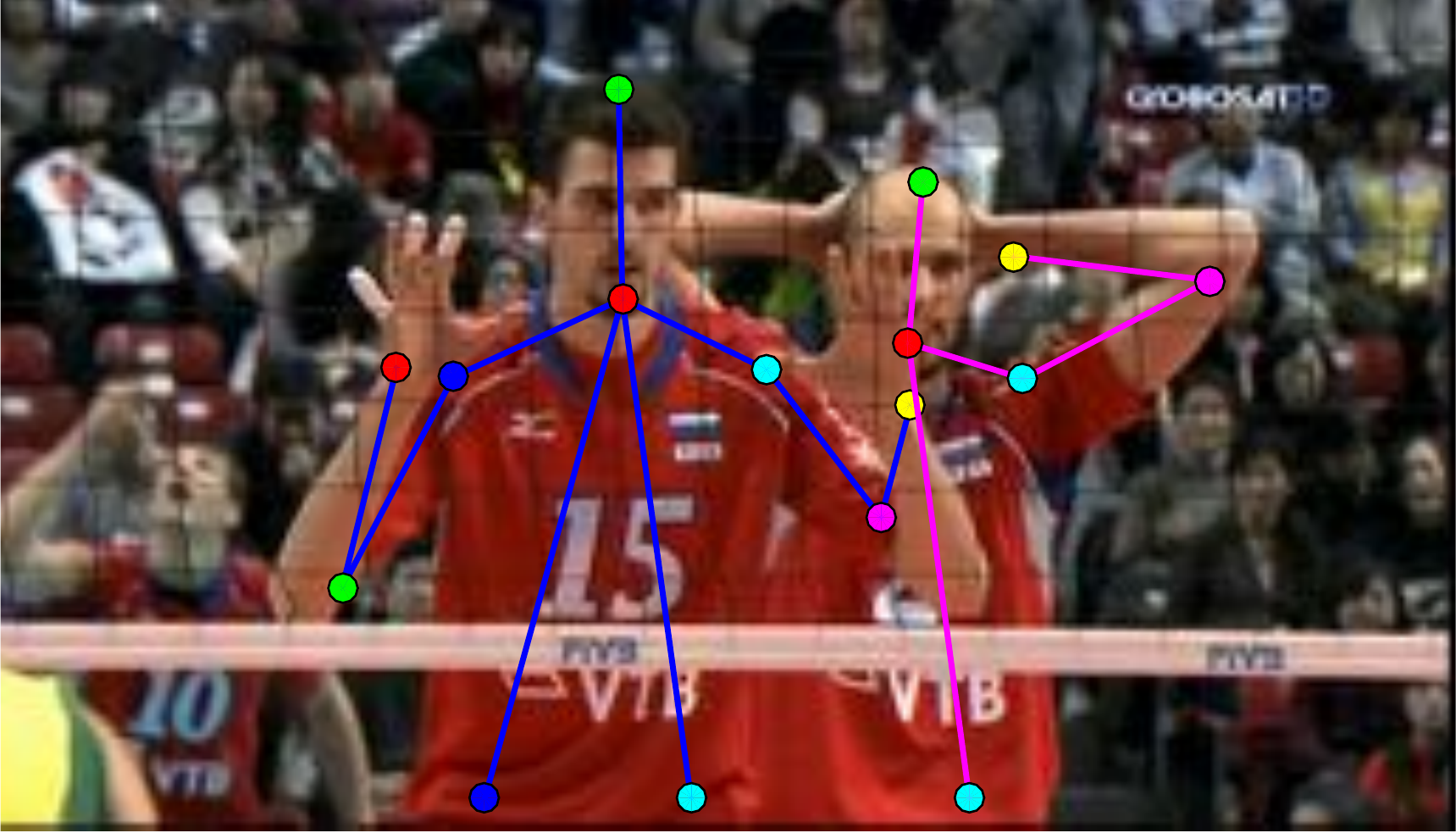}&
    \includegraphics[height=0.140\linewidth]{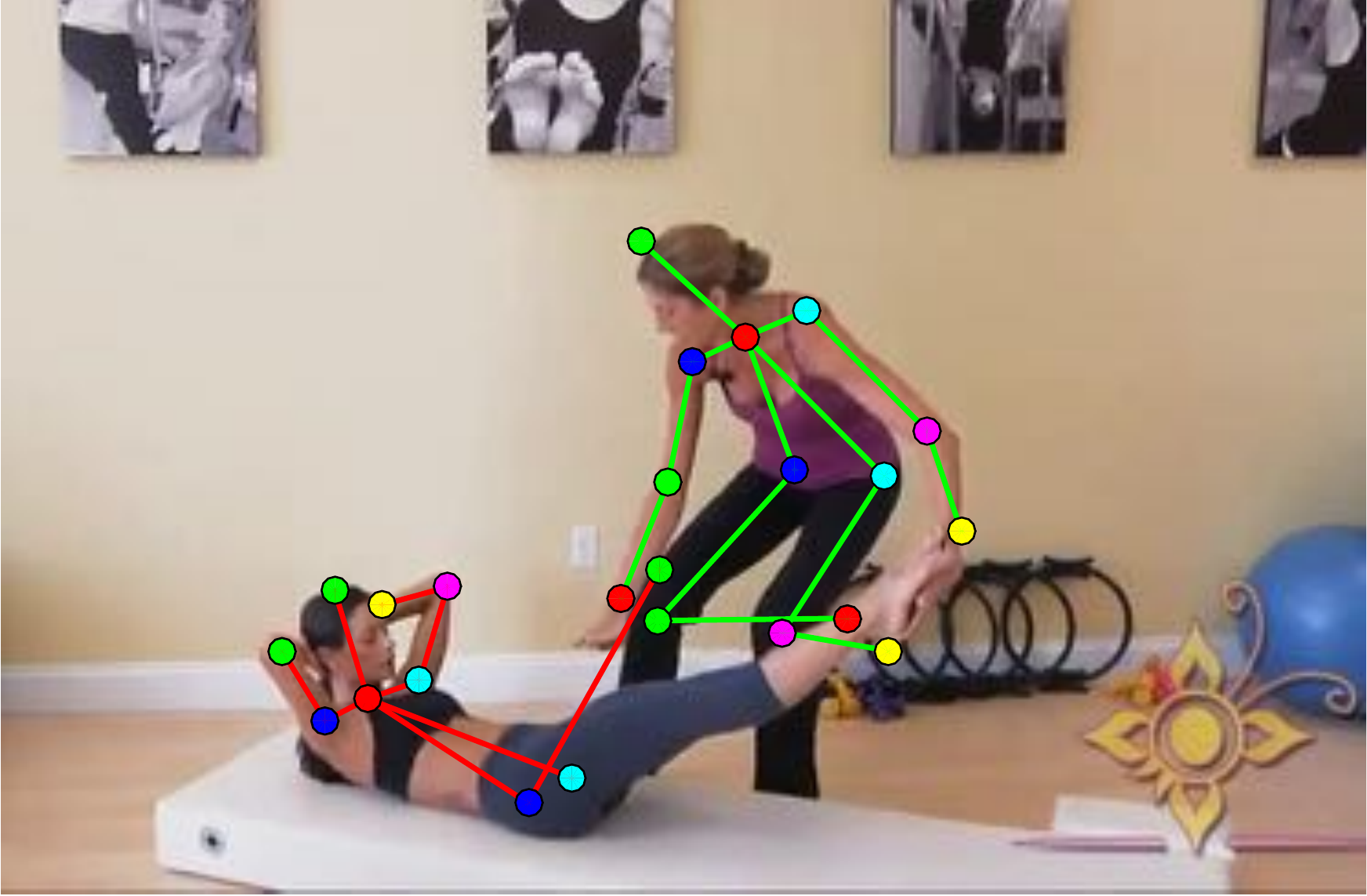}& 
    \includegraphics[height=0.140\linewidth]{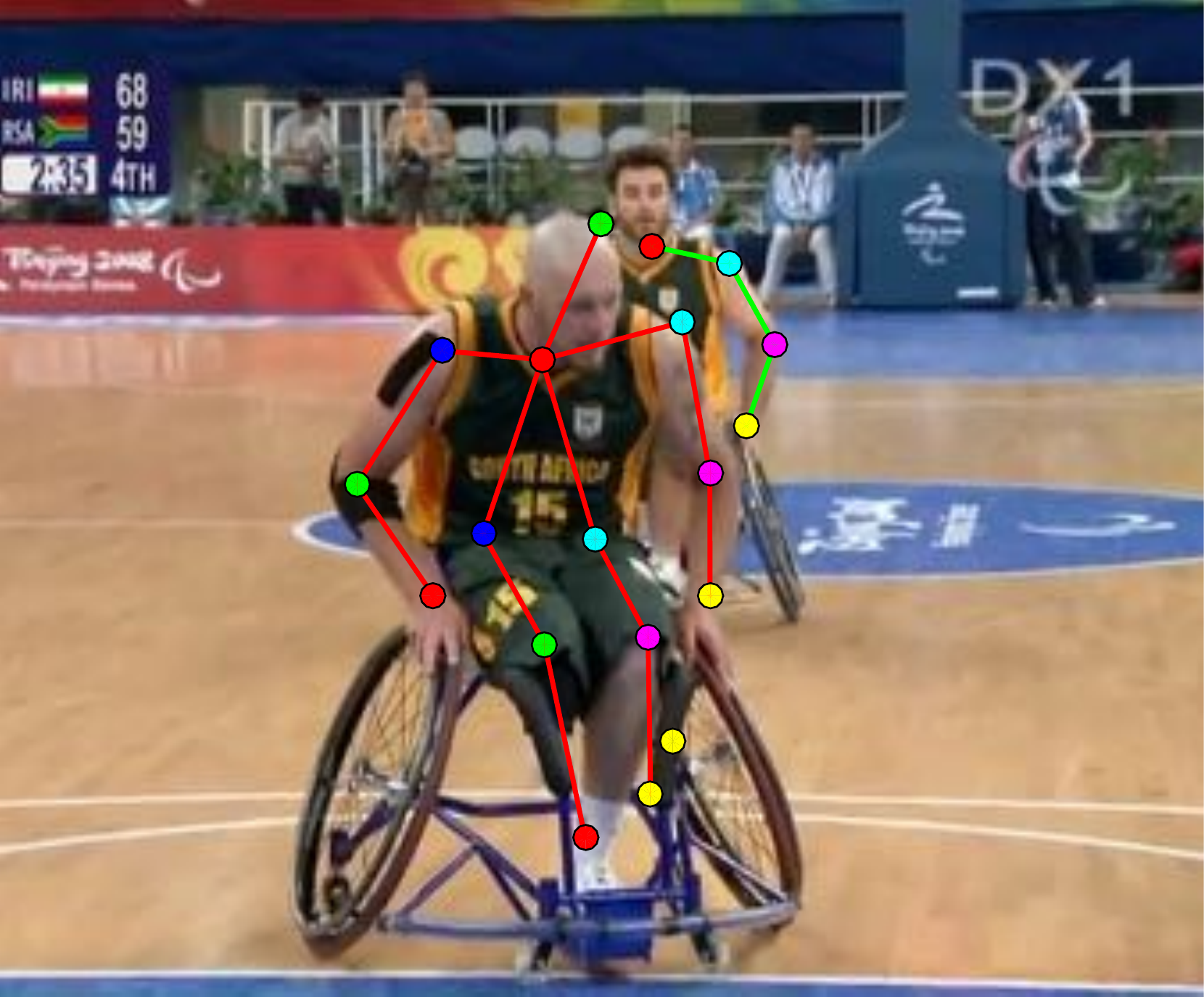}&
    \includegraphics[height=0.140\linewidth]{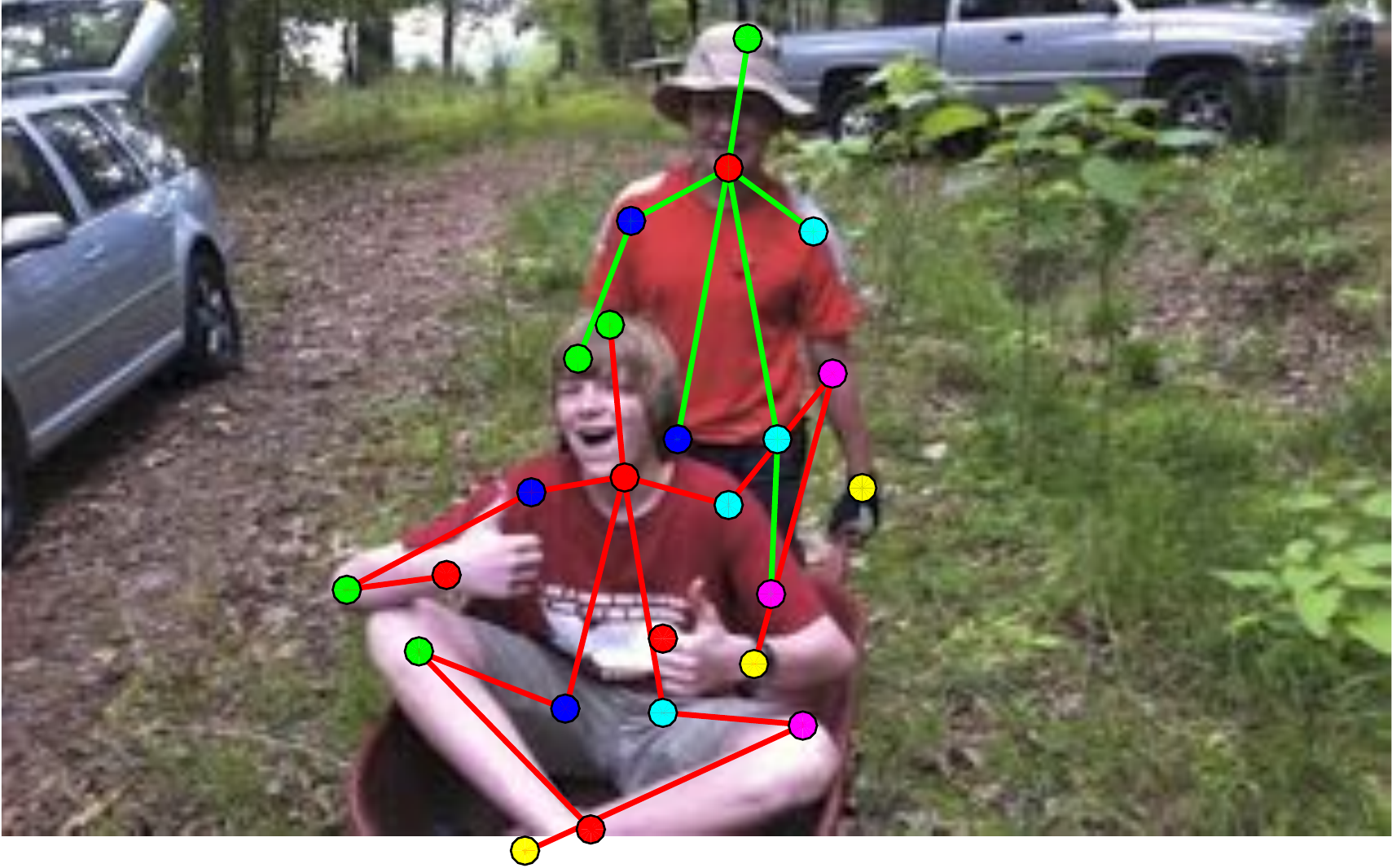}\\
    \midrule\midrule
    \begin{sideways}\bf \quad\quad$\detroi$\end{sideways}&

    \includegraphics[height=0.140\linewidth]{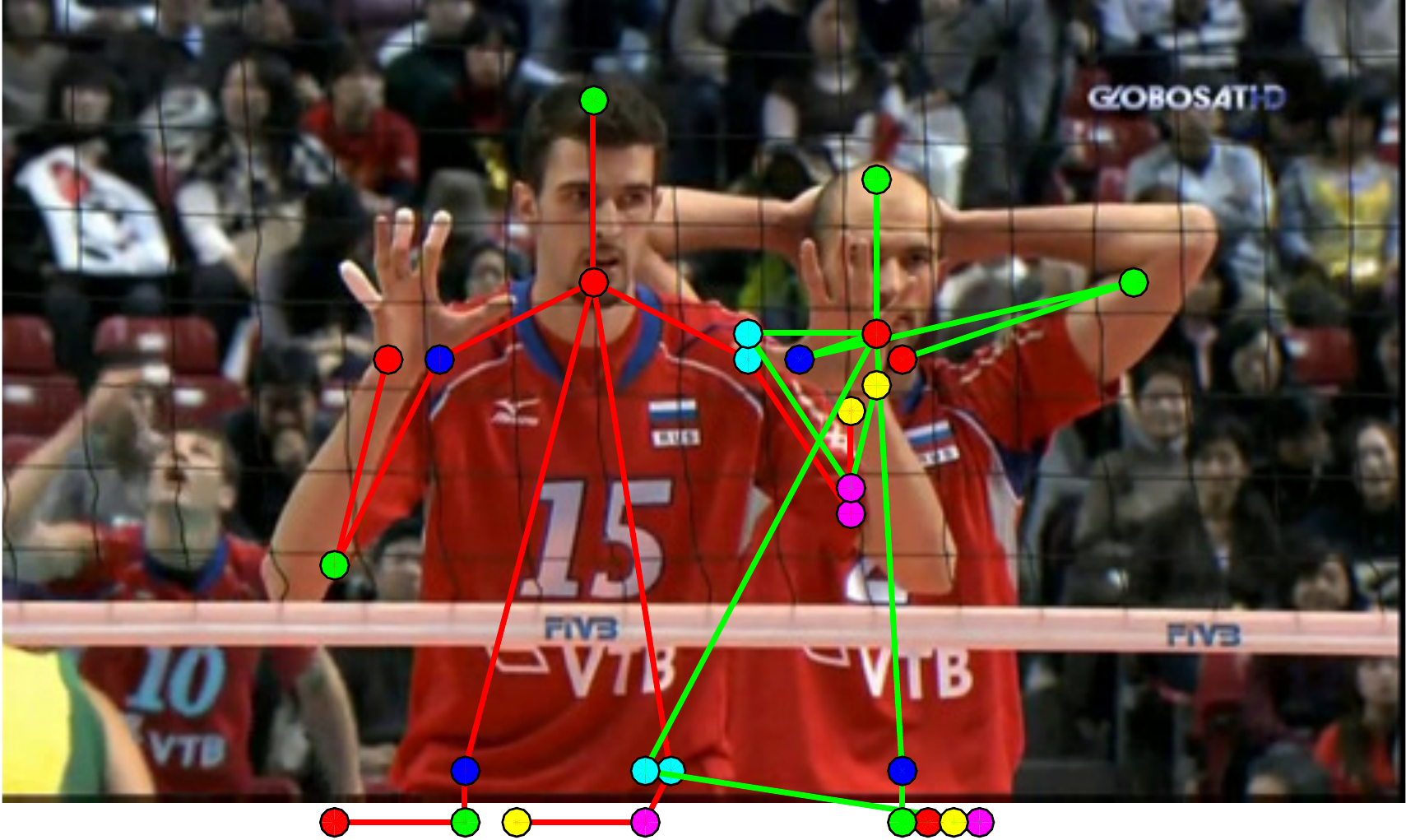}&
    \includegraphics[height=0.140\linewidth]{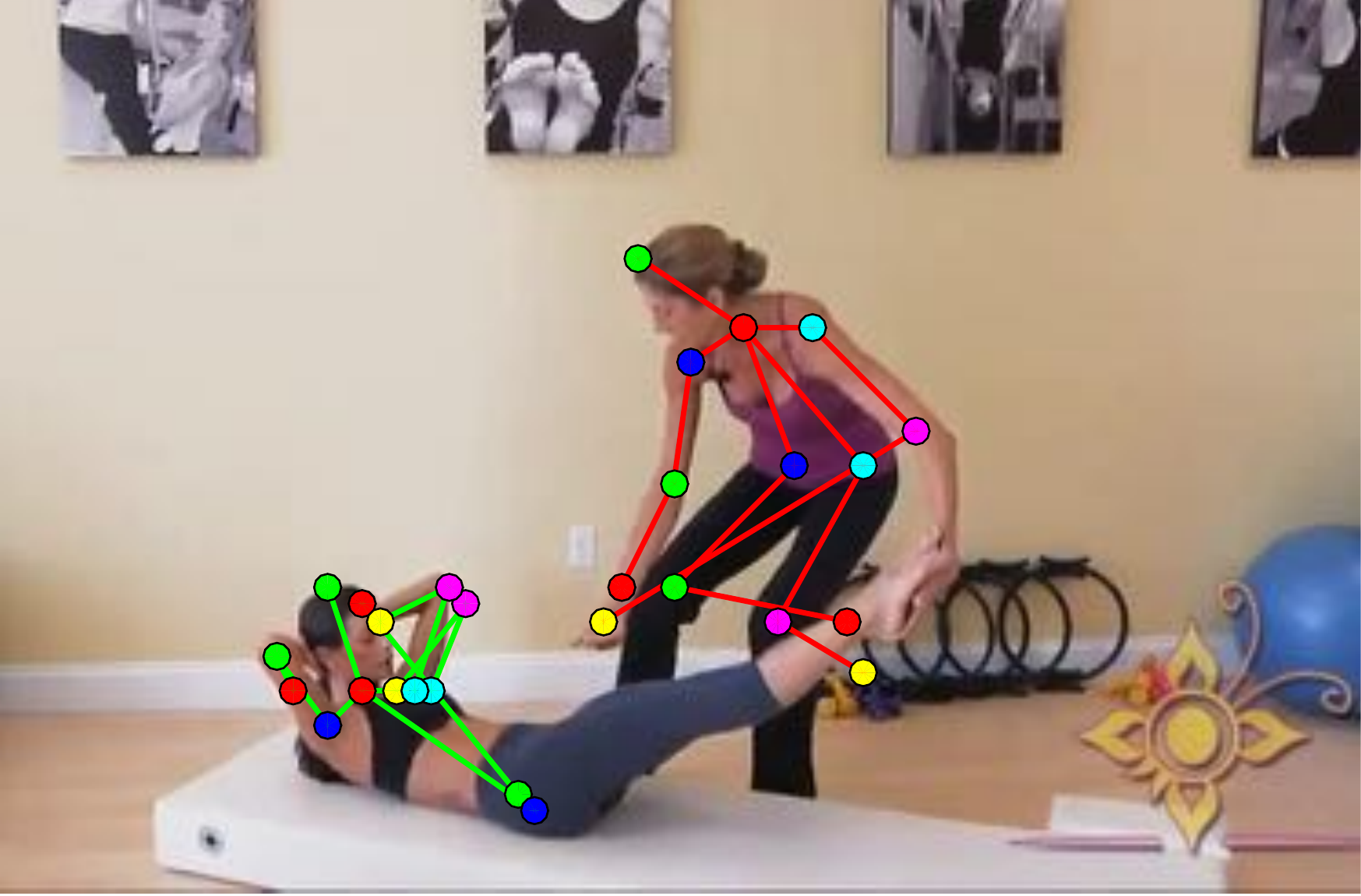}& 
    \includegraphics[height=0.140\linewidth]{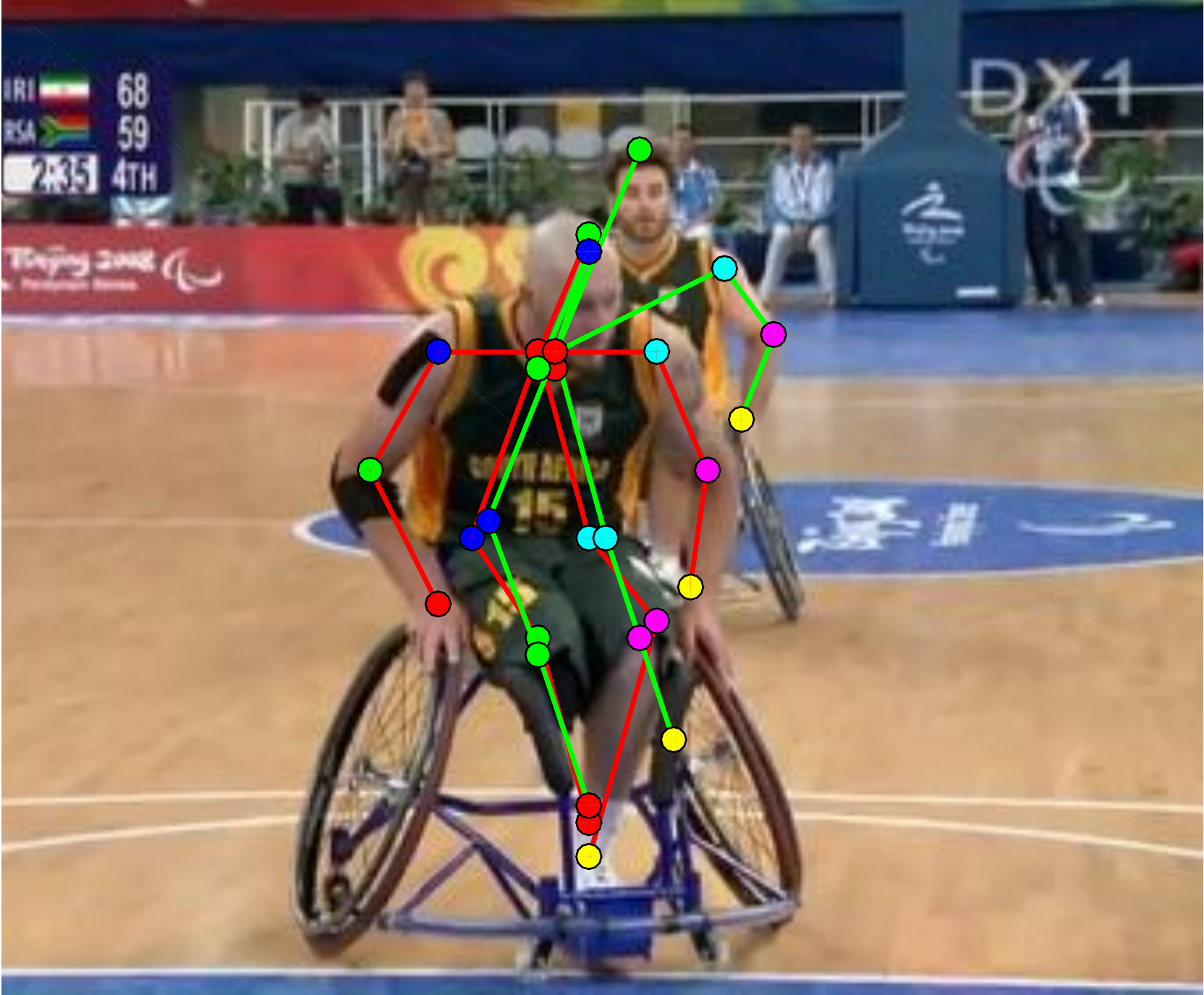}&
    \includegraphics[height=0.140\linewidth]{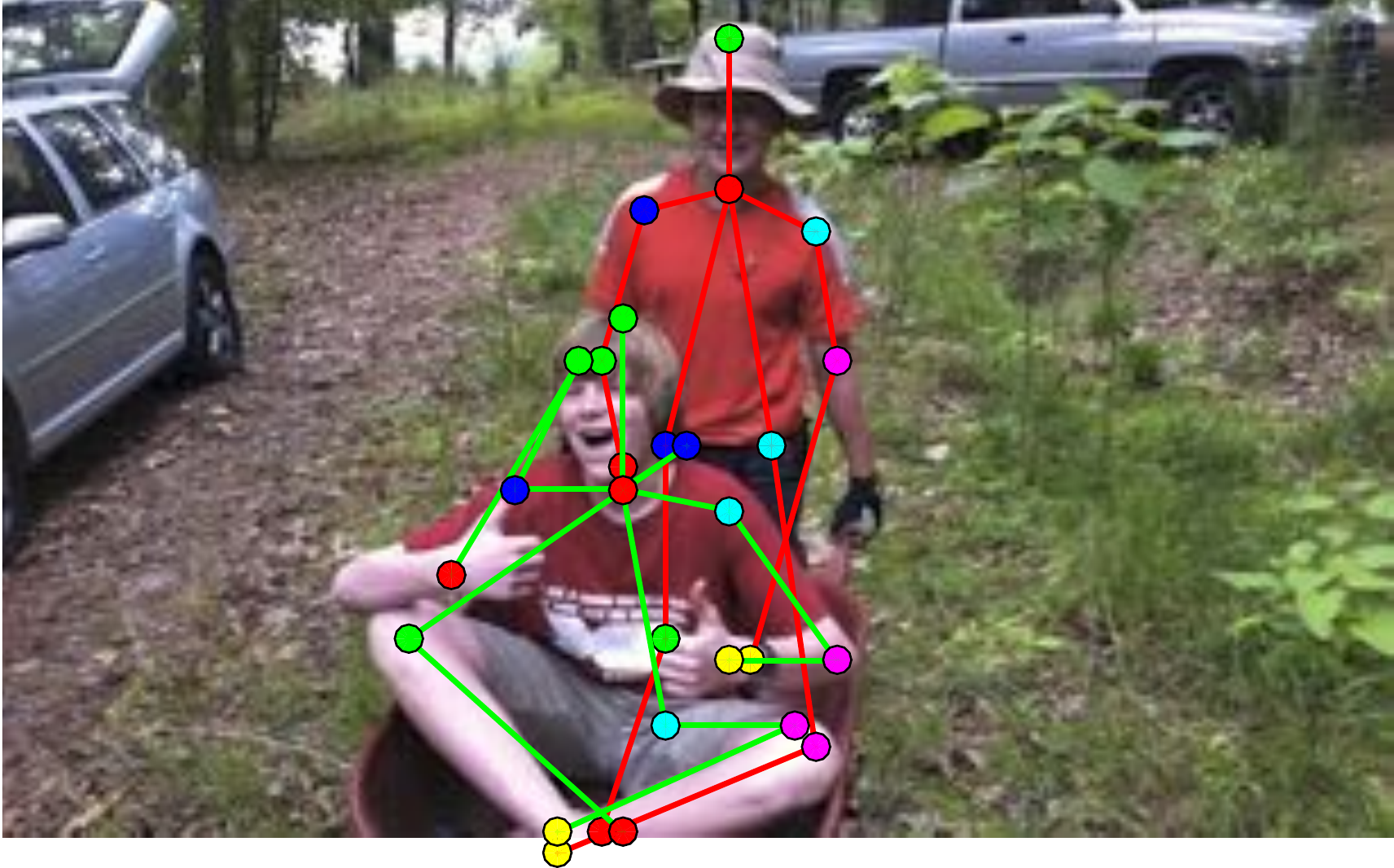}\\
  &6&7&8&9\\  
  \end{tabular}
  %\vspace{-1em}
  \caption{Qualitative comparison of our joint formulation
    $\deepcut~\multb~\dense$ (rows 1-3, 5-7) to the traditional
    two-stage approach $\dense~\detroi$ (rows 4, 8) on MPII
    %Multi-Person dataset. 
    See Fig.~\ref{fig:overview} for the color-coding explanation.
    %See Fig.~\textcolor{red}{1} in paper for the color-coding explanation.
}
  \label{fig:qualitative_mpii}
\end{figure*}

\begin{figure*}
  \centering
  \begin{tabular}{c c c c c c c}
    &
    \includegraphics[height=0.140\linewidth]{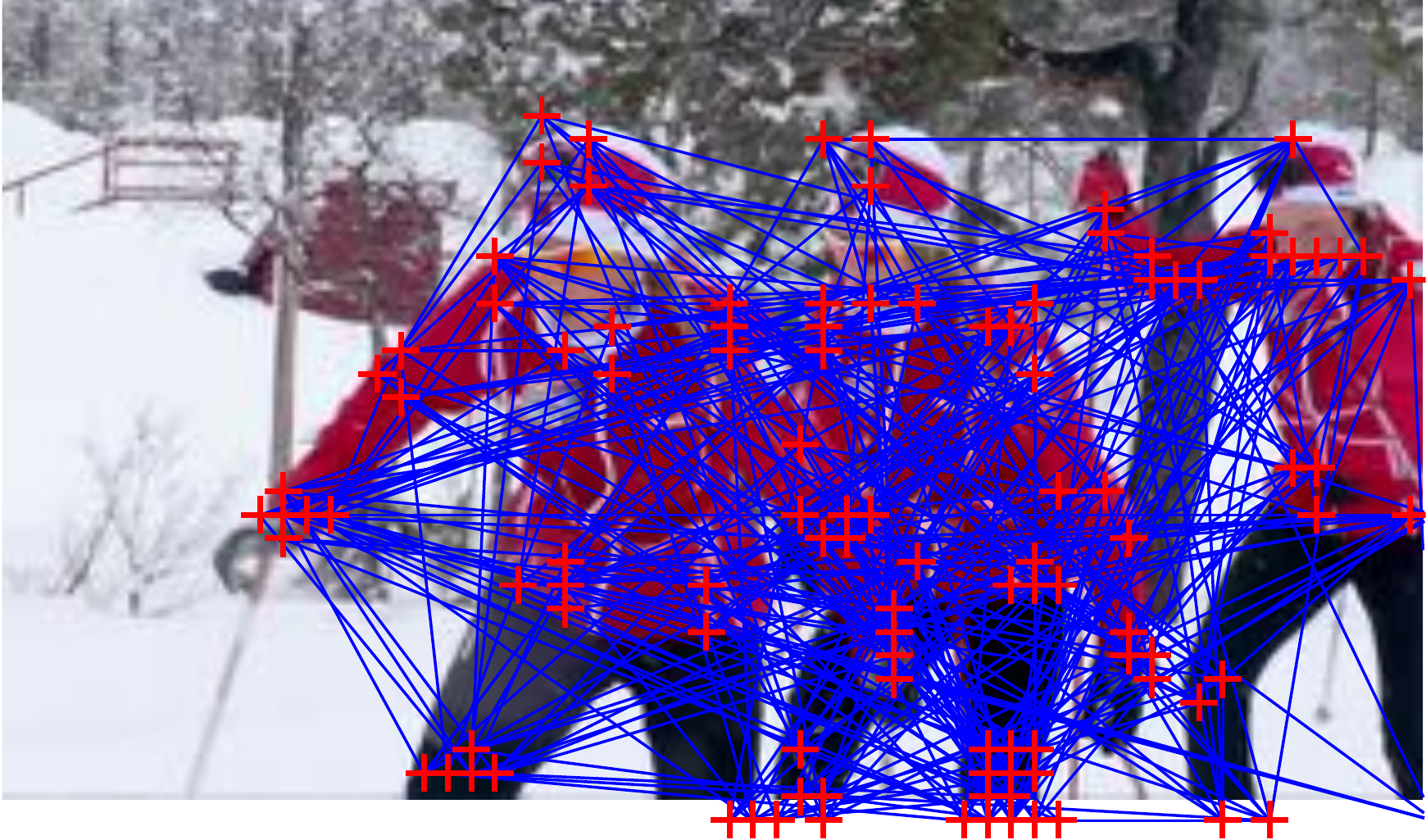}&
    \includegraphics[height=0.140\linewidth]{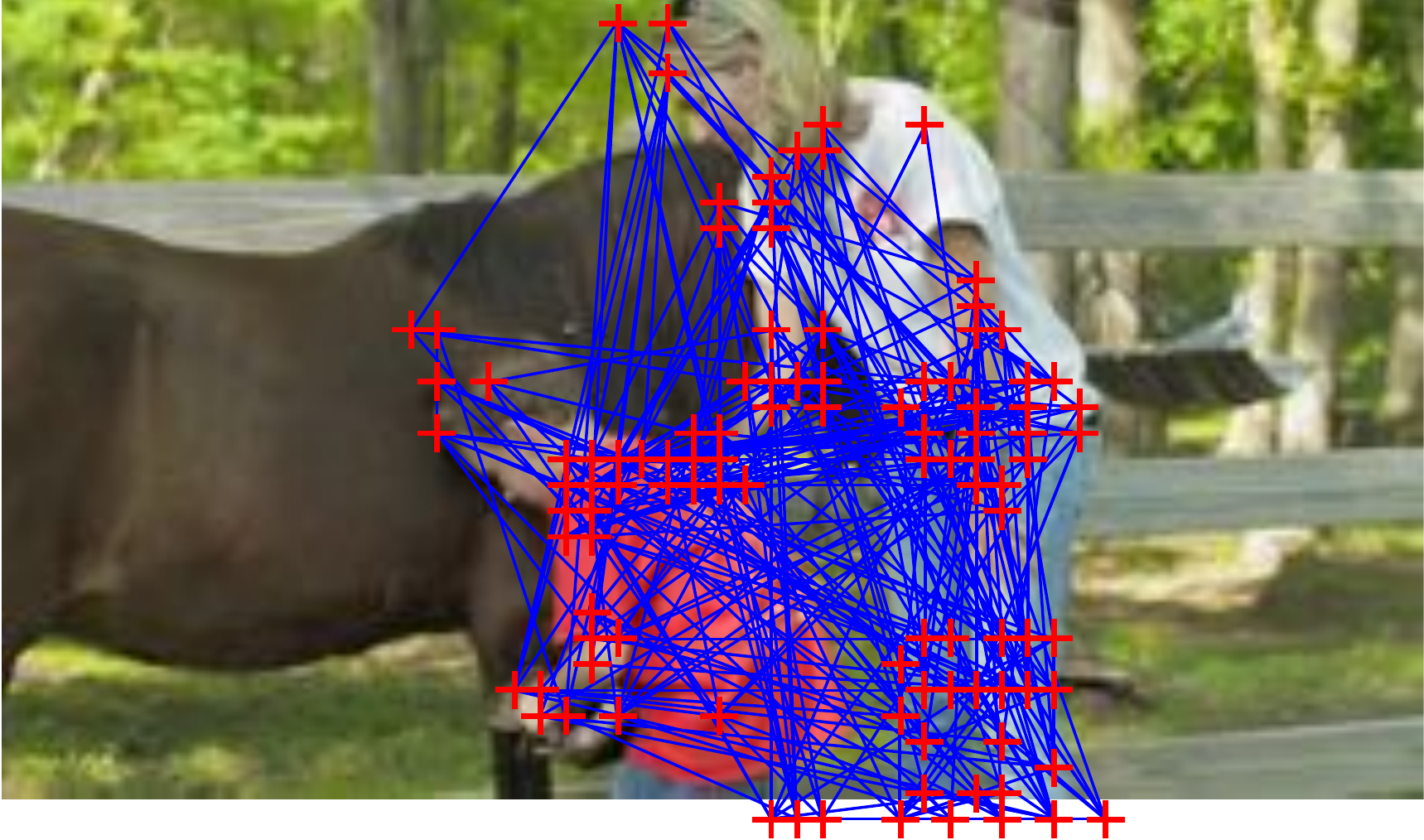}&
    \includegraphics[height=0.140\linewidth]{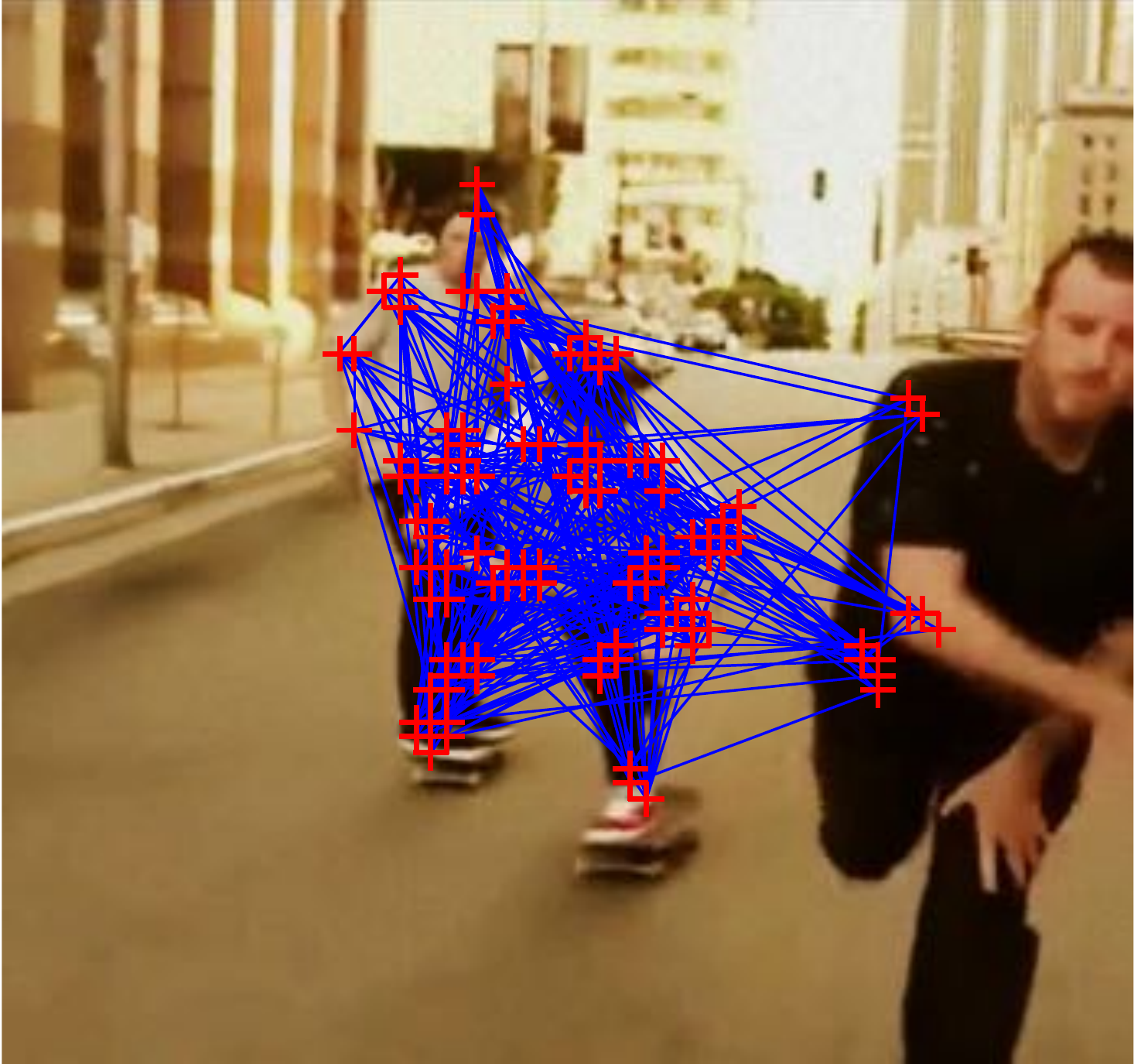}&
    \includegraphics[height=0.140\linewidth]{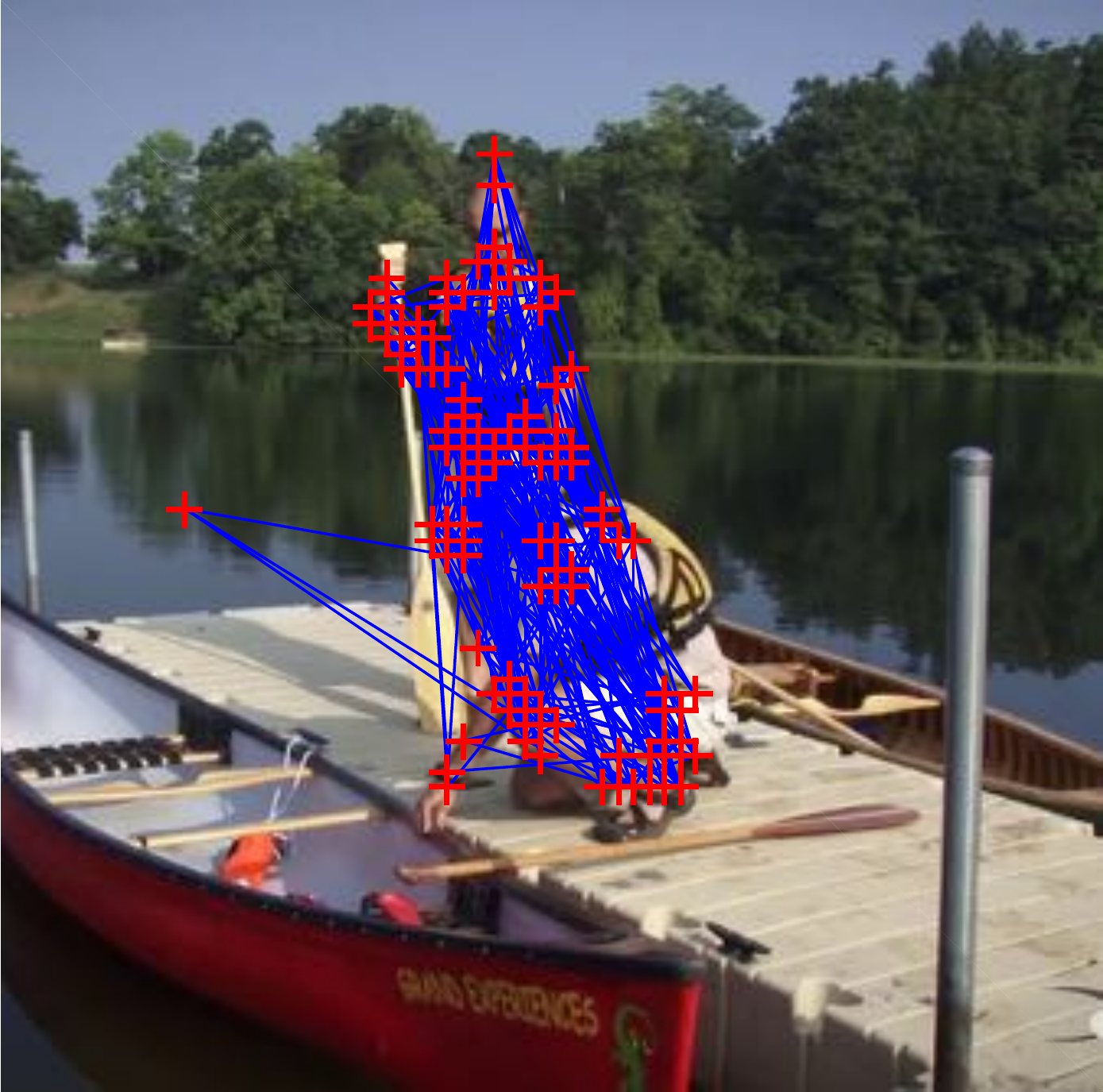}&
    \includegraphics[height=0.140\linewidth]{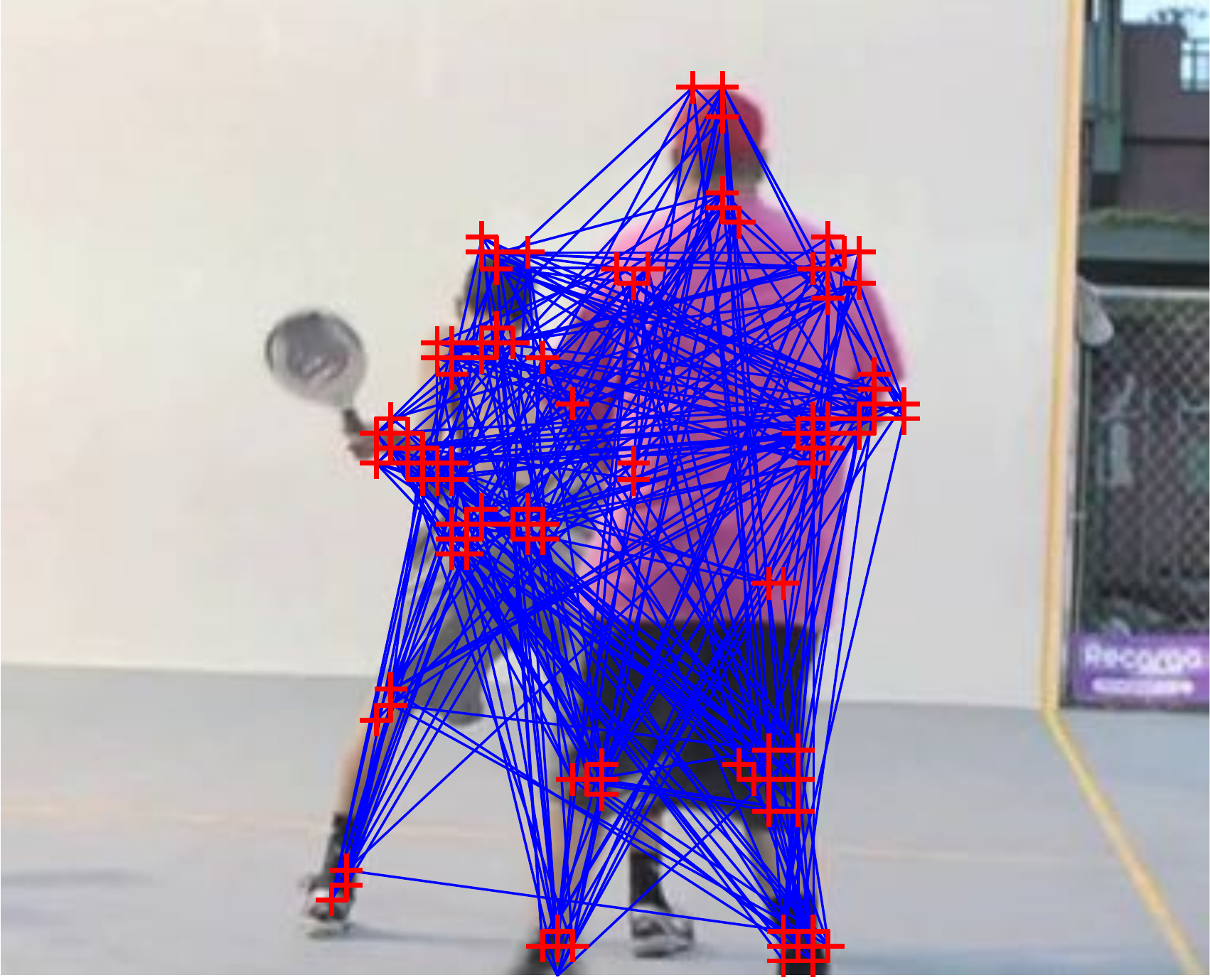}\\
    \begin{sideways}\bf\quad $\deepcut~\multb$\end{sideways}&        
    \includegraphics[height=0.140\linewidth]{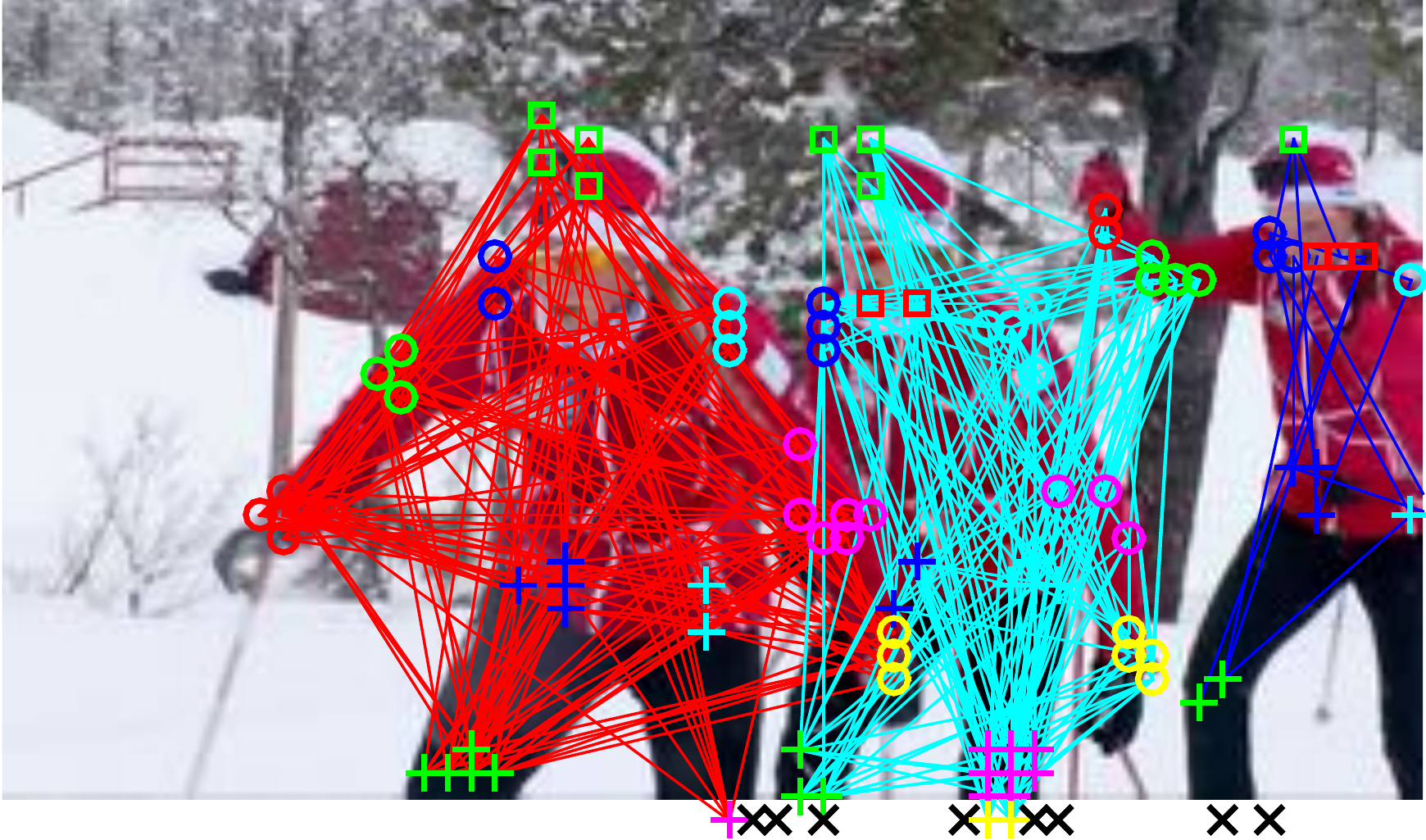}&
    \includegraphics[height=0.140\linewidth]{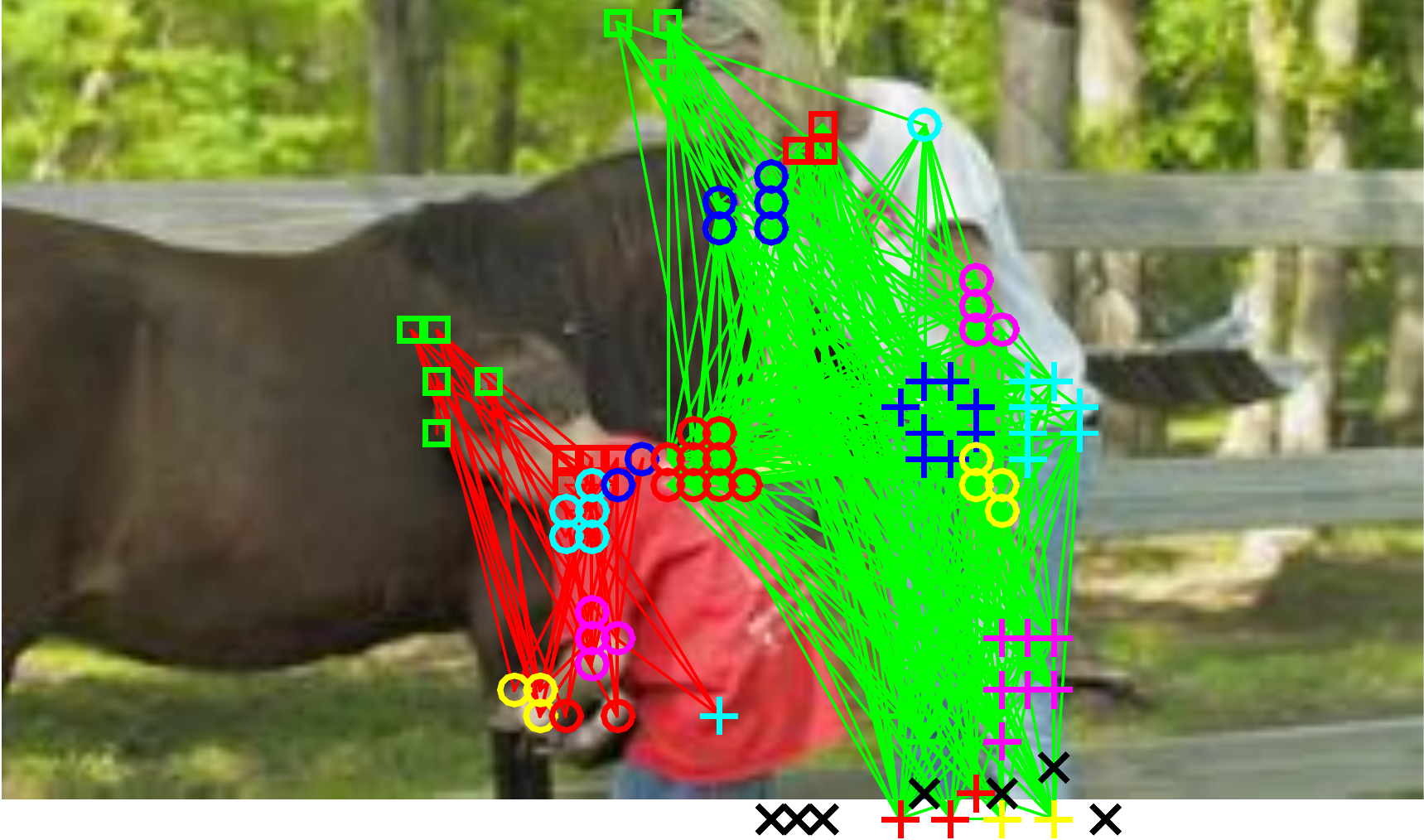}&
    \includegraphics[height=0.140\linewidth]{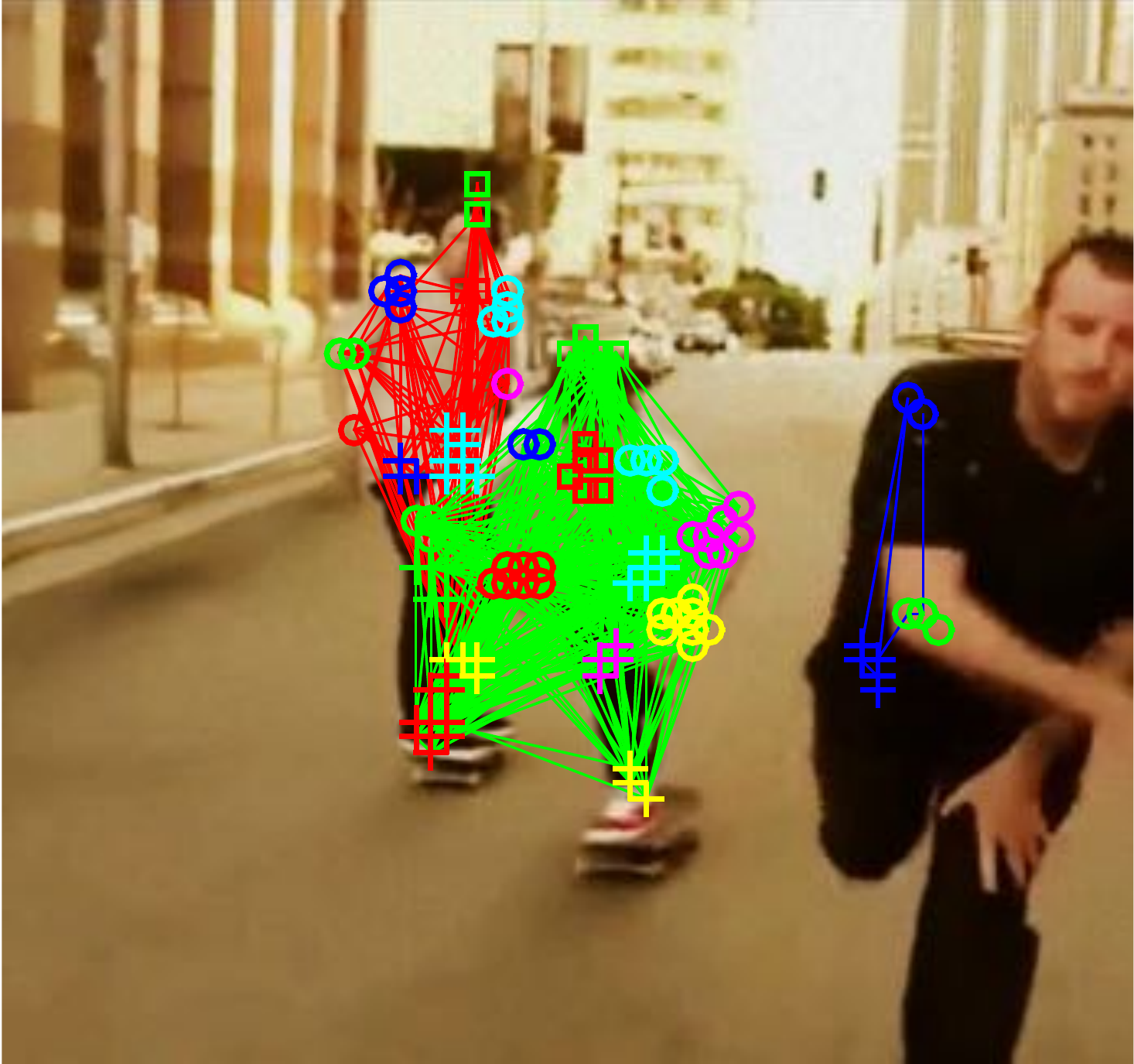}&
    \includegraphics[height=0.140\linewidth]{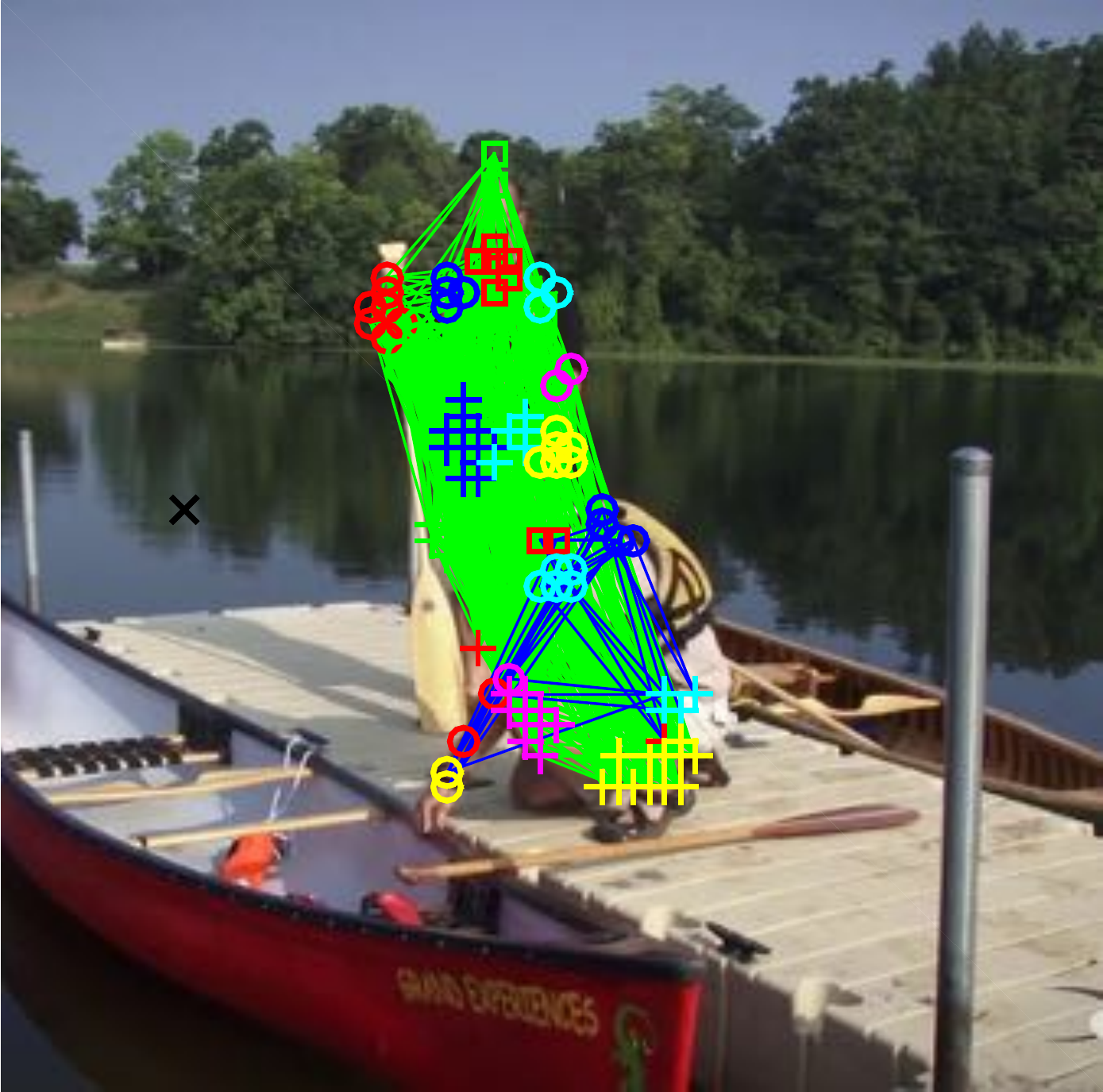}&
    \includegraphics[height=0.140\linewidth]{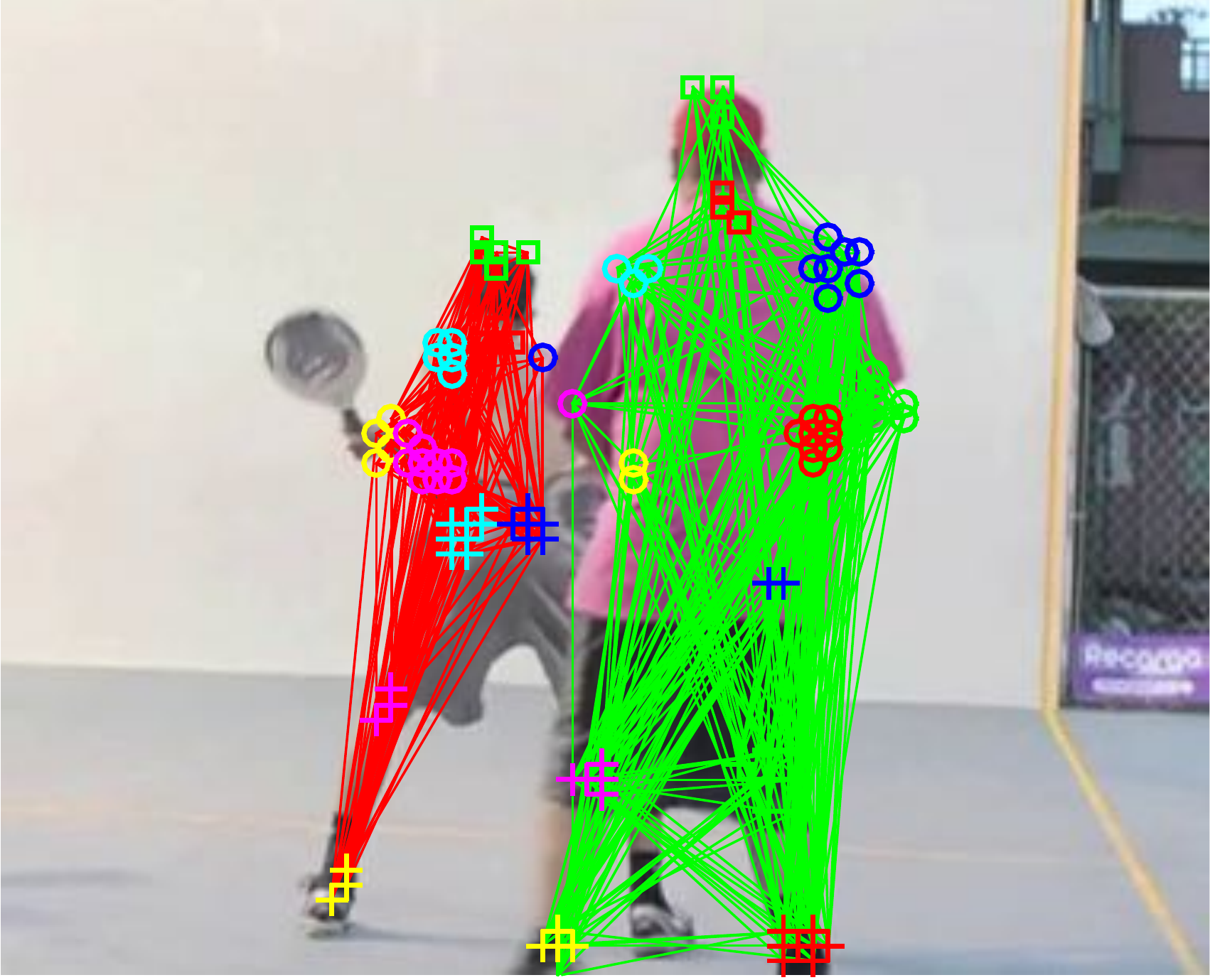}\\
    &
    \includegraphics[height=0.140\linewidth]{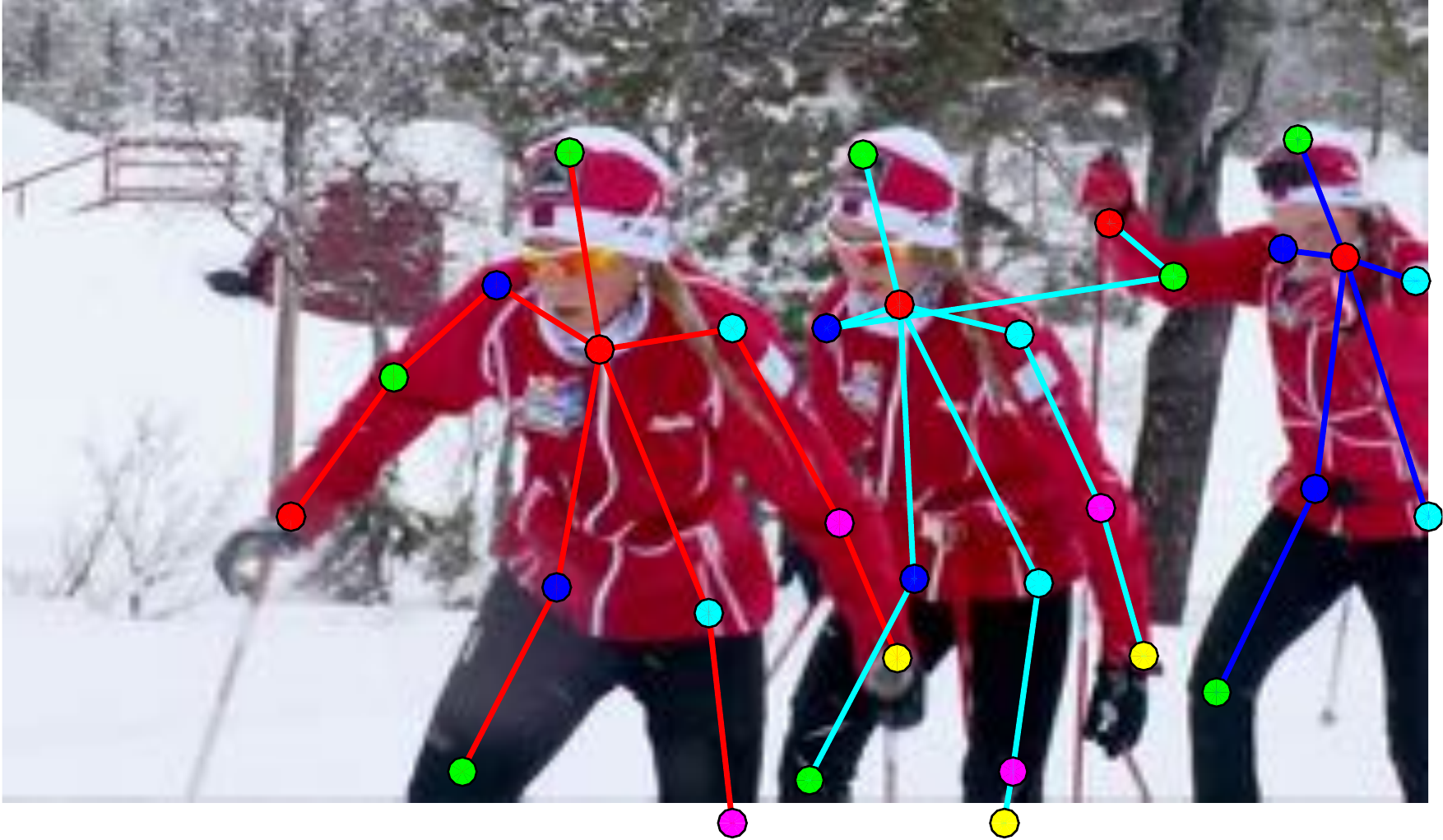}&
    \includegraphics[height=0.140\linewidth]{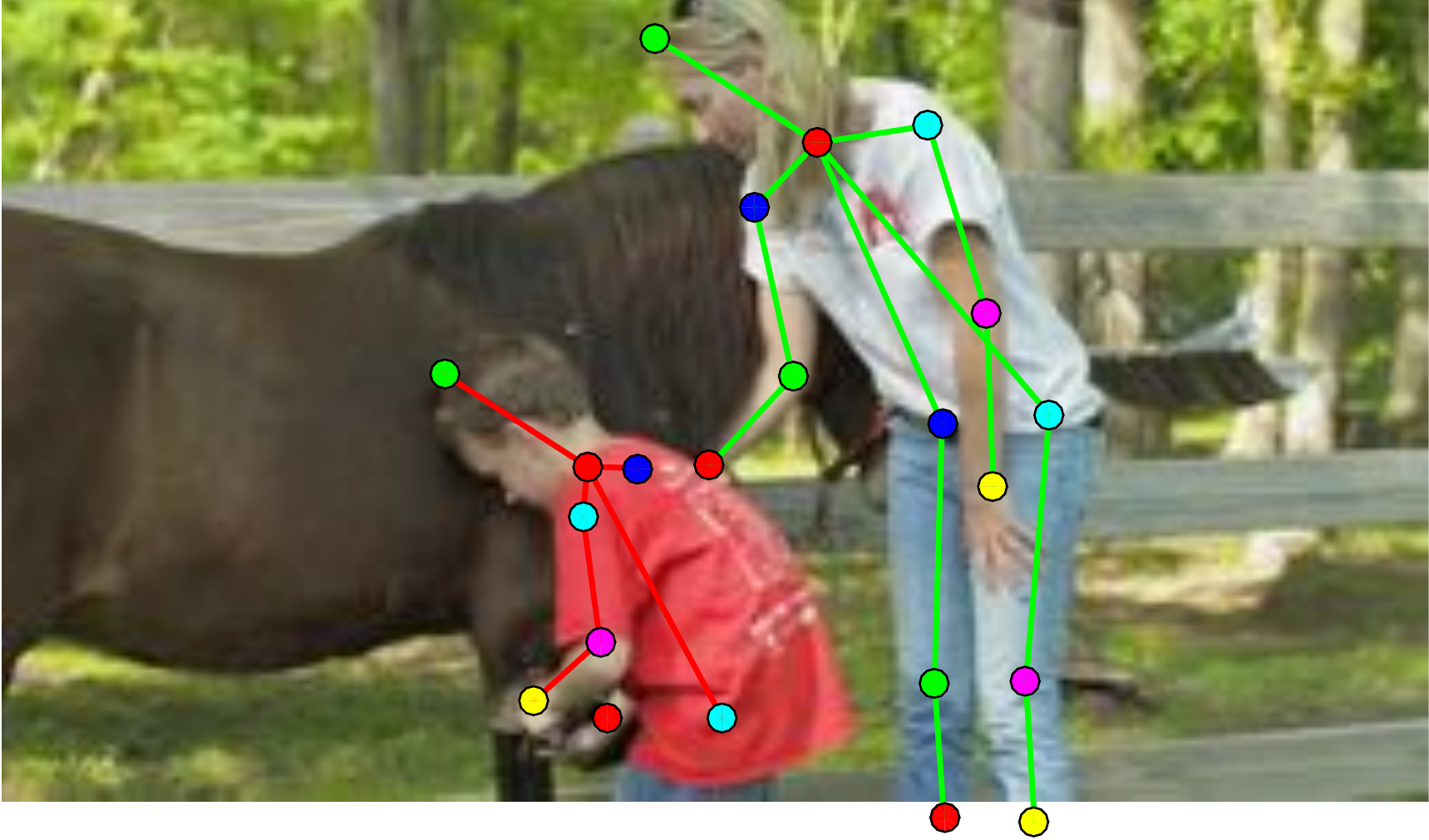}& 
    \includegraphics[height=0.140\linewidth]{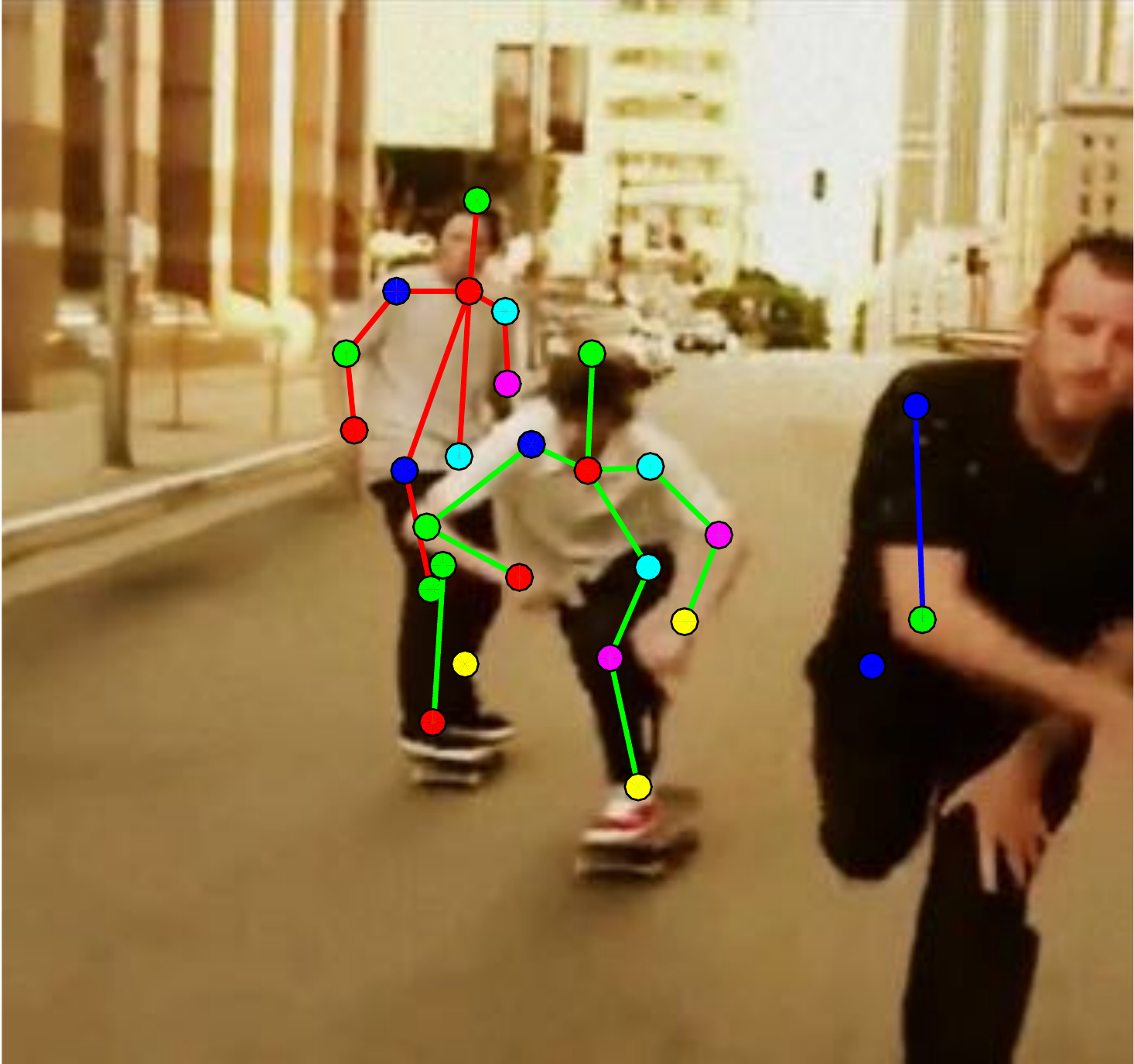}&
    \includegraphics[height=0.140\linewidth]{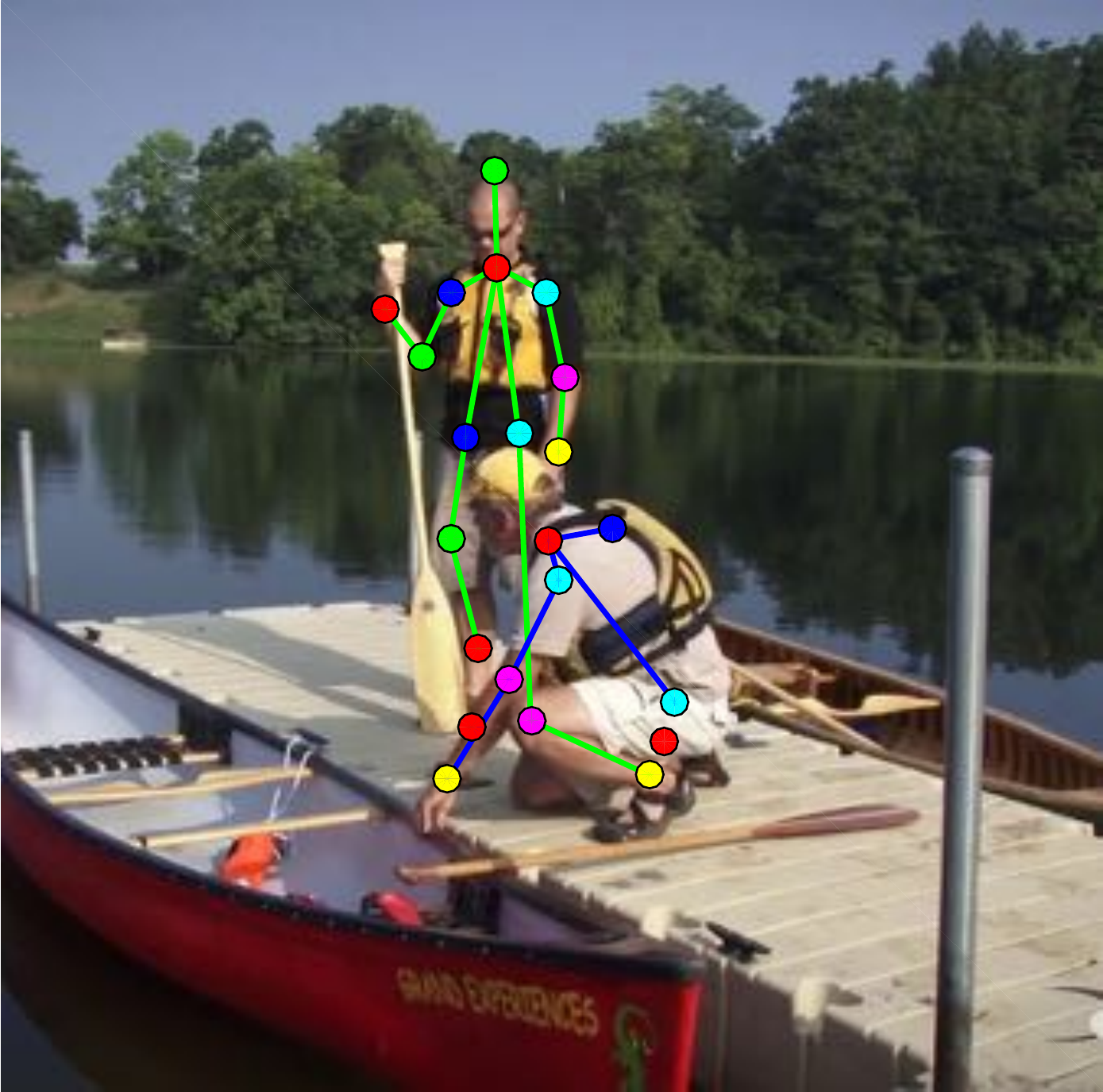}&
    \includegraphics[height=0.140\linewidth]{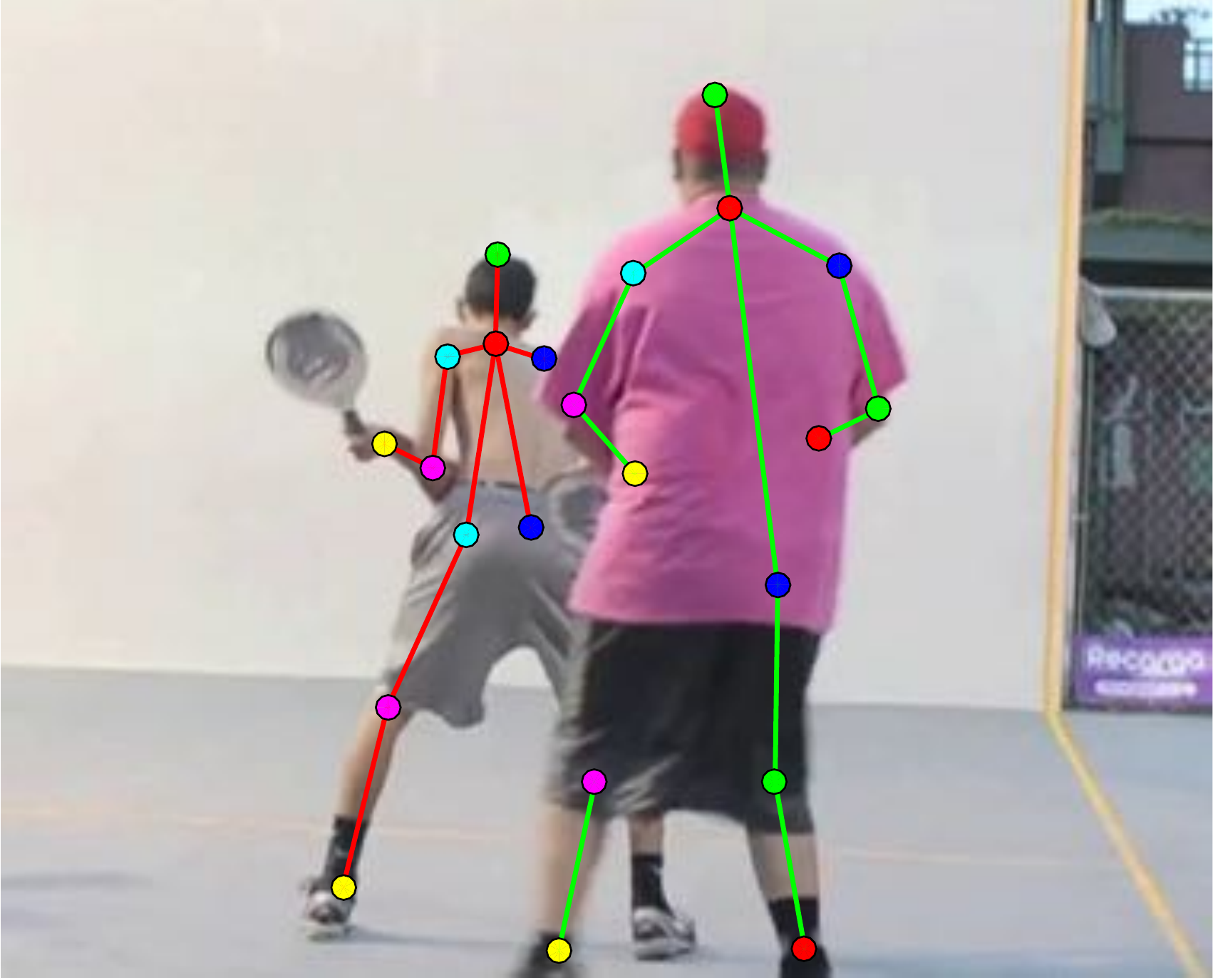}\\

    \midrule\midrule
    \begin{sideways}\bf \quad\quad$\detroi$\end{sideways}&
    \includegraphics[height=0.140\linewidth]{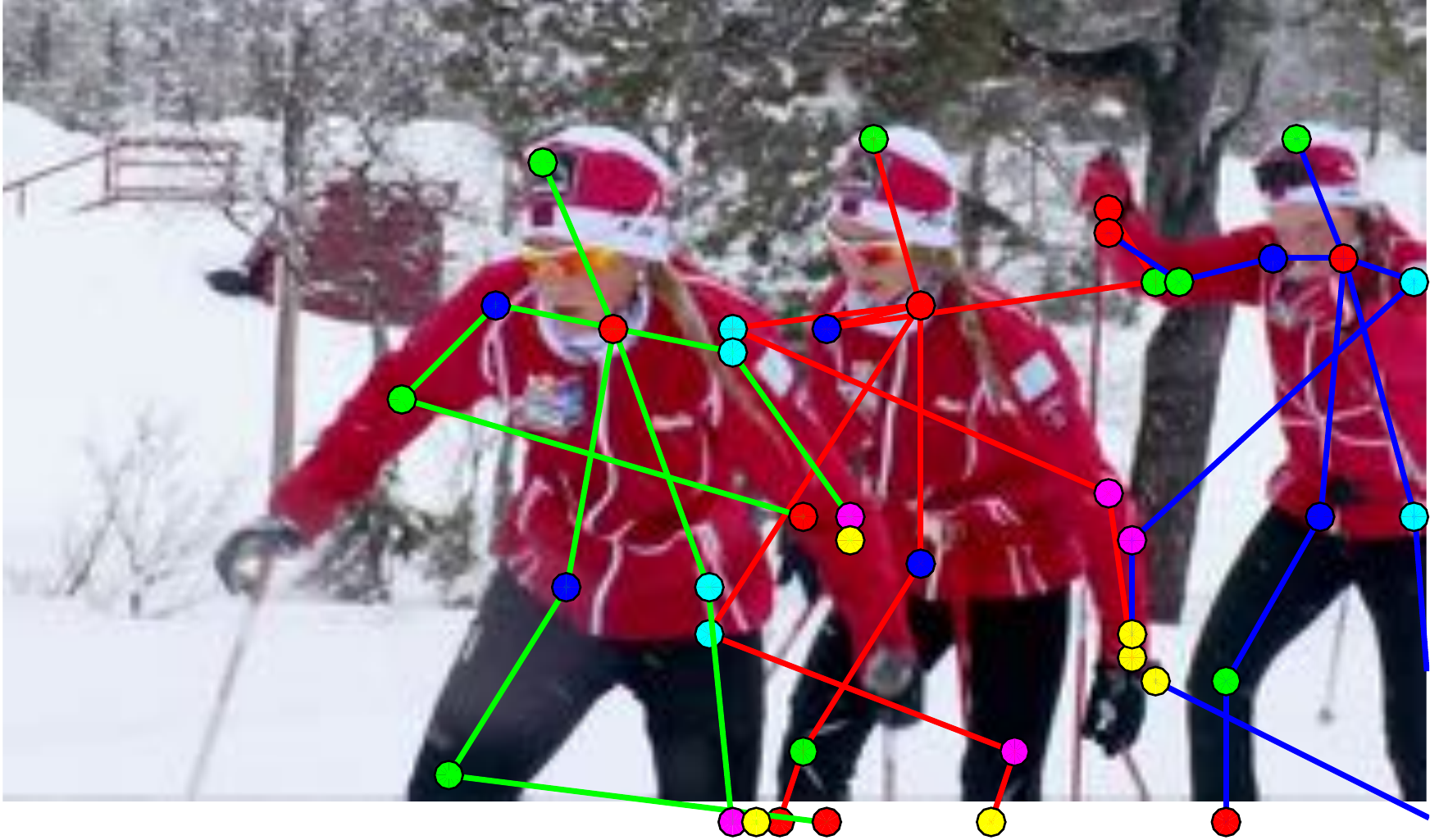}&
    \includegraphics[height=0.140\linewidth]{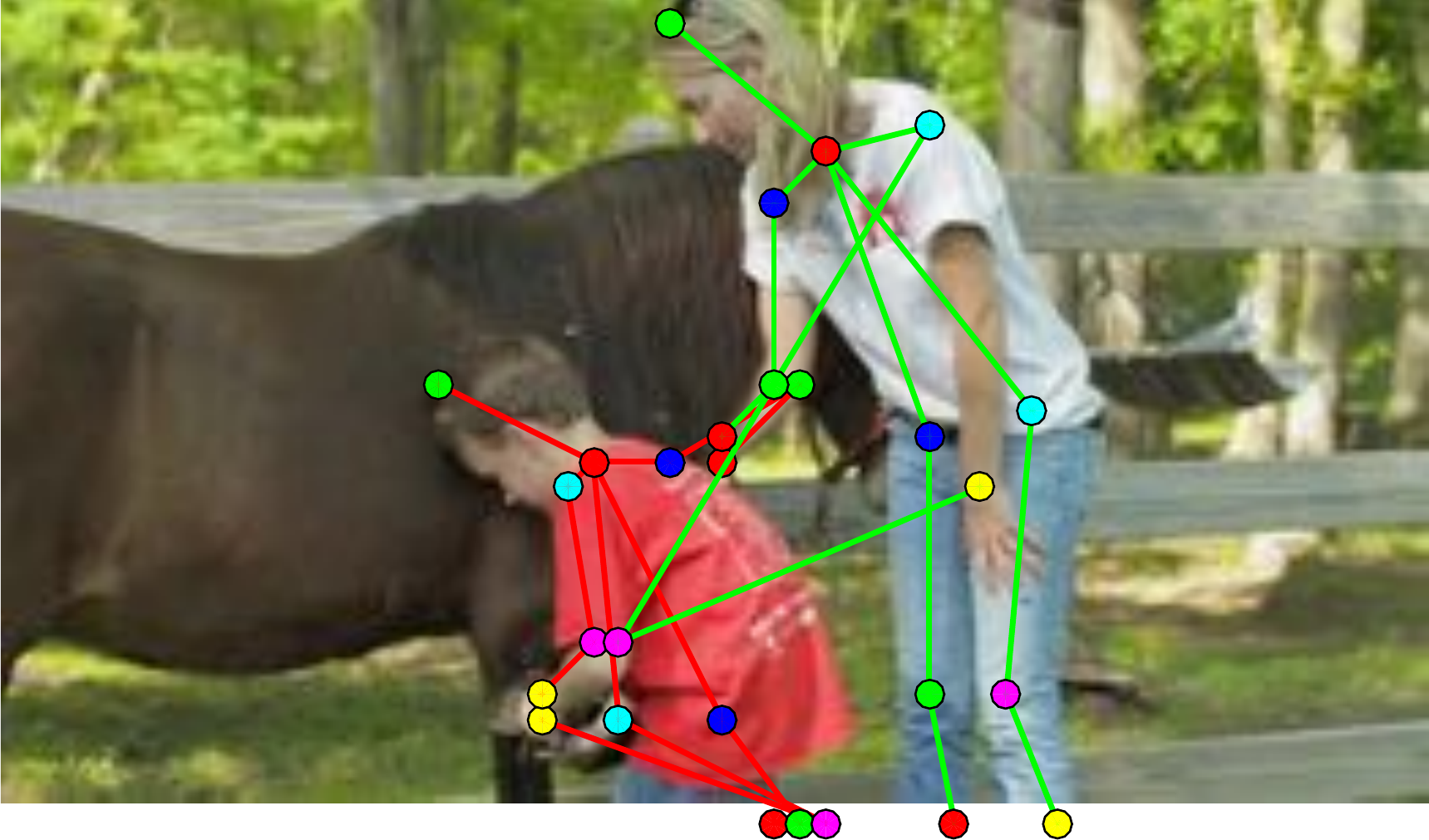}& 
    \includegraphics[height=0.140\linewidth]{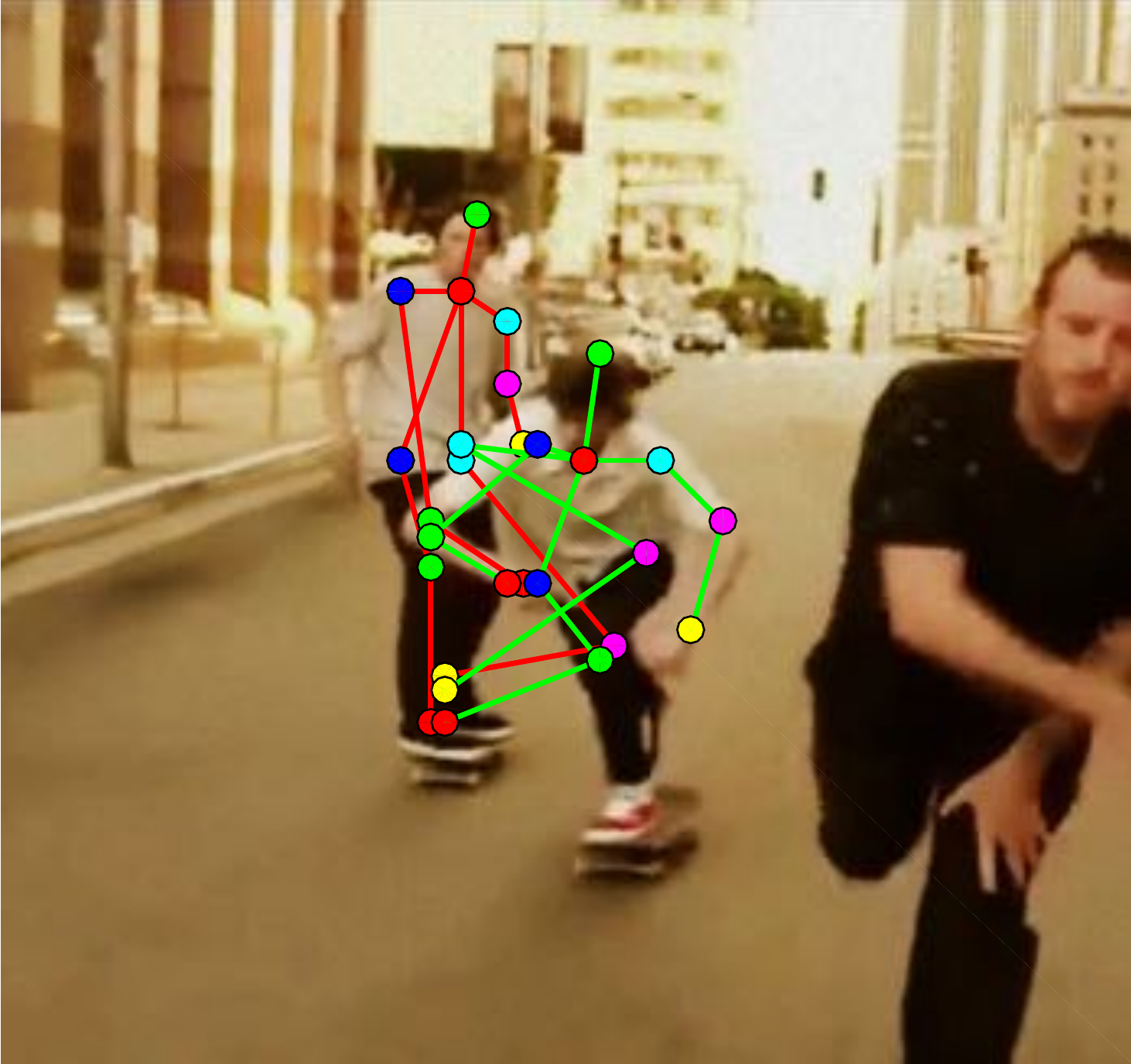}&
    \includegraphics[height=0.140\linewidth]{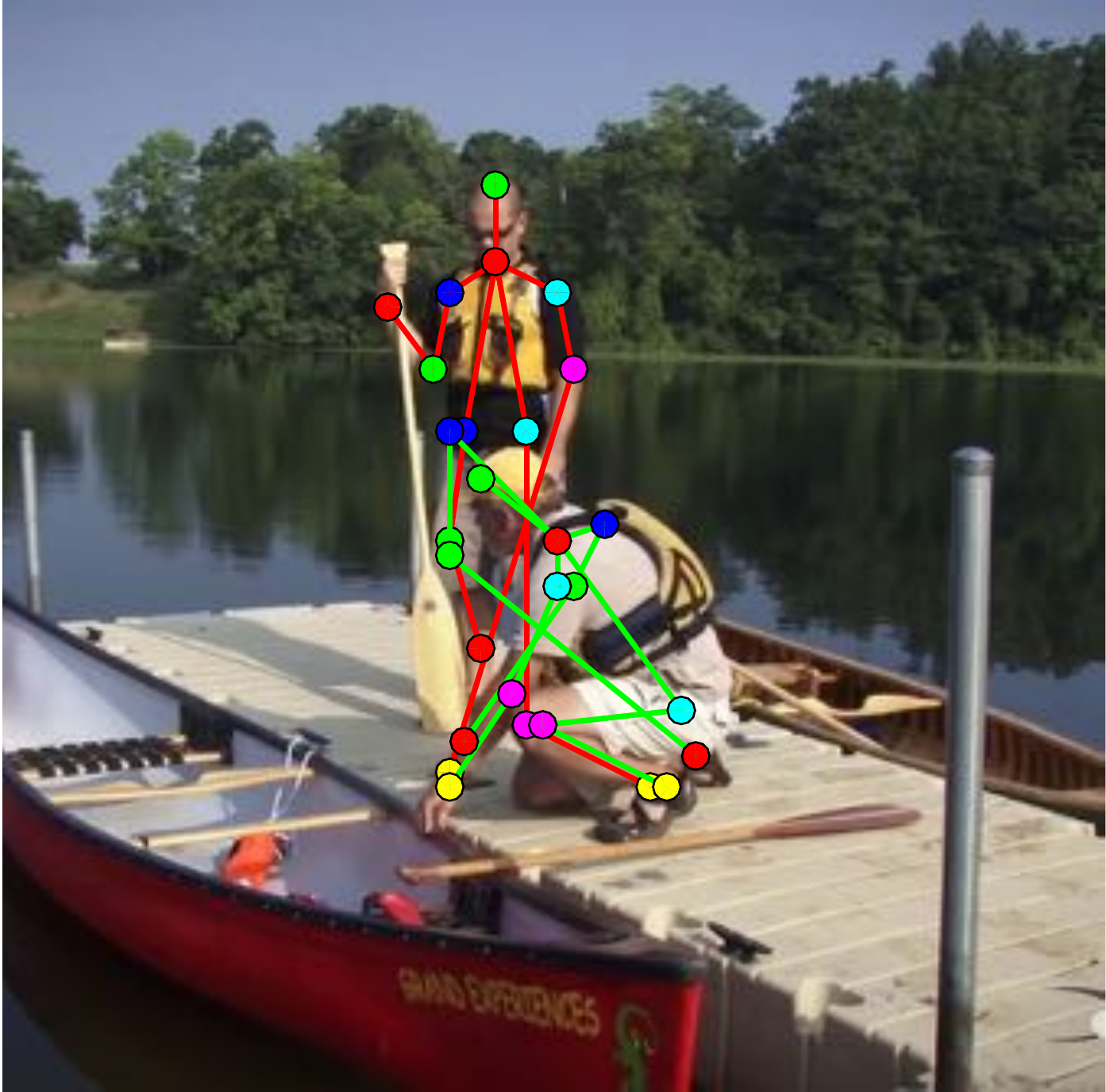}&
    \includegraphics[height=0.140\linewidth]{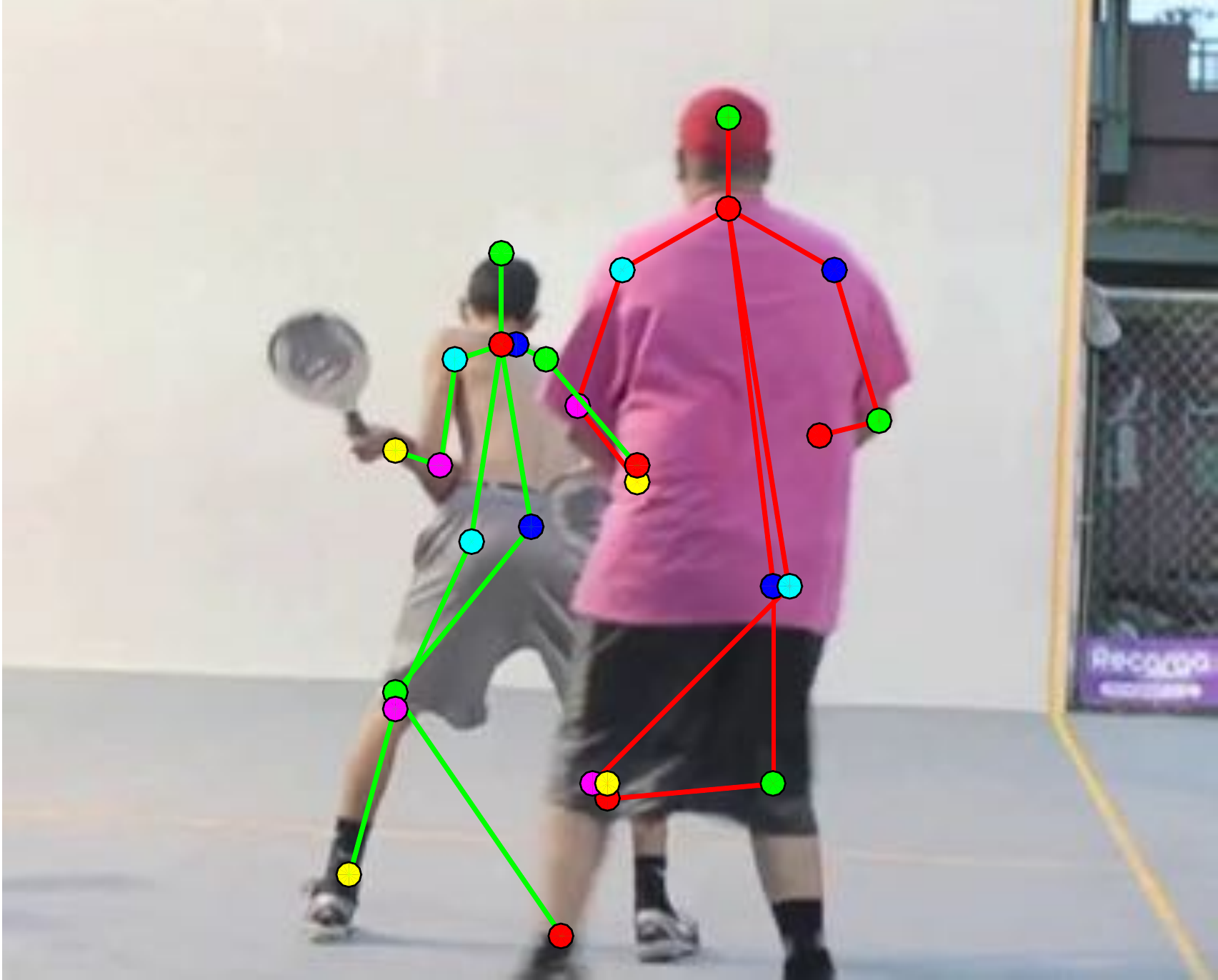}\\
    &10&11&12&13&14\\  
  \end{tabular}

  \begin{tabular}{c c c c c c c}
    \toprule
    &
    \includegraphics[height=0.140\linewidth]{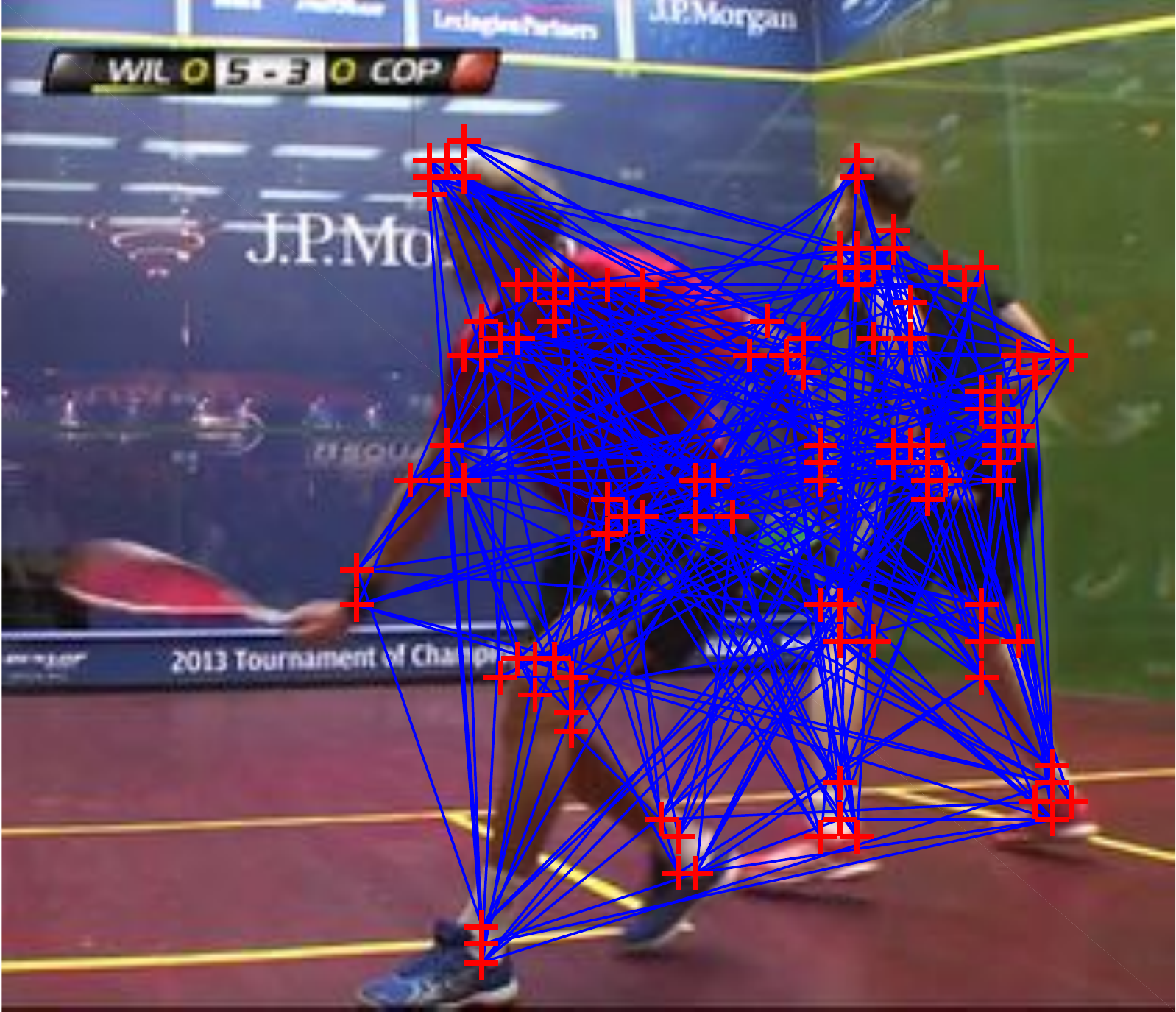}&
    \includegraphics[height=0.140\linewidth]{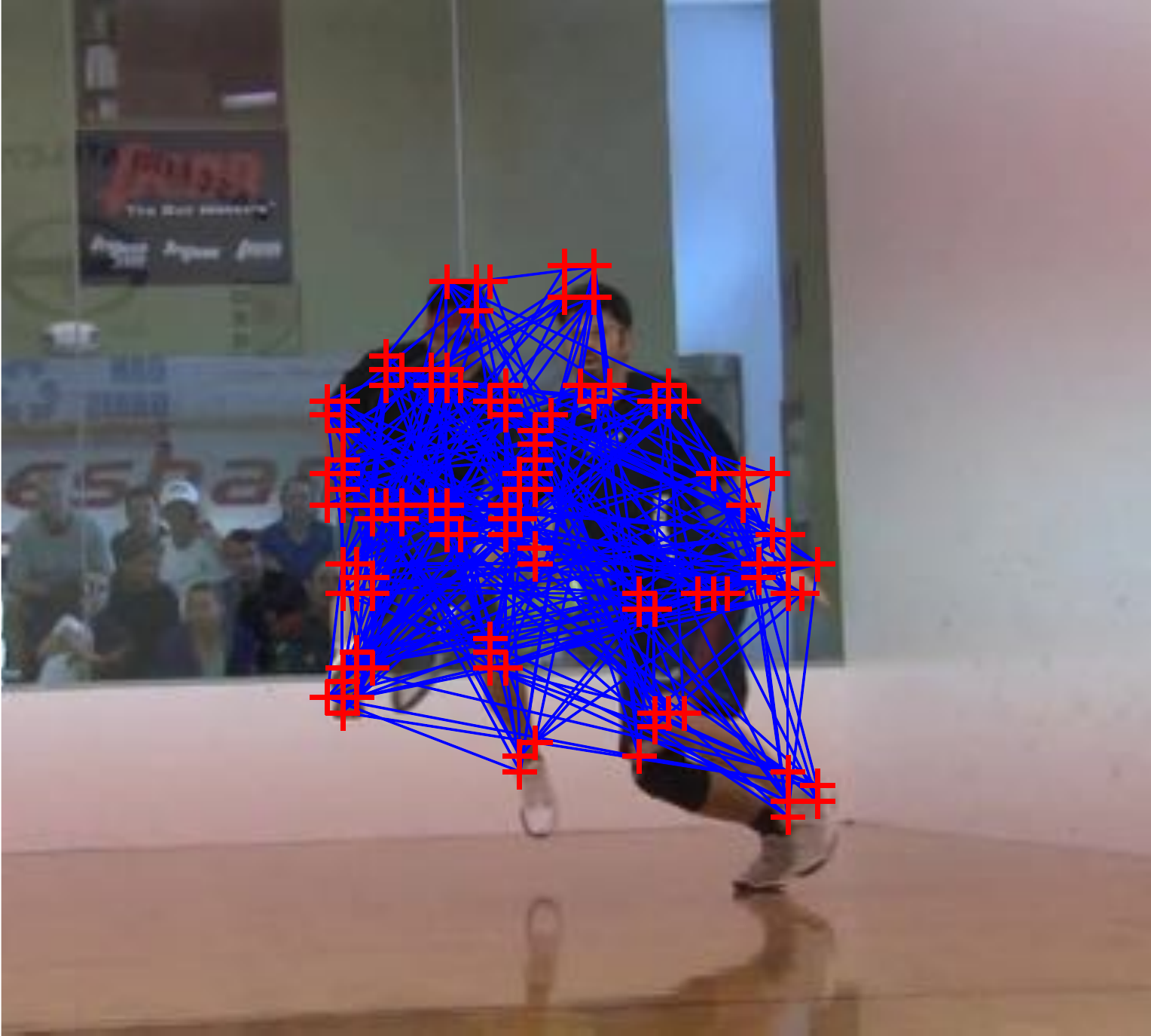}&
    \includegraphics[height=0.140\linewidth]{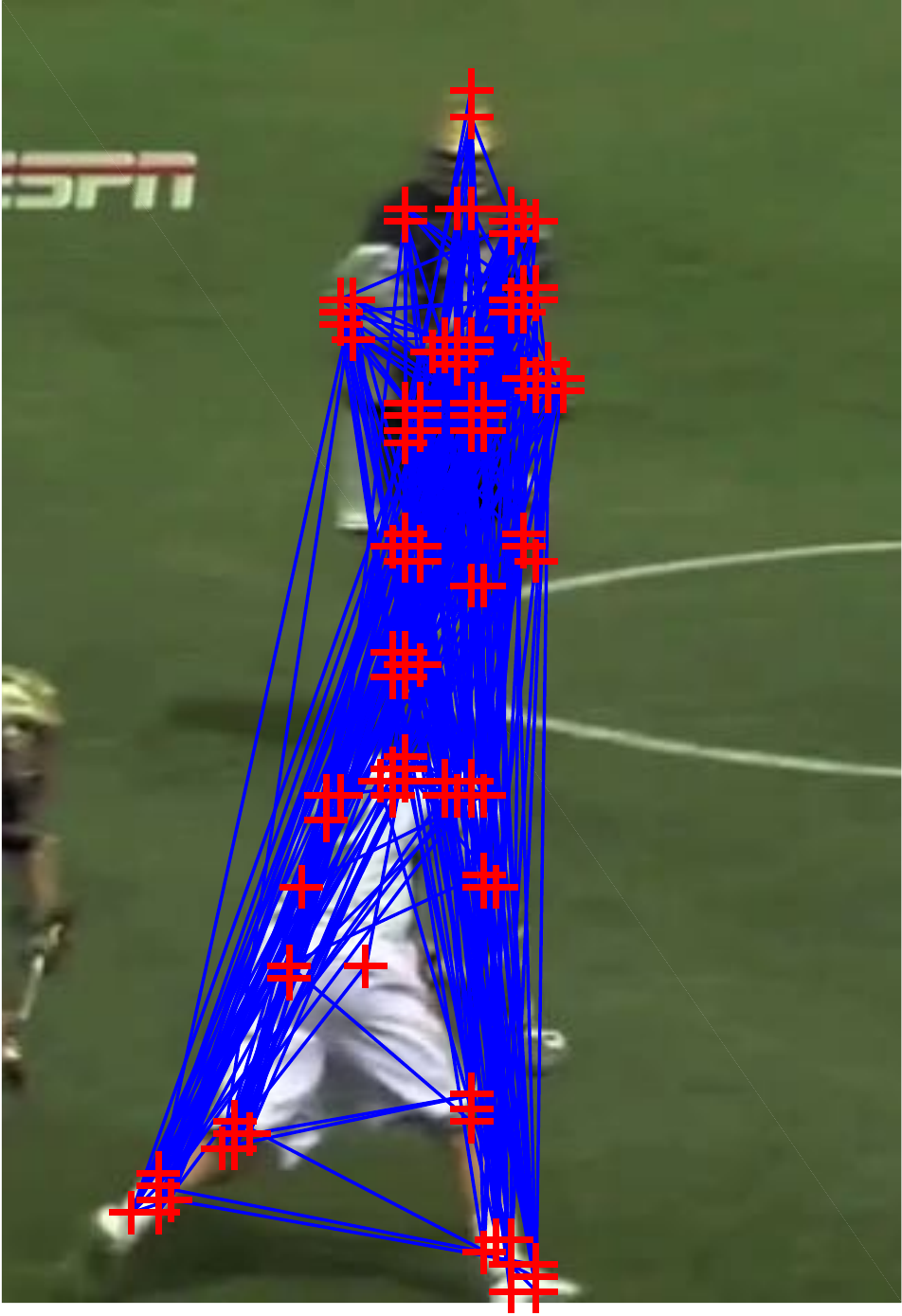}&
    \includegraphics[height=0.140\linewidth]{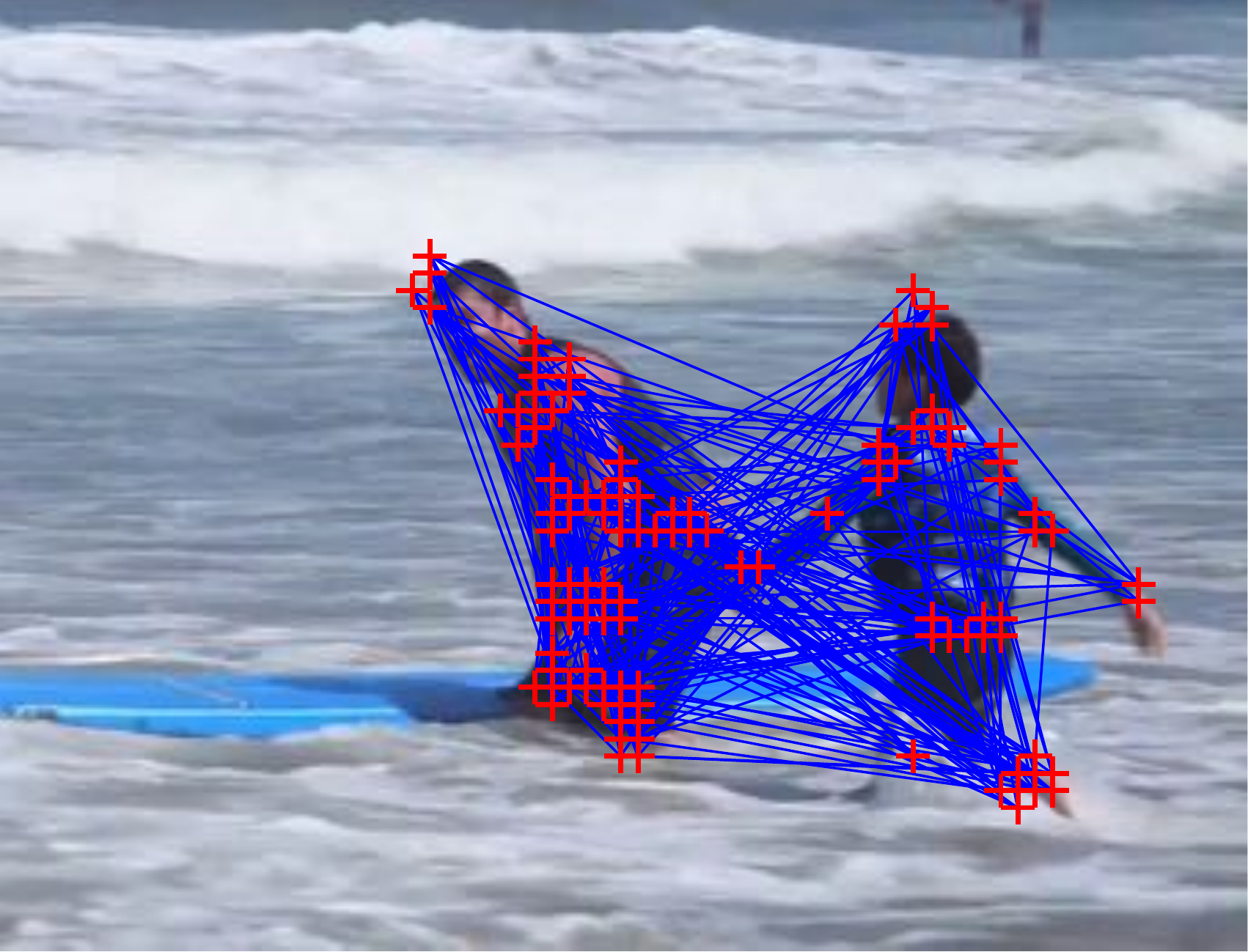}&
    \includegraphics[height=0.140\linewidth]{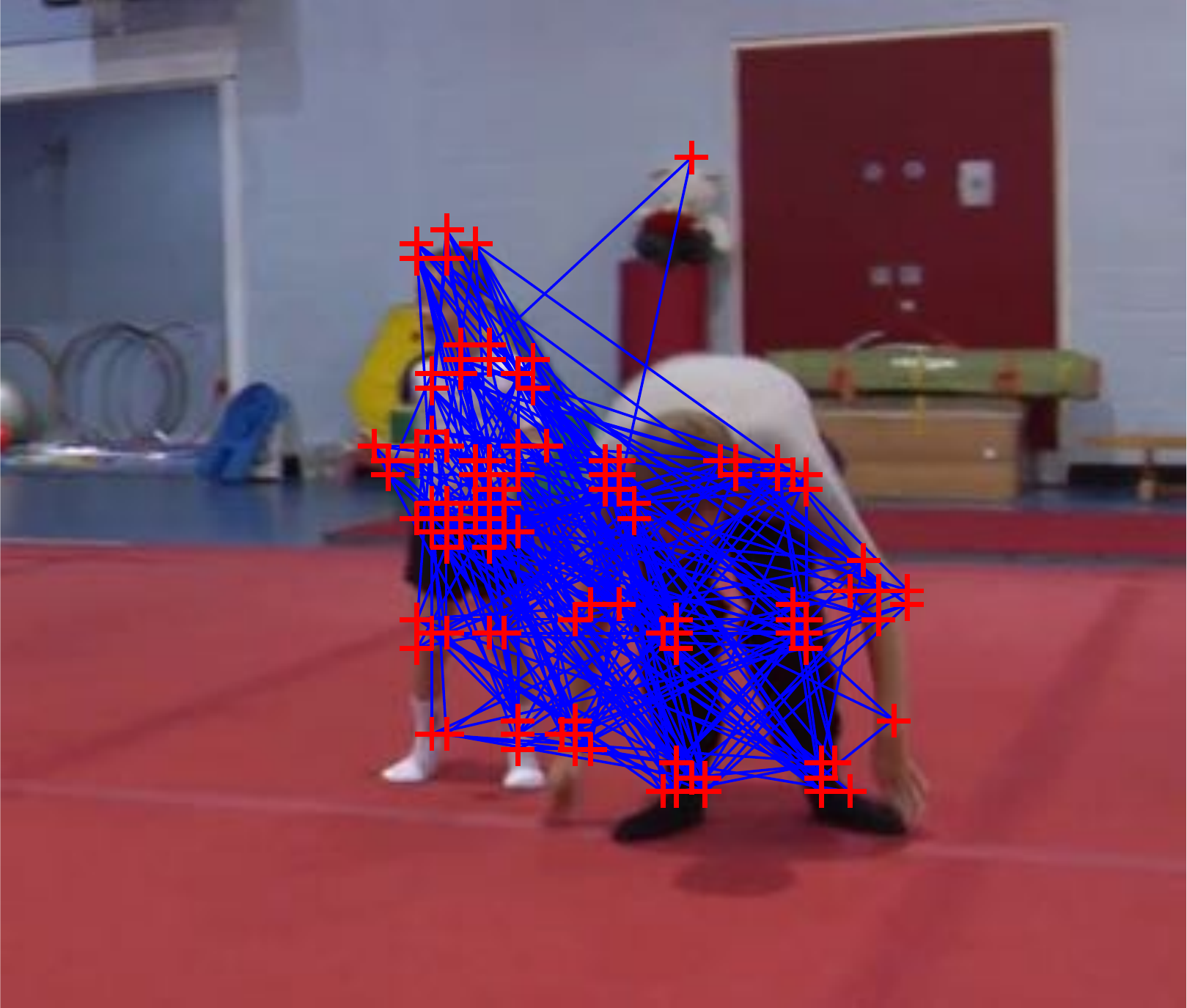}&
    \includegraphics[height=0.140\linewidth]{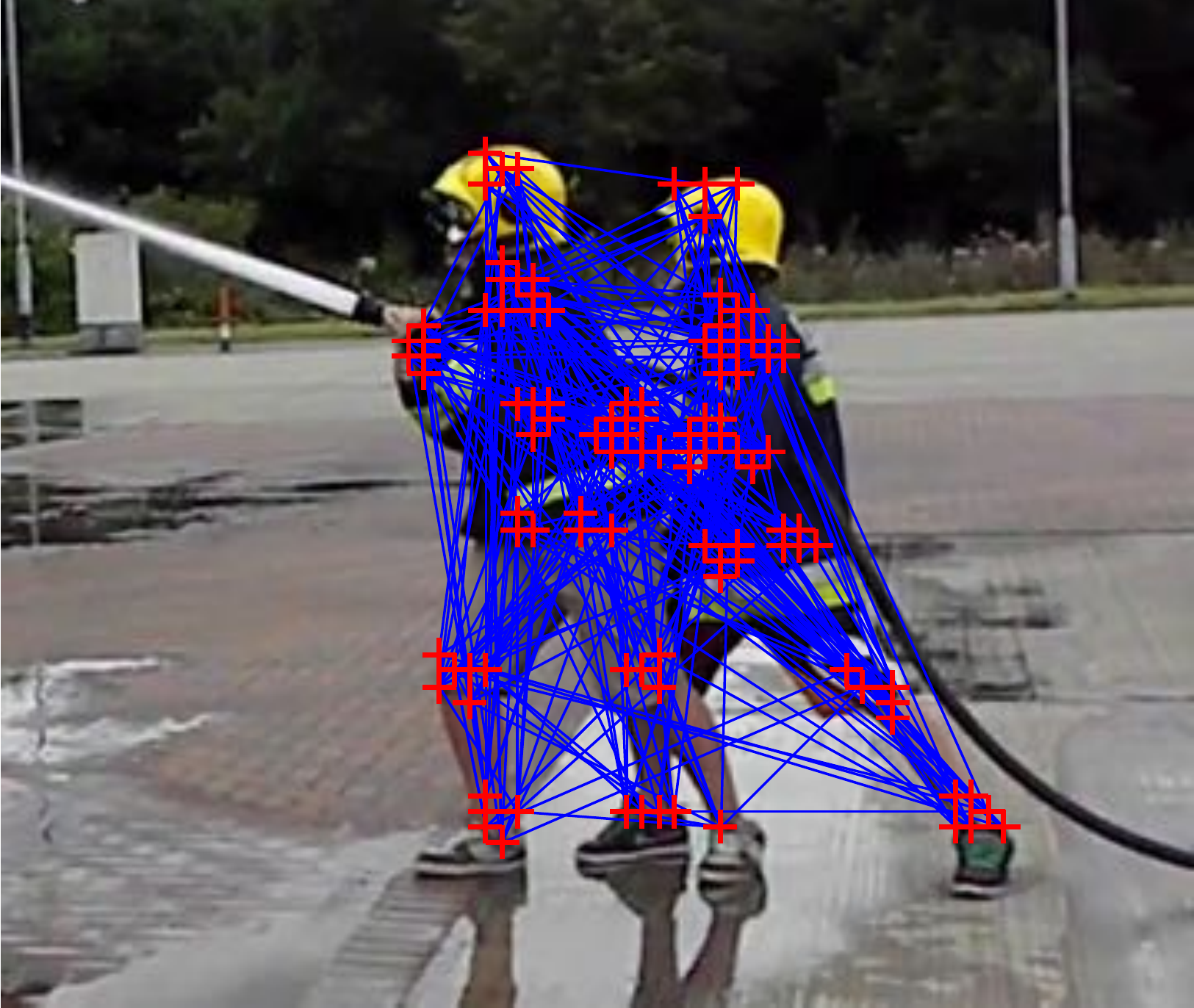}\\
    \begin{sideways}\bf\quad $\deepcut~\multb$\end{sideways}&    
    \includegraphics[height=0.140\linewidth]{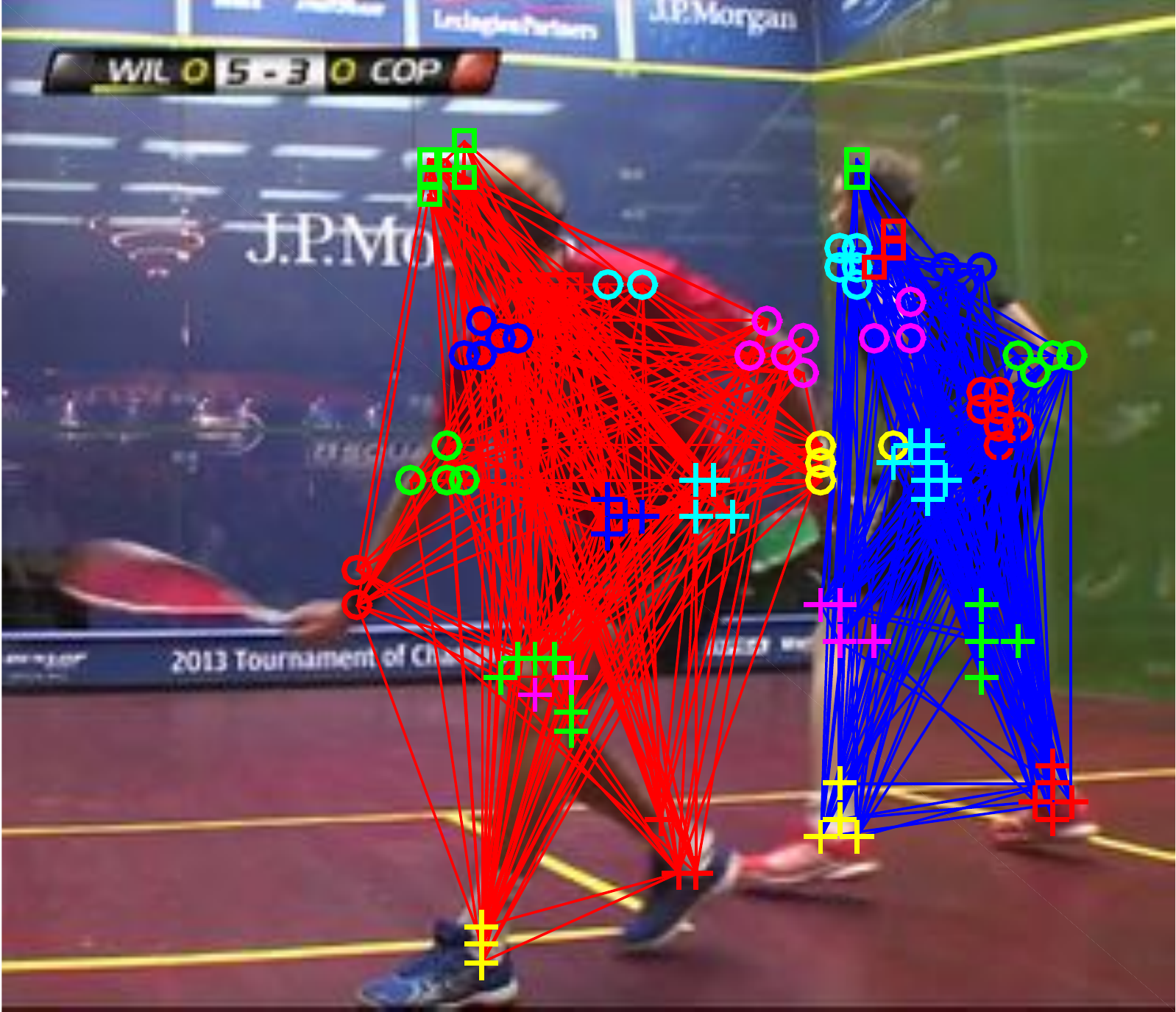}&
    \includegraphics[height=0.140\linewidth]{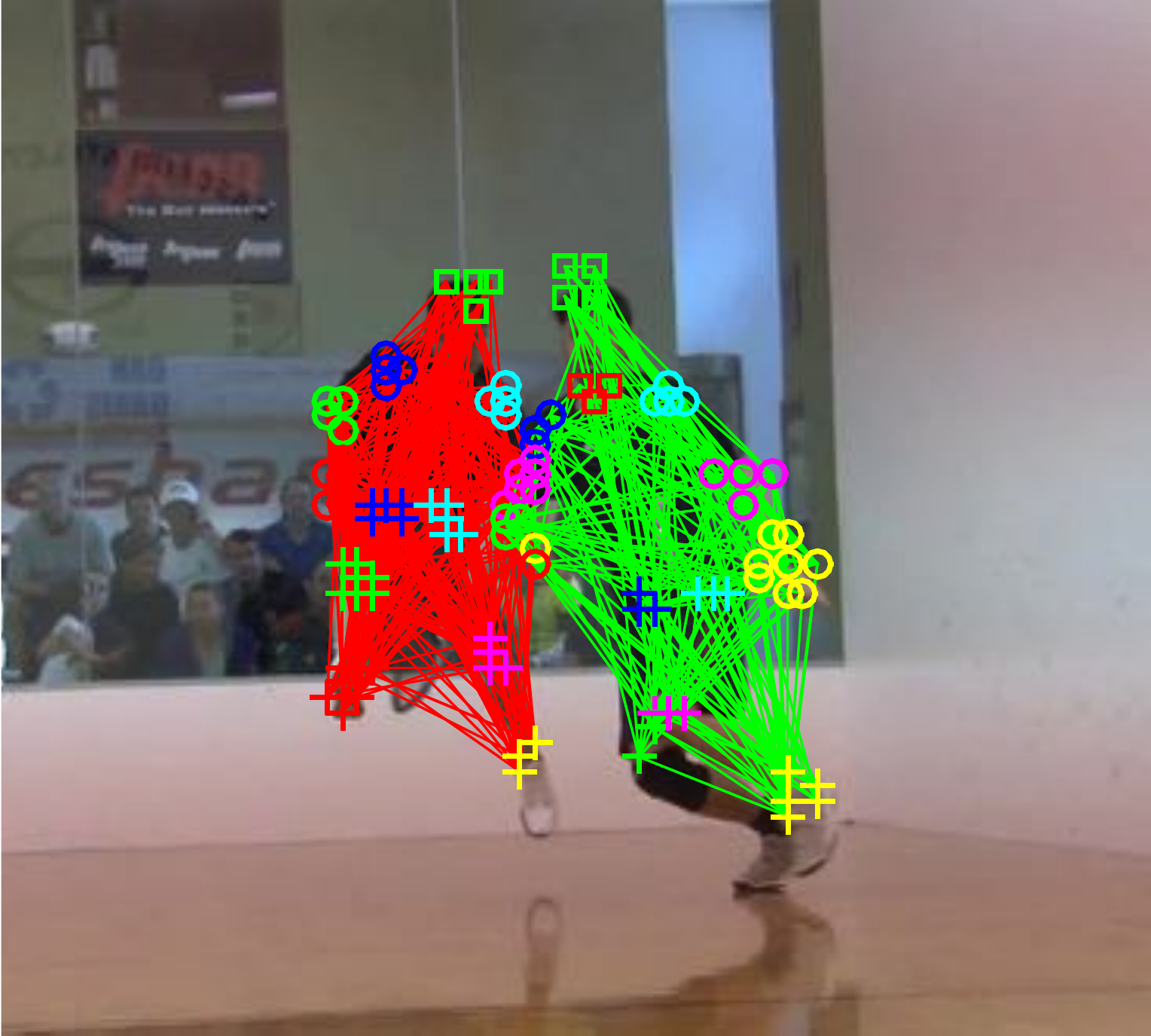}&
    \includegraphics[height=0.140\linewidth]{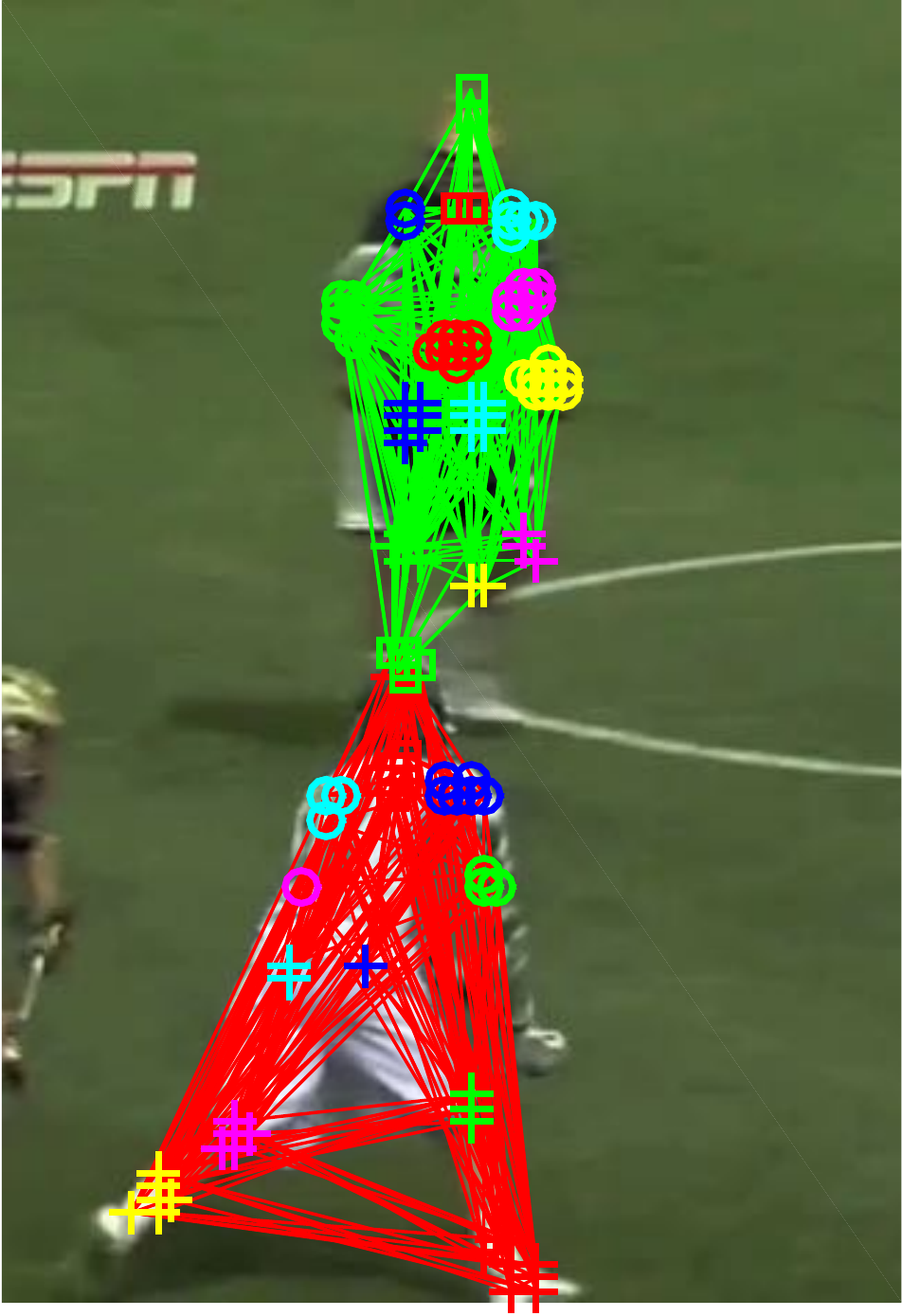}&
    \includegraphics[height=0.140\linewidth]{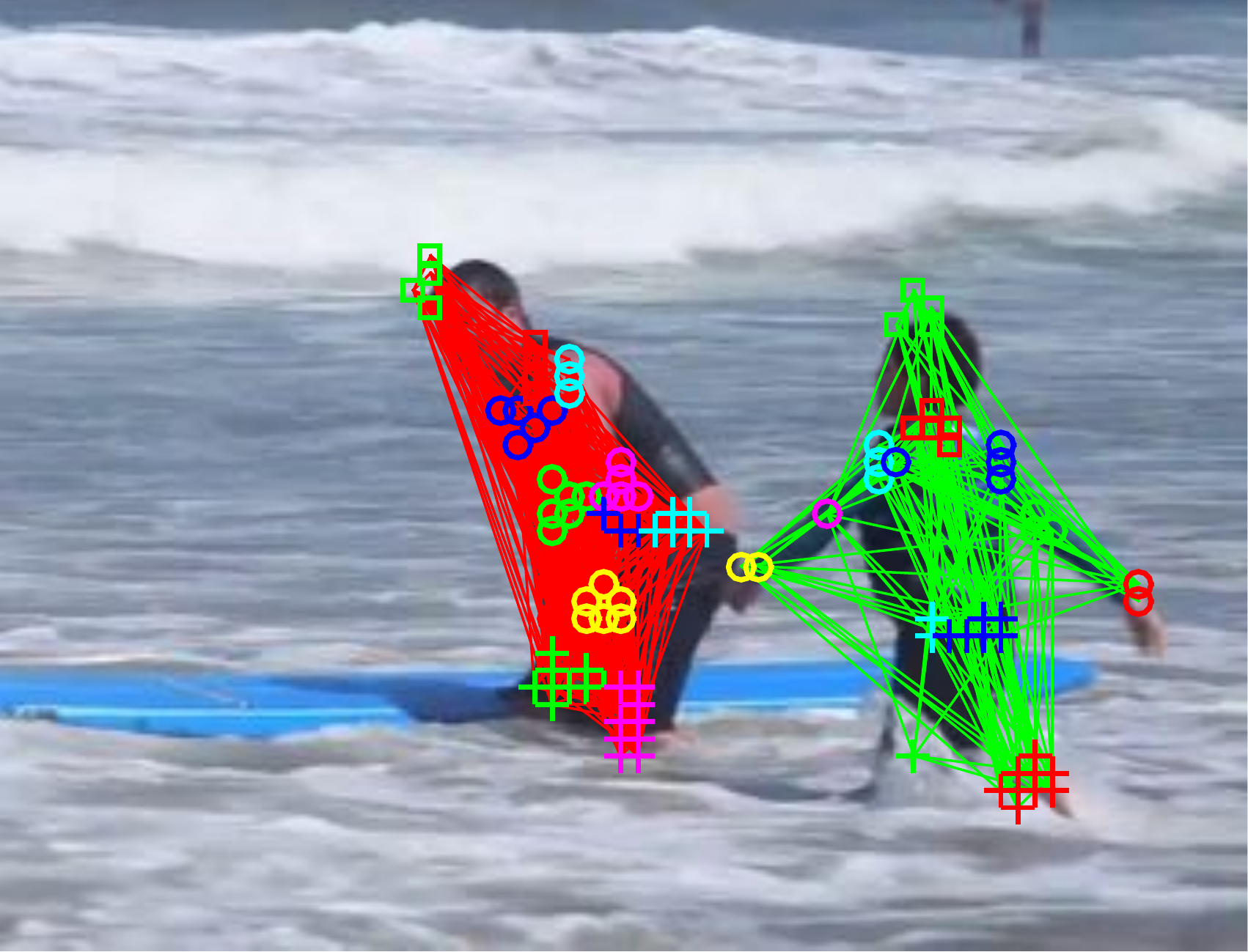}&
    \includegraphics[height=0.140\linewidth]{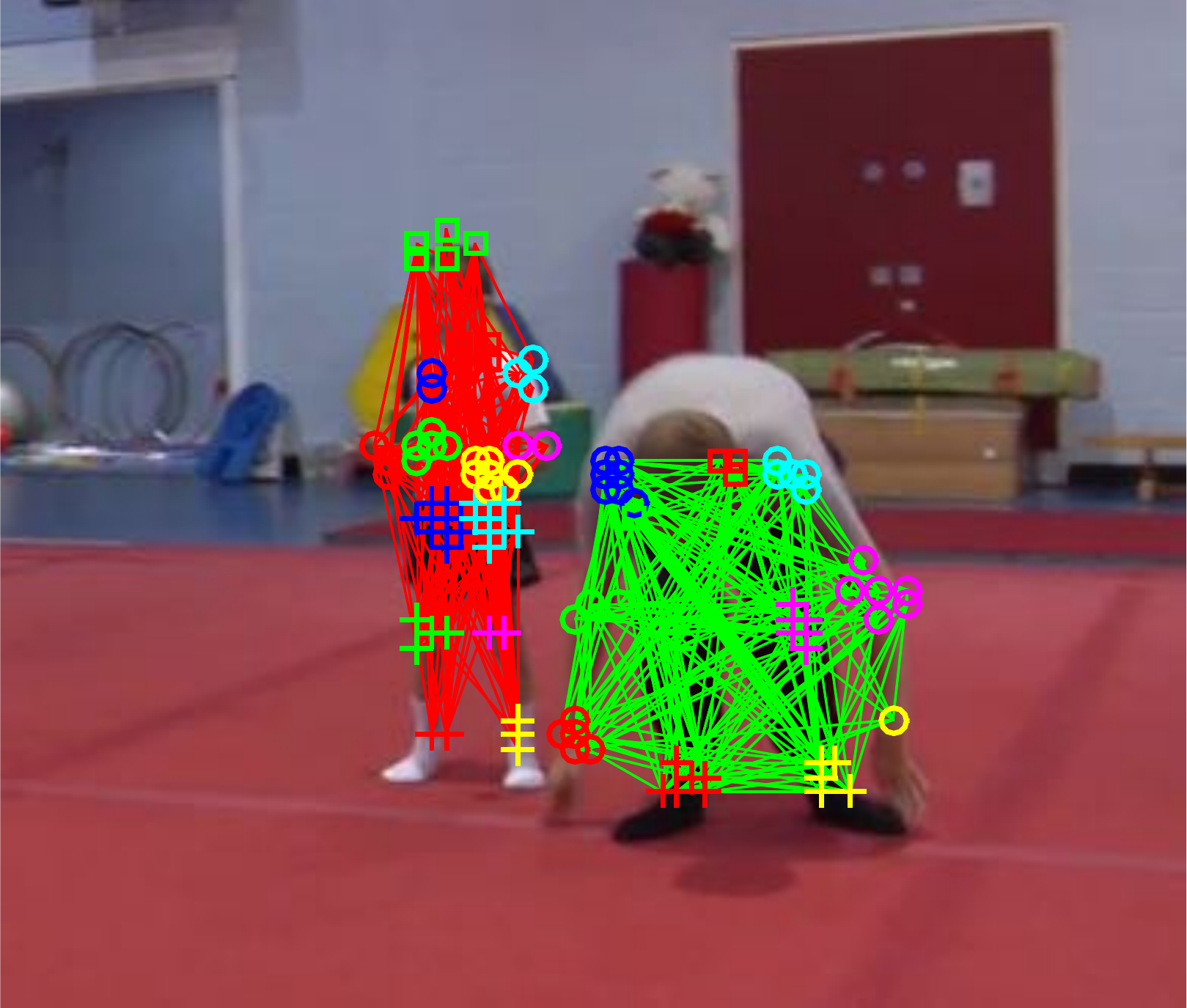}&
    \includegraphics[height=0.140\linewidth]{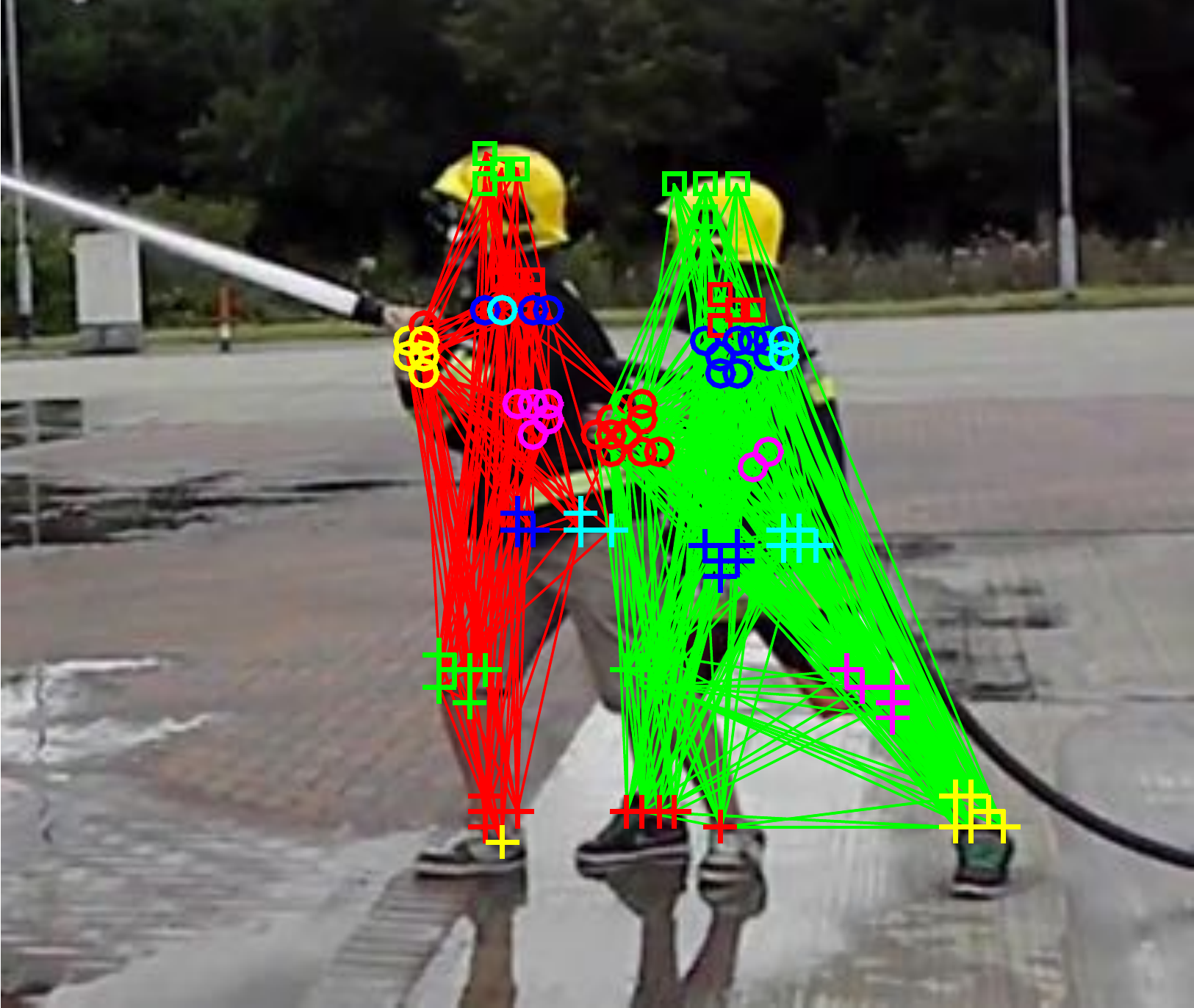}\\
    &
    \includegraphics[height=0.140\linewidth]{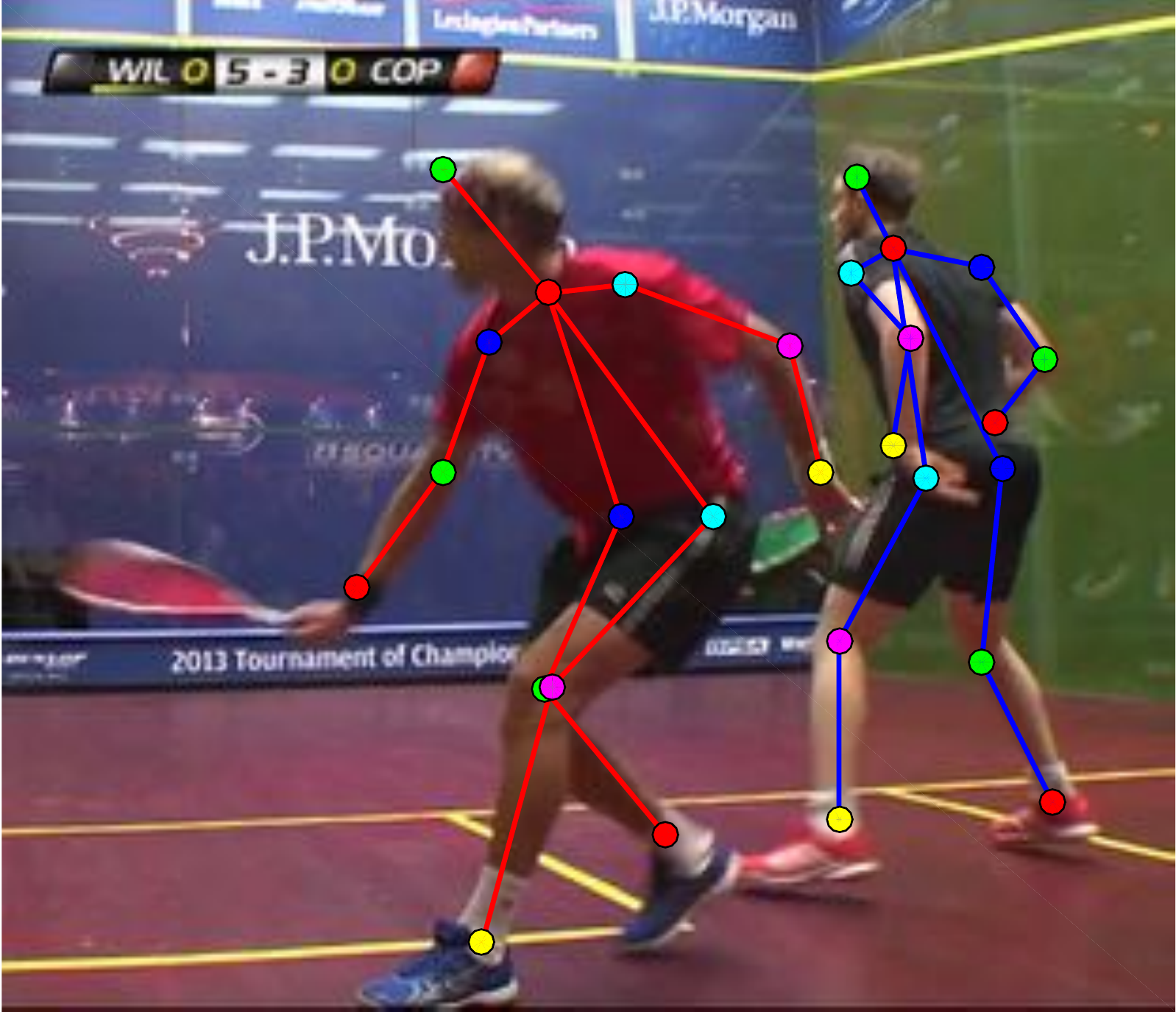}&
    \includegraphics[height=0.140\linewidth]{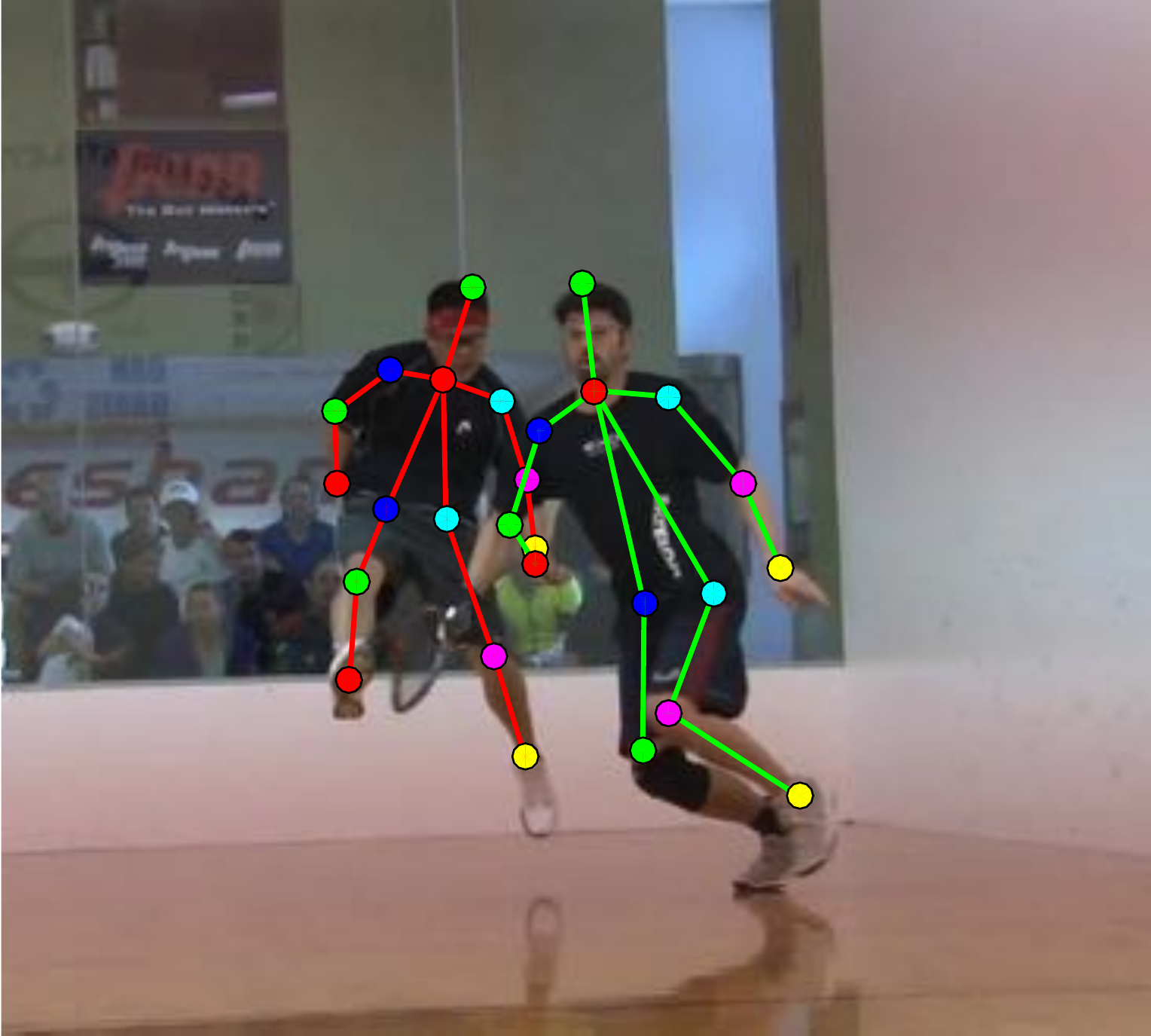}&
    \includegraphics[height=0.140\linewidth]{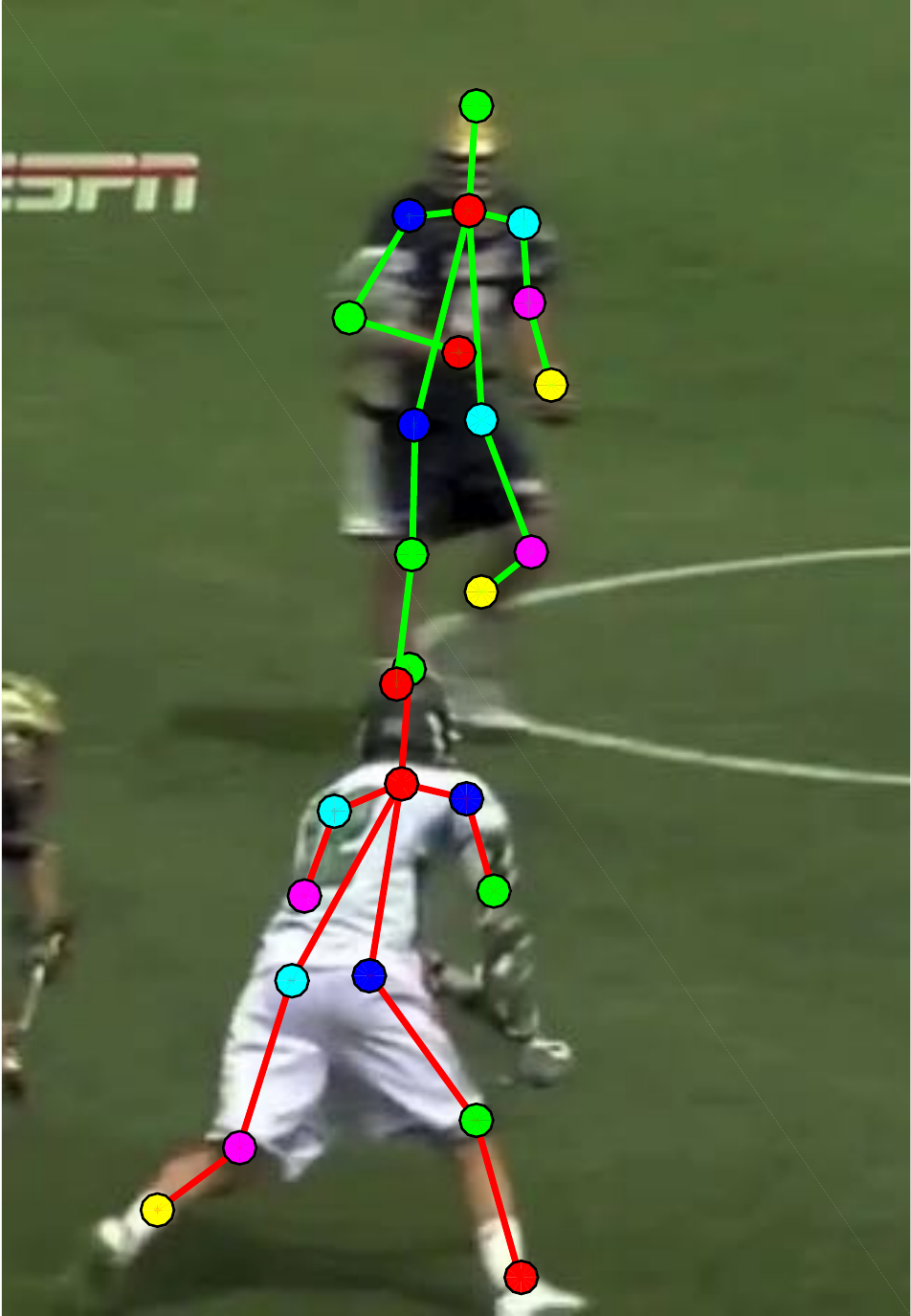}&
    \includegraphics[height=0.140\linewidth]{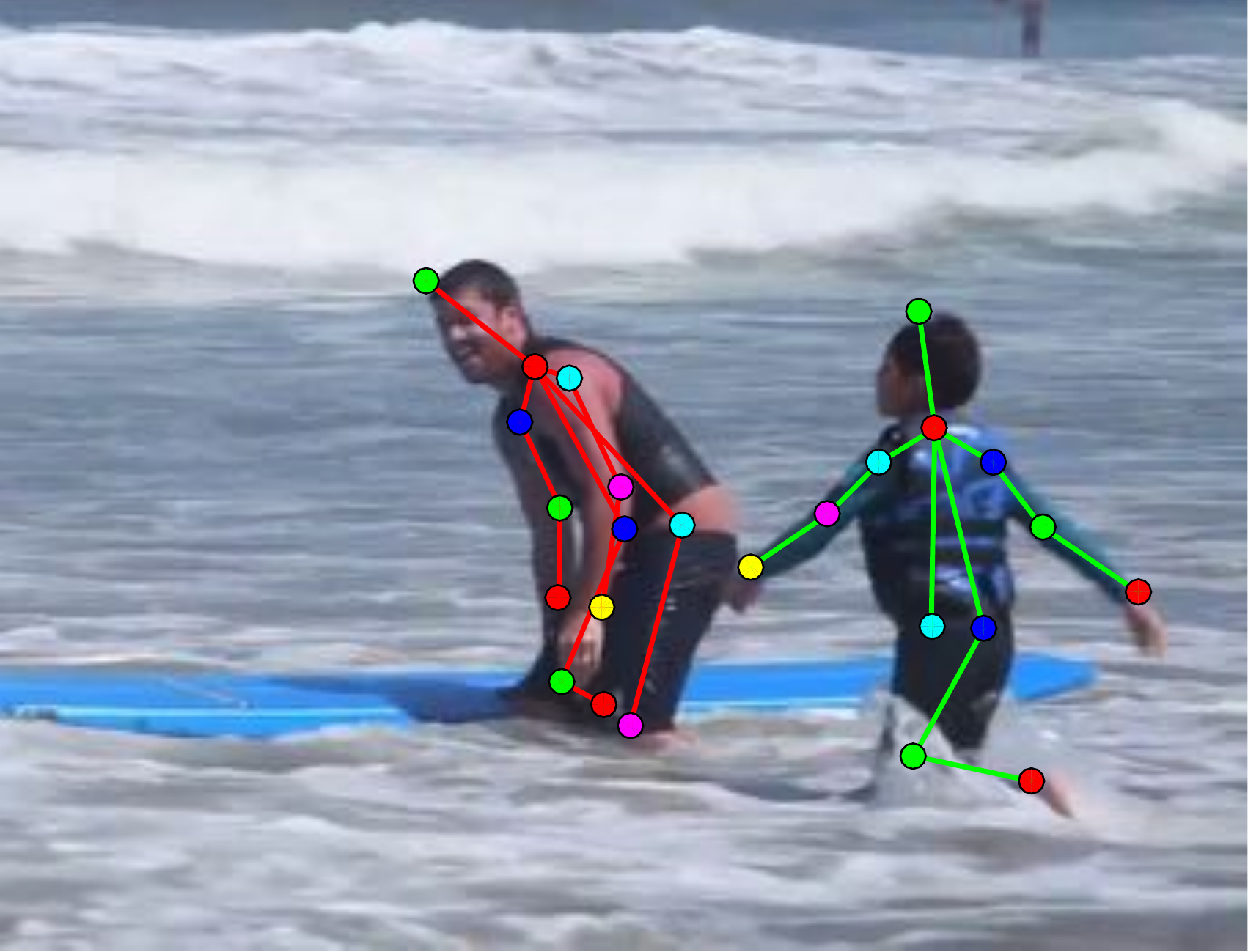}&
    \includegraphics[height=0.140\linewidth]{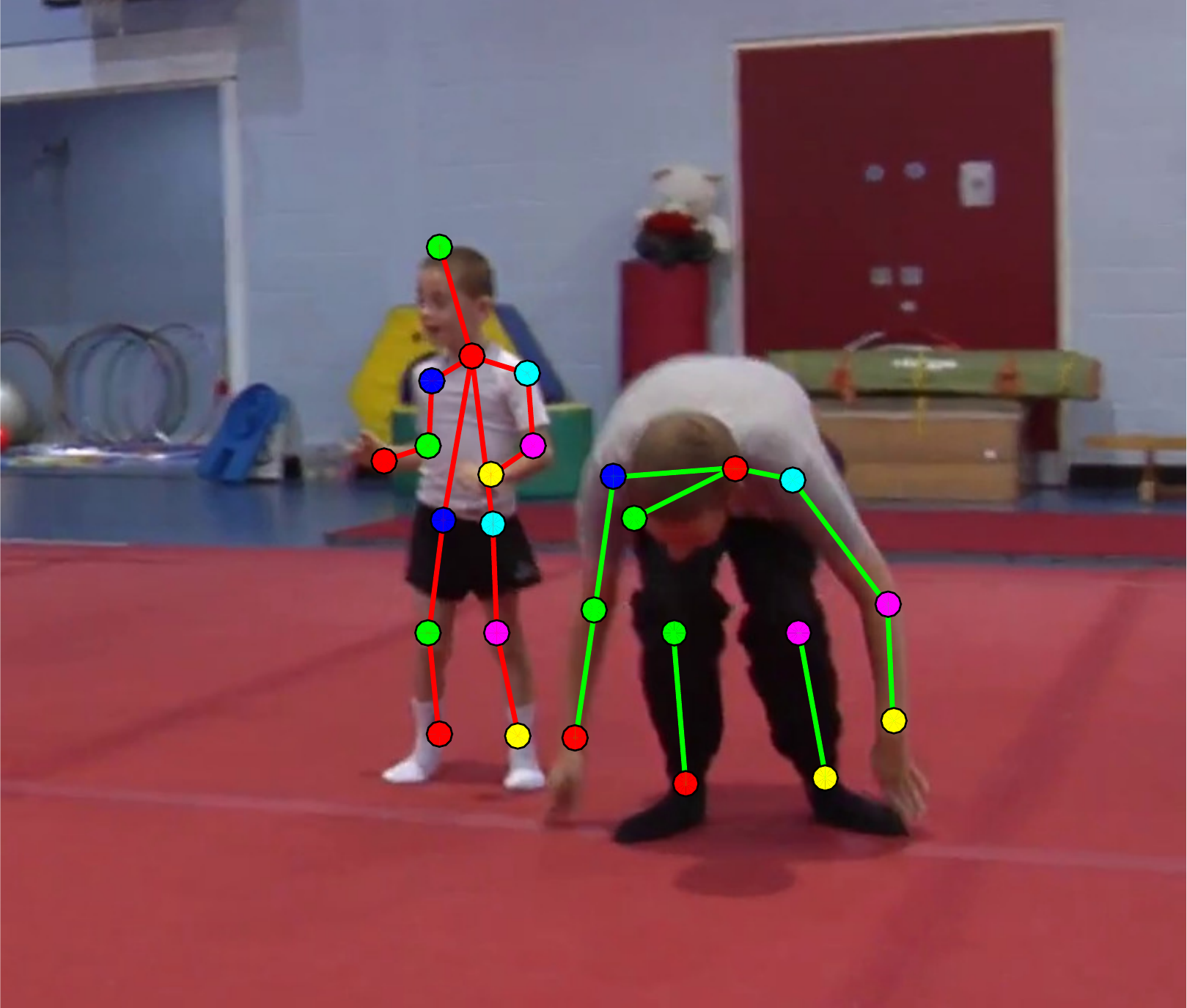}&
    \includegraphics[height=0.140\linewidth]{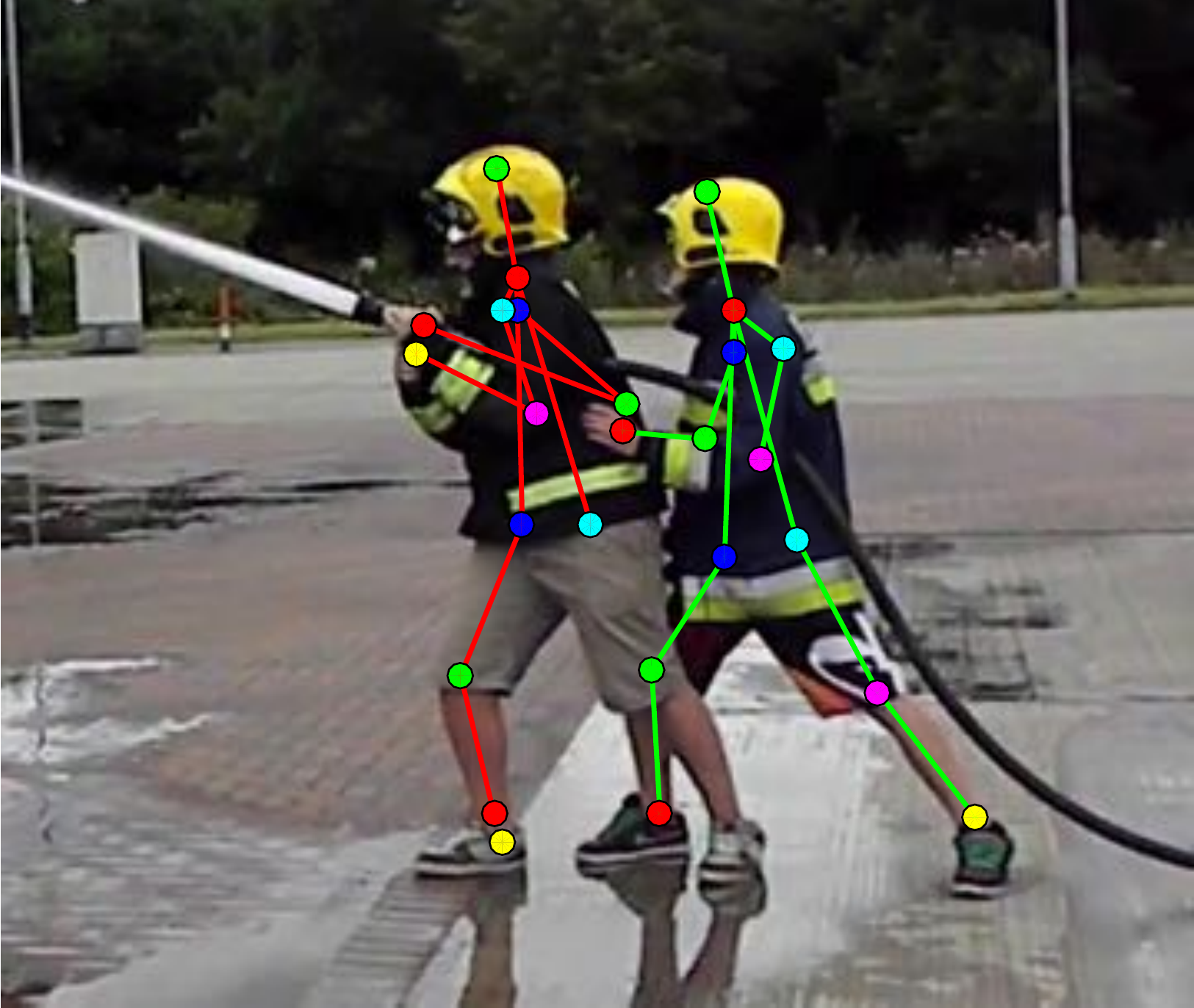}\\

    \midrule\midrule
    \begin{sideways}\bf \quad\quad$\detroi$\end{sideways}&
    \includegraphics[height=0.140\linewidth]{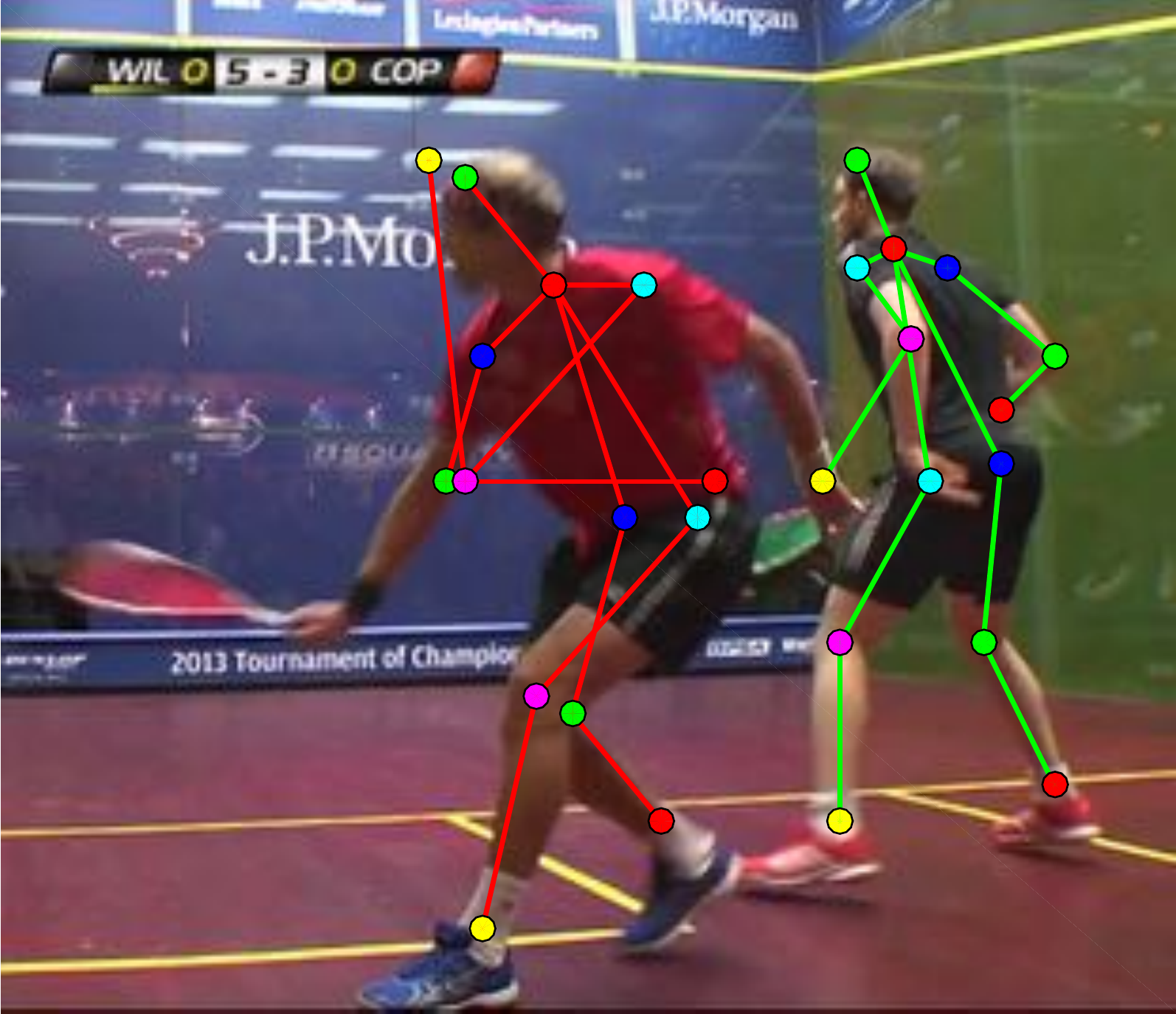}&
    \includegraphics[height=0.140\linewidth]{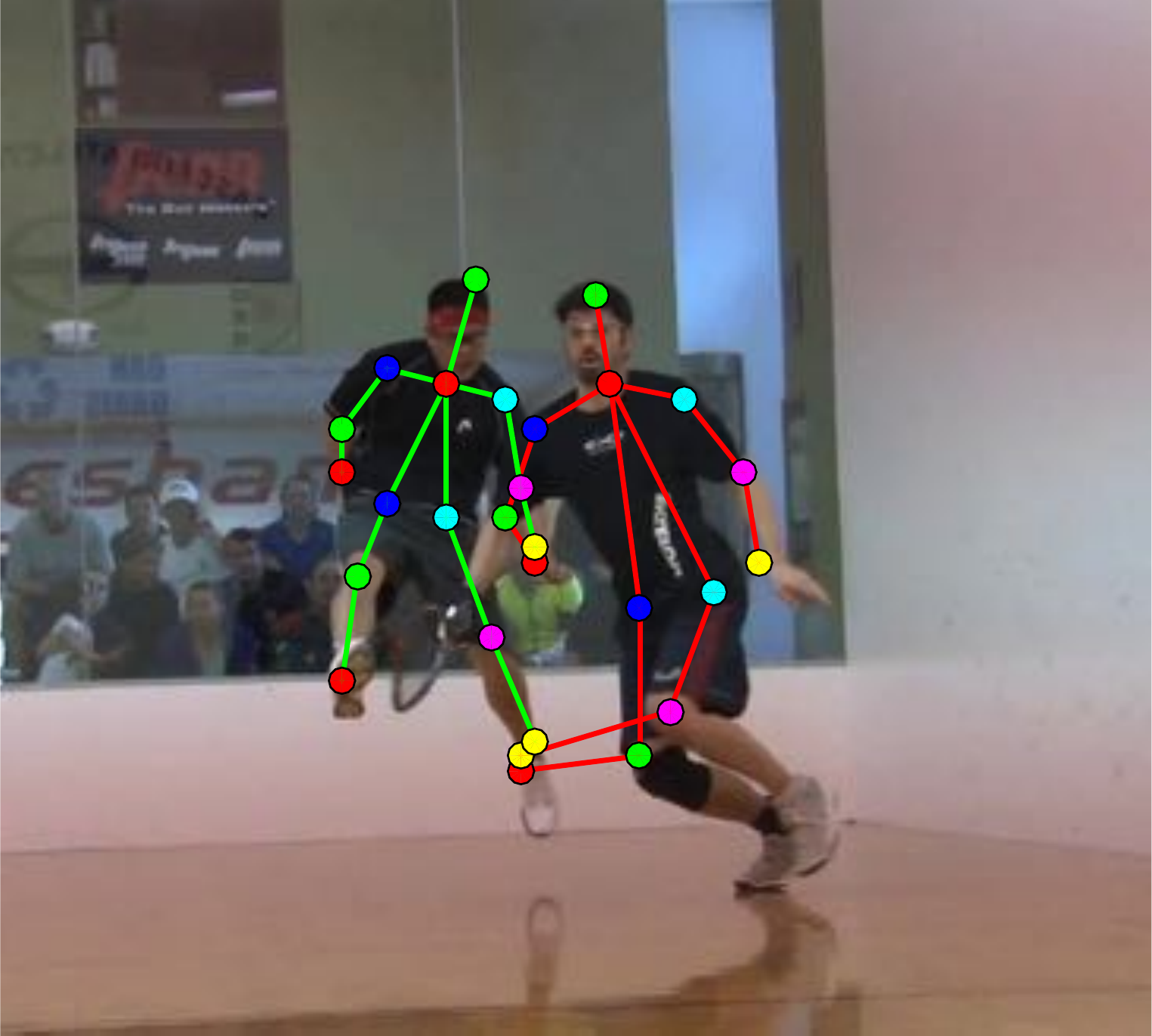}&
    \includegraphics[height=0.140\linewidth]{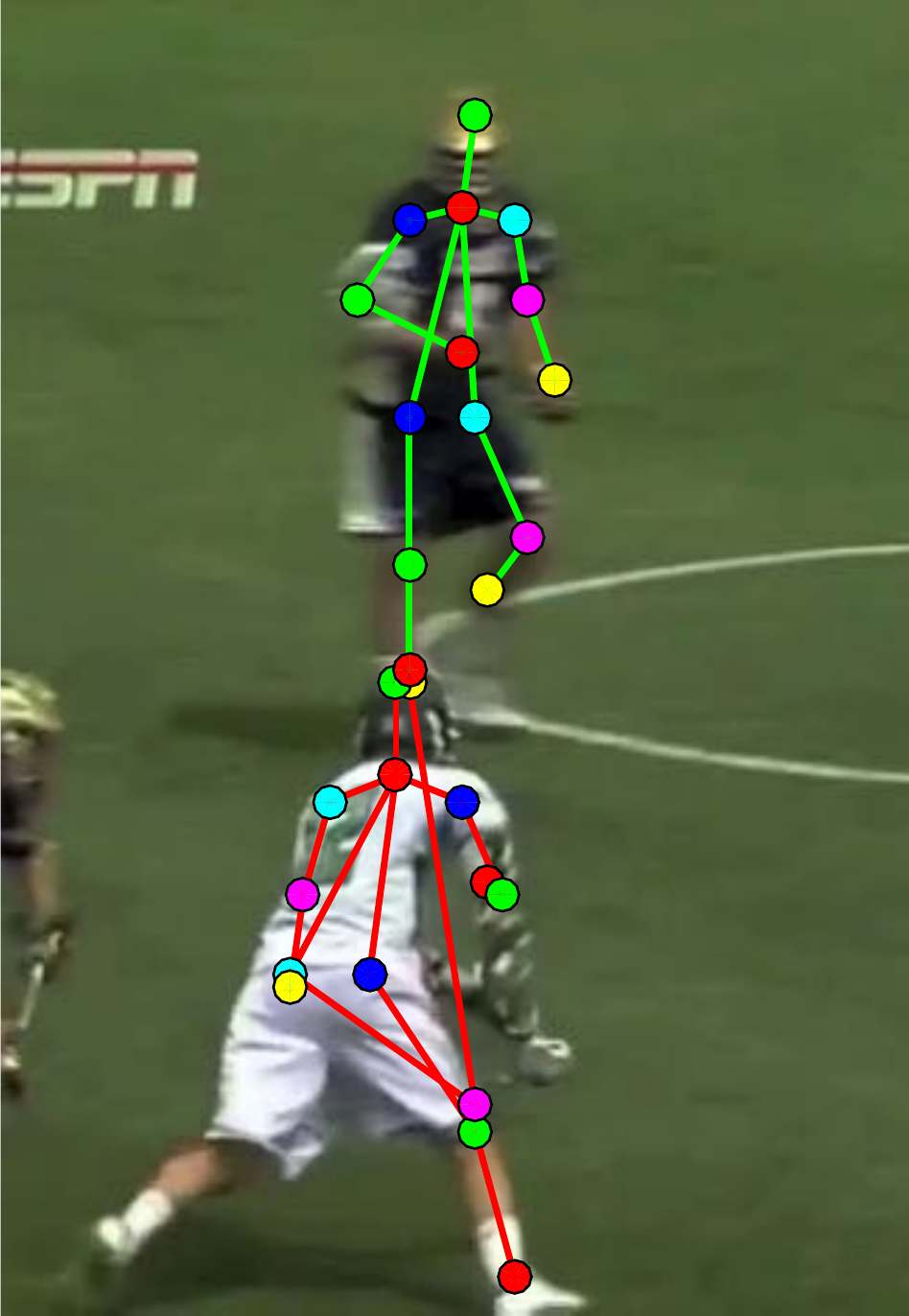}&
    \includegraphics[height=0.140\linewidth]{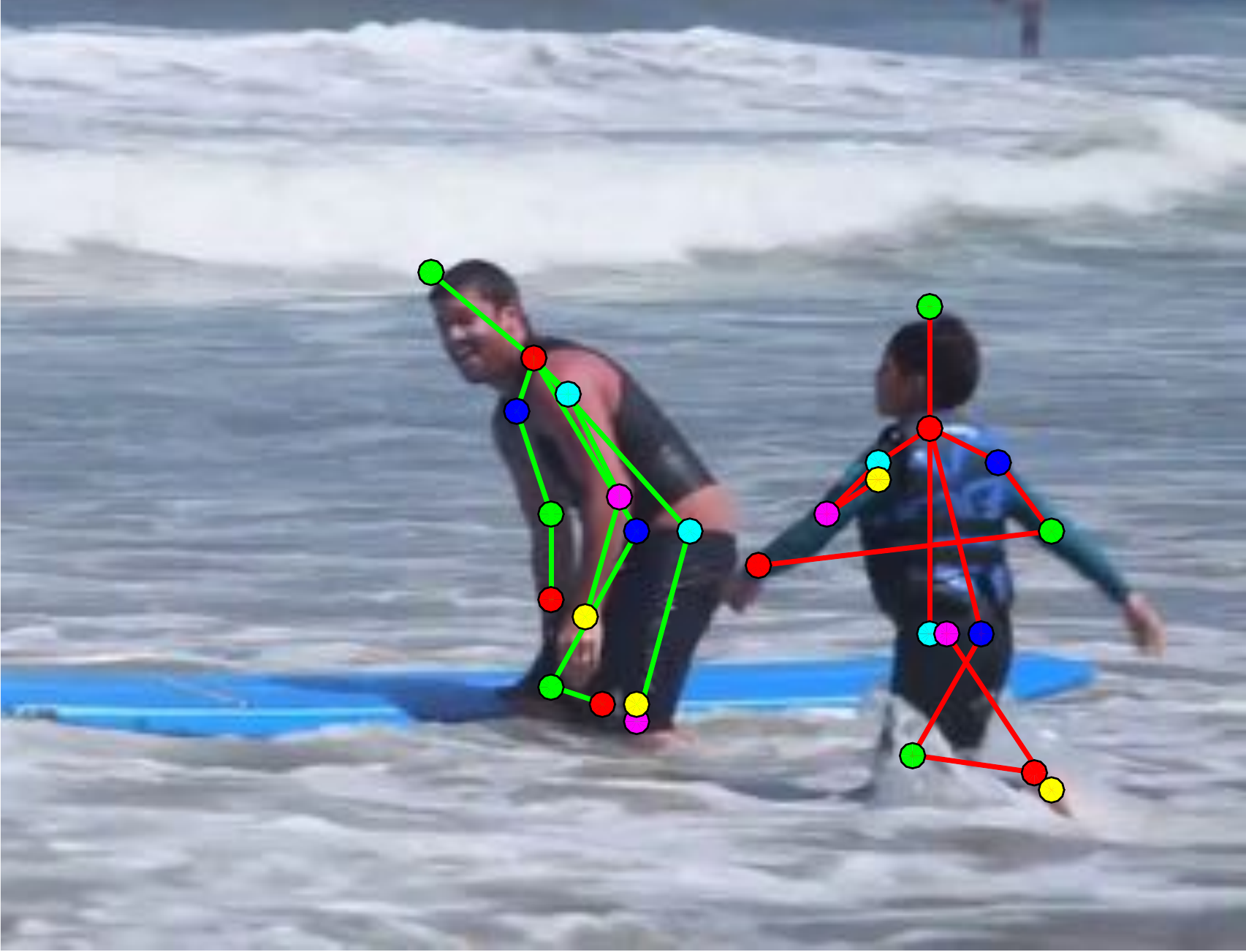}&
    \includegraphics[height=0.140\linewidth]{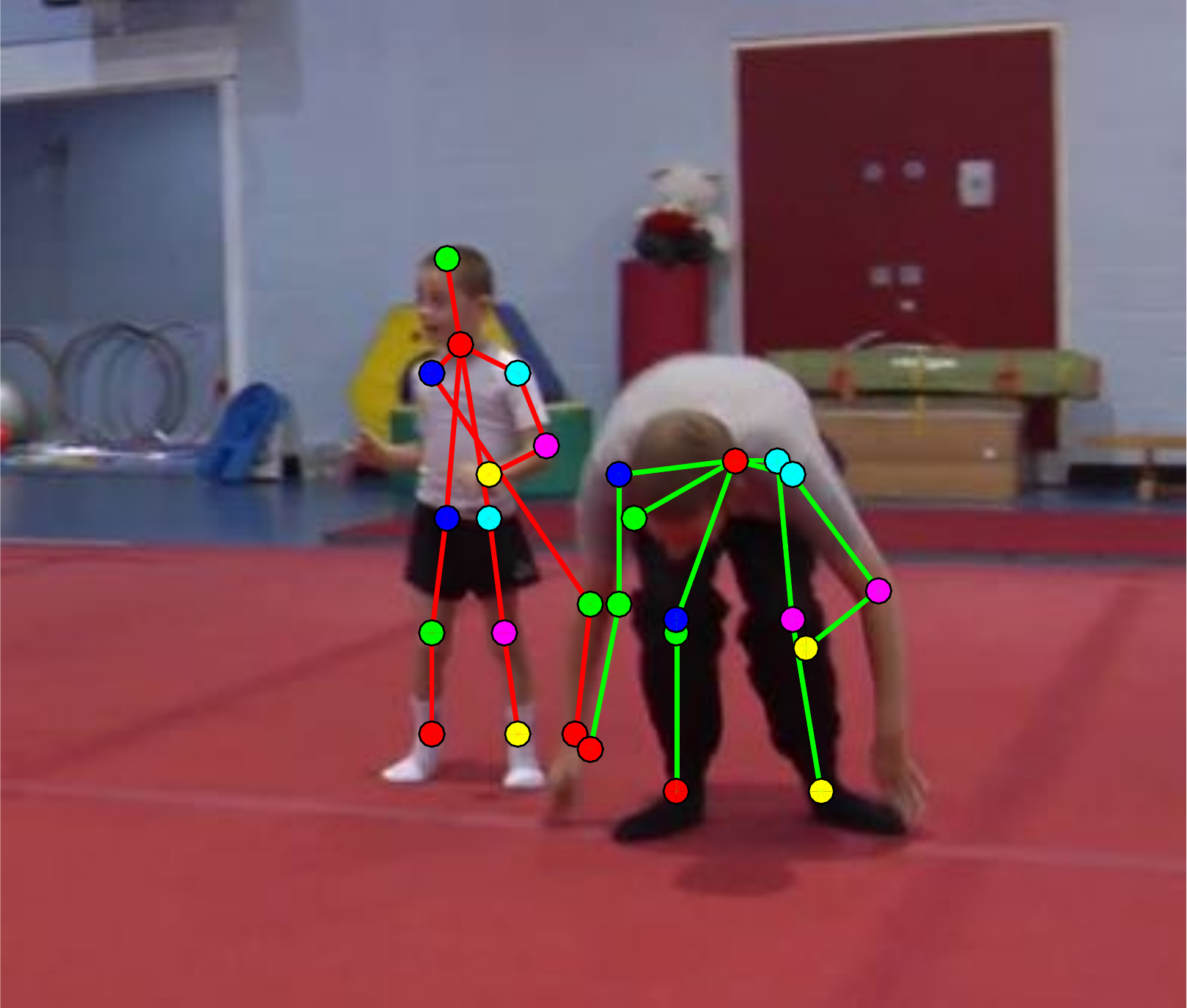}&
    \includegraphics[height=0.140\linewidth]{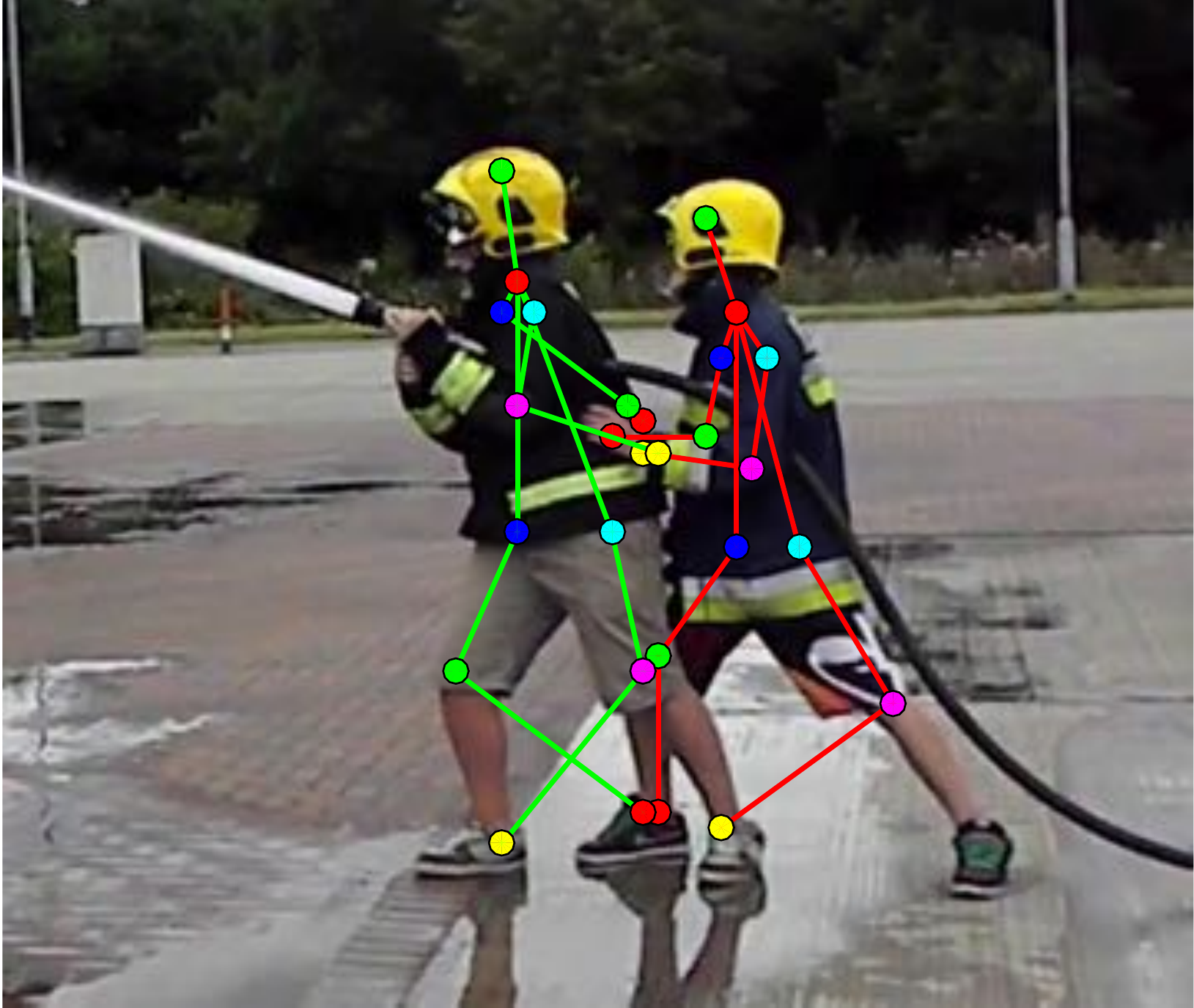}\\
    &15&16&17&18&19&20\\  
  \end{tabular}
  %\vspace{-1.0em}
  \caption{Qualitative comparison (contd.) of our joint formulation
    $\deepcut~\multb~\dense$ (rows 1-3, 5-7) to the traditional
    two-stage approach $\dense~\detroi$ (rows 4, 8) on MPII
    Multi-Person dataset. 
    See Fig.~\ref{fig:overview} for the color-coding explanation.
    %See Fig.~\textcolor{red}{1} in paper for the color-coding explanation.
    }
  \label{fig:qualitative_mpii2}
\end{figure*}

Qualitative comparison of our joint formulation
$\deepcut~\multb~\dense$ to the traditional two-stage approach
$\dense~\detroi$ on MPII Multi-Person dataset is shown in
Fig.~\ref{fig:qualitative_mpii} and~\ref{fig:qualitative_mpii2}.
$\dense~\detroi$ works well when multiple fully visible individuals
are sufficiently separated and thus their body parts can be
partitioned based on the person detection bounding box. In this case
the strong $\dense$ body part detection model can correctly estimate
most of the visible body parts (image 16, 17, 19). However,
$\dense~\detroi$ cannot tell apart the body parts of multiple
individuals located next to each other and possibly occluding each
other, and often links the body parts across the individuals (images 1-16,
19-20). In addition, $\dense~\detroi$ cannot reason about occlusions
and truncations always providing a prediction for each body part
(image 4, 6, 10). In contrast, $\deepcut~\multb~\dense$ is able to
correctly partition and label an initial pool of body part candidates
(each image, top row) into subsets that correspond to sets of mutually
consistent body part candidates and abide to mutual consistency and
exclusion constraints (each image, row 2), thereby outputting
consistent body pose predictions (each image, row 3). $c \neq c'$
pairwise terms allow to partition the initial set of part detection
candidates into valid pose configurations (each image, row 2:
person-clusters highlighted by dense colored connections). $c = c'$
pairwise terms facilitate clustering of multiple body part candidates
of the same body part of the same person (each image, row 2: markers
of the same type and color). In addition, $c = c'$ pairwise terms
facilitate a repulsive property that prevents nearby part candidates
of the same type to be associated to different people (image 1:
detections of the left shoulder are assigned to the front person
only). Furthermore, $\deepcut~\multb~\dense$ allows to either merge
or deactivate part hypotheses thus effectively performing non-maximum
suppression and reasoning about body part occlusions and truncations
(image 3, row 2: body part hypotheses on the background are
deactivated (black crosses); image 6, row 2: body part hypotheses for
the truncated body parts are deactivated (black crosses); image 1-6,
8-9, 13-14, row 3: only visible body parts of the partially occluded
people are estimated, while non-visible body parts are correctly
predicted to be occluded). These qualitative examples show that
$\deepcuts~\multb$ can successfully deal with the unknown number of
people per image and the unknown number of visible body parts per
person.

\end{appendices}
{\small
\bibliographystyle{ieee}
\bibliography{biblio}
}

\end{document}